# Machine Learning Approaches for Type 2 Diabetes Prediction and Care Management

Aloysius Lim[1], Ashish Singh[1], Jody Chiam[1], Carly Eckert[1,2], Vikas Kumar[1], Muhammad Aurangzeb Ahmad[1,3] *, Ankur Teredesai[1,4]

1. KenSci Inc.
2. Department of Epidemiology, University of Washington
3. Department of Computer Science, University of Washington Bothell
4. Department of Computer Science, University of Washington Tacoma

{aloysius, ashish, jody, carly, vikas, muhammad, ankur} @ kensci.com

* Corresponding Author: Muhammad Aurangzeb Ahmad maahmad@uw.edu

Abstract

Prediction of diabetes and its various complications has been studied in a number of settings, but a comprehensive overview of problem setting for diabetes prediction and care management has not been addressed in the literature. In this document we seek to remedy this omission in literature with an encompassing overview of diabetes complication prediction as well as situating this problem in the context of real world healthcare management. We illustrate various problems encountered in real world clinical scenarios via our own experience with building and deploying such models. In this manuscript we illustrate a Machine Learning (ML) framework for addressing the problem of predicting Type 2 Diabetes Mellitus (T2DM) together with a solution for risk stratification, intervention and management. These ML models align with how physicians think about disease management and mitigation, which comprises these four steps: Identify, Stratify, Engage, Measure.

## Introduction

Diabetes Mellitus (DM) is a group of metabolic disorders mainly caused by abnormal insulin secretion and/or action, resulting in elevated blood glucose levels (hyperglycemia) and impaired metabolism of carbohydrates, fats and proteins [1]. DM affects more than 463 million adults worldwide, and its prevalence is projected grow to 578 million by 2030 and 700 million by 2045. Its rapid rise is thought to be the result of a complex interplay of socioeconomic, demographic, environmental and genetic factors, including rising levels of obesity, unhealthy diets and widespread physical inactivity [2].

Diabetes has serious implications on individuals and health systems. Diabetes can cause numerous debilitating health complications, such as coronary artery disease, peripheral arterial disease, stroke, diabetic neuropathy, nephropathy and retinopathy [1]. Globally, 11.3% of deaths are due to diabetes. An estimated USD 760 billion is spent on treating diabetes, with over 50% of that spent on treating its complications. By 2045, this cost is projected to rise to USD 845 billion [2].

The good news is that the risk of diabetes and its complications can be reduced with therapies and lifestyle interventions, especially if the risks are detected early. A healthy diet and physically active lifestyle are some of the most important preventive factors for T2DM. There are many opportunities for ML to aid in early detection and intervention of T2DM, including predicting the risk of pre-diabetes, diabetes, and diabetic complications, identifying risk factors for T2DM, and optimizing interventions for effective management and follow-up.

## Literature Review

A large and diverse body of research exists on the applications of ML for diabetes. [1] provides a comprehensive review of the 103 articles published in the five years spanning July 2011 to July 2016, which can be classified in the following five broad categories: Biomarkers Identification and Prediction of DM (including Diagnostic and Predictive Markers, use of biomarkers for Prediction of DM), Diabetic Complications, Drugs and Therapies, Genetic Background and Environment and Health Care Management. For our purpose identification of diagnostic biomarkers, and diabetic complications are most relevant on the descriptive side. On the predictive side, the most relevant aspects of literature are (a) predicting T2DM and its complications, and (b) identify risk factors, especially controllable risk factors that can be used to design appropriate interventions.

Prediction of DM

The research on Prediction of DM constitutes diverse set of problem formulations, data sets, features and ML techniques applied to diabetes. The DM prediction problem can be framed in several ways. A summary of standard problem formulations is given in Table 1.

| **Problem Formulation** | **Reference** |
|---|---|
| Predict blood glucose level or glycemic status | [3]–[6] |
| Predict risk of diabetes (regression producing a probability) | [7] |
| Predict current pre-diabetes and/or diabetes (classification) | [8]–[17] |
| Predict onset of diabetes in a future time window | [18]–[20] |
| Predict progression from normoglycemia to prediabetes, normoglycemia to T2DM, prediabetes to T2DM | [21] |
| Generate association rules that predict diabetes and reveal risk factors | [22]–[24] |
| Predict death in diabetic patients | [25] |
| Predict comorbidities of diabetes (not necessarily complications due to diabetes) | [26] |

Table 1: Problem formulation for diabetes prediction

These studies explored different feature spaces, depending on the data sets available, intended use case, or research question. Thus, in most cases it is not possible to compare the predictive performance or the efficacy of the models across studies and settings. Many studies included the common risk factors for diabetes, including demographics (age and gender), family history of diabetes and BMI. In some cases (e.g. [10], [15]) lifestyle data were incorporated, such as smoking, alcohol consumption, coffee consumption, exercise, work stress and food preferences. Where medical records were available, diagnosis codes, prescriptions, labs (those most common being glucose and lipids) and vitals were commonly used. However, some studies (e.g. [9]) intentionally excluded labs or other medical data, as the intended use was to screen for diabetes in the least invasive way possible.

Some studies explored more interesting feature spaces. [8] constructed patient trajectories in terms of sequences of comorbidities (as indicated by diagnosis codes) to predict diabetes risk. [3] used salivary electrochemical parameters such as pH, redox potential, conductivity and concentration of sodium, potassium and calcium to predict fasting blood glucose. Anthropometric measures such as waist circumference, height, weight and BMI were used by several studies. In [5], the authors not only used the most extensive list of anthropometric measures, including circumferences at 8 body sites and 25 ratios of those, they also predicted fasting blood glucose using only these anthropometric measures. [26] used temporal patterns in EHR data, which are heterogeneous and irregularly sampled in the time domain.

Most diabetes prediction problems were framed as classification problems, with logistic regression, Naïve Bayes, support vector machines (SVMs), decision trees, random forests and artificial neural networks being the most commonly used ML techniques amongst these studies. While there is no clear winner, SVMs and tree-based models tend to outperform other models in side-by-side comparisons. It should however be noted that many of these studies do not employ gradient descent-based models for model building and comparison. Although the studies are not comparable, most models achieved AUC above 0.7, with a few models reaching 0.9. More complex models like ensembles of Bayesian models [21], Probabilistic Neural Networks [12] and LDA-MWSVM [17] did not necessarily outperform simpler models, and in some cases even underperformed them.

[7] developed models that produced consistent estimates of probability of diabetes using k nearest neighbors (kNN), bagged nearest neighbors (bNN), random forest (RF) regressors and random forest classifiers. The results demonstrated a trade-off where kNN and bNN produced more accurate probability estimates, while RF regressors and classifiers produced higher classification accuracy.

## Prediction of Diabetic Complications

Several studies have been conducted on ML techniques to predict complications associated with diabetes, using different data sets and methods. A survey of literature for prediction of diabetic complications is given in Table 2.

| Diabetes Complication | Reference |
| --- | --- |
| Cardiovascular Disease | [27], [28] |
| Coronary Heart Disease | [29] |
| Depression | [30] |
| Hypoglycemia | [27], [28], [31]–[33] |
| Ketoacidosis | [27], [28] |
| Liver Cancer | [34] |
| Microalbuminuria | [27], [28] |
| Nephropathy | [35], [36] |
| Neuropathy | [27], [28], [37]–[39] |
| Proteinuria | [27], [28] |
| Retinopathy | [27], [28], [40]–[49] |
| Maculopathy / macular edema | [50], [51] |

Table 2: Diabetes complication prediction literature survey

The table also illustrates that diabetes complications are a large heterogeneous category covering different medical conditions, and thus a wide variety of data sets, features and ML techniques were used. Literature survey also revealed that there is a sparsity of studies within each diabetic complication. This makes cross-comparison rather difficult. The exception to sparsity is diabetic retinopathy which appears to be the most widely studied complication in the list.

Most of these studies used commonly available clinical data such as demographics, labs and diagnoses. Genetic data were combined with clinical data in the prediction of diabetic nephropathy in [35], [36]. Ewing tests were used in the prediction of cardiac autonomic neuropathy (CAN) [37], [38]. Self-monitored blood glucose (SMBG) and professional continuous glucose monitoring (PCGM) data were used to predict hypoglycemia [31]–[33]. Fundus images [42], [44]–[48], [50], [51] were most commonly used to predict diabetic retinopathy (including maculopathy), but clinical records [40], [43], tear fluid proteomics [41], [42] and images of the tongue [49] were also used.

## Prediction of Non-Compliance

[52] developed ML models to predict intentional insulin omission in patients with T1DM, which happens often in adolescent females who omit or restrict insulin doses in order to lose weight.

## ML Design

Work by Lagani et al. [27], [28] serves as a useful reference for the design of the ML system described in the current manuscript. The various aspects of their approach which are similar to the approach we take are as follows: (a) identifying the minimal feature set for each risk event (in their case, one of seven diabetic complications), (b) developing models to predict future risk events, (c) applying the models to score a different population than the one the models were trained on, and (d) making the models easy to use via a web-based graphical user interface.

They developed a two-stage framework combining feature selection with regression to predict the risk (probability) for a specified diabetic complication (e.g., onset of microalbuminuria) at a future time (e.g. next 24 months). Four feature selection techniques – Survival Max-Min Parent Children, Lasso Cox Regression, Bayesian Variable Selection and Forward and Univariate Selection – were evaluated in combination with five regression techniques – Cox Regression, Ridge Cox Regression, Accelerated Failure Time, Random Survival Forest and Support Vector Machine Censored Regression. The best combination for each risk event (diabetic complication) was selected as the final model. Different models were found to be the best performer for each diabetic complication, and the number of features used ranged from 5 to 10. For example, the best model for predicting hypoglycemia was linear SVM with 5 features, while the best model for ketoacidosis was Ridge Cox Regression with 10 features.

Although the models were trained on data sets collected in the 1980's from American and Canadian patients, they were tested on contemporary European patients, demonstrating their transferability to new cohorts. Results were further improved by calibrating on the new data using Cox regression, while conserving the original predictive signatures.

Missing data is a practical issue when deploying risk prediction models on a new cohort of patients from a different EHR data source. Lagani et al. developed a missing information module that estimated the *distribution* of missing values based on other available features, using Bayesian Networks. The survival function for each risk event was then computed as the average of the original survival functions with the imputed values plugged in, weighted by the probabilities of the imputed values.

Once the models were trained, a web-based interface was developed where clinicians could select the type of risk, enter the minimal feature set for a patient (Figure 1), and get the predicted risks (probabilities) for that patient at different time horizons (Figure 2).

Although the framework was used to develop risk prediction models for diabetic complications, it can be generalized for other risk events such as the onset of pre-diabetes or diabetes. Other feature selection and regression techniques can also be used as candidate models, in addition to or in place of those used by Lagani et al.

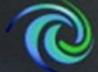

*Figure 1. REACTION platform providing a microalbuminuria risk assessment tool based on 7 predictive features* [28].

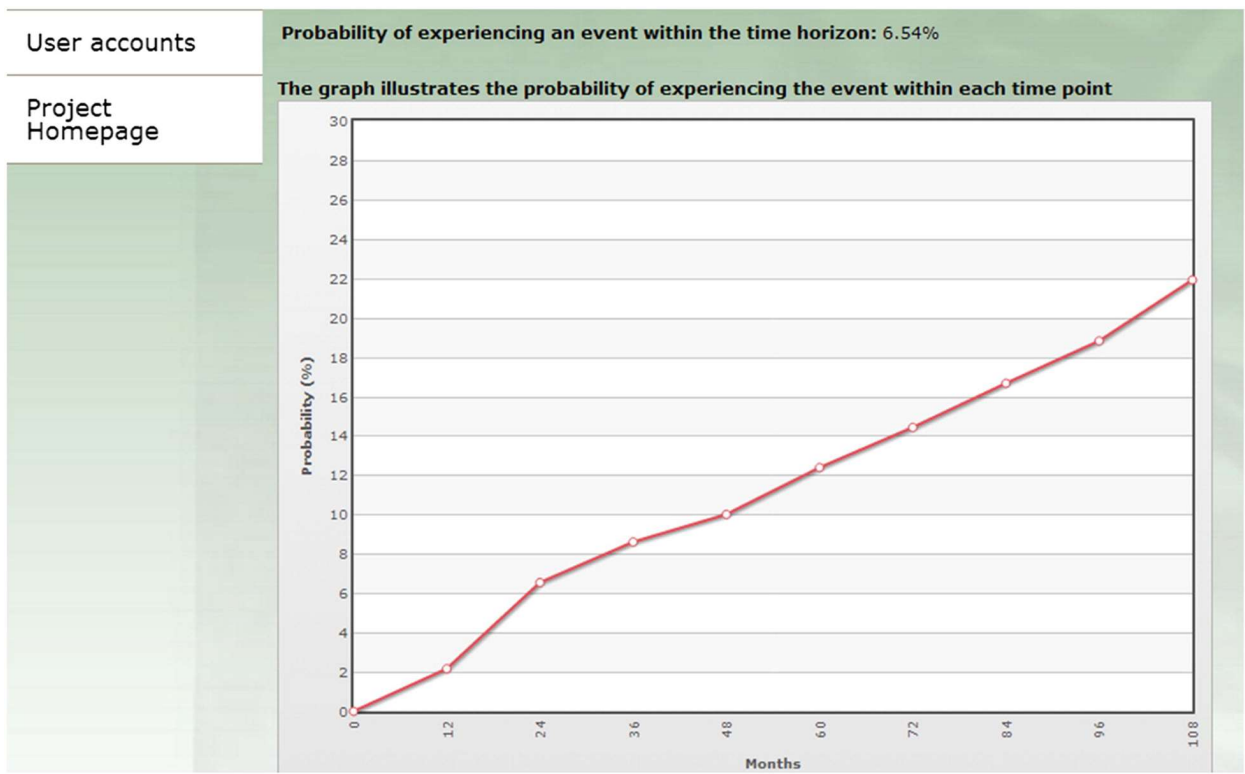

*Figure 2. REACTION platform showing microalbuminuria risk assessment for a given individual* [28].

## DM ED Utilization

Studies on ED utilization are mainly focused on identifying factors affecting ED usage of all patients in general, DM patients or across specific subgroups such as the Ambulatory Care Sensitive Condition[1] (ACSC) population [53], Hispanic and African American with diabetes [54], patients with severely uncontrolled T2DM (A1C > 10%) [55] or patients with both schizophrenia and diabetes [56]. The main difference among these studies is often the cohort criteria and the features used which will be discussed later. We considered two main bodies of literature, one focusing on all-cause ED usage and the other on DM-related ED usage. Studies on all-cause ED usage centered around understanding patients with at least one occurrence of ED use [54], [57]–[59] or those who are frequent ED users [53], [60]–[68] while DM-related ED usage studies covered DM-related ED visits in general [55], [56], ED visits relating to hypoglycemia [69]–[73] or both hypoglycemia and hyperglycemia [74]. All of the studies were framed at the patient level except for 2, which carried out the prediction at the ED visit level [56], [70]. Binary classification was most commonly used, some studies used multi-class classification [54], [56], [59], [67] and only 1 study carried out regression on the number of visits to the ED in a 12-month period [75].

Studies predicting ED usage had differing definitions of their target variable. All-cause ED usage predictions were defined with respect to a specific time frame, with a period of 12 months being the most common while studies on DM-related ED usage typically did not have a time frame. Studies in all-cause ED usage typically involves predicting the patients with $\geq x$ ED visits in the next 12 months ($x$ can range from 1 – 17, with 4 being the most common) [53], [57], [58], [60]–[66]. The cutoff $x$, was commonly determined by reviewing the distribution of ED visits and setting it based on the percentage of patients to target. Some studies created multiple models, one for each cutoff value and compared their performances before selecting a suitable cutoff [60], [62], [67]. Other studies stratified patients into 3 classes of low (0 visits), medium (1 visit) and high frequency of use ($\geq 2$ visits) [54], [59], [67]. DM-related (in general (E11) or relating to hypoglycemia (E11.64) or hyperglycemia (E11.65, E11.0, E11.1)) ED usage problems that were framed at the visit level aimed at predicting which of the ED visits were DM related among all other visits [56], [70] with the use case of identifying the relevant predictors of ED utilization whereas those at the patient level predicted whether patients will have any DM-related ED visits in the future [55], [69], [71], [73], [74].

All-cause ED usage studies which specified their cohort criteria included patients with at least 1 ED visit and considered ED visits that were made on consecutive days as the same episode by counting them only once [53], [67]. DM-related ED visit studies excluded patients who were pregnant, newly diagnosed with T2DM, type 1 diabetic and had life expectancies of < 12 months [55] while studies on Hypoglycemia-related ED visits applied stricter cohort definitions. Patients without diabetes, having type 1 diabetes or gestational diabetes as well as those with deliberate overdose of diabetes medications were excluded [69], [70], [72]. ED visits having secondary discharge diagnoses for hypoglycemia were also excluded since such events might have occurred during the ED encounter itself and is not the primary reason of the encounter [69]. ED episodes that occurred within 3 days after a prior event were excluded to avoid double counting [70]. Lastly, patients on lifestyle modifications alone were excluded while those who received at least one oral hypoglycaemic agent or insulin therapy (for more than 7 days) 3 months prior to the visit were included to ensure that the event can be attributed to the use of anti-diabetic agents [70], [71].

---

[1] ACSC refer to chronic conditions which can be managed effectively through primary care services and includes high blood pressure, diabetes, angina, asthma, COPD, CHF, etc.

Demographics, socioeconomic indicators, comorbidity, healthcare utilization, diabetes condition and treatment factors were found to be predictors of ED utilization. In terms of demographics, females [54], [60], [63], [64], [70], [71], younger (18 – 44) and older age groups (> 65) [53], [54], [57], [59], [60], [64], [70], [74], Hispanic, Black and African American [54], [56], [63], [64] were found to have higher ED utilization. Patients with less education [54], [65], [66], lower income [57], [59], [61], [65], [74], who are single parents, single or divorced [65] and those who are insured, especially with Medicaid or Medicare [63], [64], [68] are more likely to visit the ED. Patients with a history of substance abuse, mental illness [53], [57], [62], [63], [66], [68] or greater number of comorbidities [53], [59], [61]–[63], [65]–[67] were associated with all-cause ED visits. In terms of utilization, most studies found that prior ED or hospital utilization were predictive of future ED visits [53], [60], [63], [66], [67], [69]. However, while all-cause ED visit studies reported that having a usual source of care [64]–[66], [68] and high utilization of other medical services (ambulatory care visits, specialist or physician visits) [63]–[66] increase the likelihood of ED usage, one DM-related ED visit study disagreed and suggested that having a usual source of care and more primary care visits reduces the probability of hyperglycemia or hypoglycemia ED use [74]. Most of the studies found that having a longer duration of diabetes [57], [73], diabetic complications [54], [59], [71] and higher HbA1C measurements [73] contributes to ED usage. In terms of treatment, patients who require insulin or oral hypoglycaemic medications were more likely to visit the ED [54], [57], [69], [71], [73]. Specific hypoglycaemic medications such as Sulfonylurea [72] were found to increase the likelihood of hypoglycemia related ED visits while [71] reported the following diabetes-related medications, metformin, thiazolidinedione and dipeptidyl peptidase-4 inhibitor have helped to decrease hypoglycemia related ED visits. [71] also found that non-diabetes-related medications like NSAIDs, benzodiazepines, fluoroquinolones, warfarin, and trimethoprim led to increased hypoglycemia related ED visits while having a systolic blood pressure between 90-150 mmHg and a heart rate of > 110 bpm served as protective factors. Lastly, although there was no consensus on the directionality of ED use in terms of the distance between the patient's home and hospital, it was considered to be a significant factor [60], [66], [67].

The most common ML technique to model ED utilization was Logistic Regression [53], [57]–[62], [65], [68], [74] while other studies used Decision Tree, Support Vector Machine, Random Forest, and Adaboost [58], [67], [69] for classification whereas the study which performed regression made use of Negative Binomial regression [75]. Of the few studies which specified their model performance, all-cause ED usage predictions had AUC values ranging from 0.786 - 0.960 whereas DM-related ED usage predictions had AUC values between 0.759 - 0.830.

An interesting finding mentioned in a few studies predicting frequent all-cause ED usage was that most frequent users in a given year will not remain as a frequent user in the following year: patients with $\geq 4$ visits in a year have only 28 - 38% probability of being a frequent user in the next year [60], [64]. Of note, [64] also discussed predictions of chronic frequent users (frequent users over consecutive years), highest frequency users (> 20 visits in a year) and multiple ED users (treated at $\geq 5$ EDs in a year) in the literature in addition to frequent ED users. Lastly, [72] shared how the use of multiple data sources, namely hospital episode statistics and patient administration system in addition to ambulance electronic records have helped to increase the number of ED episodes that were identified by 79.4%.

## DM Visit Non-Compliance (All/DM patients)

Most of the literature surrounding visit compliance were around predictors of no-show, self-reported reasons of no-show (gathered through surveys and interviews with patients), effect of no-show on health outcomes which includes elevated A1C levels [76], [77] and significantly higher mean systolic blood pressure [78], and interventions to reduce no-show such as appointment reminders and patient education [79]. While the literature on visit compliance in general and those specific to DM have a lot in common, the main difference would be that studies specific to DM included DM related variables such as the time since the patient had DM or its complications, the patient's diabetes treatment type and intensity, and the patient's A1C measurements [80]–[82]. On the other hand, studies on visit compliance in general tend to include features such as the clinic type, medical specialty, healthcare provider as well as referral source [83]–[89]. No-show prediction was framed either at the appointment level or the patient level and binary classification was performed in all the studies (although it is possible to approach the problem using regression on the no-show rates).

Studies predicting no-show behavior had varied definitions of no-show. Those that predicted no-show at the appointment level considered no-show either as scheduled appointments that were missed (excluding any cancelled appointments) [84], [85], [88] or more commonly any missed appointments or late cancellations [90], [91]. Late cancellations were defined as cancellations that were made on the day of the appointment [92] or within 24 hours of the appointment time [86], [87]. On the other hand, in studies that modelled the problem at the patient level, greater variation was observed in how no-show was quantified. Definitions include the inability to visit a physician within 2 months after a missed appointment [80], missing more than one appointment within a pre-defined period, having no record of A1C measurements in the past 12-15 months [93] or whether a patient had missed more than 30% of appointments in a year [81]. Studies which specified their cohort criteria excluded patients that showed up without any prior appointment [89] as well as same-day appointments or appointments scheduled within 24/48 hours before the actual appointment [88], [91] since such appointments are most likely to be kept.

Determinants of no-show include patient demographics, comorbidities, treatment type, medication adherence, medical provider, patient appointment history and appointment characteristics. Characteristics of patients that are more likely to no-show include younger patients [84], [86], [87], [90], [91], [93], males [84], [90], smokers [93], patients with depressive symptoms [80]–[82] or psychiatric diagnoses [84], patients having greater co-payment for outpatient appointments [81] as well as those belonging to minority racial, ethnic, financial or insurance groups [81], [84], [93]. Those having higher comorbidity scores [81], those with family members or are married [84] or those who are not proficient in English [86] were less likely to no-show. Pharmacologic treatment of diabetes (as opposed to treatment only based on diet) and poor medication refill adherence [81] were associated with an increased risk of no-show while patients with intensive insulin therapy (> 2 insulin injections daily) and those who carry out daily self-monitoring of blood glucose were less likely to no-show. As for provider characteristics, patients with clinicians having greater expertise were less likely to no-show [84]. Multiple studies found that the risk of no-show increases with the duration between when the appointment was made and the actual appointment date [84], [86], [87], [90], [91], [94]. Also, if the patient has multiple appointments on the same day or if the appointment was a follow up, the probability of no-show is reduced [91]. Although no directional effects were mentioned for the following variables, they were found to be significant no-show predictors as well: amount of co-payment due [92], amount of diabetic medications a patient had left [94], the number of previous no-shows [84], [86], [89], the number of previous appointments [84], whether the appointment was rescheduled [92], if past kept appointments were on the same day of week as the current appointment [94], the appointment length [92] as well as patient engagements such as the number of calls and whether a phone reminder was provided [92].

In terms of machine learning algorithms employed, Logistic Regression was most commonly used [80]–[82], [86], [88], [91], [92], [94]. Others such as [83] used Logistic Regression, Support Vector Machine and Recursive Partitioning, [85] used a hybrid probabilistic model based on Logistic Regression and Bayesian Inference while [89] used a Gradient Boosting Machine. Prediction performance of the models ranged from AUC values of 0.710 to as high as 0.958.

Although many factors were found to be significant in no-show prediction, factors complicating predictions include patients who switched clinic without prior notice and extreme weather or traffic conditions which prevented appointments from being kept [84], [94]. In addition, there could be underlying causes of no-show that might be hard to detect from the data, these are self-reported reasons from patients such as forgetting the appointment, competing priorities or conflicts, the patient's health status, appointment scheduling problems or financial problems [84]. [91] included a discussion on ways to encode the appointment attendance history of patients. While common representations include describing attendance history as a rate or using the outcome of the most recent appointment, the paper proposed representing patient histories using binary sequences (e.g. a sequence of 100 indicates that a patient had a total of 3 appointments, missed the most recent appointment and showed up for the previous 2) and computing the conditional probability of no-show on the next appointment given the sequence from past appointments based on the existing sequences observed in the data. The study capped the sequence length at 10 since the represented number of appointments for each sequence degraded once it exceeds 10. An advantage of this method would be that it is capable of adjusting for the recency of past attendance. For instance, the no-show probability calculated for a patient with sequence 0000011111 which attended the 5 most recent appointments was 0.197 while a patient with sequence 1111100000 which missed the 5 most recent appointments has a no-show probability of 0.547.

## DM Medication Non-Adherence

Past studies on medication adherence were focused on the factors influencing medication adherence, patient-perceived barriers to adherence such as fear of side effects or embarrassment of insulin injection in public [95]–[98], outcomes of poor medication adherence which includes higher HbA1c levels, onset of diabetes complications, increased risk of morbidity and mortality and increased costs from outpatient care, emergency department visits and hospitalization [97]–[99]. Lastly, there were also discussions around strategies to improve adherence such as reducing medication complexity, providing better patient education, improving communication with health care providers, reminder systems and reducing out-of-pocket costs for patients [97]–[101]. All of the literature that were reviewed covered patients with T2DM that are on oral antihyperglycemic drugs (OAD) and/or insulin except for [102], [103] which excluded patients on insulin, [104], [105] which focused on patients newly initiating statins and [106] which includes patients who filled a prescription for a statin, antihypertensive, or oral antidiabetic. Medication adherence prediction was framed at the patient level and it could be either on a single medication or multiple medications. Medication adherence is made up of two separate components, compliance (the degree to which a patient takes medication as prescribed which includes filling prescriptions, taking medication on time or administering the right amount of injections) and persistence (the duration from initiation to discontinuation of therapy) [95], [97], [107]. Binary classification or multi-class classification is typically performed for predictions on compliance while survival models were used for predicting persistence.

There exists a wide variety of compliance measures such as (1) biological measurements of therapeutic outcomes (blood glucose levels, urinary glucose levels or glycosylated hemoglobin concentration) which are more sensitive but can be invasive [97], (2) self-reported measures (generic for medication adherence in chronic conditions: Morisky Medication Adherence Scale (MMAS), Medication Adherence Rating Scale (MARS) or diabetes specific: Diabetes Self-Care Inventory (SCI), Summary of Diabetes Self-Care Activities (SDSCA)) which are often inaccurate and subjective [99], [108], (3) pill counts which requires patients to return their pill bottles and are time intensive [101], [107], [108], (4) Medication Event Monitoring System (MEWS) which does electronic monitoring of the time and frequency of bottle openings on pill bottles fitted with microprocessors [107], [108]. Despite being able to provide a precise measurement, MEWS is expensive and requires significant time and effort to implement. A widely used method that is objective and suitable for measuring adherence over large populations would be utilizing (5) pharmacy or medical records to compute measures such as the Medication Possession Ratio (MPR) or the Proportion of Days Covered (PDC) which are reported in terms of the percentage of time in which a patient has medication available [95], [97], [99], [107]–[109]. These measures do not directly measure medication consumption behavior but rather they measure medication collecting behavior. Lastly, (6) measures suitable for insulin adherence such as the Ecological Momentary Assessment (EMA) and computerized logbooks were also introduced in [108].

MPR and PDC are the most widely used measures in the domain and thus require further elaboration. It should be noted that there are some caveats when measuring non-compliance for insulin. This is so because the daily dosage is dependent on the blood glucose level. In contrast, OAD doses are fixed, so the PDC/MPR measures are more accurate. The key variables required for computing both measures are the drug name, the date filled and the number of days of supply. MPR refers to the total number of days' supply in a time period divided by the number of days in the time period and while PDC is similar to MPR, it considers the days that are covered (number of days patient has access to medication) instead of the total days supplied (number of days of medication the patient has on hand). Although a time frame of 1 year is typically used [102], [103], [110], [111], a much shorter time frame of 3 months was also observed [101] and patients are considered as adherent if their MPR or PDC $\geq x$% and non-adherent otherwise ($x$ is most commonly fixed at 80% [97], [98], [101], [102], [110], [112], a cut-off at 90% was also used in [103]). While binary classification is conventionally used, [113] split PDC into 6 categories (PDC <20%, 20%≤PDC<40%, 40%≤PDC<60%, 60%≤PDC<80%, 80%≤PDC<100%, and PDC=100%) and carried out multi-class classification.

A disadvantage of MPR is that it tends to overestimate adherence in some cases, for example, patients who routinely refill their medications ahead of time will have overlapping days supplied, causing their MPR to exceed 100%. PDC on the other hand will shift the overlapping prescription to begin the day after the patient has completed his medication from the previous fill and this guarantees that PDC will not exceed 100%. In terms of handling multiple medications when using PDC, [111] first calculated the average PDC of each of the prescriptions within each oral hypoglycemic class and subsequently computed the group mean of the averages. Other approaches include [113] which required only 1 medication to be available to the patient for a day to be considered as covered while [114] defined a stricter approach in which all medications have to be available to be considered as covered. Unlike PDC, MPR is not as flexible and the only way to calculate MPR for multiple medications would be to take an average of the MPR of each drug. Furthermore, skewed values will be produced upon computing the average should the MPR of any of the drugs exceed 100% [114]. Given the advantages that PDC has over MPR, it is becoming the preferred measure of compliance [114].

A methodology that has been gaining traction in medication adherence is Group-Based Trajectory Modelling (GBTM) which is able to capture the dynamic patterns of medication adherence over time as compared to a single static measure of adherence like PDC or MPR. For example, it is possible for patients with initial consistent use but poor subsequent use, poor initial use and consistent subsequent use or intermittent adherence throughout to share the same PDC value. GBTM provides an alternative method for representing long-term medication adherence that accounts for the fluid nature of adherence and it can be used to segment groups of patients with similar adherence patterns, analyze how adherence patterns change across time, predict adherence by fitting models with the trajectory groups as labels or even as variables in subsequent predictions (risk of hospitalization or ED visits) [106], [115]–[117].

First, monthly measures of adherence (e.g. PDC≥80%) for each patient over some pre-defined period, known as trajectories are calculated. The trajectories, along with a set of covariates are fitted into a mixture model which is used to identify subgroups of patients with similar trajectories, summarized by a set of polynomial functions of the covariates that were determined through maximum likelihood estimation using a quasi-Newton procedure [105], [118], [119]. GBTM then gives an approximate trajectory curve for each group, estimates the probability of group membership for each patient and assigns them to the group for which they have the highest membership probability. The most common adherence trajectory groups identified in the literature were (1) consistent, high adherence, (2) early and consistent adherence, (3) declining adherence, and (4) initial non-adherence followed by a slight increase in adherence [115]. The advantages of GBTM are being able to capture changes in adherence as well as customize interventions based on adherence trajectory memberships [115]. However, GBTM requires a longer observational period of at least 6-12 months, is more complicated to implement and measuring adherence of multiple medications is challenging and less researched upon [116].

Although most of the literature were focused on the compliance aspect of medication adherence, a few discussed persistence as well. Predictions on persistence could be framed as a binary classification problem on whether a patient persists with treatment after 1 year [107] or as a survival model to predict the time from initiation to discontinuation of medication [97], [112], where discontinuation is defined as a gap of more than $x$ days without any available drug supply ($x$ is commonly fixed between 30-90 days) [98]. In [112], the case in which a patient switched to a different medication within the same drug class was also considered as a discontinuation.

Studies which specified their cohort criteria included patients that are diagnosed with T2DM and having at least 1 prescription of an OAD for at least 6-12 months prior to the prediction [102], [103], [120]. Some studies excluded patients receiving insulin, pramlintide, or exenatide prescriptions alone or those only on dietary treatments [102], [103], [111] while [103], [111], [121] excluded patients with < 2 OAD fills which includes patients with primary non-adherence that never filled their initial prescription. Additionally, patients with type 1 diabetes [121], gestational diabetes [111], [116], a new T2DM diagnosis [120], a life expectancy of < 1 year [120], Metformin use due to polycystic ovary syndrome [111], [116] as well as evidence of prolonged institution (their medications will be administered by nurses or caregivers) [111], [116] were excluded. In some cases, patients receiving their medication by mail were also excluded since the distribution of PDC for such patients will be different from that of patients that collect their medication from the pharmacy [105]. Lastly, [106], [117] adjusted the PDC/MPR denominator to account for the days in which the patient was hospitalized.

Predictors of medication compliance include patient demographics, socioeconomic indicators, comorbidity, healthcare utilization, diabetes condition, treatment burden, and treatment-related beliefs. Characteristics of patients with lower medication compliance include younger patients (25 - 44 years old) [99], [101], [105], [112], [115], [116], [121]–[123], non-whites [101], [115], [116], those who are obese [97], professionally active [99], [115], [121] or with lack of support from their family members [97], [99], [100], [121], as well as those with lower income [97], [101], [105], [121], lower educational level [105], [115] or those with depression, alcohol abuse or drug abuse [100], [110], [115], [120]. Patients with more comorbidities were less compliant [102], [112], [115], [123], while [113] found that those with a longer duration of hypertension or dyslipidemia were associated with higher medication compliance. Patients with a higher number of past hospitalizations or ED use [102], [116], fewer diabetes-related tests [116], those with longer duration of diabetes [100], [103], diabetes complications [112], [121] or hbA1c measurements of > 8% [113], [121] tend to be less compliant.

On the other hand, patients who experienced a recent critical medical event [115] or those with higher compliance to previous medications [104], [115] were more compliant. In terms of treatment burden, patients with higher co-pays [49, 60, 58, 45, 47, 62, 48, 65], greater complexity of medication regimen in terms of more daily number of tablets, number of co-medications, higher dosages (e.g. short-acting insulin requiring multiple injections vs. once daily injection of insulin glargine) [97]–[100], [103], [108], [109], [112], [115], [120], patients experiencing side effects from medications [96], [99], [101], [120], especially a history of a hypoglycemic event [98] were associated with lower medication compliance since patients might deliberately keep their blood glucose levels high to prevent further hypoglycemic events from occurring. Conversely, those using a mail order pharmacy as compared to retail pharmacy [99], [112], those using a pen device instead of conventional syringe and needles for insulin injections [95], [98], and those having prescription of medication for one other chronic condition [99] tend to have higher levels of compliance.

As for treatment-related beliefs, perceived treatment efficacy [96], [98]–[100], [120]–[122] as well as having trust and a good relationship with one's health care provider [97], [98], [100], [121] led to improved compliance. Although directional effects were not mentioned or there was no consensus on the directionality for the following variables, they were found to be significant predictors for medication compliance as well: gender, length of fill, number of available refills, number of phone calls made by patient, and having at least one missed appointment [95], [100], [101], [106], [112], [115], [124]. Lastly, while most of the studies were focused on the predictors of medication compliance, [112] stated that the factors resulting in better or worse persistence mirrored that of the factors identified by compliance.

The most common machine learning technique used to model medication compliance was Logistic Regression [103], [104], [112], [117], [120], [121], [123], [124] while other studies used ordinal regression [113] or boosting algorithms [106]. Studies modelling the time to discontinuation of the drug made use of Cox Proportional Hazard models [112]. Of the few studies which specified their model performance, medication compliance predictions had AUC values ranging from 0.695 – 0.881 when PDC was used.

Of note, [125] suggested a method to adjust for the number of days supplied for insulin to reflect a more accurate time between insulin prescriptions, thus addressing the challenges in using refill data to measure insulin use since patients are required to adaptively adjust the insulin dosage according to their blood glucose readings [110]. The authors found that patients receiving a 30 days' supply of insulin tend to fill their prescription after an average of 45 days from the previous fill. As such, they adjusted the days supplied by a factor of 45/30 to obtain a more appropriate measure of compliance.

## Domain Problem Description

**Motivation:** Primary care providers (PCP) are burdened with numerous non-clinical tasks and laborious chronic disease management activities that do not require direct physician intervention. Patients with DM in particular require significant condition-specific management. To support PCPs, care managers manage the condition-specific care of patients with DM. Because of the large number of patients with DM, the patient panel size of each care manager is so large that they can only focus on the uncontrolled diabetic patients (those with A1C>7%), with the priority being to bring their diabetes back under control.

**Scope:** The goal of the intended use case solutions for T2DM is to use machine learning to assist care managers in caring for patients with DM. This will improve patient care, reduce the burden on PCPs, allow care managers to perform their job more efficiently, and improve overall staff satisfaction. With the assistance of ML predictions and optimizations, care managers may eventually expand their work into the proactive care of controlled or pre-diabetic patients (A1C <7%). Furthermore, we aim to improve quality of diabetes care for patients and reduce complications.

In particular, care managers want to identify risks of undesirable clinical events, such as disease progression, development of complications and non-compliance, in order to identify appropriate evidence-based interventions that may mitigate these risks. The needs of care managers may be classified as follows:

1. **Risk Stratification**: Who do I need to pay attention to?
2. **Care Plan Optimization**: What can I do for each patient?
3. **Outreach Optimization**: How do I get in touch with them?

Traditional care management is typically reactionary: cases appear in a work queue after a patient hits a threshold on utilization or clinical metrics or has an exacerbation event that necessitates high-touch engagement, such as an acute hospitalization or a recent avoidable ED encounter. Furthermore, recent hemoglobin A1c is the only quickly accessible measure care managers use to identify patients for interventions. This means that care managers have limited insight into patients without a recent A1c measurement (or who are just below the threshold of 7%) who may be nonetheless at risk for disease progression and complications.

**Goals:** The goal of predictive models for diabetes is to flag patients for engagement *before* a trigger event that requires a care manager's intervention occurs, allowing proactive correction. Additionally, predictions can help care managers prioritize activities by identifying *modifiable* risk factors (e.g., if the model identifies a patient to be at high risk due to smoking, smoking cessation could be recommended).

# Machine Learning Problem Description

To address the needs of care managers, the following ML goals can be defined with respect to the three dimensions described above:

| Need | Domain Goal | Modeling Goal |
|---|---|---|
| **Risk Stratification** | Enable care managers to identify the patients who are pre-diabetic and at risk of developing T2DM (A1C >=6.5%) | Predict risk and/or time to T2DM from pre-T2DM |
| | Enable care managers to identify the patients who are at risk of developing uncontrolled T2DM (A1C >=7%) | Predict risk and/or time to uncontrolled T2DM from T2DM |
| | Enable care managers to identify the patients who are at risk of developing microvascular complications associated with T2DM | Predict risk and/or time to event of developing microvascular complications associated with T2DM<br>a. Neuropathy<br>b. Nephropathy<br>c. Retinopathy |
| | Enable care managers to identify the DM patients who are at risk of utilizing the ED for any reason | Predict all-cause ED utilization among DM patients |
| | Enable care managers to identify the patients who are not likely to show up for their scheduled appointments | Predict visit non-compliance (all patients) |
| | Enable care managers to identify the DM patients who are not likely to show up for their scheduled appointments | Predict visit non-compliance (DM patients) |
| | Enable care managers to identify the patients who are at risk of being non-adherent to medications | Predict medication non-adherence |
| **Care Plan Optimization** | Enable care managers to see the risk factors and the strength of their association with a predicted risk; show risk factors and actions in a ranked order based on strength of association with a predicted risk | Measure and rank features associated with each of the predicted risks (above), globally |
| | Enable care managers to see modifiable factors and the strength of their association with a predicted risk | Measure and rank features representing *modifiable risk factors* associated with each of the predicted risks (above), *per patient* |
| **Outreach Optimization** | Enable care managers to identify the optimal time of day to engage with patients via telephone visits, and whether patient reads email | Predict probability of response to a contact event, across different channels |

*Table 1. Care Manager optimization goals mapped to machine learning prediction problems*

# Machine Learning Design Constraints

The machine learning solution has the following design constraints for the use cases that we seek to address in this manuscript:

1. **Training and scoring on disparate populations**: The trained machine learning models will need to be deployed in an environment where the underlying population may be different from what is encountered in the training setting. Training, tuning and evaluation will be done centrally in a pre-deployment environment using a data set and population to be selected, but deployment and scoring will be done in a new environment on an unseen data set. It is therefore important to clearly state the assumptions about the underlying population that may affect scoring outcomes, characterize the cohorts on which the model was trained on, and measure model performance and biases on different cohorts.
2. **Minimal, readily available feature set with clear definitions (Bring Your Own Features)**: In the target deployment scenario, the deployment environment will compute the features required for the models to score the patients. Model tuning should be optimized to select the *minimal* set of features (say < 15) that give reasonable model performance, so as not to burden the end user with many features to compute and manage. Feature selection will be a critical part of the model training process. The selected features should be *easy* to compute based on *readily available* EHR data that all providers will have. The end user should be provided documentation on the required features with simple, unambiguous definitions so that the end user can compute the features to use the models.
3. **Modifiable vs unmodifiable risk factors**: Features used need to be clearly classified as modifiable (e.g., smoking) vs unmodifiable (e.g., gender). This distinction is important in the generation of interventions that can help to reduce risk. However, it is important to note that since we are not dealing with causal models, taking an action on a modifiable feature will not necessarily have an impact on the patient's risk. At best the modifiable factors can be used to inform the generation of risk for risk reduction. Additionally, we note that many of these types of features that are salient to this problem space (smoking, insurance, etc.) may be encoded differently across populations or health systems.
4. how do we make use of these features when the encoding of them differs across systems?
5. **Robust to missing features**: The trained models must be robust in the face of missing features. Not all required features may be available at different sites. The models should still be able to score the provided instances (albeit at a lower level of model performance) if some features are missing. The robustness of models to missing features should also be measured.
6. **Easily packaged and deployable in any environment**: Models should be able to be packaged and easily deployed in any environment, which could be as simple as a single virtual machine. There should not be any specialized technical requirements such as GPUs or clusters. There are a number of technical solutions that can be explored to enable easy and flexible deployment, such as packaging the models in containers with REST API interfaces exposed. Additionally, there is a need to understand the actions that end users are most likely to take upon these model outputs. This is important as there is the possibility for unfairness or bias (explicit or implicit) to creep in during delivery. Consequently, there is need to put guardrails upon use of these models before we package and deploy them.

7. **Explainable globally and locally**: The models and their underlying features should be easily explained at both the global and local levels, due to the business need for identifying modifiable risk factors.

The machine leaning models that we consider can be organized into the model prediction flow depicted in Figure 3.

1. Predict risk of pre-diabetes for normal (healthy) patients [10], [13], [15], [21].
2. Predict risk of diabetes [4], [7]–[13], [16]–[20], [22] for (a) normal [21], [24] (b) pre-diabetes [21] patients.
3. Predict risk of uncontrolled diabetes for (a) normal, (b) pre-diabetes and (c) diabetes patients.
4. Predict risk of specific diabetic complications for (a) diabetes and (b) uncontrolled diabetes patients.

We define the patient risk as follows:

a. *The likelihood of the risk event occurring in a specified time frame* [18]–[20], [27], [28], e.g., "30% probability of diabetic retinopathy in 6 months, 50% in 12 months". This can be predicted with probability scores from classification models. As an extension, this can also be presented as the likelihood of *early vs. late onset* of the risk event, where the relevant time frames (e.g., 6 vs. 12 months) can be defined for early or late onset.
b. *The time to the risk event occurring*, e.g., "onset of diabetic retinopathy is likely to happen in 6 months". This can be predicted using regression models that estimate the time of onset for the risk event.

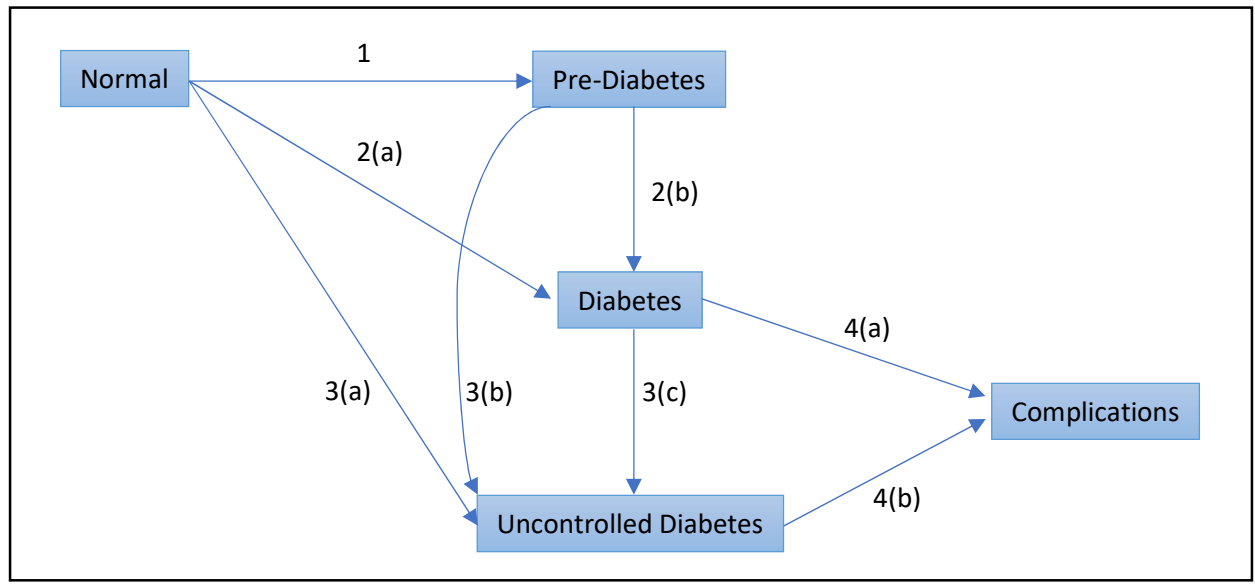

*Figure 3. Machine Learning prediction flow set of problems*

## Data Requirements

The dataset that we consider contains patient encounters from January 2016 to June 2020. Temporal outliers are removed i.e., data with dates from the future and distant past relative to the majority of the encounters in the data set. Table 2 summarizes the features available for model training and testing. A full list can be found in Appendix F: Features for Model Training). Specific features used in each model will depend on the goals and usage setting of each model.

| Category | Number of Features |
| --- | --- |
| Demographic | 4 |
| Diagnosis / Conditions | 398 |
| Lab | 50 |
| Lifestyle | 1 |
| Specialty | 242 |
| Utilization | 12 |
| Vitals | 47 |

Table 2. Available features for model training.

## DM Disease Prediction

The relevant conditions can be identified by ICD codes. ICD code corresponds to the International Statistical Classification of Diseases and Related Health Problems (ICD), a medical classification list by the World Health Organization (WHO). The set codes contain codes for diseases, symptoms, abnormal conditions, complaints, social circumstances, and external causes of injury or diseases. It is widely used in the encoding healthcare data including the data that we plan to use for model building described in this manuscript. ICD codes for T2DM and its complications are as follows:

| Condition | ICD-10 (WHO) | ICD-10-CM (U.S.) |
| --- | --- | --- |
| **T2DM** | E11 | E11 |
| **Diabetic Neuropathy** | E11.4<br>G73.0?<br>G99.0?<br>G59.0?<br>G63.2? | E11.4 |
| **Diabetic Nephropathy** | E11.2<br>N08.3 | E11.21 |
| **Diabetic Retinopathy** | E11.3<br>H36.0 | E11.31 – E11.35 |

Table 3. Relevant ICD-10 codes

Diagnostic criteria as defined by the American Diabetes Association (except for Uncontrolled Diabetes) is described as any one of the three criteria need to be satisfied for a diagnosis [126] as given in Table 4.

| Cohort | A1C | Fasting Plasma Glucose (FPG) | 2-h Plasma Glucose (PG) during 75-g Oral Glucose Tolerance Test (OGTT) |
| --- | --- | --- | --- |
| **Healthy** | <= 5.6%<br>(<= 38 mmol / mol) | <= 99 mg / dL<br>(<= 5.5 mmol / L) | <= 139 mg / dL<br>(<= 7.7 mmol / L) |

| Cohort | A1C | Fasting Plasma Glucose (FPG) | 2-h Plasma Glucose (PG) during 75-g Oral Glucose Tolerance Test (OGTT) |
|---|---|---|---|
| **Pre-T2DM** | 5.7% to 6.4% (39 to 47 mmol / mol) | 100 to 125 mg / dL (5.6 to 6.9 mmol / L) | 140 to 199 mg / dL (7.8 to 11.0 mmol / L) |
| **T2DM** | 6.5% to 6.9% (48 to ? mmol / mol) | >= 126 mg / dL (>= 7.0 mmol / L) | >= 200 mg / dL (>= 11.1 mmol / L) |
| **Uncontrolled T2DM** | >= 7.0% | | |

*Table 4. Diagnostic criteria for Type 2 Diabetes Mellitus*

DM ED Utilization

For the ED utilization problems, the following sources of data need to be considered.

1. ED encounter data to would be used to define the target labels for each DM patient.
2. Diagnosis and problem list. Table 5 gives the relevant conditions which are required construct features on the type of historical ED visits that were made by a patient.

| Condition | ICD-10 |
|---|---|
| **T2DM** | E11 |
| **Hypoglycemia** | E11.64 |
| **Hyperglycemia** | E11.65, E11.0 (hyperosomolarity), E11.1 (ketoacidosis) |

*Table 5. ICD-10 Codes for T2DM, Hypoglycemia and Hyperglycemia*

3. NYU ED Algorithm (EDA) [127], [128] to identify the types of ED visits to exclude. The EDA is a tool that can be used to classify ED visits based on ICD diagnosis codes into the following categories:
    a. Non-emergent: visits that do not require immediate medical care within 12 hours.
    b. Emergent – primary care treatable: visits that require treatment within 12 hours that could have been effectively treated in a primary care setting.
    c. Emergent – preventable / avoidable: visits that require emergency department care but could have been avoided with appropriate primary care management.
    d. Emergent – not preventable / avoidable: visits that require emergency department care and could not be prevented through ambulatory care treatment.

Due to the unique clinical needs of each patient, where patients sharing the same diagnosis code can have conditions with differing levels of severity (e.g., a diagnosis of "abdominal pain" might be given to patients complaining of stomach pain as well as those complaining of chest pain (possibly a heart attack)), the algorithm assigns a specific percentage to each of the categories above for each diagnosis code. The EDA also separately assigned each of the diagnosis codes to categories of mental health, alcohol, substance abuse and injury.

DM Visit Non-Compliance (All/DM patients)
1. The scheduled appointment data is used to:
    o Identify the appointments that will form the train and test data.
    o Create the target labels.
    o Create features using information from the current appointment.
    o Create features using past appointment information of each patient.

2. Calendar data is required to identify if appointment falls on / close to a holiday.

### DM Medication Non-Adherence

The following data is needed to establish DM medication non-adherence.

1. A table of diabetic drug classes and drug classes related to the medical management of diabetes as well as the name of the medications within each drug class.
2. Medication order data of medications to be taken in the ambulatory setting.
3. Medication order dispensing data of medications to be taken in the ambulatory setting.

Records from 2 and 3 is used to calculate the 1-year PDC for each patient and drug class to form the target label.

## Model Specifications

### Risk Prediction Model Framework

Many of the risk prediction models in this solution described in this manuscript attempt to predict the incidence of a risk event, most commonly incidence of disease. As such, a common framework can be used to construct such models. In predicting risk events, one is often interested to know (a) the likelihood of the risk event occurring at some future time, or (b) when the risk event is most likely to occur. These problems can therefore be framed as either (a) a classification or survival problem or (b) a regression problem, respectively. In both instances, the prediction concerns a future event that has not happened. During model training, the training data set will contain examples for which the risk event has happened and others for which it has not happened. This does not mean that the event *will not* happen, just that it *has not* happened by the end of the study period in the data set. Such data is said to be *right-censored*, where the observed time-to-event is less than or equal to the true time-to-event [129]. Furthermore, the length of time for which data is collected per patient is usually variable in real-world medical data sets, with some patients having a significantly longer history of data than others. Supervised ML algorithms assume that the status of the risk event is known for all patients (positive or negative) [130]. One approach to apply supervised ML on censored data is to discard censored observations. However, this causes a loss of highly valuable information – the fact that those patients went a certain amount of time without experiencing the risk event is itself informative [131]. Another approach is to treat these censored observations as the negative class. This is not accurate, since some patients may experience the risk event shortly after the end of the study period; the resulting model will tend to underestimate the risks.

An alternative approach commonly used in healthcare is using survival analysis methods to model the likelihood and time-to-event of specific risks. [129] provides a useful taxonomy of statistical and machine learning methods for survival analysis. Of particular interest are the machine learning methods, which are able to model nonlinear relationships with overall higher quality of predictions. In a survival analysis problem, a binary indicator $\delta$ indicates whether the risk event has occurred for a given patient by the end of the study period. The time to the event of interest $T$ is known only for patients where the risk event has occurred ($\delta = 1$) during the study period. For other patients ($\delta = 0$), only the censored time $C$ is known, which typically represents the end of the observation period in the data set. Each patient therefore only has $T$ or $C$ but not both. Thus, the input to a ML survival model for patient $i$ is the triplet $(x_i, y_i, \delta_i)$, where $x_i$ is the feature vector and $y_i$ is the observed time, which is equal to $T_i$ for an uncensored instance and $C_i$ for a censored instance. The goal of survival analysis is to estimate the time to the risk event $T_j$ (non-negative and continuous) for a new patient $j$ with features $x_j$.

There following table summarizes three ways to formulate the ML problem for risk prediction:

| Output | ML Problem | Label | Pros | Cons |
|---|---|---|---|---|
| **Risk of event (%) in x months** | Binary classification | Binary indicator $\delta$ of incidence of the risk event. | Can provide longitudinal predictions (e.g. risk in 3, 6, 12 months). | Treats censored data as negative class, introduces bias. |
| **Risk of event (%) in x months and/or time to event (depending on model)** | Survival | Binary indicator $\delta$ with observed time $y$, where: $y = \begin{cases} T, & \delta = 1 \\ C, & \delta = 0 \end{cases}$ $T$ is the actual time-to-event $C$ is the censored time. | Can provide longitudinal predictions. Works with censored data. | Evaluation metrics (e.g. C-index, Brier Score) are not widely understood. |
| **Time to event (e.g. months)** | Regression | $T$, the actual time-to-event for non-censored cases. | | Ignores censored data, losing potentially valuable information. Cannot provide longitudinal predictions. Difficult to justify a single point-in-time prediction (e.g. "patient is likely to get T2DM in 4.5 months"). |

*Table 6. Machine learning problem formulations for diabetes risk prediction*

The recommended approach for risk predictions in this solution is to frame them as **survival analysis** problems based on the limitations of the other approaches described previously. In order to train ML models for survival analysis, model training data needs to be transformed into the appropriate format. For this solution, existing encounter-level features will be used. The data set comprises 33.2 million encounters for 755,000 patients, with a mean of 44 encounters per patient.

For each patient, the encounters are sorted by date. The date of first incidence of each type of risk (e.g., pre-T2DM, T2DM, diabetic neuropathy) are also identified and compared to the dates of each encounter. This may result in the following hypothetical timeline of three patients' encounters and risk events illustrated below, where:

- $t_i$ is the $i$th time period
- $E_j^p$ is the $j$th encounter for patient $p$

- $R_r^p$ is the first occurrence of risk event $r$ for patient $p$

| Time | $t_1$ | $t_2$ | $t_3$ | $t_4$ | $t_5$ | $t_6$ | $t_7$ | $t_8$ |
|---|---|---|---|---|---|---|---|---|
| Patient 1 | $E_1^1$ | | $E_2^1$ | | $r_{T2DM}^1$ | $E_3^1$ | | $E_4^1$ |
| Patient 2 | | $E_1^2$ | | $E_2^2$ | | | | |
| Patient 3 | | | $E_1^3$ | $E_2^3$ | | $E_3^3$ | | $r_{T2DM}^3$ |

In this example, there are a total of 9 encounters, each with a set of computed features such as diagnoses, labs and vitals. We can make the following observations:

- The time of the first and last encounter for each patient is not fixed. This is reflective of real-world clinical data, where different patients enter or leave the data set at different times.
- The total observation period (date of first to last encounter) is different for each patient.
- Data is right-censored for some patients (e.g. Patient 2), where the risk event is not observed during the observation period.
- For other patients, the risk event is observed, and the time-to-event from each encounter can be computed.

To form the training set for ML survival analysis models, the time to event (for observed risk event) or time to censoring (for no risk event observed) can be computed, provided the risk event has not happened before the encounter date. Encounters that occur after the risk event are not used, as the risk event has occurred and does not need to be predicted ($P(T > t) = 0$ for all future time $t$). The features and labels for each encounter can thus be computed as illustrated below:

| Encounter Features | Binary Indicator $\delta$ | Time to Event $T$ | Censored Time $C$ | Observed Time $y$ | Remarks |
|---|---|---|---|---|---|
| $E_1^1$ | 1 | 4 | N.A. | 4 | |
| $E_2^1$ | 1 | 2 | N.A. | 2 | |
| $E_3^1$ | 1 | -1 | N.A. | -1 | Not used ($y \leq 0$) |
| $E_4^1$ | 1 | -3 | N.A. | -3 | Not used ($y \leq 0$) |
| $E_1^2$ | 0 | N.A. | 2 | 2 | |
| $E_2^2$ | 0 | N.A. | 0 | 0 | |
| $E_1^3$ | 1 | 5 | N.A. | 5 | |
| $E_2^3$ | 1 | 4 | N.A. | 4 | |
| $E_3^3$ | 1 | 2 | N.A. | 2 | |

Thus, readily available encounter-level features can be transformed into the features and labels as illustrated above for ML survival models, as long as the encounter and event dates are known. Before describing the model specifications and data description for the various prediction problems we first give an overview of the various challenges and risks involved in deployment.

# Challenges and Risks

## Deployment

### Model Training Pipeline

The following general model training pipeline can be applied to most, if not all, of the models in this document. A model training pipeline would facilitate code reuse, automate repetitive steps for greater productivity, and provide a consistent foundation on which a collection of models in this solution can be trained, deployed, retrained and maintained.

1. 70-30 train-test split
2. Feature pre-processing
    a. One-hot encoding of categorical features
    b. Normalization of continuous features
3. Missing value imputation – single value or distribution (see [28] for example)
4. Feature selection (recursive or univariate) to target <= 15 features per model
5. Model training with hyperparameter tuning in a 10-fold cross-validation – train multiple candidate models (depending on problem type) for comparison
    a. Candidate models for survival
        i. Cox Proportional Hazard (CoxPH) [132] with elastic net [133]
        ii. Accelerated Failure Time (AFT) [134]
        iii. Multi-Task Logistic Regression (MTLR) [135]
        iv. Random Survival Forests (RSF) [136]
        v. Survival Support Vector Machine (SSVM) [137]
    b. Candidate models for classification
        i. Naïve Bayes (NB)
        ii. Logistic Regression (LR)
        iii. Generalized Additive Model with Interactions (GA2M)
        iv. SVM with Linear Kernel (SVM-Linear)
        v. SVM with RBF Kernel (SVM-RBF)
        vi. Decision Tree (DT)
        vii. Random Forest (RF)
        viii. xgBoost (XGB)
    c. Candidate models for regression
        i. Linear Regression (LinR)
        ii. SVR with Linear Kernel (SVR-Linear)
        iii. SVR with RBF Kernel (SVR-RBF)
        iv. Decision Tree (DT)
        v. Random Forest (RF)
        vi. xgBoost (XGB)
6. Model evaluation (on test set)
    a. Metrics for survival
        i. Concordance Index [138]
        ii. Unbiased Concordance Index [139]
        iii. Brier Score / Integrated Brier Score [140]

b. Metrics for classification
             i. AUC
            ii. Accuracy
           iii. Balanced Accuracy
           iv. Precision
            v. Recall
           vi. F1
        c. Metrics for regression
             i. RMSE
            ii. MAE
           iii. R2
  7. Model selection
  8. Model explanation
        a. Global: Top global risk factors for each model, and a measure of the strength of association between the risk factor and the target risk
        b. Local: A means to compute the top risk factors for each individual (instance) as they are scored by the model

The whole pipeline, from feature pre-processing and including the final selected model, should then be serialized into a deployable format (e.g., pickle file containing a full scikit-learn pipeline).

# Modelling Pipeline

We consider six prediction tasks which are framed as survival analysis tasks, with different target events. Predictions were made for each encounter, so that when the models are deployed at a provider, patient risk profiles can be updated after every encounter with the latest patient data. Since these models are designed to be used in an outpatient setting, we assumed that prediction happens *after* each inpatient or outpatient encounter, and complete information of each encounter is available. The following pipeline was used to train and test models for all tasks. The pipeline is configuration-driven, and therefore is able to support different survival analysis tasks by specifying different data sets and labels as input. For details, refer to the following sections in the rest of this document.

1. **Label and feature preparation**: Selected features were used for modelling which were curated to ensure repeatability across providers, so that the Diabetes models can be widely deployed at multiple customers (see Appendix A). Labels were also constructed for the five prediction tasks using the underlying Epic data on encounters, labs and diagnoses.
2. **Cohort selection**: Filtering criteria were applied to select appropriate cohorts for each prediction task (see the cohort criteria used for each task in the following sections).
3. **Feature pre-processing**: Patient Flow features were already partially processed (e.g., one-hot encoded). Outliers with numeric features greater than 3 standard deviations above the mean were removed. Missing values were imputed using different strategies depending on the type of feature:
    a. "daysSince" features were imputed using the maximum value among all instances.
    b. Labs and vitals were imputed using the mean values among other instances with the same age bucket and sex.

c. All remaining features were imputed with zeros, as they are binary or counts of events.
4. **Model signature generation**: A backward feature elimination loop was used to select the best subset of up to 30 features per prediction task, including the mandatory features. This was conducted in two stages. Starting with the full set of about 1,200 features (excluding mandatory features), Stage 1 eliminated 5 percent (about 60 features) per iteration, to efficiently narrow down the search to the top 30 features. In stage 2, one feature was eliminated per iteration, to fine-tune the search for the best subset of features. Mandatory features were protected from elimination, thereby guaranteeing that they remain in the final set of selected features. At each iteration of the loop, a Cox proportional hazards [4], [5] model was trained on the subset of features, and its performance measured by Concordance Index [6]. The smallest set of features that achieved a cross-validation score within 0.01 of the maximum score was selected for the model signature. This margin allowed fewer features to be selected when there is a plateau in the performance
5. curve, without sacrificing much performance. The selected features (including mandatory features) became the "model signature" of each task.
6. **Training and test data preparation**: The data for each task was randomly divided into a 70% training set and 30% test set.
7. **Parameter optimization on 70% training set**: Available open-source implementations of survival ML algorithms from the PySurvival [7] package were used. The following list of candidate models were explored and optimized for each task:
   a. Cox proportional hazards (CoxPH) [8]
   b. DeepSurv Cox proportional hazards deep neural network (DeepSurv) [9]
   c. Random survival forest (RSF) [10]
   d. Conditional survival forest (CSF) [11]
   e. Extra survival trees (EST)

Distributed parameter search was performed on a Spark cluster using Hyperopt [12], which optimizes the search using Tree-structured Parzen Estimators [4]. For each task and candidate model, the search was performed for up to 200 parameter combinations and 12 hours of maximum run time. Each trained model was evaluated using Concordance Index in a 5-fold cross-validation set up, and the cross-validation scores were reported.

8. **Model training on 70% training set:** Each candidate model was re-trained on the full training set using the best set of model parameters found, i.e. the one resulting in the highest cross-validated Concordance Index during parameter search.
9. **Model evaluation on 30% test set**: The trained models were evaluated on the test set using the following metrics:
   a. Concordance Index (C-Index) [5], which measures the probability that the predicted risks for any given pair of test examples are in the correct rank order, adjusted for censoring times to achieve unbiased and consistent estimates [13]. CIdx produces real values in [0, 1], with 1 being the best score and 0.5 being the expected score for random predictions.
   b. Integrated Brier Score (IBS) [10]. The Brier Score (BS) measures the average squared distance between the observed survival status and predicted survival probability at any given time $t$, adjusted for censoring. The Integrated Brier Score gives an overall measure by computing the integral of BS over all available times. Both BS and IBS produce real values in [0, 1], with 0 being the best score 0.25 being the expected score for random predictions.
   c. Similarity of the average survival function curve to the Kaplan Meier (KM) survival curve.
   d. Brier Score over time.

10. **Selection of best model**: One best model with the highest CIdx score on the test set was selected. Detailed evaluation metrics were further computed for the selected model, including model performance by different sub-cohorts such as demographic groups.
11. **Model explanation**: Global and local model explanations were generated for each of the five selected models. The **global explanations** were generated using permutation importance where each feature is randomly shuffled and the change in concordance index is measured – a greater change would mean that the feature is more important. As for **local explanations (individual patient level explanations)**, because available explanation packages only work for regression and classification tasks, the predicted survival functions were converted into classification probabilities at different future times – 3, 6, 9, 12, 18 and 24 months – and local surrogate models (LIME) [14] were trained to explain the predictions for each future time and example. 3 examples were selected for local explanations by sampling the risk scores of the test examples in the first quartile ("low risk"), second to third quartiles ("medium risk") and fourth quartile ("high risk"). Additional notes:
    a. Explanations of up to a maximum of 20 global and 10 local features are shown
    b. Probabilities were generated using 1 – S(t), where S(t) is the survival function, which translates to the probability of having the disease or complication at time t.
    c. Risk scores of examples are not comparable across models as they do not have a fixed scale; they only indicate *relative* predicted risk between encounters.
12. **Model calibration (experimental)**: After the trained models are deployed in a customer environment, they may need to be calibrated as the characteristics of the target patient population may be different from that used to train the models. The goal of calibration is to optimize model performance with respect to a target population, in an efficient way that does not involve re-training the underlying pre-trained model. In Appendix D, we experiment with a calibration method proposed by [15] for survival models under a proportional hazards assumption.

# Inclusion and Exclusion Criteria for Assigning Class Labels

The following criteria were used to detect the first occurrence of each of the following events, per patient. The dates of the events for each patient (or lack thereof indicating censoring) were then used to construct the training labels (binary and time to event or censoring) for the survival models, as follows:

- Binary indicator $\delta = \begin{cases} 1, & Event \\ 0, & No\ Event \end{cases}$
- Time to event (or censoring) $y = \begin{cases} T, & \delta = 1 \\ C, & \delta = 0 \end{cases}$

See the respective sections in the rest of this document for the "Table 1" statistics for each label / task.

## Pre-DM

Pre-DM events were detected based on diagnostic criteria published in [14], i.e. any one test meeting the following criteria:

- From Labs: 100 mg/dL $\leq$ Fasting Plasma Glucose (FPG) $\leq$ 125 mg/dL (outpatient only) OR
- From Labs: 140 mg/dL $\leq$ 2-h Plasma Glucose (2hPG) $\leq$ 199 mg/dL during Oral Glucose Tolerance Test (OGTT) (outpatient only) OR
- From Labs: 140 mg/dL $\leq$ Random Plasma Glucose (RPG) $\leq$ 199 mg/dL (outpatient only) OR
- From Labs: 5.7% $\leq$ A1C $\leq$ 6.4%

## DM

DM events were detected based on diagnostic criteria published in [14], i.e. two abnormal test results from the same sample or separate test samples, in the last 4 weeks (0 to 27 days, inclusive):

- Fasting Plasma Glucose (FPG) $\geq$ 126 mg/dL (outpatient only)
- 2-h Plasma Glucose (2hPG) $\geq$ 200 mg/dL during Oral Glucose Tolerance Test (OGTT) (outpatient only)
- Random Glucose (RG) $\geq$ 200 mg/dL (outpatient only)
- A1C $\geq$ 6.5%

## Uncontrolled DM

Uncontrolled DM events were detected based on A1C $\geq$ 9.0%.

## Diabetic Nephropathy

Diabetic nephropathy events were detected based on ICD-10 codes:

- E11.2: Type 2 diabetes mellitus with kidney complications
- E11.21: Type 2 diabetes mellitus with diabetic nephropathy
- E11.22: Type 2 diabetes mellitus with chronic kidney disease
- E11.29: Type 2 diabetes mellitus with other diabetic kidney complication

### Diabetic Neuropathy

Diabetic neuropathy events were detected based on ICD-10 codes:

- E11.4: Type 2 diabetes mellitus with neurological complications
- E11.40: Type 2 diabetes mellitus with diabetic neuropathy, unspecified
- E11.41: Type 2 diabetes mellitus with diabetic mononeuropathy
- E11.42: Type 2 diabetes mellitus with diabetic polyneuropathy
- E11.43: Type 2 diabetes mellitus with diabetic autonomic (poly)neuropathy
- E11.44: Type 2 diabetes mellitus with diabetic amyotrophy
- E11.49: Type 2 diabetes mellitus with other diabetic neurological complication

### Diabetic Retinopathy

Diabetic retinopathy events were detected based on ICD-10 codes:

- E11.3: Type 2 diabetes mellitus with ophthalmic complications
- E11.31*: Type 2 diabetes mellitus with unspecified diabetic retinopathy
- E11.32*: Type 2 diabetes mellitus with mild nonproliferative diabetic retinopathy
- E11.33*: Type 2 diabetes mellitus with moderate nonproliferative diabetic retinopathy
- E11.34*: Type 2 diabetes mellitus with severe nonproliferative diabetic retinopathy
- E11.35*: Type 2 diabetes mellitus with proliferative diabetic retinopathy
- E11.36*: Type 2 diabetes mellitus with diabetic cataract
- E11.37*: Type 2 diabetes mellitus with diabetic macular edema, resolved following treatment
- E11.39: Type 2 diabetes mellitus with other diabetic ophthalmic complication

### T1DM

T1DM events were detected based on ICD-10 codes:

- E10*: Type 1 diabetes mellitus

This label was used to exclude T1DM patients from the cohorts.

## Model Descriptions and Results

In the next few sections, we describe the data requirements, model signatures, model evaluation and model explanations for a series of models related to prediction of Type 2 Diabetes and its complications

# T2DM to Microvascular Complications Prediction

The following apply to each of the complications of interest, namely diabetic nephropathy, diabetic neuropathy and diabetic retinopathy.

| Item | Specification |
|---|---|
| **ML Class** | Survival |
| **Usage Setting** | Outpatient |
| **Instances for Prediction** | Encounter |
| **Labels for Instances** | <ul><li>Binary indicator $\delta = \begin{cases} 1, & Complication \\ 0, & No\ Complication \end{cases}$</li><li>Time to event (or censoring) $y = \begin{cases} T, & \delta = 1 \\ C, & \delta = 0 \end{cases}$</li></ul> |
| **Cohort Criteria** | <ul><li>$18 \leq age \leq 110$ (adults without outliers)</li><li>Met T2DM diagnostic criteria before encounter date</li><li>Never diagnosed with the target complication before encounter date</li></ul> |
| **Input Features** | <ul><li>Demographic</li><li>Diagnoses EXCEPT the target complication</li><li>Labs EXCEPT those used to diagnose the target complication</li><li>Vitals</li></ul> |
| **Optimization Metric** | Concordance Index |
| **Evaluation Metrics** | <ul><li>Concordance Index</li><li>Integrated Brier Score</li></ul> |

Preliminary data exploration of encounters meeting the cohort criteria for **T2DM to diabetic nephropathy prediction**:

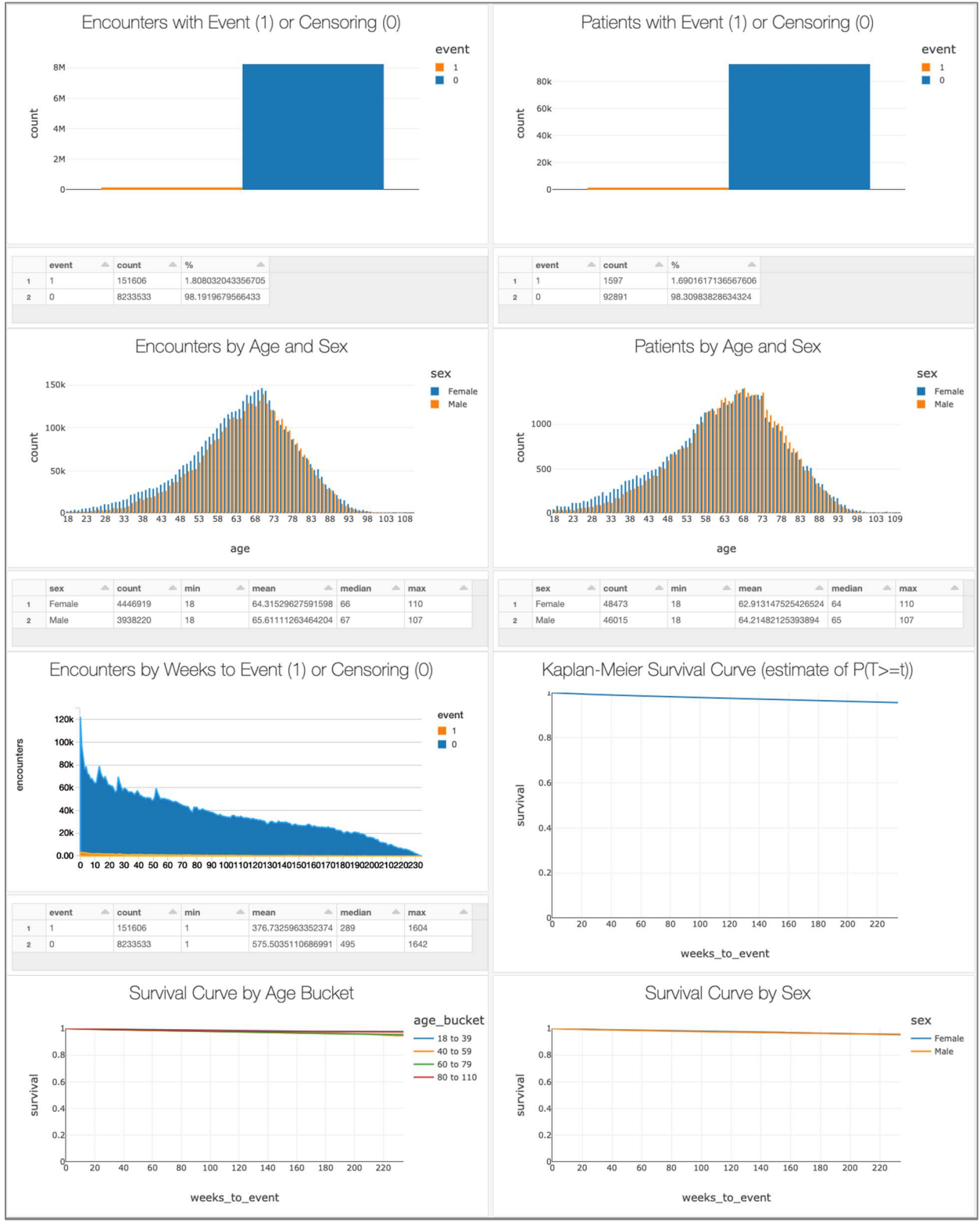

Preliminary data exploration of encounters meeting the cohort criteria for **T2DM to diabetic neuropathy prediction**:

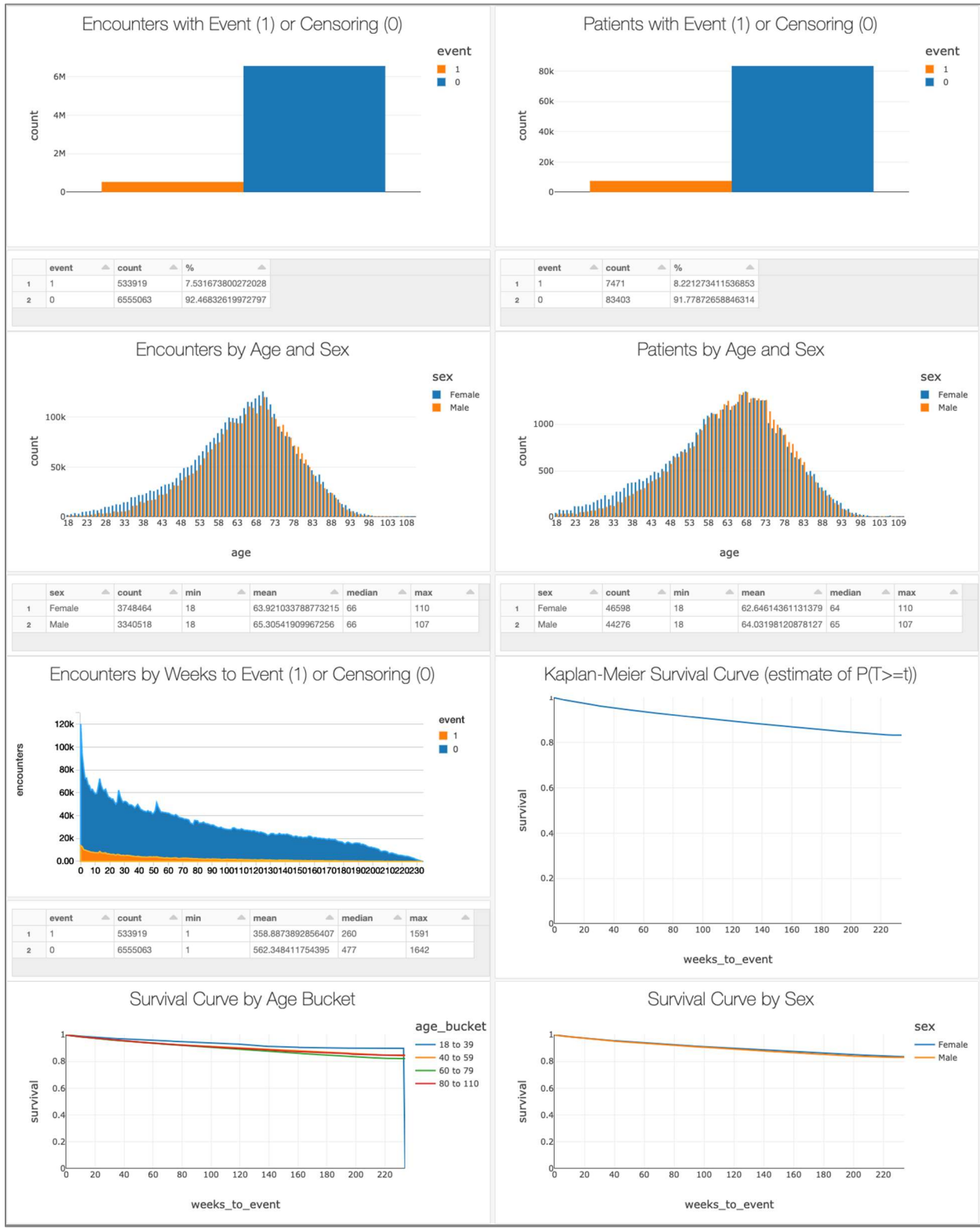

Preliminary data exploration of encounters meeting the cohort criteria for **T2DM to diabetic retinopathy prediction**:

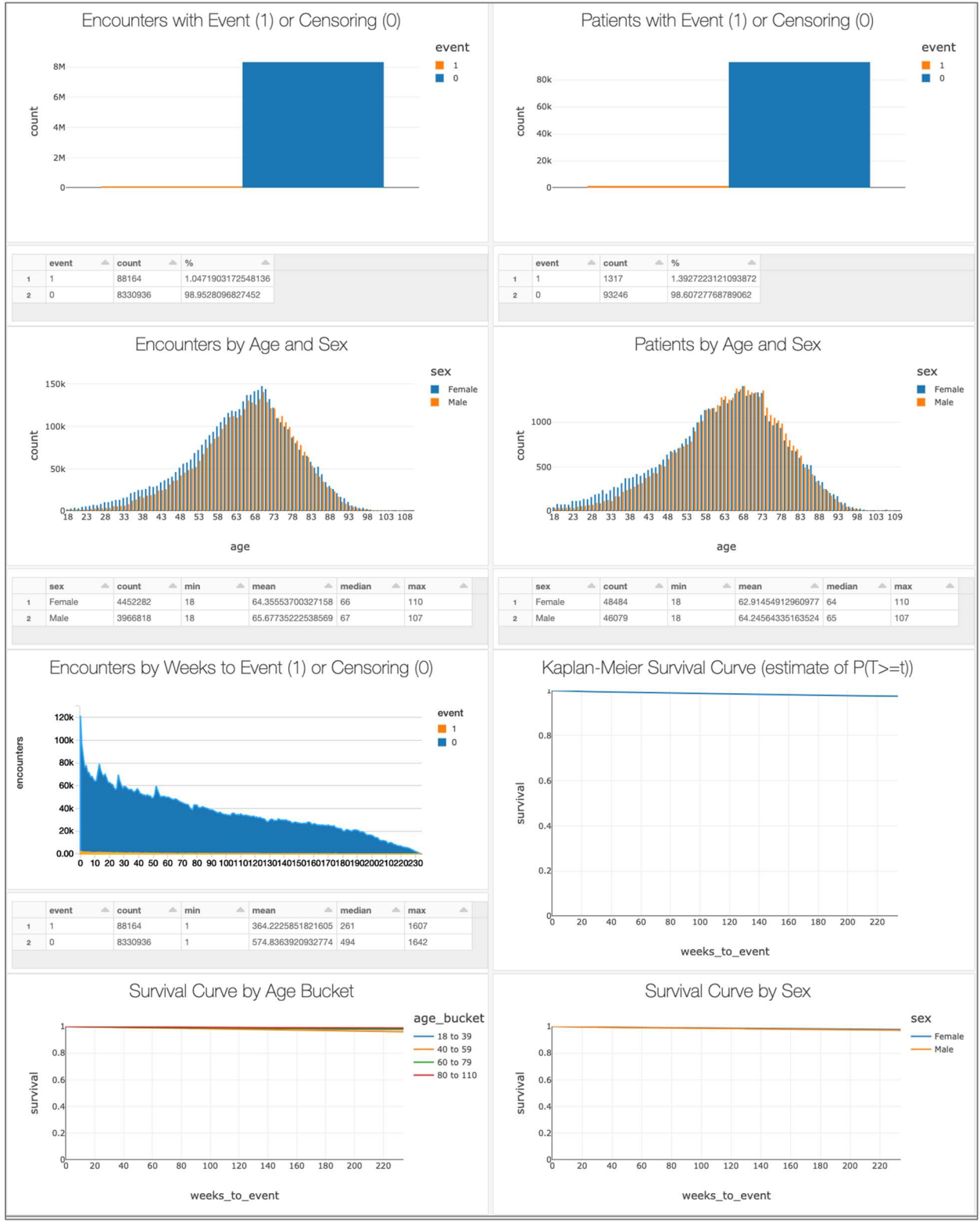

## T2DM ED Utilization Prediction

| Item | Specification |
|---|---|
| **ML Class** | Classification |
| **Usage Setting** | Outpatient |
| **Instances for Prediction** | Patient |
| **Labels for Instances** | <ul><li>Binary indicator $$y = \begin{cases} 1, & \geq 3 \text{ all cause ED visits in next } 12 \text{ months} \\ 0, & \text{otherwise} \end{cases}$$</li><li>Remove ED visits occurring within 3 days after a previous ED visit to avoid double counting</li><li>Exclude ED visits relating to injury</li><li>Construct 2 models:<ul><li>Include all remaining ED visits ("Non-emergent", "Emergent – primary care treatable", "Emergent – preventable / avoidable", "Emergent – not preventable / avoidable")</li><li>From the remaining ED visits, exclude the "Emergent – not preventable / avoidable" ED visits</li></ul></li></ul> |
| **Cohort Criteria** | <ul><li>$18 \leq age \leq 110$ (adults without outliers)</li><li>Met the diagnostic criteria of DM, Uncontrolled DM or any of the DM Complications before the prediction</li><li>Include pregnant women and those with gestational diabetes</li></ul> |
| **Input Features** | <ul><li>Demographic</li><li>Diabetes treatment type and medications</li><li>Diabetes duration and complications</li><li>Healthcare utilization</li><li>Diagnoses, labs, vitals</li></ul>See Appendix E for details |
| **Optimization Metric** | AUC |
| **Evaluation Metrics** | <ul><li>AUC</li><li>Accuracy</li><li>Balanced Accuracy</li><li>Precision</li><li>Recall</li></ul> |

## Visit Non-Compliance Prediction (All patients)

| Item | Specification |
|---|---|
| **ML Class** | Classification |
| **Usage Setting** | Outpatient |
| **Instances for Prediction** | Appointment |
| **Labels for Instances** | <ul><li>Binary indicator $$y = \begin{cases} 1, & \text{No Show or Late Cancellation} \\ 0, & \text{Completed Appointment} \end{cases}$$ Late Cancellation is defined as within 24 hours prior to appointment time</li></ul> |

| Item | Specification |
|---|---|
| | - $y = 1$, *No Show or Late Cancellation*<br>  o Status == "No Show", or<br>  o Status == "Canceled" & Cancel initiator == "Patient" & Cancel reason != "Deceased" & cancellation was late<br>- $y = 0$, *Completed Appointment*<br>  o Status == "Completed" & Cancel reason == "No Reason" |
| **Cohort Criteria** | - $18 \leq age \leq 110$ (adults without outliers)<br>- Exclude visits without any prior appointment (walk_in_yn == "Y")<br>- Exclude appointments made within 48 hours of actual appointment |
| **Input Features** | - Demographic<br>- Medication adherence<br>- Patient appointment history<br>- Current appointment characteristics<br>- Diagnoses, labs, vitals<br><br>See Appendix E for details |
| **Optimization Metric** | AUC |
| **Evaluation Metrics** | - AUC<br>- Accuracy<br>- Balanced Accuracy<br>- Precision<br>- Recall |

## T2DM Visit Non-Compliance Prediction (DM patients)

| Item | Specification |
|---|---|
| **ML Class** | Classification |
| **Usage Setting** | Outpatient |
| **Instances for Prediction** | Appointment |
| **Labels for Instances** | - Binary indicator<br>$y = \begin{cases} 1, & \text{No Show or Late Cancellation} \\ 0, & \text{Completed Appointment} \end{cases}$<br>Late Cancellation is defined as within 24 hours prior to appointment time<br>- $y = 1$, *No Show or Late Cancellation*<br>  o Status == "No Show", or<br>  o Status == "Canceled" & Cancel initiator == "Patient" & Cancel reason != "Deceased" & cancellation was late<br>- $y = 0$, *Completed Appointment*<br>  o Status == "Completed" & Cancel reason == "No Reason" |
| **Cohort Criteria** | - $18 \leq age \leq 110$ (adults without outliers)<br>- Met the diagnostic criteria of DM, Uncontrolled DM or any of the DM Complications before the appointment date |

| Item | Specification |
|---|---|
|  | - Exclude visits without any prior appointment (walk_in_yn == "Y")<br>- Exclude appointments made within 48 hours of appointment time |
| **Input Features** | - Demographic<br>- Diabetes treatment type<br>- Medication adherence<br>- Patient appointment history<br>- Current appointment characteristics<br>- Diagnoses, labs, vitals<br><br>See Appendix E for details |
| **Optimization Metric** | AUC |
| **Evaluation Metrics** | - AUC<br>- Accuracy<br>- Balanced Accuracy<br>- Precision<br>- Recall |

Additional Notes

**Usage Settings:** x days before appointment date / at time of scheduling / any time between appointment schedule and appointment time

**Instance for prediction:** Appointment vs. patient level prediction

Pros for predicting on appointments:

- It will be easier to model appointment specific characteristics.
- Definition of the response variable is more straightforward whereas patient level prediction requires setting arbitrary cut-offs for the percentage of missed appointments as well as the time period over which the percentage will be computed.

Cons for predicting on appointments:

- In some cases, the care manager might want to understand the patient's visit compliance in general rather than with respect to a specific appointment.

## T2DM Medication Non-Adherence Prediction

| Item | Specification |
|---|---|
| **ML Class** | Classification |
| **Usage Setting** | Outpatient |
| **Instances for Prediction** | Patient |
| **Labels for Instances** | - Binary indicator<br>$$y = \begin{cases} 1, & PDC \text{ over } 1 \text{ year} \geq 80\% \\ 0, & otherwise \end{cases}$$ |

| Item | Specification |
|---|---|
| | $$PDC = \frac{No.\,of\,days\,in\,period^*\,covered}{No.\,of\,days\,in\,period^*} \times 100\%$$ A day is considered to be covered only when the patient has access to all the antidiabetic medications specified.<br><br>*Exclude PDC calculation for days which the patient is hospitalized |
| **Cohort Criteria** | - $18 \leq age \leq 110$ (adults without outliers)<br>- Met the diagnostic criteria of DM, Uncontrolled DM or any of the DM Complications before the prediction<br>- Include patients with at least 1 prescription of an OAD or insulin<br>- Exclude patients using Metformin for polycystic ovary syndrome only |
| **Input Features** | - Demographic<br>- Diabetes treatment type and medications<br>- Diabetes duration and complications<br>- Healthcare utilization<br>- Diagnoses, labs, vitals<br><br>See Appendix E for details |
| **Optimization Metric** | AUC |
| **Evaluation Metrics** | - AUC<br>- Accuracy<br>- Balanced Accuracy<br>- Precision<br>- Recall |

Additional Notes

**Usage Settings**

What are the specific use cases for medication adherence prediction? Will we be scoring daily using a rolling window?

**Labels for Instances**

1. How should we handle multiple medications?
2. What would be a meaningful time interval for compliance with diabetic medications? Is 90 days too long / short?
3. Do we want to focus only on adherence relating diabetic medications only or all medications that are prescribed / medications typically taken by diabetic patients to control blood pressure for example?
4. The 80% threshold for PDC is arbitrary, should we come up with certain heuristics to determine the threshold?

**Cohort Criteria**

1. Do we want to include prediction for primary non-adherence (patients that never filled their first prescription)? (e.g., Exclude patients with less than 2 prescription fills for a medication)
2. Do we want to exclude patients receiving their medication by mail?
3. Are we able to find out patients with prolonged institution and exclude them?

## Pre-DM to DM Prediction

| Item | Specification |
|---|---|
| **Business Goal** | Enable care managers to identify the patients who are pre-diabetic and at risk of developing DM |
| **Usage Setting** | Outpatient |
| **ML Task** | Predict risk and/or time from pre-DM to DM |
| **ML Class** | Survival |
| **Instances for Prediction** | Encounters |
| **Labels for Instances** | Binary indicator and time to event or censoring for DM <br> • Binary indicator $\delta = \begin{cases} 1, & T2DM \\ 0, & No\ T2DM \end{cases}$ <br> • Time to event (or censoring) $y = \begin{cases} T, & \delta = 1 \\ C, & \delta = 0 \end{cases}$ |
| **Cohort Criteria** | • 2016-01-01 $\leq$ encounter date $\leq$ 2020-06-30 (available Epic data, excluding outliers) <br> • Encounter date is not within the first 90 days of when the patient entered the data set, to adjust for left-censoring <br> • 18 $\leq$ age $\leq$ 110 (adults without outliers) <br> • No T1DM diagnosis <br> • Not pregnant <br> • No "Do Not Resuscitate" diagnosis <br><br> • Pre-DM event before encounter date <br> • No DM or uncontrolled DM event before encounter date <br> • No DM or uncontrolled DM event up to 6 days after encounter date (encounters where lab results confirm event within the week) |

| Item | Specification |
|---|---|
| **Input Features** | - Demographic<br>- Diagnosis, except:<br>  - hasDiabetes*<br>  - has_END002* (Diabetes mellitus without complication)<br>  - has_END003* (Diabetes mellitus with complication)<br>  - has_END004* (Diabetes mellitus, Type 1)<br>  - has_END005* (Diabetes mellitus, Type 2)<br>  - has_END006* (Diabetes mellitus, due to underlying condition, drug or chemical induced, or other specified type)<br>- Labs<br>- Utilization<br>- Vitals<br><br>See Appendix E for details. |
| **Evaluation Metrics** | - Concordance Index<br>- Integrated Brier Score |

Data

The following charts summarize the key characteristics of the data after applying the cohort criteria stated above, along with selected features (see Model Signature below).

| Category | Variable | count | mean | stddev | min | 25% | 50% | 75% | max |
|---|---|---|---|---|---|---|---|---|---|
| Demographic | AgeBucket_18_to_39 | 258142.0 | 0.069903 | 0.254985 | 0.00 | 0.000 | 0.000 | 0.000 | 1.00 |
|  | AgeBucket_40_to_59 | 258142.0 | 0.248332 | 0.432046 | 0.00 | 0.000 | 0.000 | 0.000 | 1.00 |
|  | AgeBucket_60_to_79 | 258142.0 | 0.543267 | 0.498125 | 0.00 | 0.000 | 1.000 | 1.000 | 1.00 |
|  | AgeBucket_80_to_109 | 258142.0 | 0.138497 | 0.345422 | 0.00 | 0.000 | 0.000 | 0.000 | 1.00 |
|  | Sex_Female | 258142.0 | 0.627515 | 0.483467 | 0.00 | 0.000 | 1.000 | 1.000 | 1.00 |
|  | Sex_Male | 258142.0 | 0.372485 | 0.483467 | 0.00 | 0.000 | 0.000 | 1.000 | 1.00 |
|  | Ethnicity_Hispanic_or_Latino | 258142.0 | 0.054687 | 0.227369 | 0.00 | 0.000 | 0.000 | 0.000 | 1.00 |
|  | Ethnicity_Not_Hispanic_or_Latino | 258142.0 | 0.945313 | 0.227369 | 0.00 | 1.000 | 1.000 | 1.000 | 1.00 |
| Encounter | EncounterType_Emergency | 258142.0 | 0.183232 | 0.386858 | 0.00 | 0.000 | 0.000 | 0.000 | 1.00 |
|  | EncounterType_Inpatient | 258142.0 | 0.079848 | 0.271058 | 0.00 | 0.000 | 0.000 | 0.000 | 1.00 |
|  | EncounterType_Outpatient | 258142.0 | 0.736920 | 0.440306 | 0.00 | 0.000 | 1.000 | 1.000 | 1.00 |
| Label | Time | 258142.0 | 15.740631 | 11.314076 | 0.00 | 6.000 | 14.000 | 24.000 | 50.00 |
|  | Event | 258142.0 | 0.064608 | 0.245833 | 0.00 | 0.000 | 0.000 | 0.000 | 1.00 |
| Feature | age | 258142.0 | 64.481131 | 14.647176 | 18.00 | 56.000 | 66.000 | 75.000 | 107.00 |
|  | Male_sex | 258142.0 | 0.372485 | 0.483467 | 0.00 | 0.000 | 0.000 | 1.000 | 1.00 |
|  | Hispanic_or_Latino_ethnicity | 258142.0 | 0.054687 | 0.227369 | 0.00 | 0.000 | 0.000 | 0.000 | 1.00 |
|  | maxhbA1CInPast365Days | 258142.0 | 6.144908 | 0.536921 | 3.00 | 5.900 | 6.121 | 6.228 | 9.80 |
|  | lasthbA1CInPast365Days | 258142.0 | 6.034896 | 0.491028 | 2.20 | 5.800 | 6.022 | 6.103 | 8.90 |
|  | meandiastolicinPast365Days | 258142.0 | 74.510641 | 7.465089 | 43.63 | 71.000 | 74.678 | 77.830 | 107.67 |
|  | meansystolicinPast365Days | 258142.0 | 130.507721 | 11.237634 | 81.00 | 125.900 | 130.162 | 134.500 | 180.00 |
|  | BMI | 258142.0 | 32.052144 | 6.993803 | 2.71 | 28.009 | 31.894 | 35.243 | 326.11 |
|  | meantriglyceroidsInPast365Days | 258142.0 | 141.515197 | 48.161881 | 11.00 | 119.839 | 142.732 | 147.721 | 468.50 |
|  | meanldlcholesterolInPast365Days | 258142.0 | 95.182333 | 23.631750 | 2.80 | 84.968 | 97.608 | 107.000 | 201.00 |
|  | meanhdlcholesterolInPast365Days | 258142.0 | 47.882752 | 10.369947 | 6.00 | 42.348 | 47.000 | 52.436 | 98.00 |
|  | has_BLD005_Past12Months | 258142.0 | 0.001189 | 0.034465 | 0.00 | 0.000 | 0.000 | 0.000 | 1.00 |
|  | has_CIR007_Past12Months | 258142.0 | 0.165409 | 0.371550 | 0.00 | 0.000 | 0.000 | 0.000 | 1.00 |
|  | has_CIR008_Past12Months | 258142.0 | 0.009239 | 0.095675 | 0.00 | 0.000 | 0.000 | 0.000 | 1.00 |
|  | has_NEO016_Past12Months | 258142.0 | 0.003146 | 0.055997 | 0.00 | 0.000 | 0.000 | 0.000 | 1.00 |
|  | has_PRG029 | 258142.0 | 0.015968 | 0.125352 | 0.00 | 0.000 | 0.000 | 0.000 | 1.00 |
|  | has_FAC006 | 258142.0 | 0.040509 | 0.197149 | 0.00 | 0.000 | 0.000 | 0.000 | 1.00 |
|  | has_DIG018 | 258142.0 | 0.022488 | 0.148263 | 0.00 | 0.000 | 0.000 | 0.000 | 1.00 |
|  | has_SKN003 | 258142.0 | 0.052475 | 0.222983 | 0.00 | 0.000 | 0.000 | 0.000 | 1.00 |
|  | has_RSP015 | 258142.0 | 0.002750 | 0.052372 | 0.00 | 0.000 | 0.000 | 0.000 | 1.00 |
|  | has_NEO017 | 258142.0 | 0.011966 | 0.108734 | 0.00 | 0.000 | 0.000 | 0.000 | 1.00 |
|  | has_INJ057 | 258142.0 | 0.001708 | 0.041297 | 0.00 | 0.000 | 0.000 | 0.000 | 1.00 |

(Percentages for binary variables can be read from the "mean" column.)

## Encounters and Patients

| | Examples | Encounters | Patients |
|---|---|---|---|
| 1 | 258142 | 258142 | 45851 |

## Train and Test Sets

| | Set | No Event | No Event % | Event | Event % | Total | Total % |
|---|---|---|---|---|---|---|---|
| 1 | Test | 72364 | 30 | 4958 | 29.7 | 77322 | 30 |
| 2 | Train | 169100 | 70 | 11720 | 70.3 | 180820 | 70 |
| 3 | Total | 241464 | 100 | 16678 | 100 | 258142 | 100 |

### Encounters with Event (1) or Censoring (0)

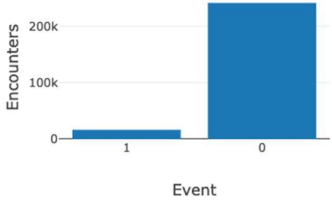

### Patients with Event (1) or Censoring (0)

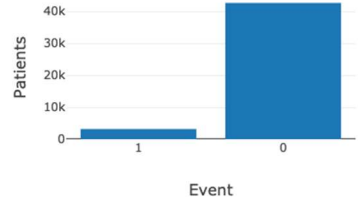

### Encounters by Age and Sex

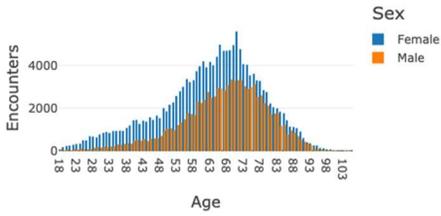

### Patients by Age and Sex

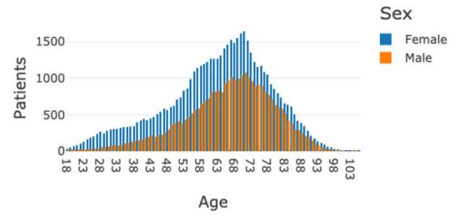

### Encounters by Ethnicity

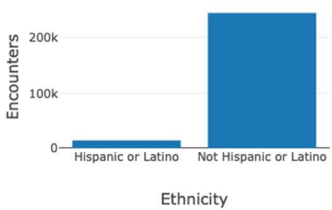

### Patients by Ethnicity

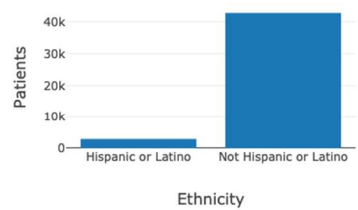

### Encounters by Encounter Type

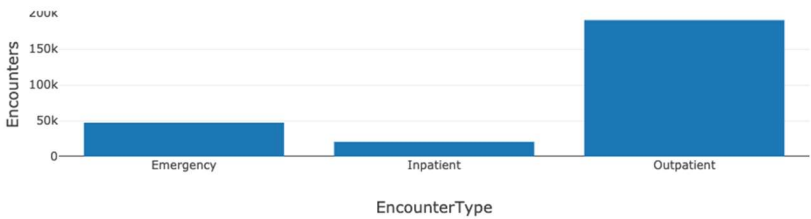

### Encounters by Time to Event (1) or Cen…

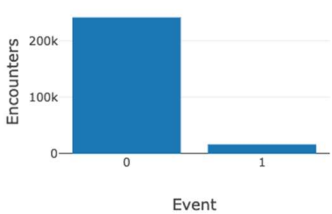

### Kaplan-Meier Estimate of Survival Functi…

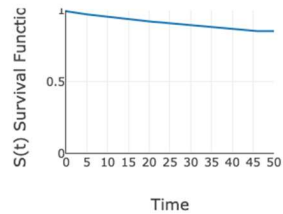

## Model Signature

The model signature has 22 features, comprising of 12 mandatory features and 10 other selected features. These are the selected features, in rank order (the last feature to be eliminated is ranked 1):

1. has_BLD005_Past12Months (Sickle cell trait/anemia)
2. has_PRG029 (Uncomplicated pregnancy, delivery or puerperium)
3. has_NEO016_Past12Months (Gastrointestinal cancers - anus)
4. has_INJ057 (Effect of foreign body entering opening, subsequent encounter)
5. has_DIG018 (Hepatic failure)
6. maxhbA1CInPast365Days
7. has_RSP015 (Mediastinal disorders)
8. has_NEO017 (Gastrointestinal cancers - liver)
9. has_SKN003 (Pressure ulcer of skin)
10. has_FAC006 (Encounter for antineoplastic therapies)
11. Male_sex
12. lasthbA1CInPast365Days
13. has_CIR008_Past12Months (Hypertension with complications and secondary hypertension)
14. Hispanic_or_Latino_ethnicity
15. has_CIR007_Past12Months (Essential hypertension)
16. meanhdlcholesterolInPast365Days
17. meansystolicinPast365Days
18. BMI
19. meandiastolicinPast365Days
20. age
21. meanldlcholesterolInPast365Days
22. meantriglyceroidsInPast365Days

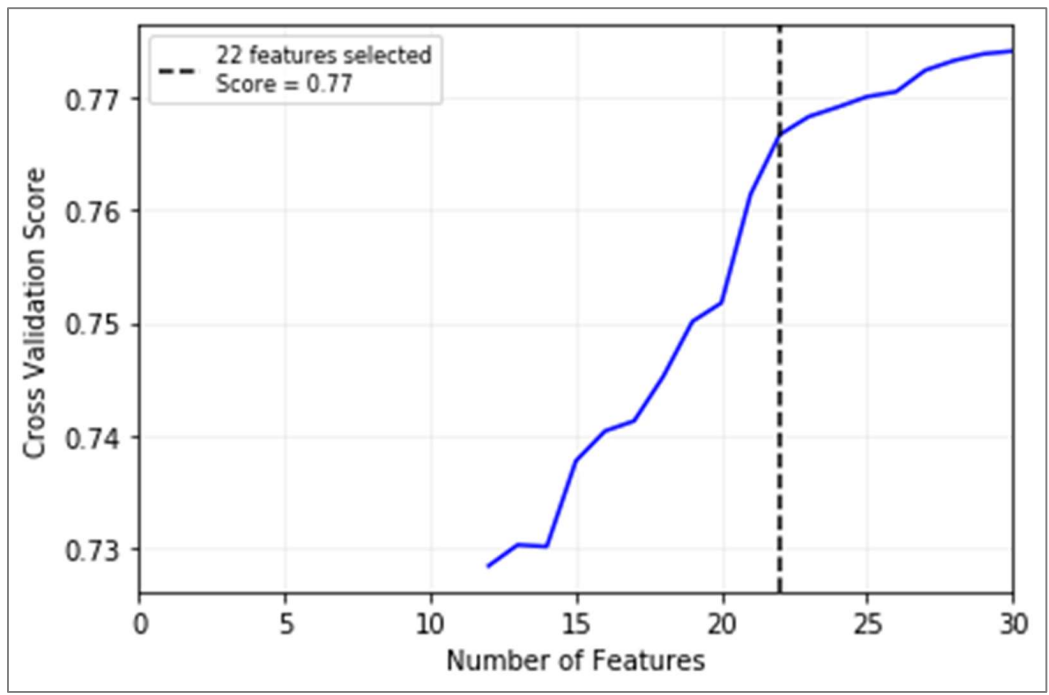

## Model Performance

The following table and chart summarize the performance of all candidate models on the test set for this prediction task in terms of the Concordance Index and the Integrated Brier Score.

| Model | No. of Parameter Combinations Successfully Tested | Concordance Index | Integrated Brier Score |
| --- | --- | --- | --- |
| **CoxPH** | 89 | 0.78 | 0.07 |
| **DeepSurv** | 60 | **0.81** | **0.06** |
| **RSF** | Timeout | - | - |
| **CSF** | 7 | 0.76 | 0.07 |
| **EST** | Timeout | - | - |

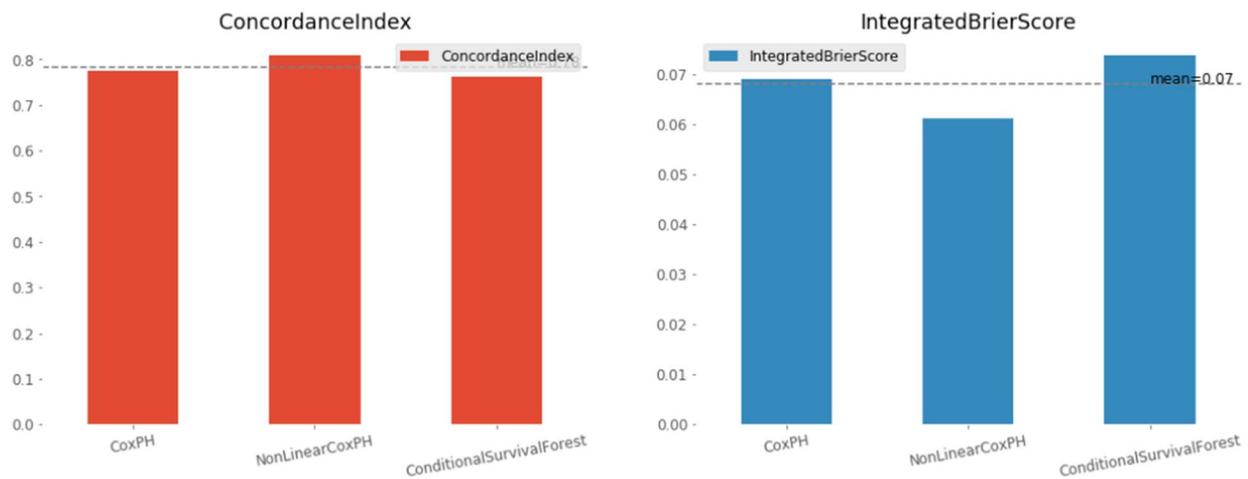

The following chart shows how the average survival function curves of the candidate models compare to the KM survival curve, the more similar their curves are to the KM survival curve the better.

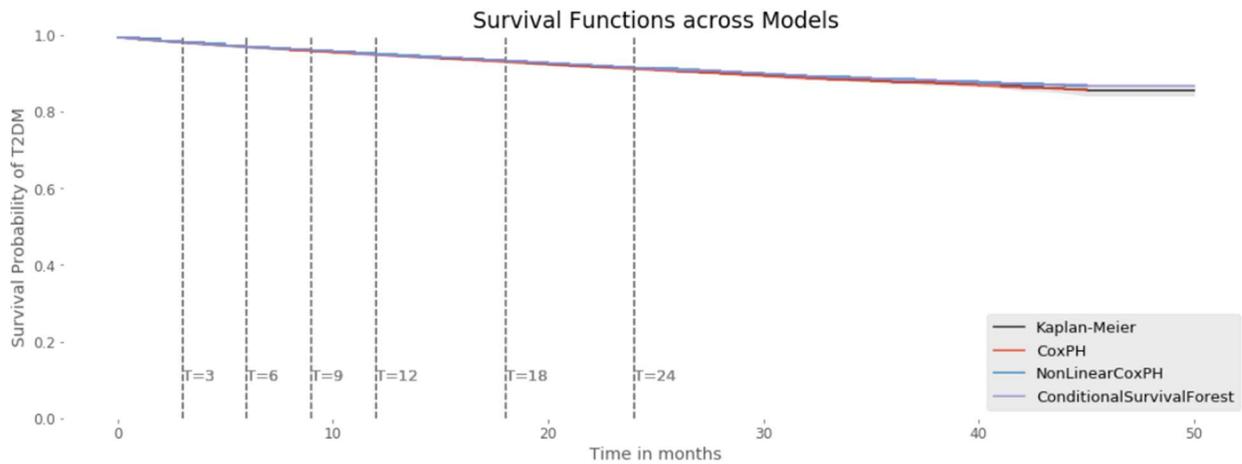

This chart shows the change in the Brier Score over time for all candidate models, the closer the scores are to 0 the better.

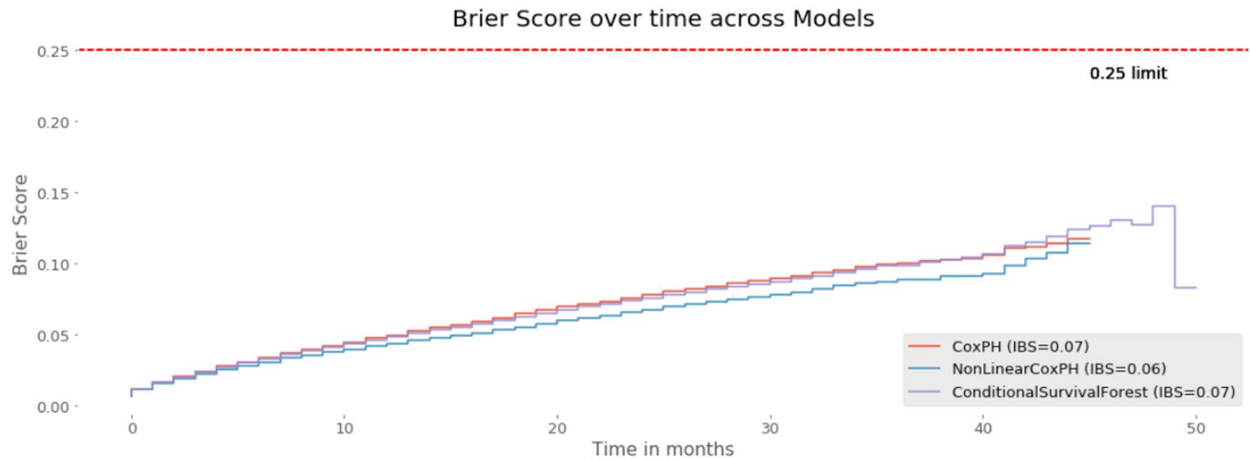

## Model Evaluation of Selected Model (DeepSurv)

Overall

This chart shows the change in the Brier Score over time for the selected model, the closer the scores are to 0 the better.

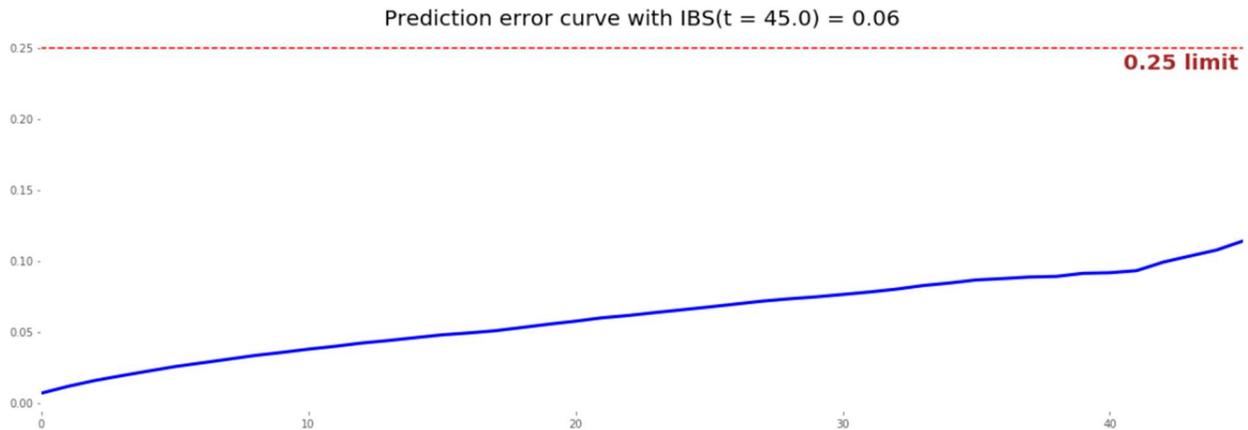

The following chart shows the actual vs. predicted density functions, i.e. number of instances that get the disease / complication at each time point and the RMSE, Median Absolute Error and Mean Absolute Error across the time points.

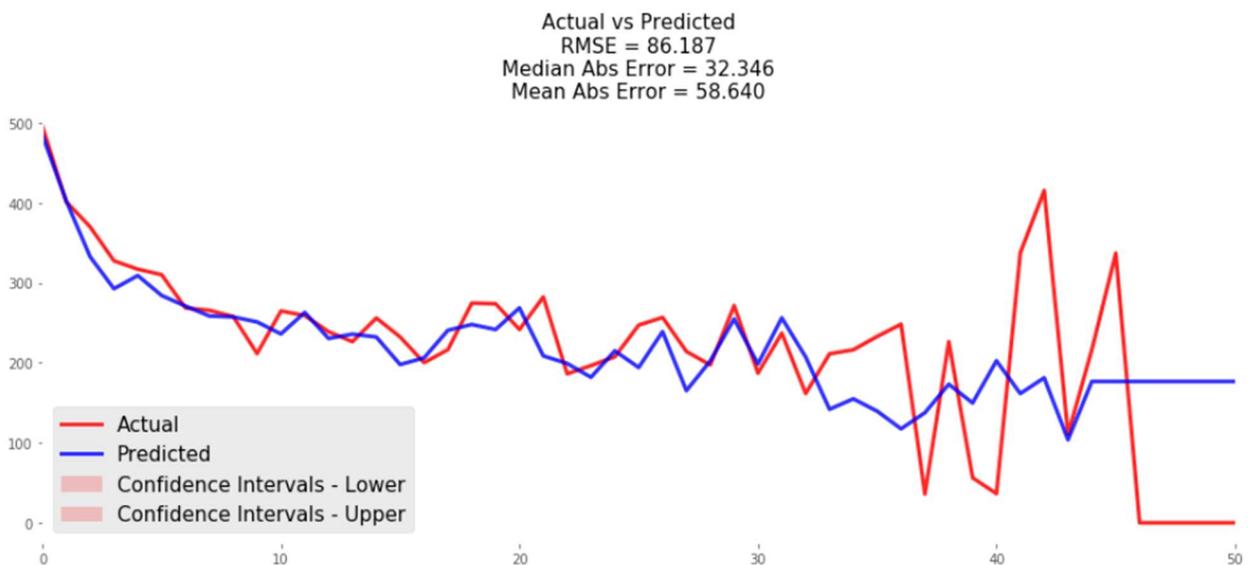

The following chart shows the actual vs. predicted survival functions, i.e. the number of instances that have not had the disease / complication by each time point and the RMSE, Median Absolute Error and Mean Absolute Error across the time points.

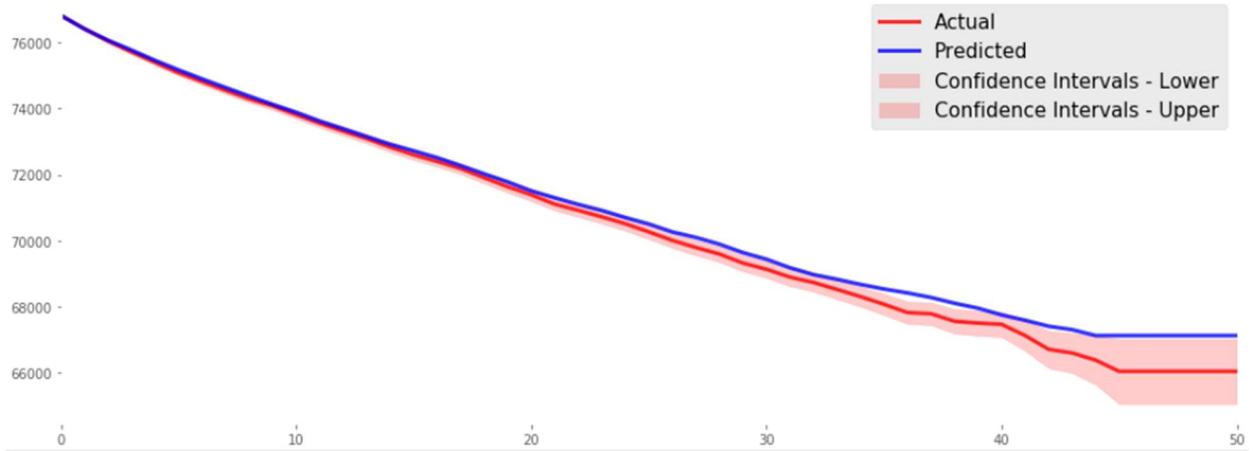

Risk Stratification

The low, medium and high risk groups are defined as examples with predicted risk scores belonging to the first quartile, second to third quartiles, and fourth quartile respectively.

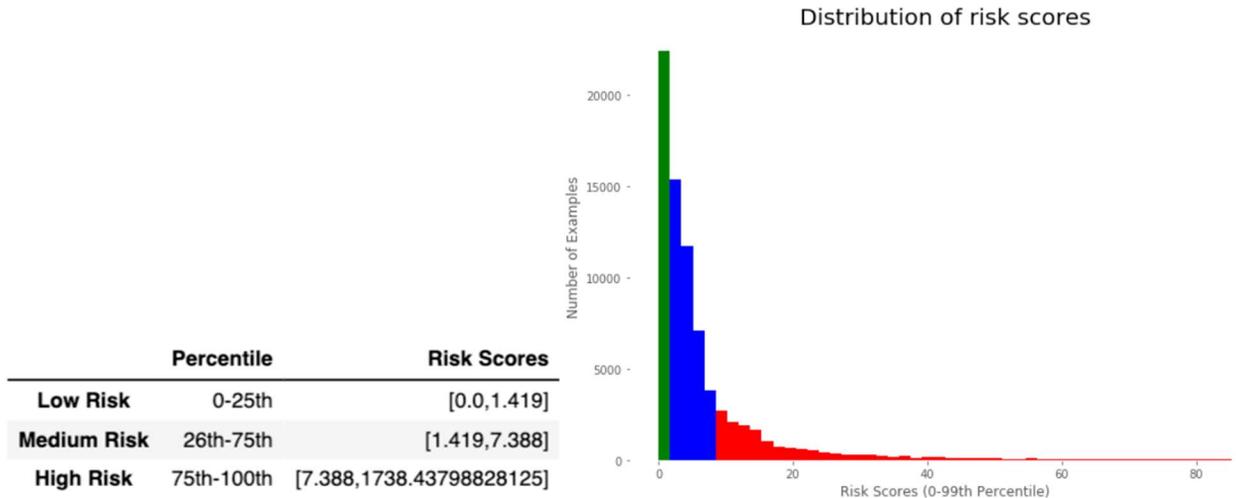

|  | Percentile | Risk Scores |
|---|---|---|
| Low Risk | 0-25th | [0.0, 1.419] |
| Medium Risk | 26th-75th | [1.419, 7.388] |
| High Risk | 75th-100th | [7.388, 1738.43798828125] |

Summary Metrics across Subgroups

The table below displays the summary metrics across subgroups of risk, age, sex, ethnicity and patient history.

| Category | Subgroup | Cohort Size | Concordance Index | Brier Score | Mean AUC | Mean Specificity | Mean Sensitivity | S(t), t=3 | S(t), t=6 | S(t), t=9 | S(t), t=12 | S(t), t=18 | S(t), t=24 |
|---|---|---|---|---|---|---|---|---|---|---|---|---|---|
| NaN | Overall | 77322.00 | 0.81 | 0.06 | 0.82 | 0.79 | 0.69 | 0.98 | 0.97 | 0.96 | 0.95 | 0.93 | 0.91 |
| Risk | Low | 19331.00 | 0.61 | 0.01 | 0.61 | 1.00 | 0.00 | 1.00 | 1.00 | 1.00 | 1.00 | 0.99 | 0.99 |
| Risk | Medium | 38660.00 | 0.61 | 0.04 | 0.62 | 1.00 | 0.01 | 0.99 | 0.98 | 0.98 | 0.97 | 0.96 | 0.95 |
| Risk | High | 19331.00 | 0.70 | 0.14 | 0.72 | 0.12 | 0.97 | 0.94 | 0.91 | 0.88 | 0.85 | 0.81 | 0.76 |
| Age Bucket | 18 to 39 | 5354.00 | 0.86 | 0.03 | 0.87 | 0.93 | 0.53 | 0.99 | 0.98 | 0.98 | 0.98 | 0.97 | 0.96 |
| Age Bucket | 40 to 59 | 19187.00 | 0.78 | 0.06 | 0.80 | 0.82 | 0.63 | 0.98 | 0.97 | 0.96 | 0.95 | 0.93 | 0.92 |
| Age Bucket | 60 to 79 | 42112.00 | 0.82 | 0.07 | 0.82 | 0.75 | 0.74 | 0.98 | 0.96 | 0.95 | 0.94 | 0.92 | 0.90 |
| Age Bucket | 80 to 109 | 10669.00 | 0.82 | 0.05 | 0.81 | 0.87 | 0.58 | 0.98 | 0.98 | 0.97 | 0.96 | 0.95 | 0.93 |
| Sex | Male | 28842.00 | 0.78 | 0.08 | 0.80 | 0.70 | 0.75 | 0.97 | 0.96 | 0.95 | 0.93 | 0.91 | 0.89 |
| Sex | Female | 48480.00 | 0.83 | 0.05 | 0.83 | 0.84 | 0.63 | 0.98 | 0.97 | 0.97 | 0.96 | 0.94 | 0.93 |
| Ethnicity | Hispanic or Latino | 4197.00 | 0.87 | 0.06 | 0.87 | 0.83 | 0.70 | 0.98 | 0.97 | 0.96 | 0.95 | 0.93 | 0.92 |
| Ethnicity | Not Hispanic or Latino | 73125.00 | 0.81 | 0.06 | 0.82 | 0.79 | 0.69 | 0.98 | 0.97 | 0.96 | 0.95 | 0.93 | 0.91 |
| History Bucket | <= 6 | 2528.00 | 0.80 | 0.07 | 0.80 | 0.74 | 0.71 | 0.98 | 0.96 | 0.95 | 0.94 | 0.92 | 0.90 |
| History Bucket | 7 to 12 | 6661.00 | 0.83 | 0.07 | 0.82 | 0.76 | 0.74 | 0.97 | 0.96 | 0.95 | 0.94 | 0.92 | 0.90 |
| History Bucket | 13 to 24 | 21015.00 | 0.81 | 0.06 | 0.82 | 0.79 | 0.69 | 0.98 | 0.97 | 0.96 | 0.95 | 0.93 | 0.91 |
| History Bucket | 25 to 36 | 22240.00 | 0.81 | 0.04 | 0.82 | 0.82 | 0.66 | 0.98 | 0.97 | 0.96 | 0.95 | 0.94 | 0.92 |
| History Bucket | 37 to 60 | 24878.00 | 0.83 | 0.03 | nan | nan | 0.70 | 0.98 | 0.97 | 0.96 | 0.95 | 0.94 | 0.92 |

Concordance Index & Integrated Brier Score

The following charts show how the Concordance Index and Integrated Brier Score varies among subgroups of risk, age, sex, ethnicity and patient history.

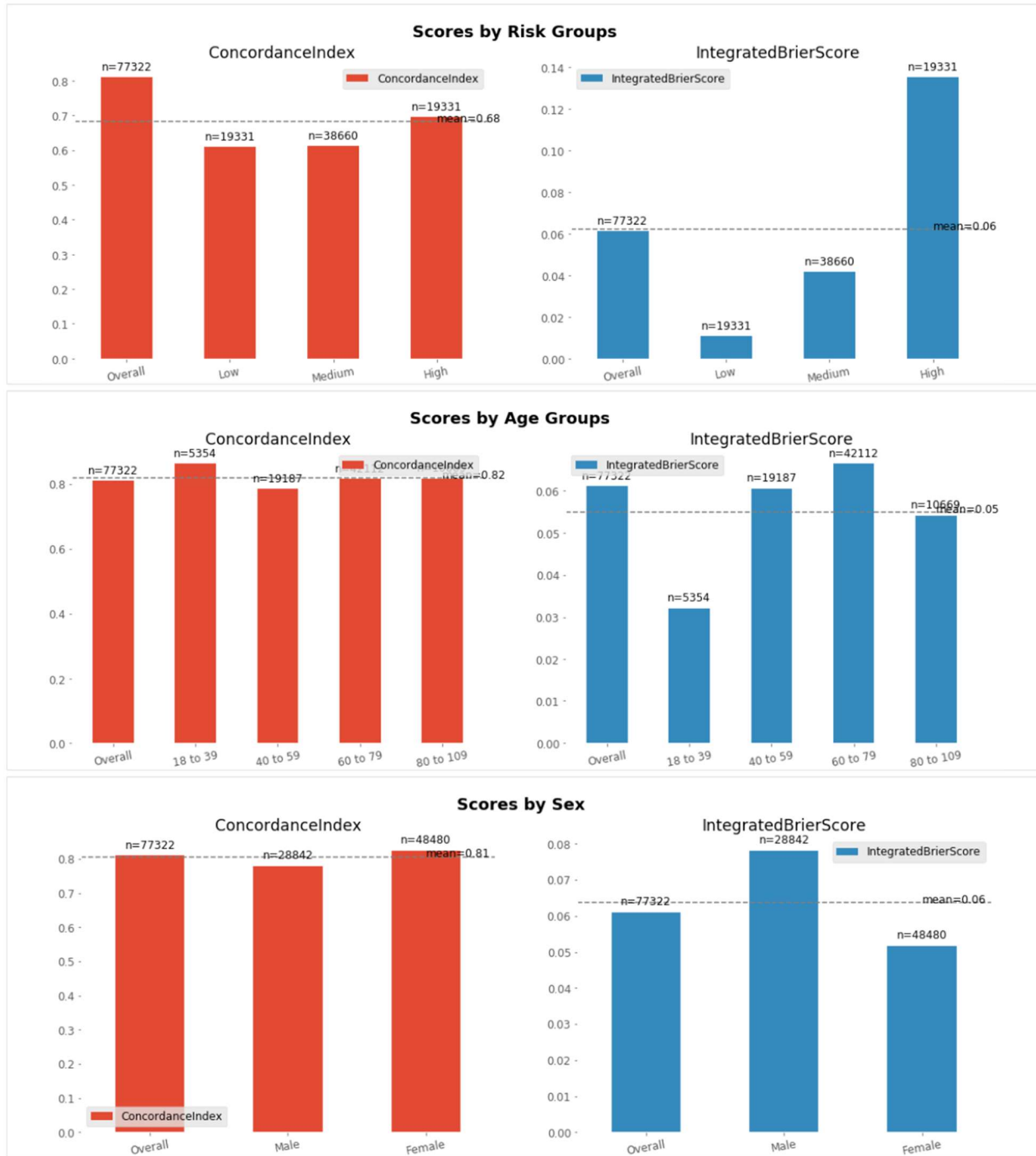

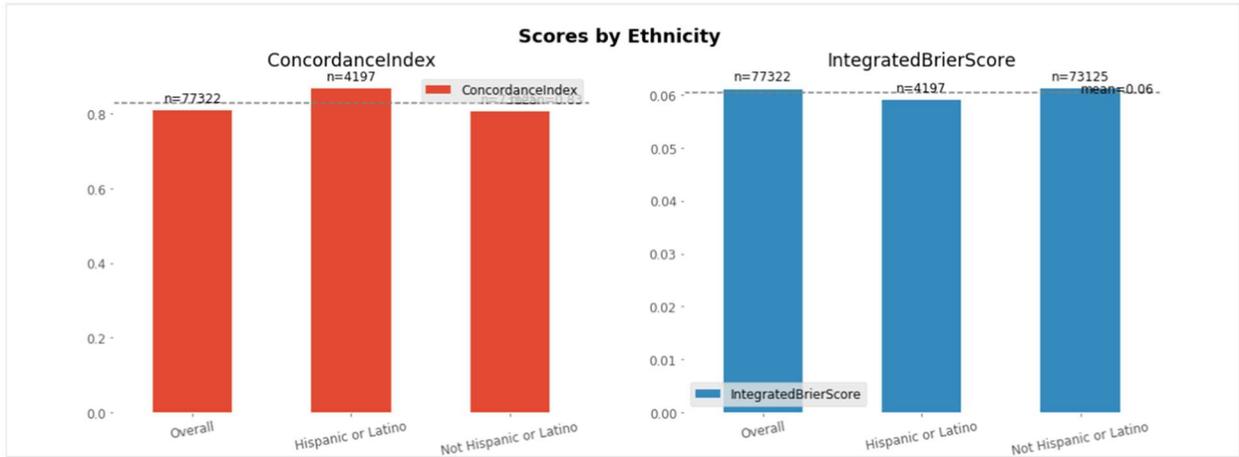
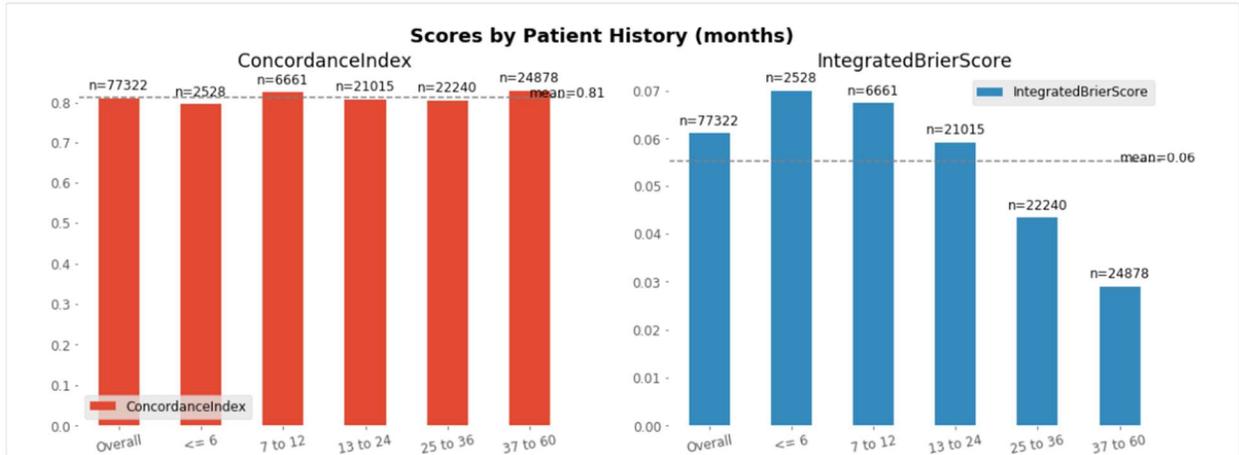

Average Survival Function Curves

The following charts show how the average survival function curve varies among subgroups of risk, age, sex, ethnicity and patient history.

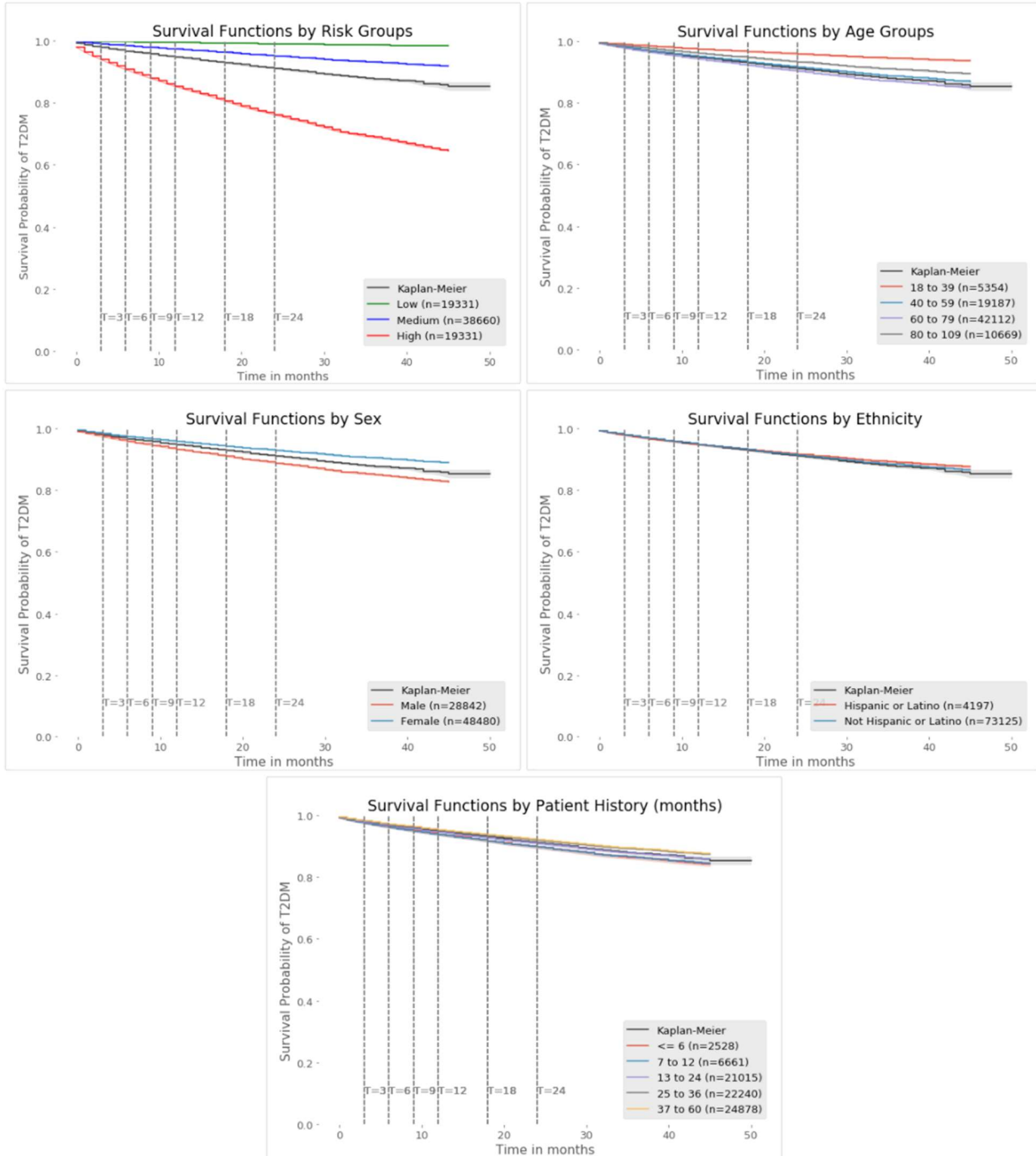

Time-dependent AUC

The following charts show how the AUC across time varies among subgroups of risk, age, sex, ethnicity and patient history.

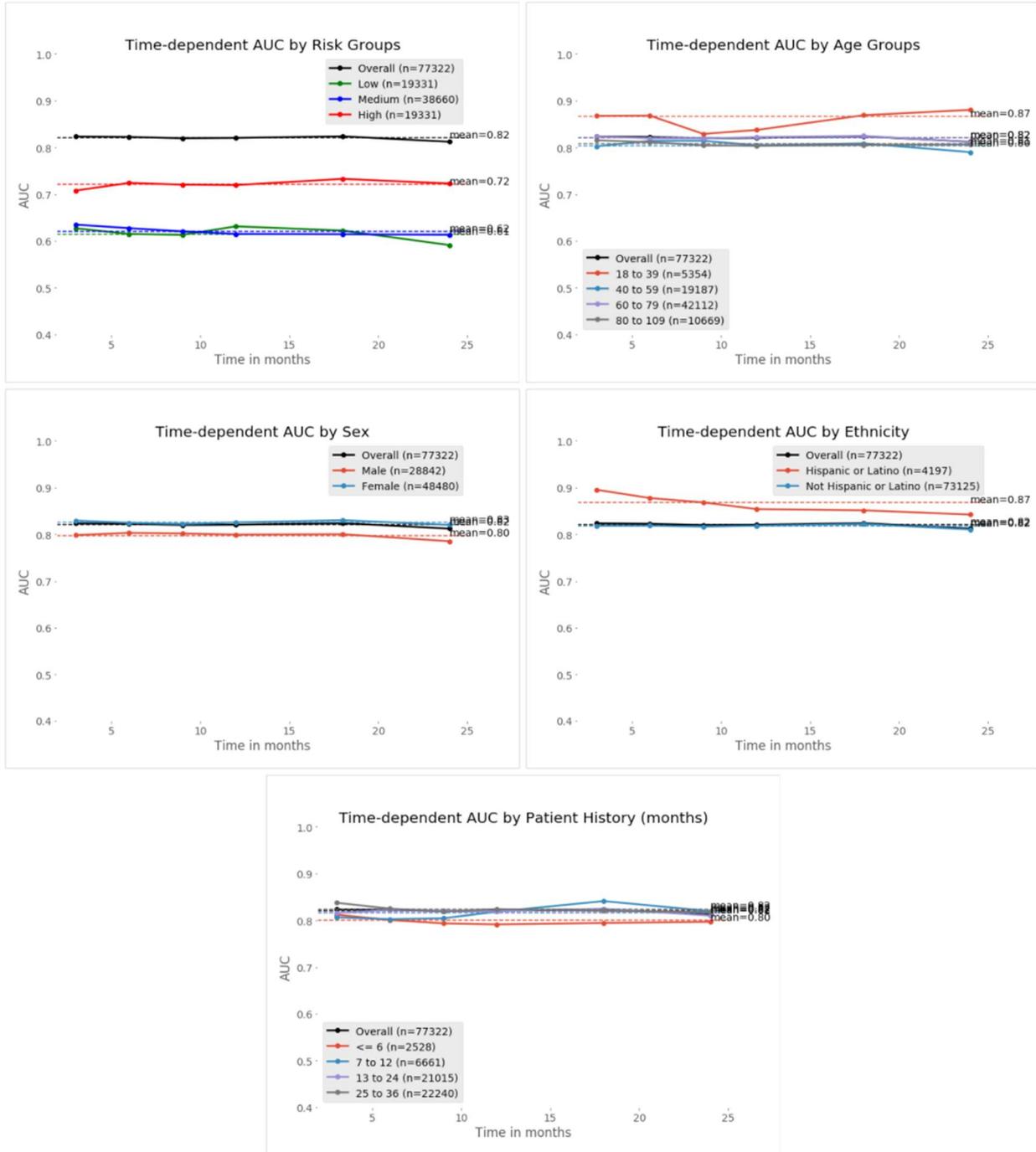

Time-dependent Specificity

The following charts show how the specificity across time varies among subgroups of risk, age, sex, ethnicity and patient history.

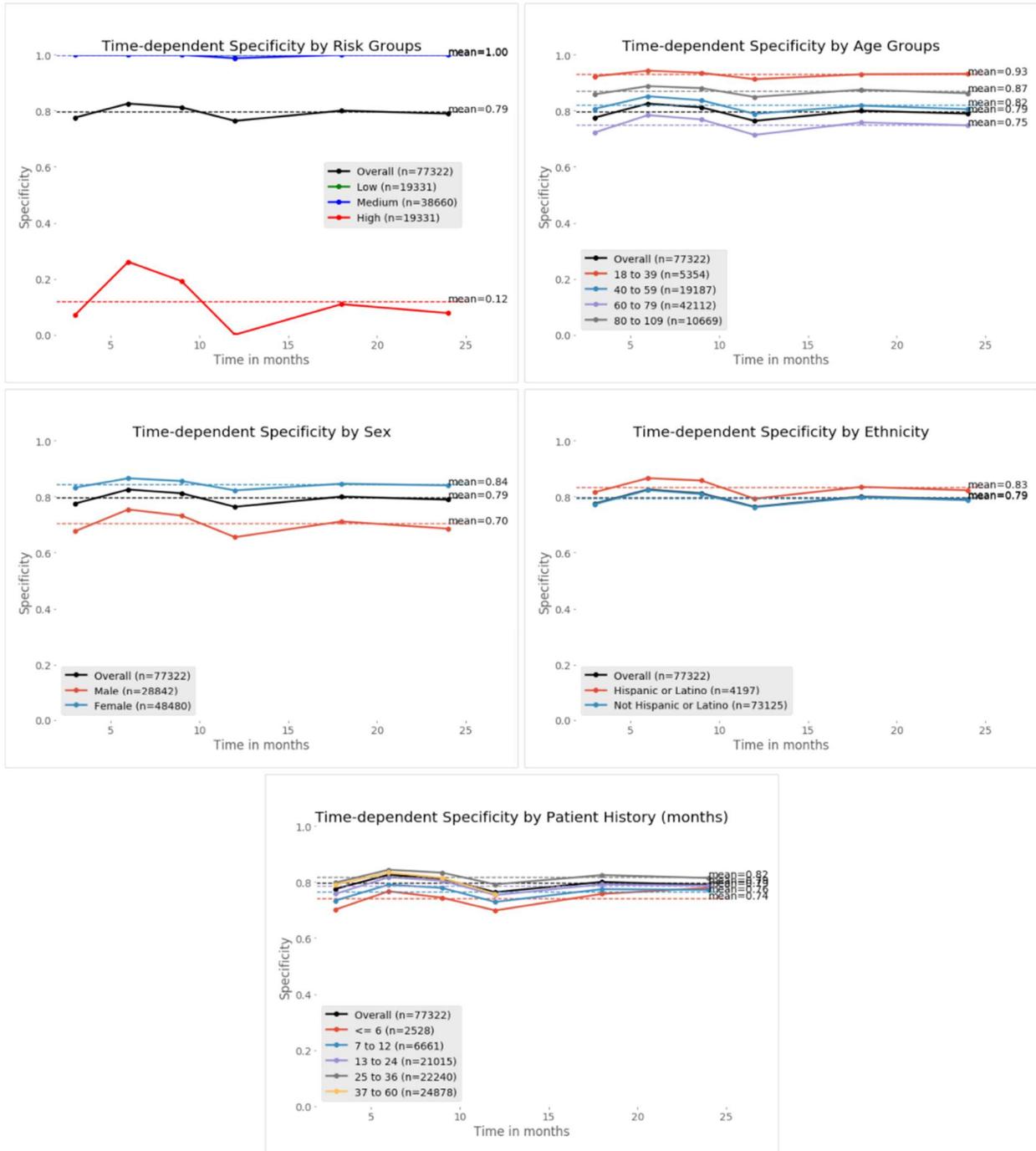

Time-dependent Sensitivity

The following charts show how the sensitivity across time varies among subgroups of risk, age, sex, ethnicity and patient history.

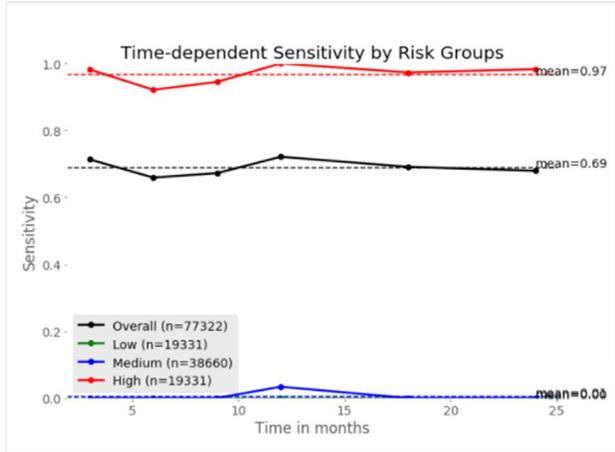
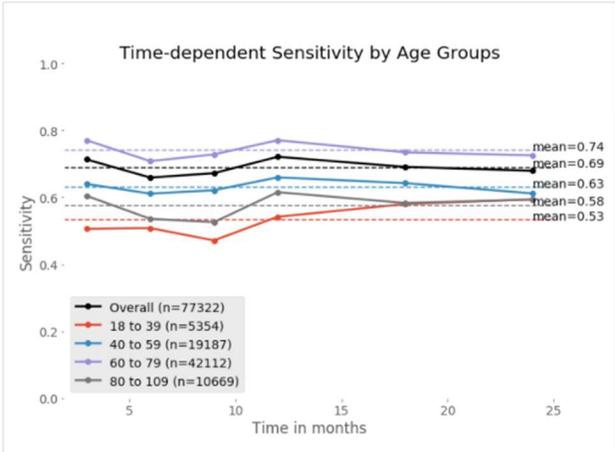
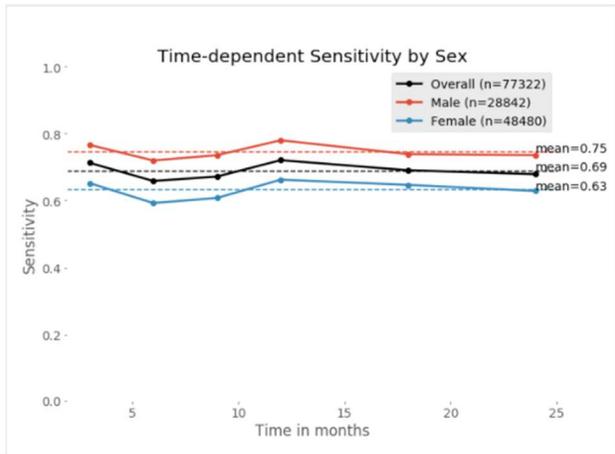
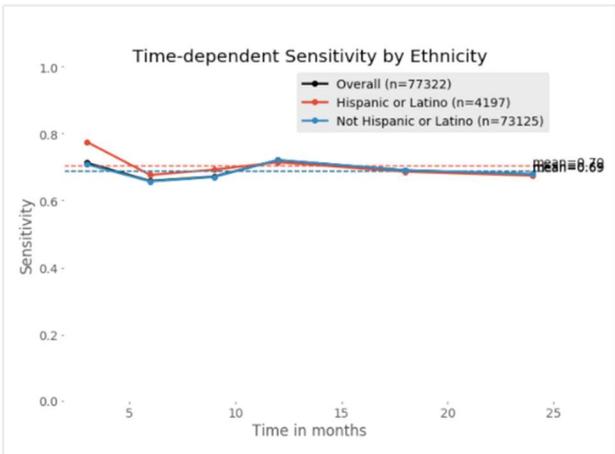
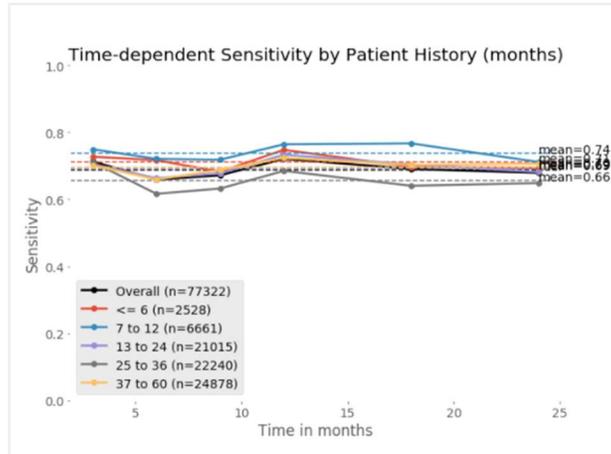

## Model Explanation (DeepSurv)

### Global

The following plots show the SHAP values of each instance in the training set for each future time (3, 6, 9, 12, 18 and 24 months). The features are sorted by the total magnitude of the SHAP values over all instances and the distribution of the effect that each feature has on the model's output can be observed.

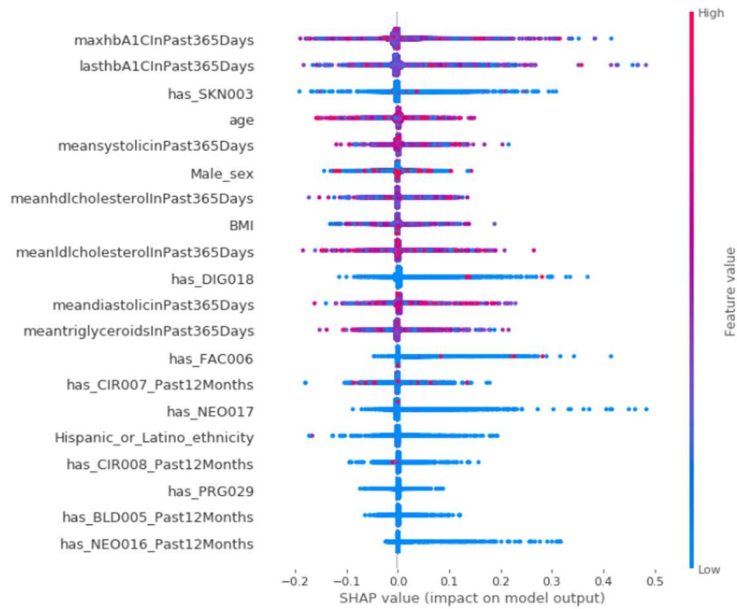

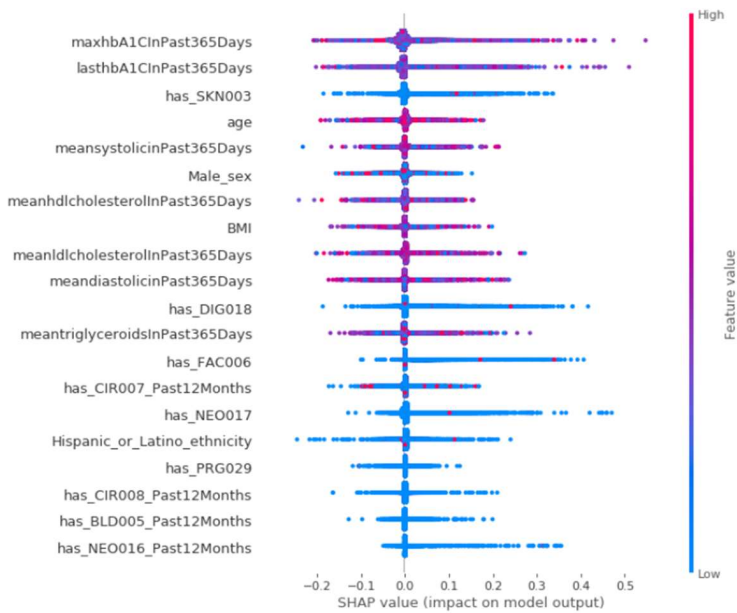

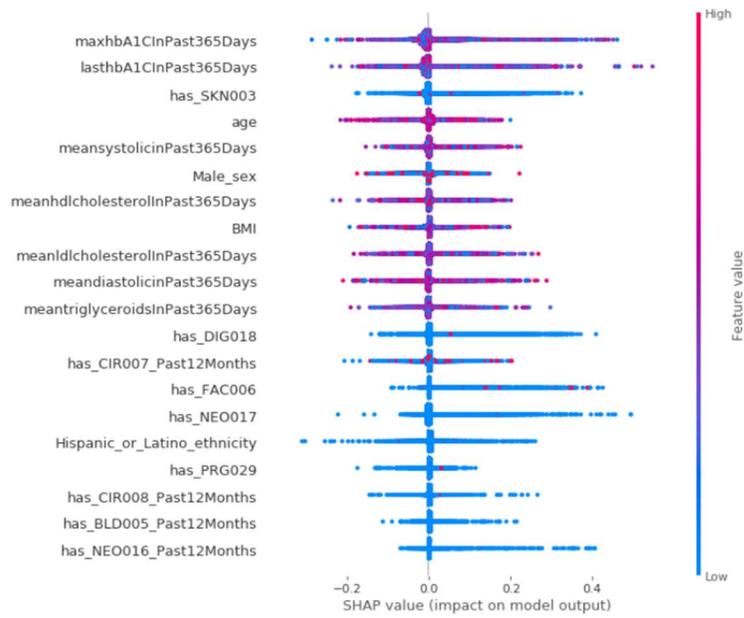

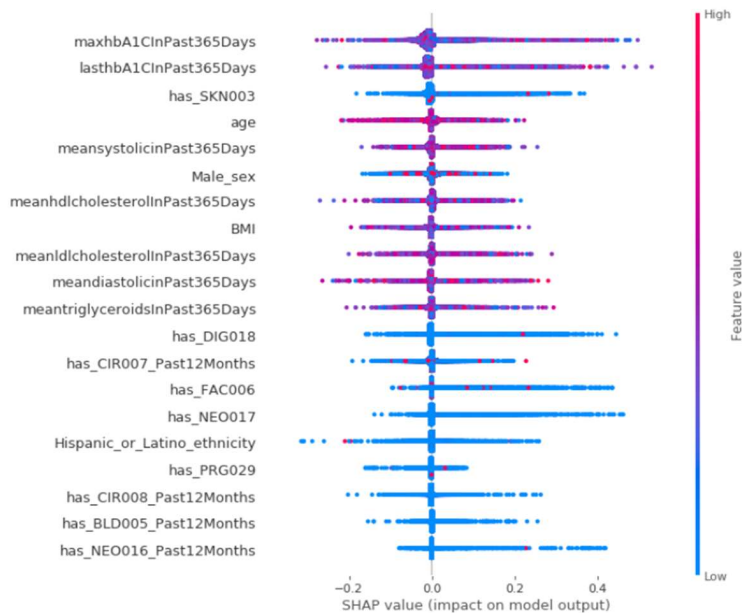

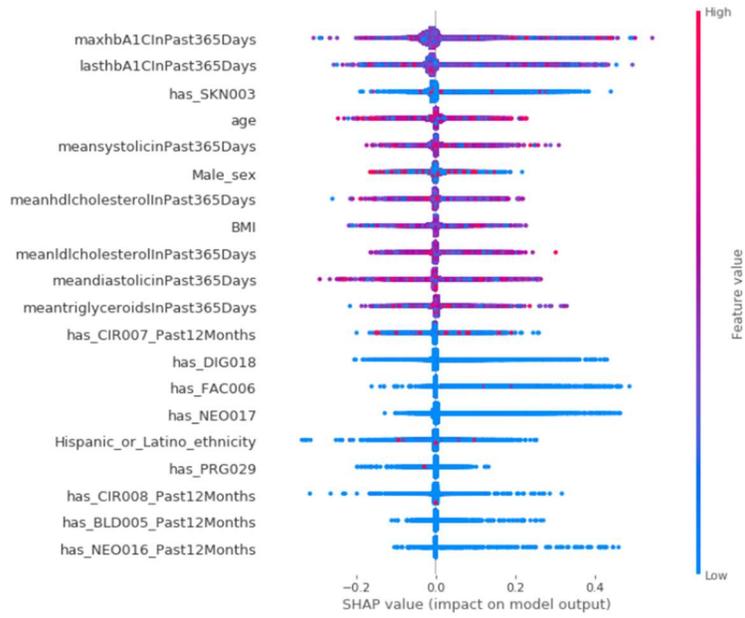
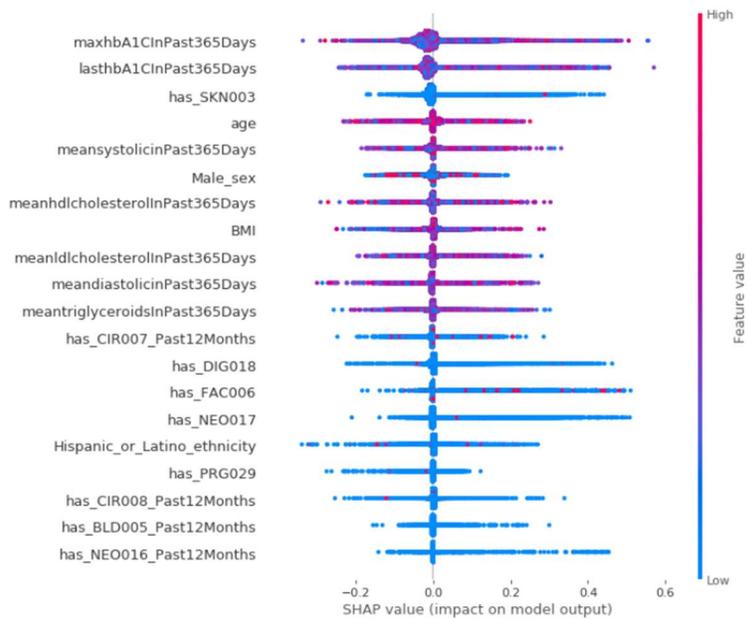

Local

SHAP values were also generated to explain the predictions of individual examples for each future time (3, 6, 9, 12, 18 and 24 months). A total of 3 examples were selected by sampling of risk scores at the 5th, 50th and 95th percentile to represent instances at low, medium and high risks respectively.

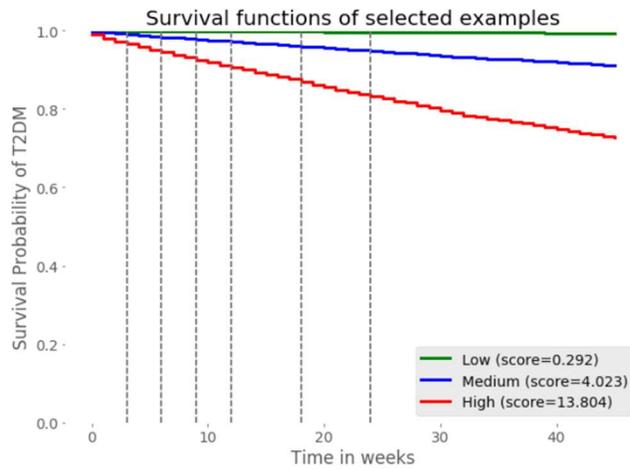

**Low Risk**
Risk Score: 0.292

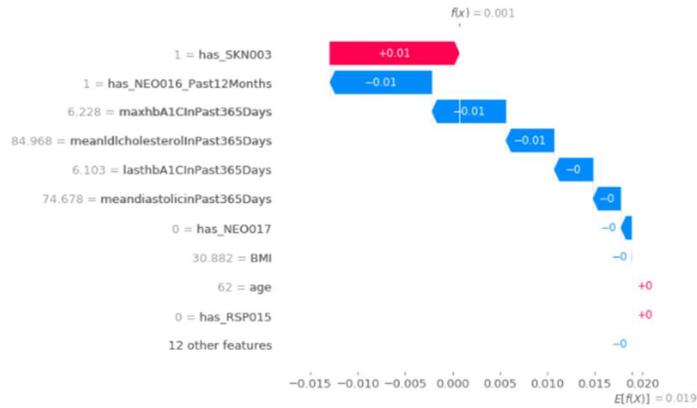

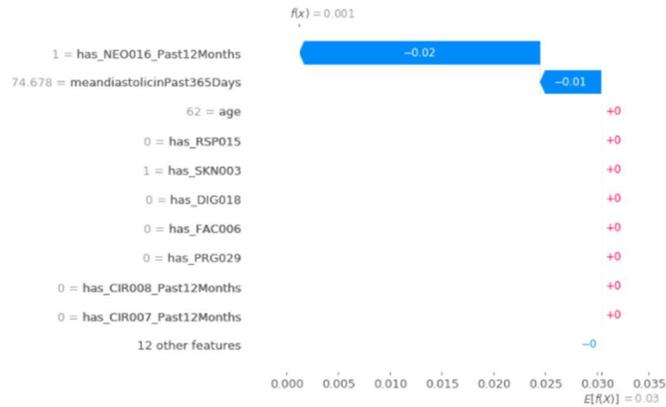

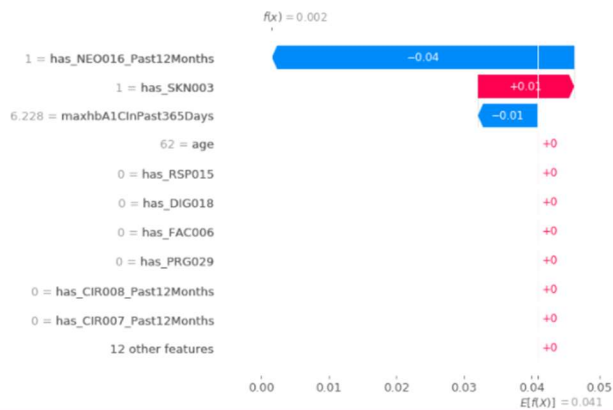

**Low Risk**
Risk Score: 0.292

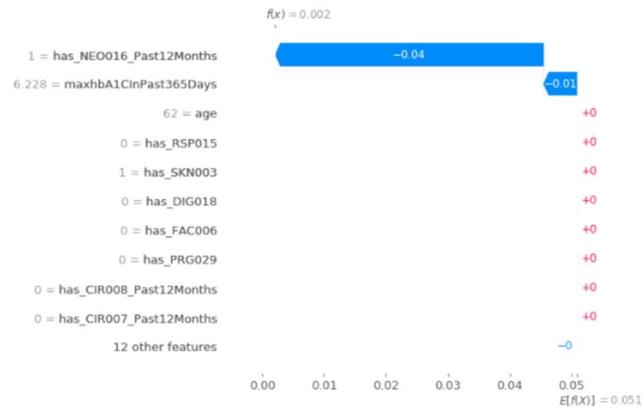

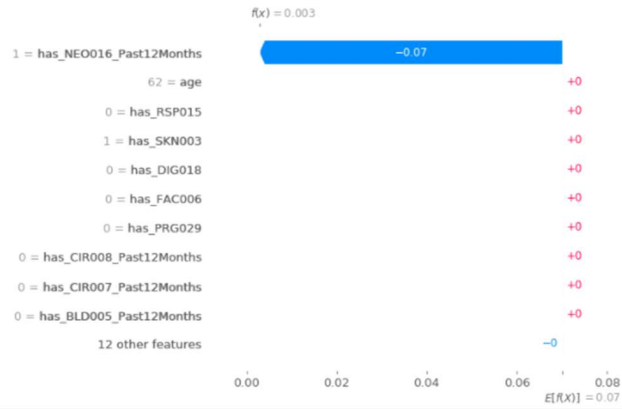

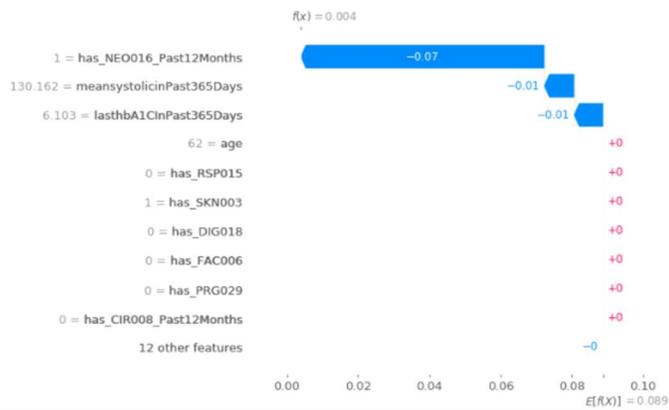

**Medium Risk**
Risk Score: 4.023

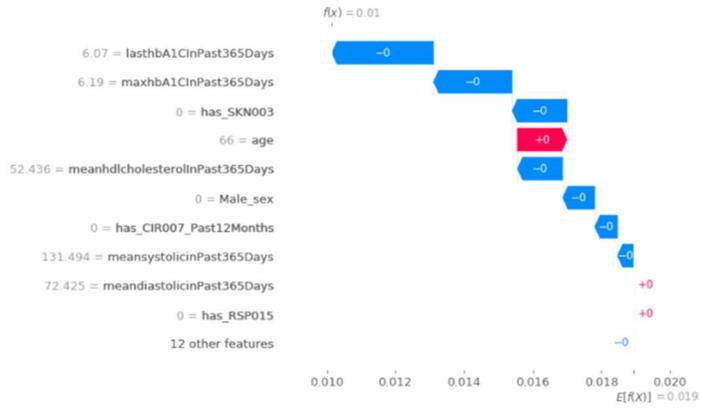

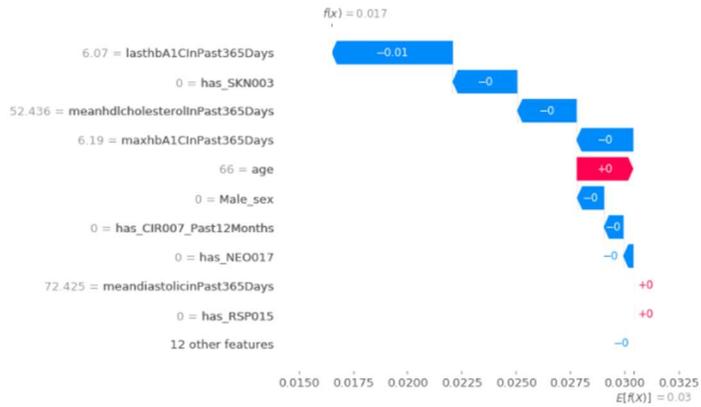

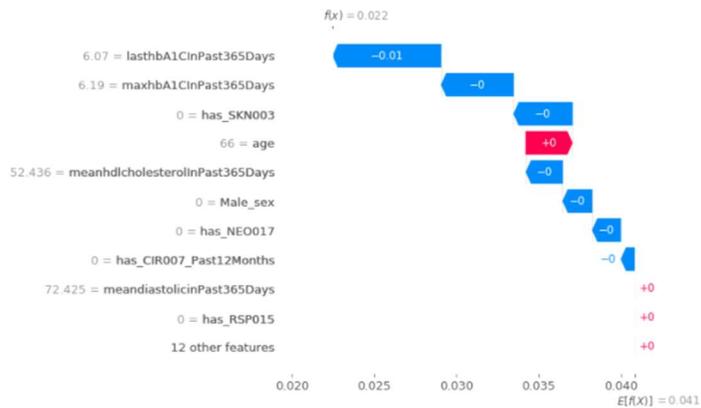

**Medium Risk**
Risk Score: 4.023

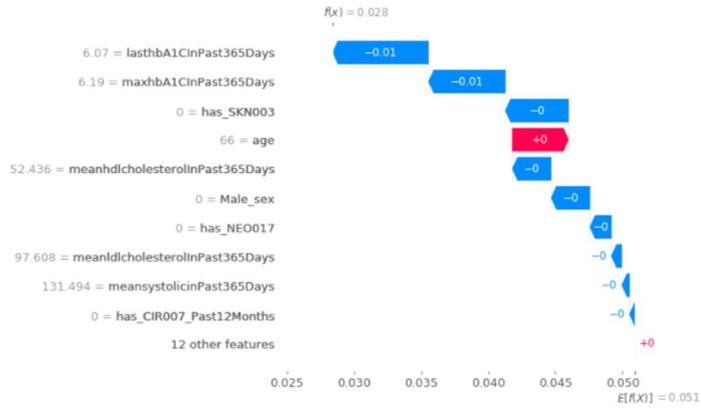

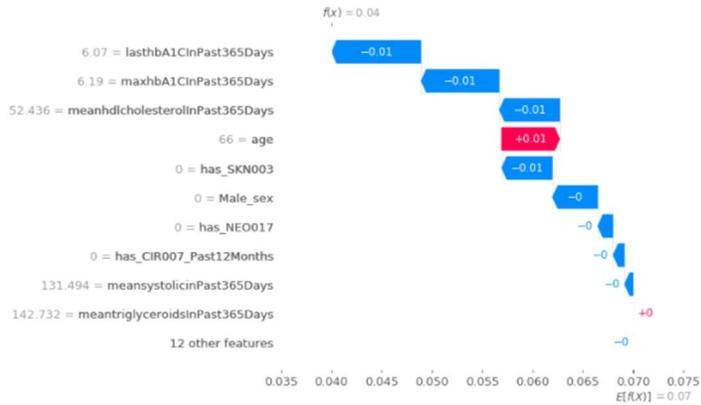

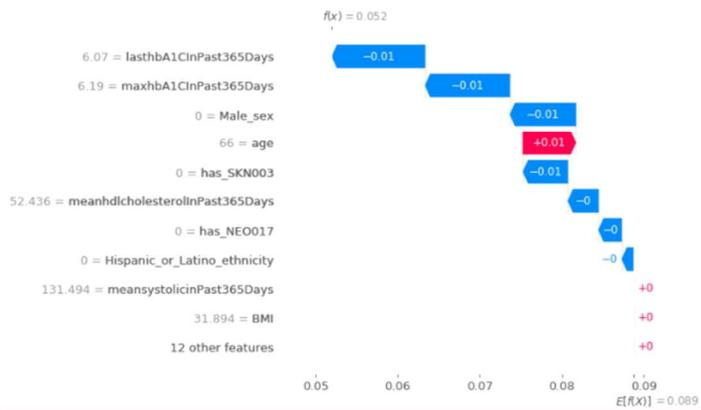

**High Risk**
Risk Score: 13.804

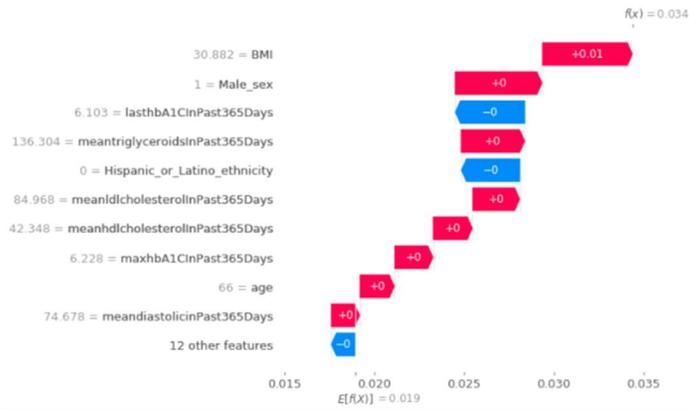

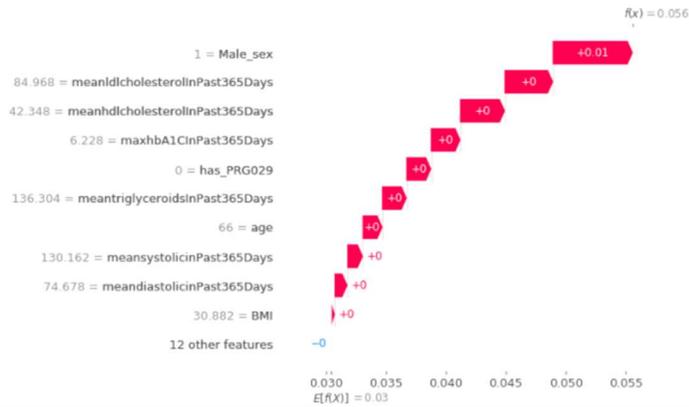

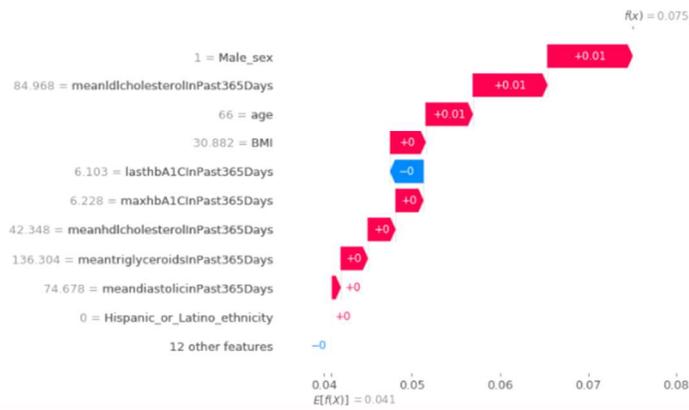

**High Risk**
Risk Score: 13.804

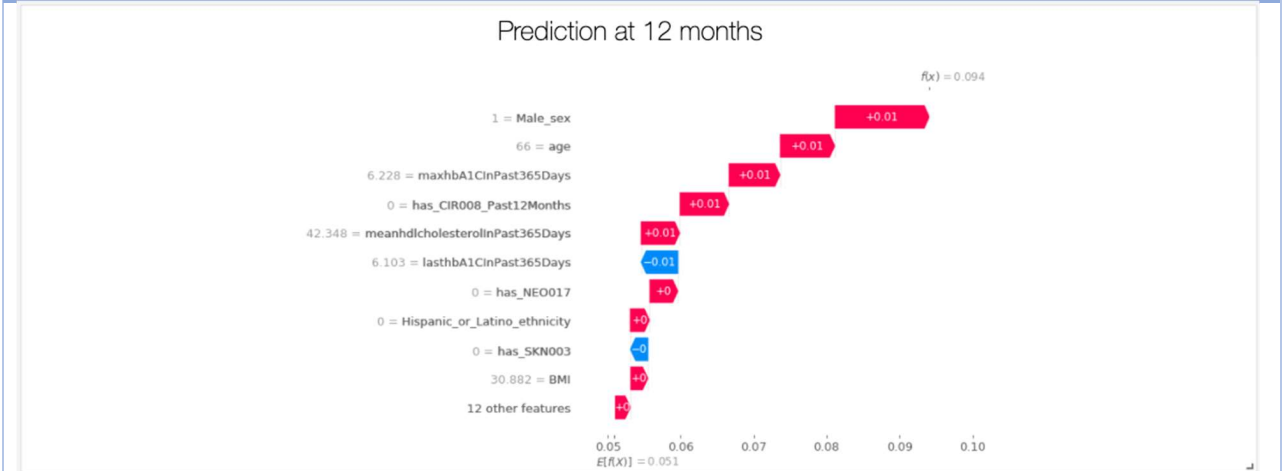

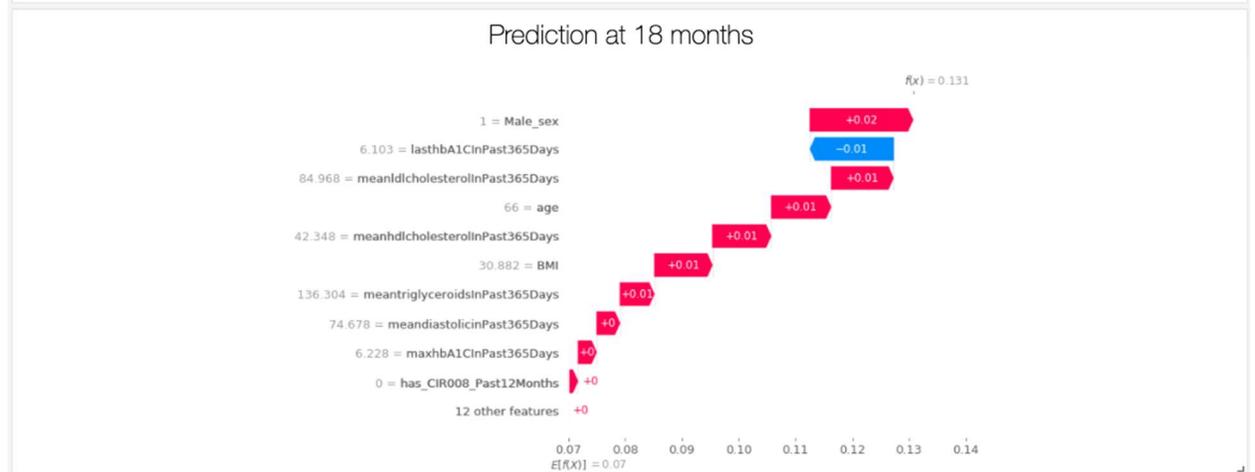

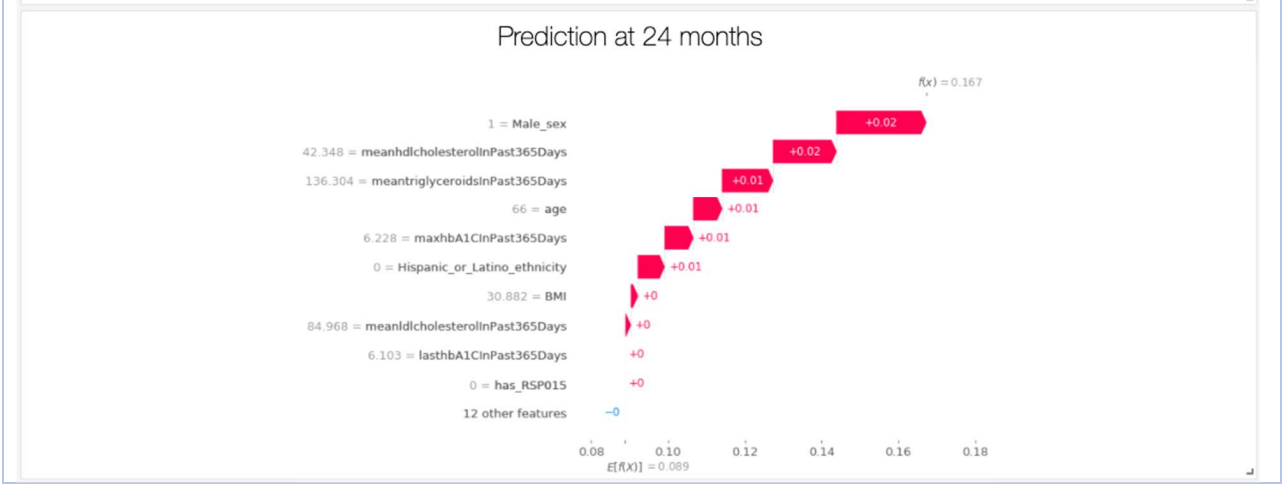

# DM to Uncontrolled DM Prediction

| Item | Specification |
|---|---|
| **Business Goal** | Enable care managers to identify the patients who are at risk of developing uncontrolled DM |
| **Usage Setting** | Outpatient |
| **ML Task** | Predict risk and/or time from DM to uncontrolled DM |
| **ML Class** | Survival |
| **Instances for Prediction** | Encounters |
| **Labels for Instances** | Binary indicator and time to event or censoring for uncontrolled DM <ul><li>Binary indicator $\delta = \begin{cases} 1, & Uncontrolled\ T2DM \\ 0, & No\ Uncontrolled\ T2DM \end{cases}$</li><li>Time to event (or censoring) $y = \begin{cases} T, & \delta = 1 \\ C, & \delta = 0 \end{cases}$</li></ul> |
| **Cohort Criteria** | <ul><li>2016-01-01 $\leq$ encounter date $\leq$ 2020-06-30 (available Epic data, excluding outliers)</li><li>Encounter date is not within the first 90 days of when the patient entered the data set, to adjust for left-censoring</li><li>18 $\leq$ age $\leq$ 110 (adults without outliers)</li><li>No T1DM diagnosis</li><li>Not pregnant</li><li>No "Do Not Resuscitate" diagnosis</li><li>DM event before encounter date</li><li>No uncontrolled DM event before encounter date</li><li>No uncontrolled DM event up to 6 days after encounter date (encounters where lab results confirm event within the week)</li></ul> |
| **Input Features** | <ul><li>Demographic</li><li>Diagnosis, except:<ul><li>hasDiabetes*</li><li>has_END002* (Diabetes mellitus without complication)</li><li>has_END003* (Diabetes mellitus with complication)</li><li>has_END004* (Diabetes mellitus, Type 1)</li><li>has_END005* (Diabetes mellitus, Type 2)</li><li>has_END006* (iabetes mellitus, due to underlying condition, drug or chemical induced, or other specified type)</li></ul></li><li>Labs</li><li>Utilization</li><li>Vitals</li></ul> See Appendix E for details. |
| **Evaluation Metrics** | <ul><li>Concordance Index</li><li>Integrated Brier Score</li></ul> |

## Data

The following charts summarize the key characteristics of the data after applying the cohort criteria stated above, along with selected features (see Model Signature below).

| Category | Variable | count | mean | stddev | min | 25% | 50% | 75% | max |
|---|---|---|---|---|---|---|---|---|---|
| Demographic | AgeBucket_18_to_39 | 49849.0 | 0.038035 | 0.191282 | 0.00 | 0.000 | 0.000 | 0.000 | 1.00 |
| | AgeBucket_40_to_59 | 49849.0 | 0.226785 | 0.418757 | 0.00 | 0.000 | 0.000 | 0.000 | 1.00 |
| | AgeBucket_60_to_79 | 49849.0 | 0.584325 | 0.492843 | 0.00 | 0.000 | 1.000 | 1.000 | 1.00 |
| | AgeBucket_80_to_109 | 49849.0 | 0.150856 | 0.357912 | 0.00 | 0.000 | 0.000 | 0.000 | 1.00 |
| | Sex_Female | 49849.0 | 0.549780 | 0.497521 | 0.00 | 0.000 | 1.000 | 1.000 | 1.00 |
| | Sex_Male | 49849.0 | 0.450220 | 0.497521 | 0.00 | 0.000 | 0.000 | 1.000 | 1.00 |
| | Ethnicity_Hispanic_or_Latino | 49849.0 | 0.062288 | 0.241681 | 0.00 | 0.000 | 0.000 | 0.000 | 1.00 |
| | Ethnicity_Not_Hispanic_or_Latino | 49849.0 | 0.937712 | 0.241681 | 0.00 | 1.000 | 1.000 | 1.000 | 1.00 |
| Encounter | EncounterType_Emergency | 49849.0 | 0.212341 | 0.408969 | 0.00 | 0.000 | 0.000 | 0.000 | 1.00 |
| | EncounterType_Inpatient | 49849.0 | 0.113864 | 0.317649 | 0.00 | 0.000 | 0.000 | 0.000 | 1.00 |
| | EncounterType_Outpatient | 49849.0 | 0.673795 | 0.468828 | 0.00 | 0.000 | 1.000 | 1.000 | 1.00 |
| Label | Time | 49849.0 | 15.095669 | 11.302945 | 0.00 | 5.000 | 13.000 | 23.000 | 49.00 |
| | Event | 49849.0 | 0.106682 | 0.308712 | 0.00 | 0.000 | 0.000 | 0.000 | 1.00 |
| Feature | age | 49849.0 | 66.416297 | 13.014329 | 18.00 | 59.000 | 68.000 | 75.000 | 102.00 |
| | Male_sex | 49849.0 | 0.450220 | 0.497521 | 0.00 | 0.000 | 0.000 | 1.000 | 1.00 |
| | Hispanic_or_Latino_ethnicity | 49849.0 | 0.062288 | 0.241681 | 0.00 | 0.000 | 0.000 | 0.000 | 1.00 |
| | lasthbA1CInPast365Days | 49849.0 | 6.871925 | 0.766836 | 3.60 | 6.600 | 6.866 | 7.200 | 8.90 |
| | meandiastolicinPast365Days | 49849.0 | 72.694129 | 7.925292 | 43.62 | 68.888 | 73.000 | 76.790 | 107.56 |
| | meansystolicinPast365Days | 49849.0 | 131.636895 | 11.790046 | 81.00 | 127.316 | 131.065 | 136.050 | 180.00 |
| | BMI | 49849.0 | 32.705948 | 6.495782 | 10.37 | 29.190 | 32.440 | 35.570 | 80.03 |
| | meantriglyceroidsInPast365Days | 49849.0 | 161.155430 | 52.310612 | 19.00 | 138.000 | 163.096 | 170.874 | 458.00 |
| | meanldlcholesterolInPast365Days | 49849.0 | 85.611683 | 21.682910 | 8.00 | 80.720 | 85.917 | 95.500 | 200.00 |
| | meanhdlcholesterolInPast365Days | 49849.0 | 43.558758 | 9.079637 | 5.00 | 39.813 | 43.000 | 47.864 | 97.00 |
| | has_MAL004_Past12Months | 49849.0 | 0.000622 | 0.024930 | 0.00 | 0.000 | 0.000 | 0.000 | 1.00 |
| | has_CIR007_Past12Months | 49849.0 | 0.196774 | 0.397564 | 0.00 | 0.000 | 0.000 | 0.000 | 1.00 |
| | has_NEO061_Past12Months | 49849.0 | 0.001063 | 0.032590 | 0.00 | 0.000 | 0.000 | 0.000 | 1.00 |
| | has_CIR008_Past12Months | 49849.0 | 0.011134 | 0.104928 | 0.00 | 0.000 | 0.000 | 0.000 | 1.00 |
| | has_EXT030 | 49849.0 | 0.001324 | 0.036363 | 0.00 | 0.000 | 0.000 | 0.000 | 1.00 |
| | has_NEO029 | 49849.0 | 0.004674 | 0.068208 | 0.00 | 0.000 | 0.000 | 0.000 | 1.00 |
| | has_END013 | 49849.0 | 0.005055 | 0.070921 | 0.00 | 0.000 | 0.000 | 0.000 | 1.00 |
| | has_CIR038 | 49849.0 | 0.006038 | 0.077472 | 0.00 | 0.000 | 0.000 | 0.000 | 1.00 |
| | has_MUS029 | 49849.0 | 0.007182 | 0.084441 | 0.00 | 0.000 | 0.000 | 0.000 | 1.00 |

(Percentages for binary variables can be read from the "mean" column.)

## Encounters and Patients

| | Examples | Encounters | Patients |
|---|---|---|---|
| 1 | 49849 | 49849 | 8064 |

## Train and Test Sets

| | Set | No Event | No Event % | Event | Event % | Total | Total % |
|---|---|---|---|---|---|---|---|
| 1 | Test | 13317 | 29.9 | 1636 | 30.8 | 14953 | 30 |
| 2 | Train | 31214 | 70.1 | 3682 | 69.2 | 34896 | 70 |
| 3 | Total | 44531 | 100 | 5318 | 100 | 49849 | 100 |

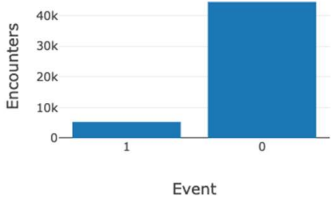

Encounters with Event (1) or Censoring (0)

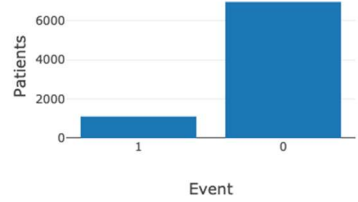

Patients with Event (1) or Censoring (0)

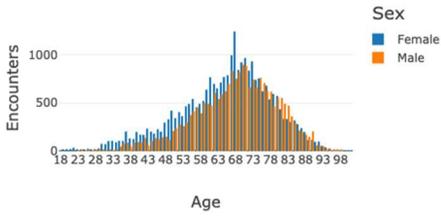

Encounters by Age and Sex

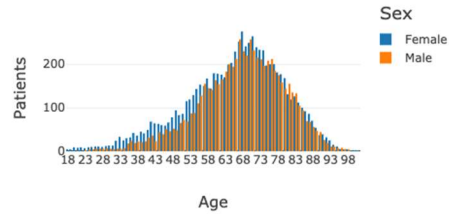

Patients by Age and Sex

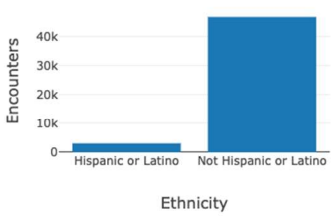

Encounters by Ethnicity

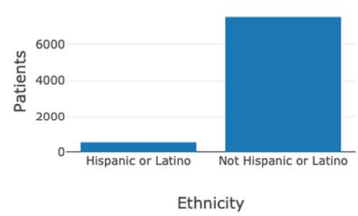

Patients by Ethnicity

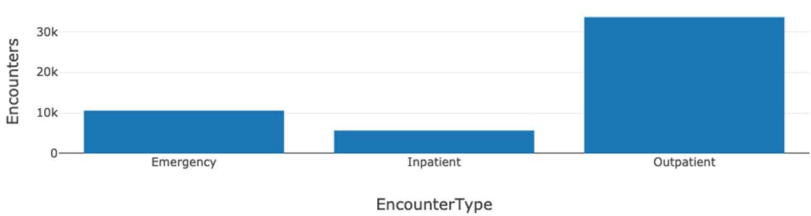

Encounters by Encounter Type

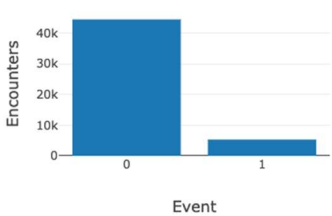

Encounters by Time to Event (1) or Cen…

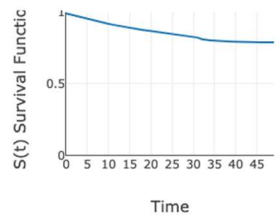

Kaplan-Meier Estimate of Survival Functi…

Model Signature

The model signature has 19 features, comprising of 12 mandatory features and 7 other selected features. These are the selected features, in rank order (the last feature to be eliminated is ranked 1):

1. has_CIR038 (Postprocedural or postoperative circulatory system complication)
2. has_END013 (Pituitary disorders)
3. has_MAL004_Past12Months (Nervous system congenital anomalies)
4. has_EXT030 (External cause codes: sequela)
5. has_NEO061_Past12Months (Leukemia - chronic lymphocytic leukemia (CLL))
6. has_MUS029 (Disorders of jaw)
7. has_NEO029 (Breast cancer - ductal carcinoma in situ (DCIS))
8. lasthbA1CInPast365Days
9. Hispanic_or_Latino_ethnicity
10. has_CIR008_Past12Months (Hypertension with complications and secondary hypertension)
11. has_CIR007_Past12Months (Essential hypertension)
12. BMI
13. Male_sex
14. age
15. meansystolicinPast365Days
16. meanhdlcholesterolInPast365Days
17. meandiastolicinPast365Days
18. meanldlcholesterolInPast365Days
19. meantriglyceroidsInPast365Days

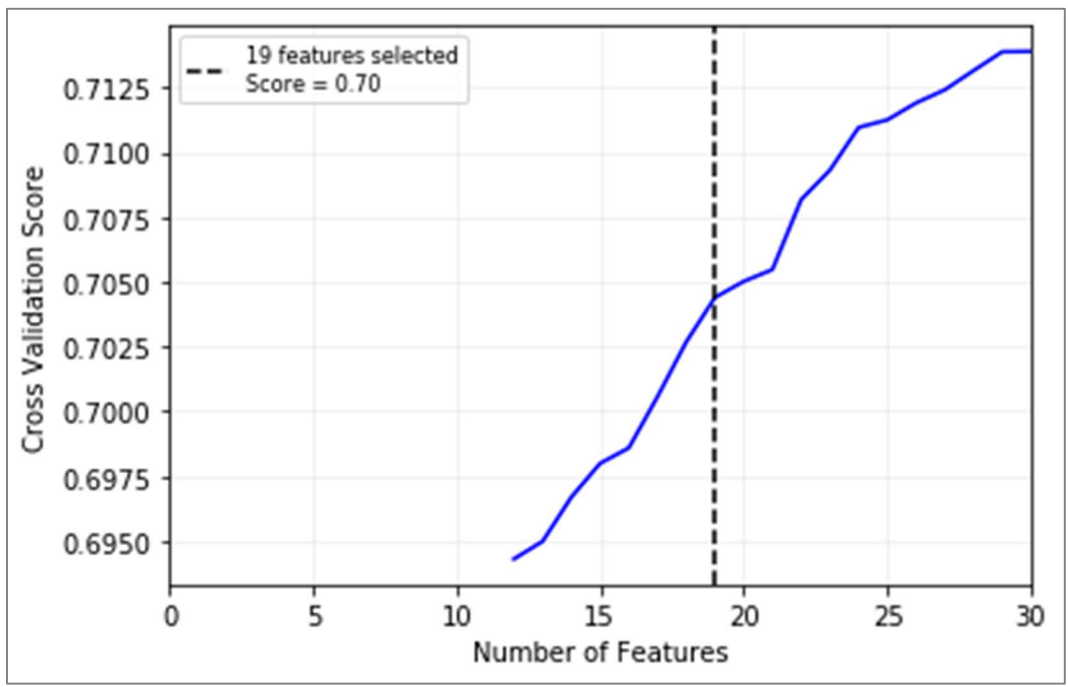

## Model Performance

The following table and chart summarize the performance of all candidate models on the test set for this prediction task in terms of the Concordance Index and the Integrated Brier Score.

| Model | No. of Parameter Combinations Successfully Tested | Concordance Index | Integrated Brier Score |
|---|---|---|---|
| **CoxPH** | 94 | 0.70 | 0.10 |
| **DeepSurv** | 161 | **0.79** | **0.08** |
| **RSF** | 143 | 0.72 | 0.10 |
| **CSF** | 82 | 0.71 | 0.11 |
| **EST** | 71 | 0.73 | 0.10 |

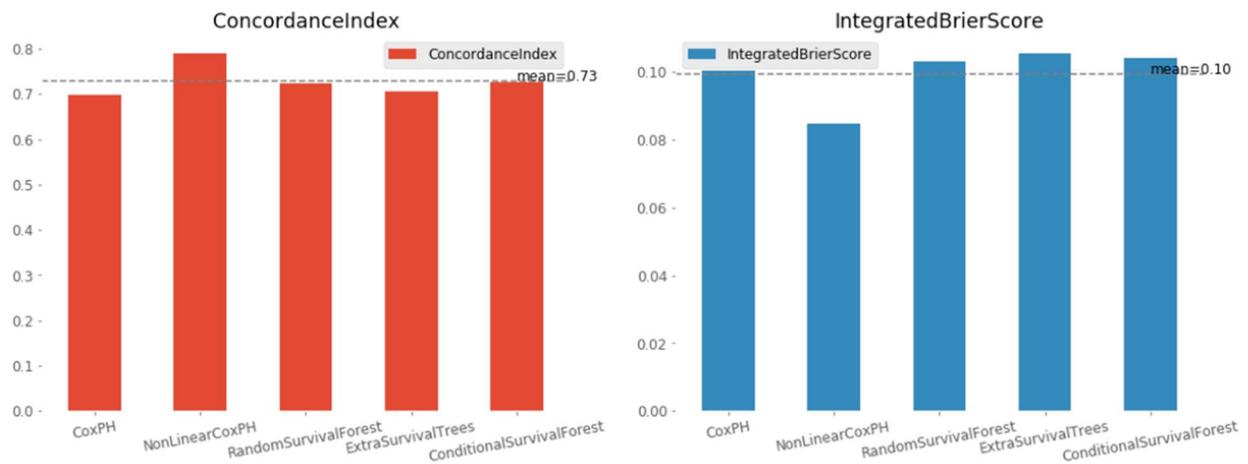

The following chart shows how the average survival function curves of the candidate models compare to the KM survival curve, the more similar their curves are to the KM survival curve the better.

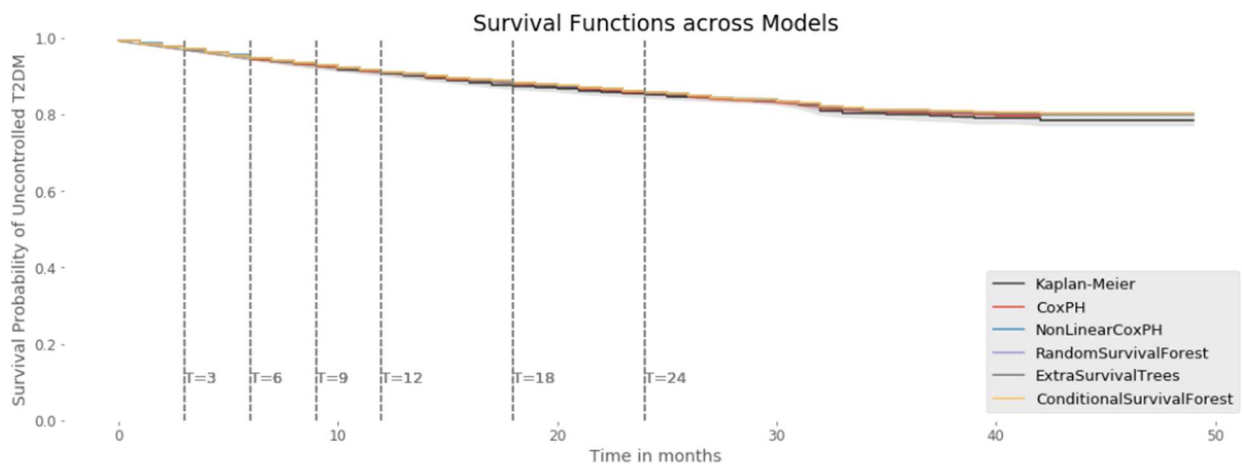

This chart shows the change in the Brier Score over time for all candidate models, the closer the scores are to 0 the better.

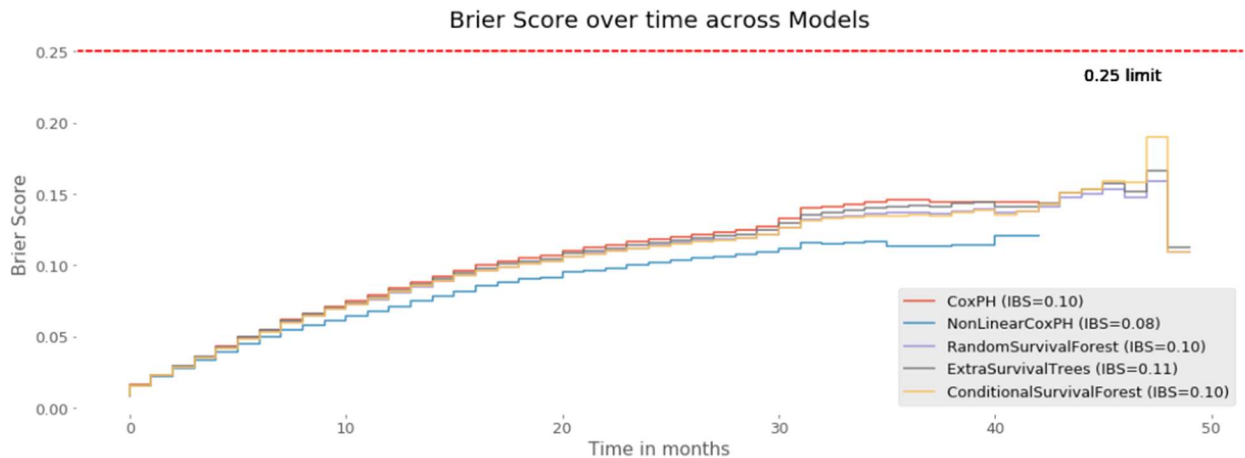

## Model Evaluation of Selected Model (DeepSurv)

Overall

This chart shows the change in the Brier Score over time for the selected model, the closer the scores are to 0 the better.

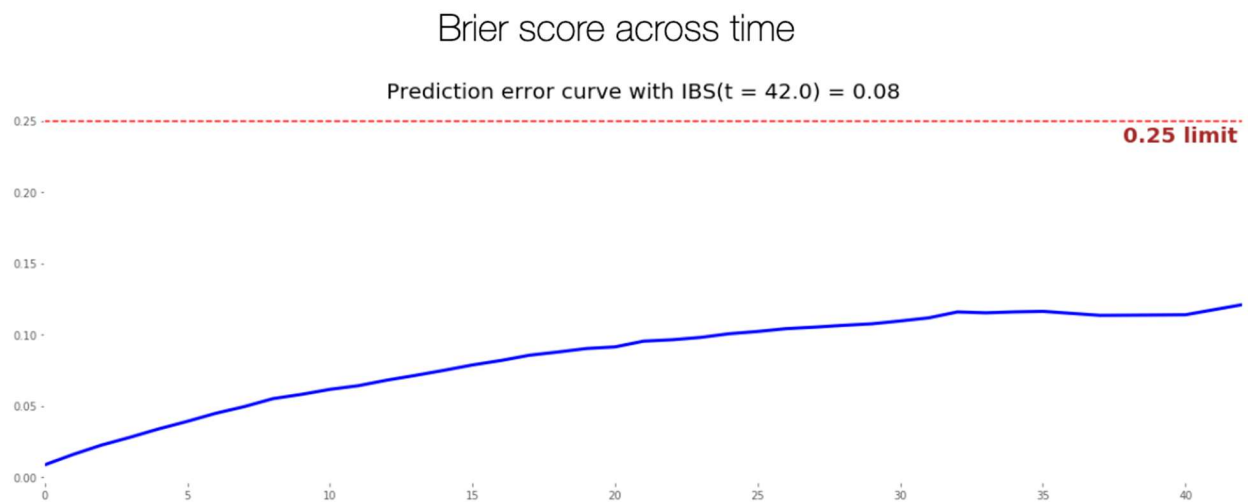

The following chart shows the actual vs. predicted density functions, i.e. number of instances that get the disease / complication at each time point and the RMSE, Median Absolute Error and Mean Absolute Error across the time points.

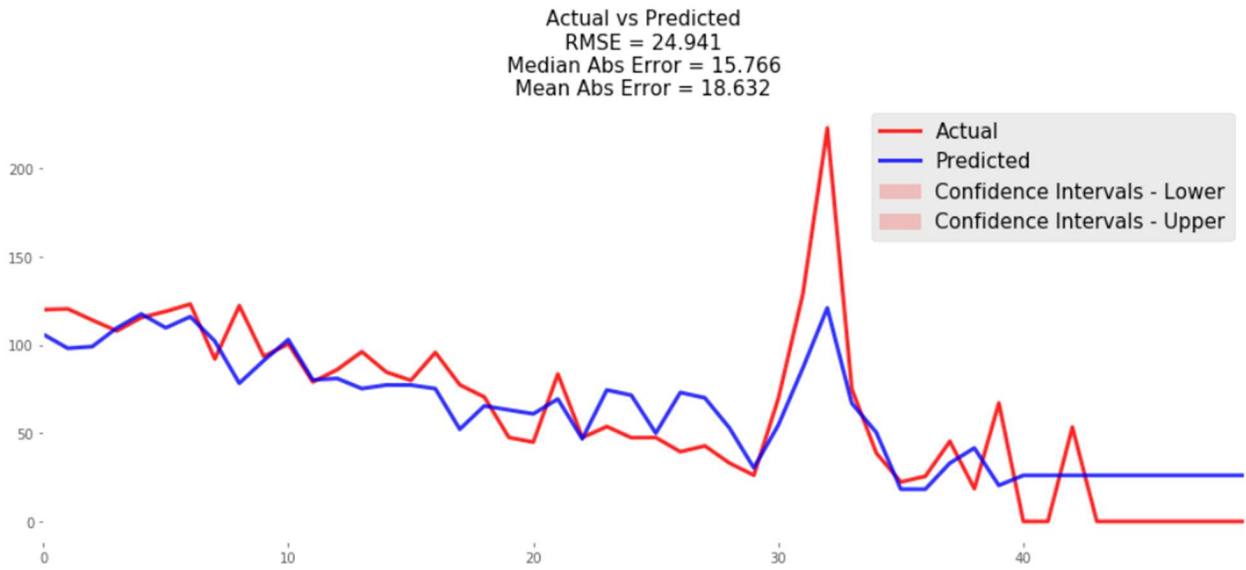

The following chart shows the actual vs. predicted survival functions, i.e. the number of instances that have not had the disease / complication by each time point and the RMSE, Median Absolute Error and Mean Absolute Error across the time points.

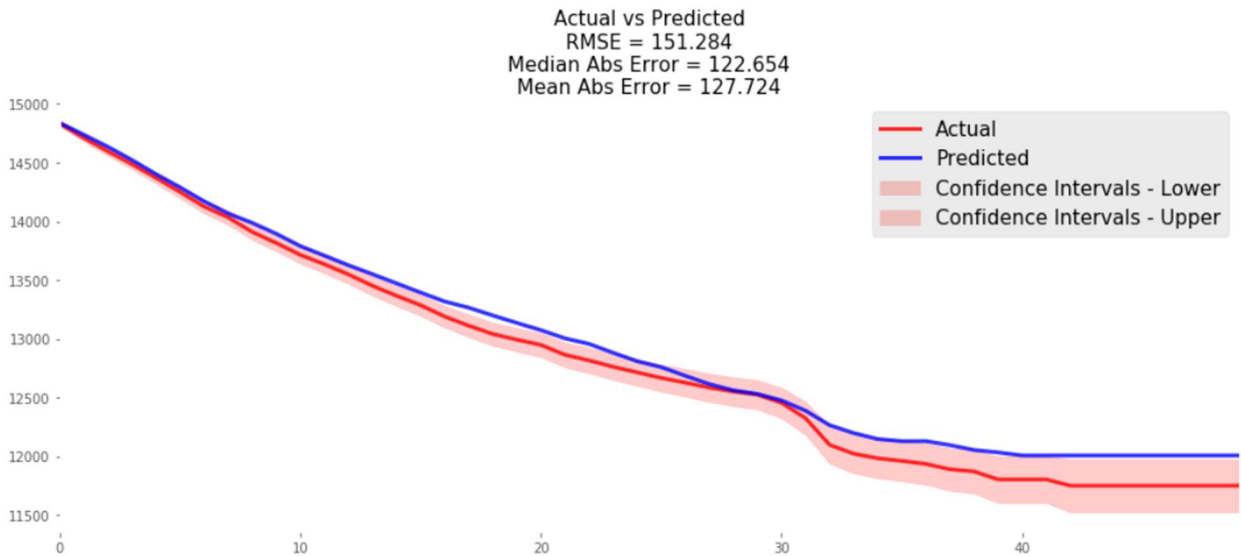

Risk Stratification

The low, medium and high risk groups are defined as examples with predicted risk scores belonging to the first quartile, second to third quartiles, and fourth quartile respectively.

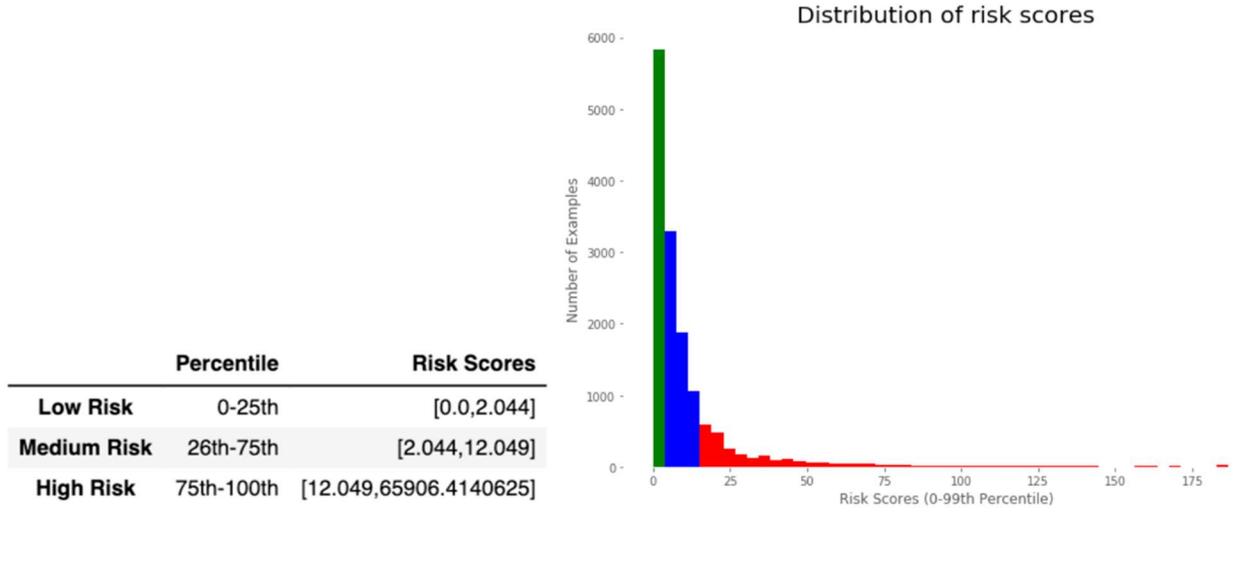

|  | Percentile | Risk Scores |
|---|---|---|
| Low Risk | 0-25th | [0.0, 2.044] |
| Medium Risk | 26th-75th | [2.044, 12.049] |
| High Risk | 75th-100th | [12.049, 65906.4140625] |

Summary Metrics across Subgroups

The table below displays the summary metrics across subgroups of risk, age, sex, ethnicity and patient history.

| Category | Subgroup | Cohort Size | Concordance Index | Brier Score | Mean AUC | Mean Specificity | Mean Sensitivity | S(t), t=3 | S(t), t=6 | S(t), t=9 | S(t), t=12 | S(t), t=18 | S(t), t=24 |
|---|---|---|---|---|---|---|---|---|---|---|---|---|---|
| NaN | Overall | 14953.00 | 0.79 | 0.08 | 0.80 | 0.73 | 0.73 | 0.97 | 0.94 | 0.92 | 0.91 | 0.87 | 0.85 |
| Risk | Low | 3738.00 | 0.61 | 0.02 | 0.61 | 1.00 | 0.00 | 1.00 | 1.00 | 1.00 | 0.99 | 0.99 | 0.99 |
| Risk | Medium | 7476.00 | 0.61 | 0.07 | 0.63 | 0.89 | 0.25 | 0.99 | 0.98 | 0.96 | 0.95 | 0.93 | 0.91 |
| Risk | High | 3739.00 | 0.69 | 0.17 | 0.72 | 0.00 | 1.00 | 0.91 | 0.84 | 0.79 | 0.74 | 0.67 | 0.61 |
| Age Bucket | 18 to 39 | 593.00 | 0.85 | 0.07 | 0.86 | 0.65 | 0.88 | 0.96 | 0.93 | 0.91 | 0.89 | 0.86 | 0.83 |
| Age Bucket | 40 to 59 | 3391.00 | 0.78 | 0.11 | 0.78 | 0.63 | 0.80 | 0.96 | 0.93 | 0.91 | 0.89 | 0.85 | 0.82 |
| Age Bucket | 60 to 79 | 8657.00 | 0.80 | 0.08 | 0.81 | 0.75 | 0.71 | 0.97 | 0.95 | 0.93 | 0.92 | 0.89 | 0.86 |
| Age Bucket | 80 to 109 | 2312.00 | 0.73 | 0.07 | 0.76 | 0.82 | 0.57 | 0.98 | 0.96 | 0.95 | 0.94 | 0.91 | 0.89 |
| Sex | Male | 6749.00 | 0.77 | 0.08 | 0.79 | 0.73 | 0.70 | 0.97 | 0.95 | 0.93 | 0.91 | 0.89 | 0.86 |
| Sex | Female | 8204.00 | 0.81 | 0.09 | 0.81 | 0.72 | 0.76 | 0.97 | 0.95 | 0.93 | 0.91 | 0.88 | 0.85 |
| Ethnicity | Hispanic or Latino | 952.00 | 0.84 | 0.08 | 0.83 | 0.74 | 0.73 | 0.96 | 0.92 | 0.90 | 0.88 | 0.85 | 0.82 |
| Ethnicity | Not Hispanic or Latino | 14001.00 | 0.79 | 0.08 | 0.80 | 0.73 | 0.73 | 0.97 | 0.95 | 0.93 | 0.91 | 0.89 | 0.86 |
| History Bucket | <= 6 | 405.00 | 0.72 | 0.09 | 0.65 | 0.67 | 0.62 | 0.96 | 0.94 | 0.92 | 0.90 | 0.86 | 0.83 |
| History Bucket | 7 to 12 | 1524.00 | 0.81 | 0.07 | 0.81 | 0.74 | 0.74 | 0.97 | 0.95 | 0.93 | 0.91 | 0.88 | 0.85 |
| History Bucket | 13 to 24 | 4374.00 | 0.80 | 0.09 | 0.80 | 0.71 | 0.73 | 0.97 | 0.94 | 0.92 | 0.90 | 0.87 | 0.85 |
| History Bucket | 25 to 36 | 4229.00 | 0.78 | 0.07 | 0.79 | 0.74 | 0.70 | 0.97 | 0.95 | 0.94 | 0.92 | 0.89 | 0.87 |
| History Bucket | 37 to 60 | 4421.00 | 0.83 | 0.05 | nan | nan | 0.79 | 0.97 | 0.95 | 0.93 | 0.91 | 0.88 | 0.86 |

Concordance Index & Integrated Brier Score

The following charts show how the Concordance Index and Integrated Brier Score varies among subgroups of risk, age, sex, ethnicity and patient history.

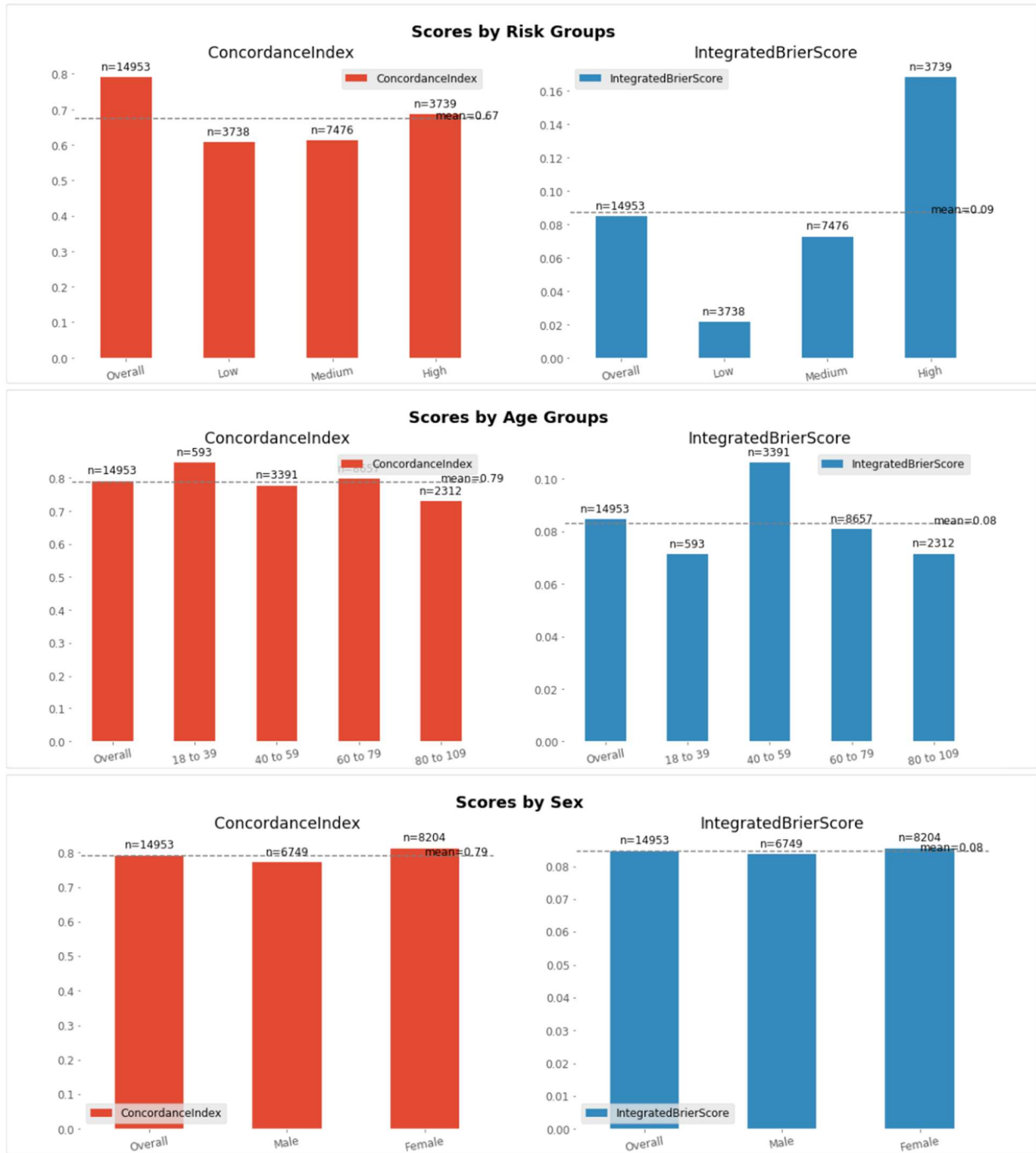

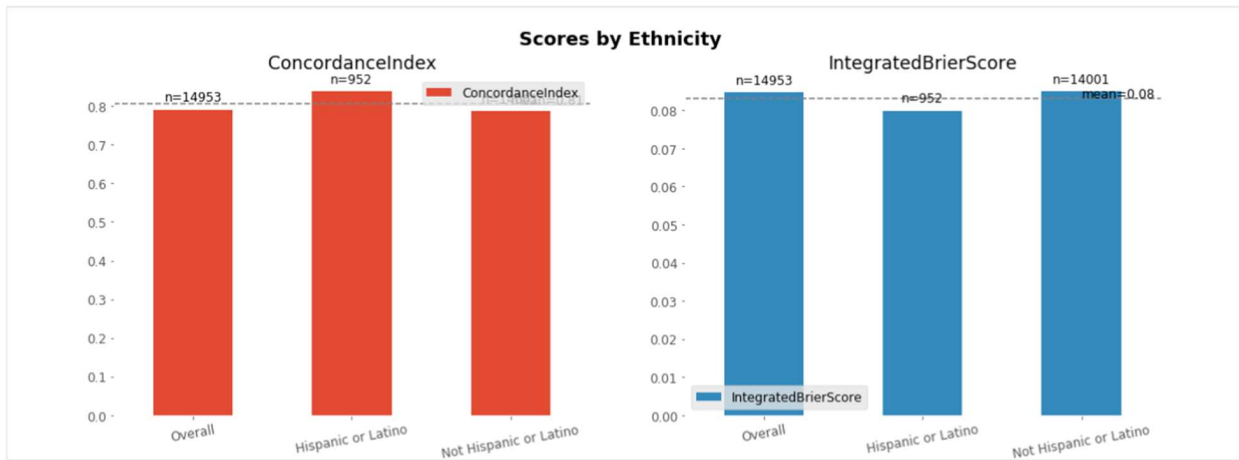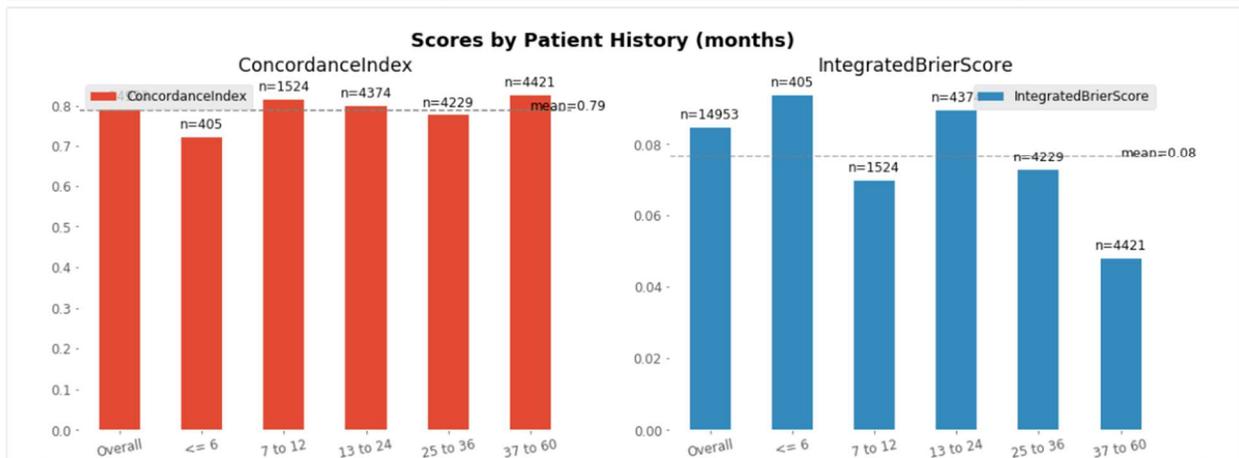

Average Survival Function Curves

The following charts show how the average survival function curve varies among subgroups of risk, age, sex, ethnicity and patient history.

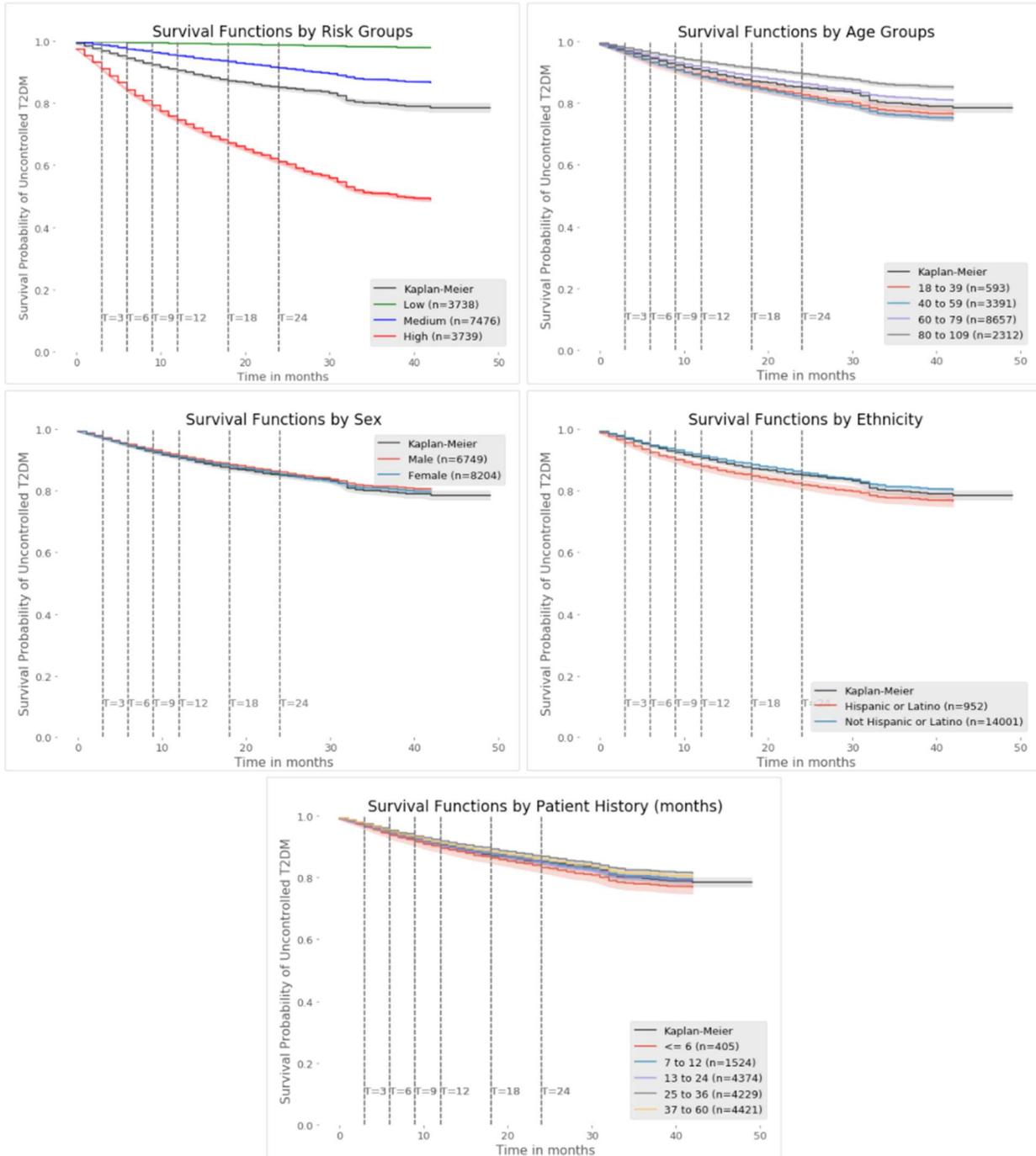

Time-dependent AUC

The following charts show how the AUC across time varies among subgroups of risk, age, sex, ethnicity and patient history.

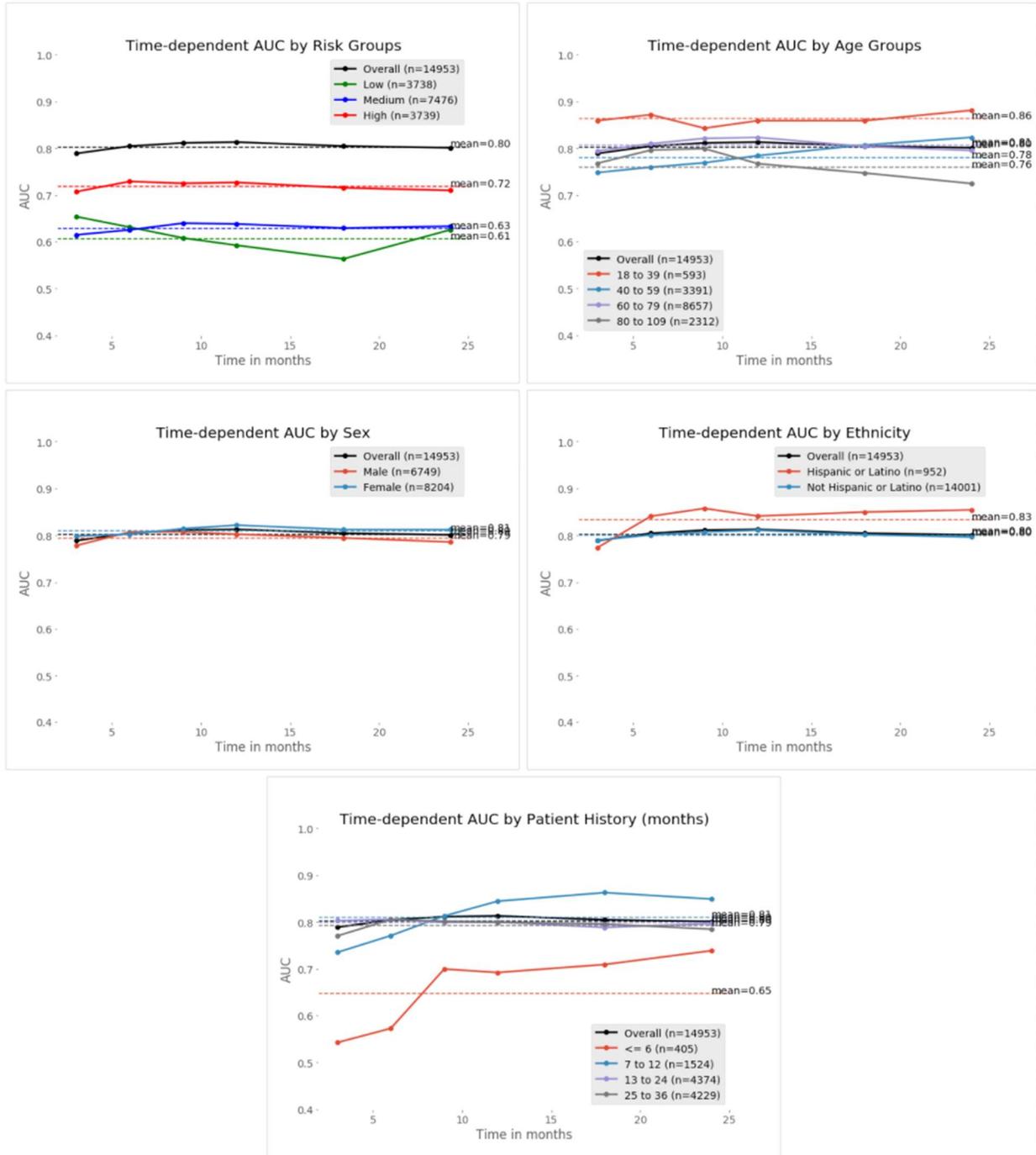

Time-dependent Specificity

The following charts show how the specificity across time varies among subgroups of risk, age, sex, ethnicity and patient history.

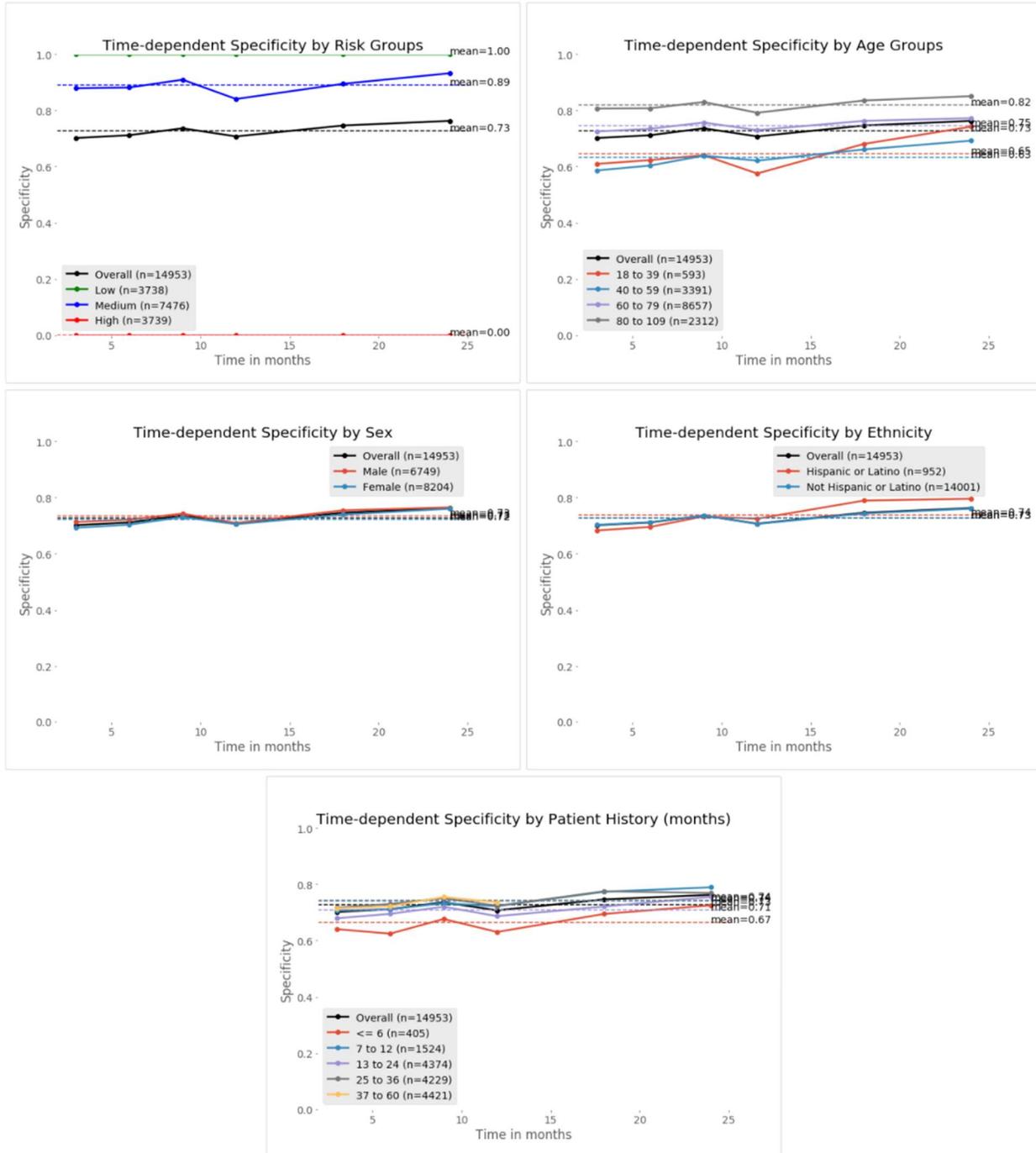

Time-dependent Sensitivity

The following charts show how the sensitivity across time varies among subgroups of risk, age, sex, ethnicity and patient history.

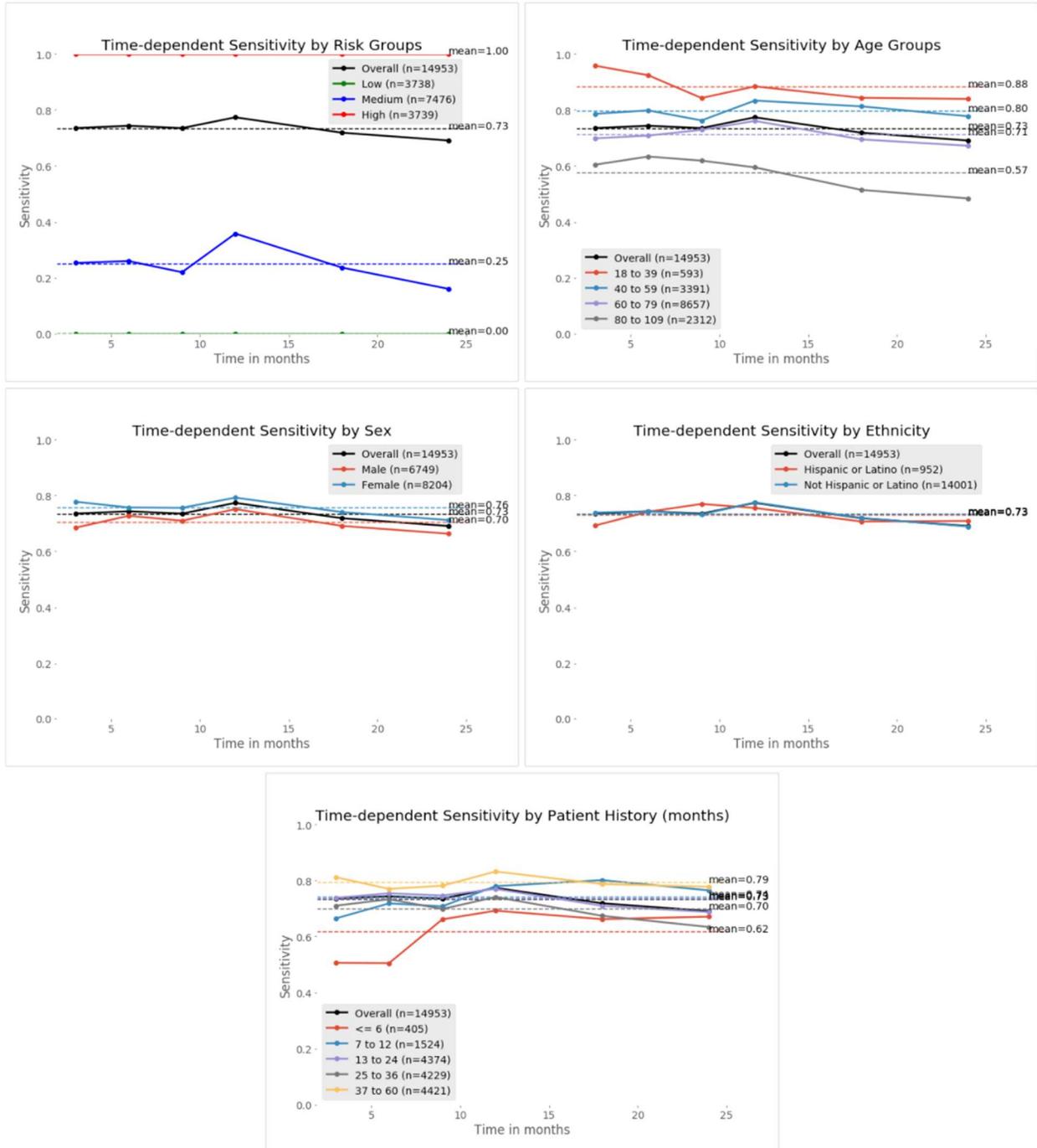

## Model Explanation (DeepSurv)
### Global
The following plots show the SHAP values of each instance in the training set for each future time (3, 6, 9, 12, 18 and 24 months). The features are sorted by the total magnitude of the SHAP values over all instances and the distribution of the effect that each feature has on the model's output can be observed.

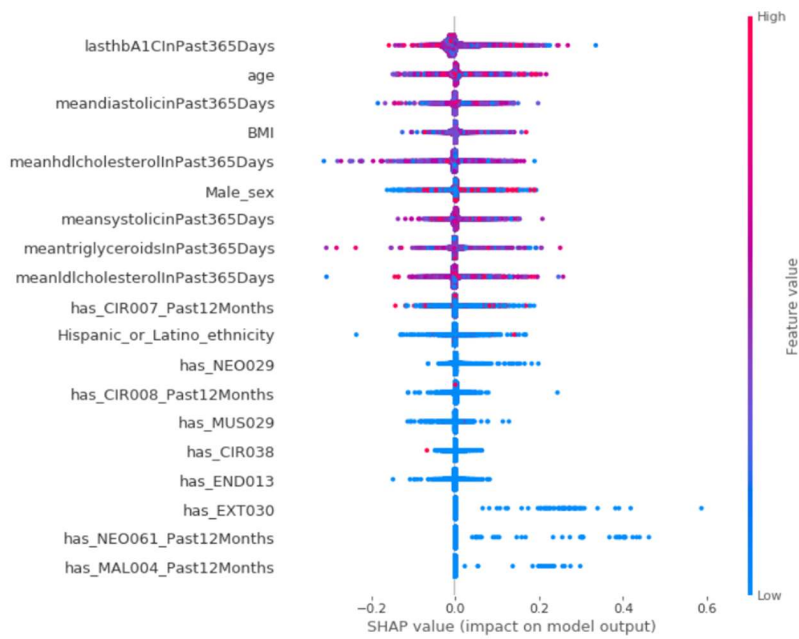

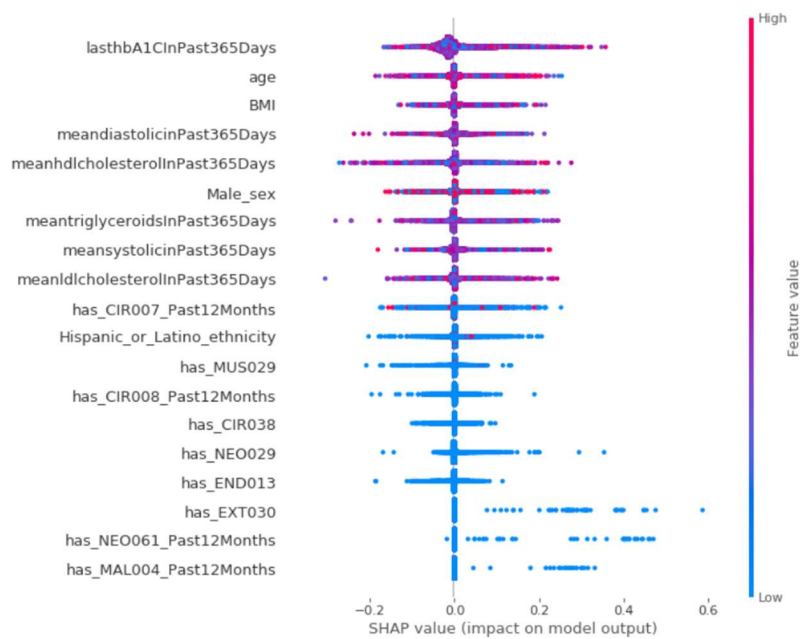

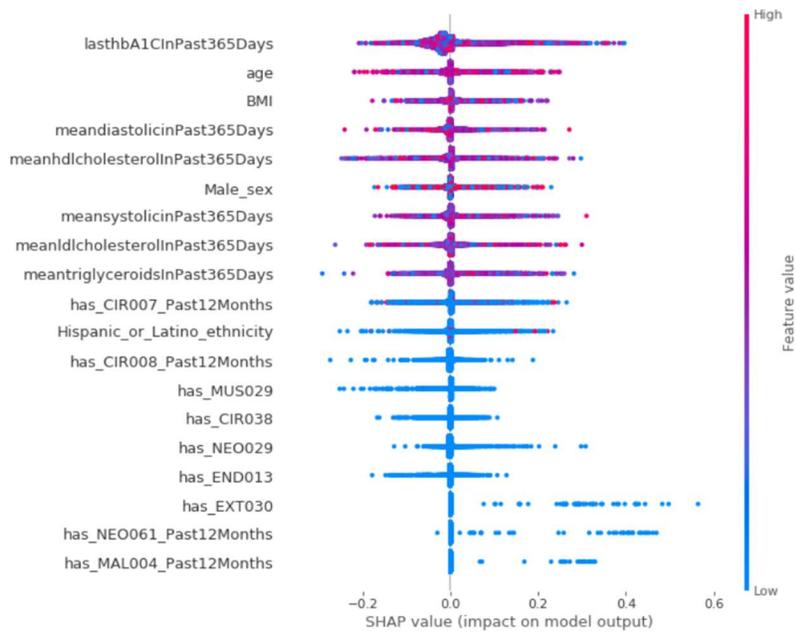

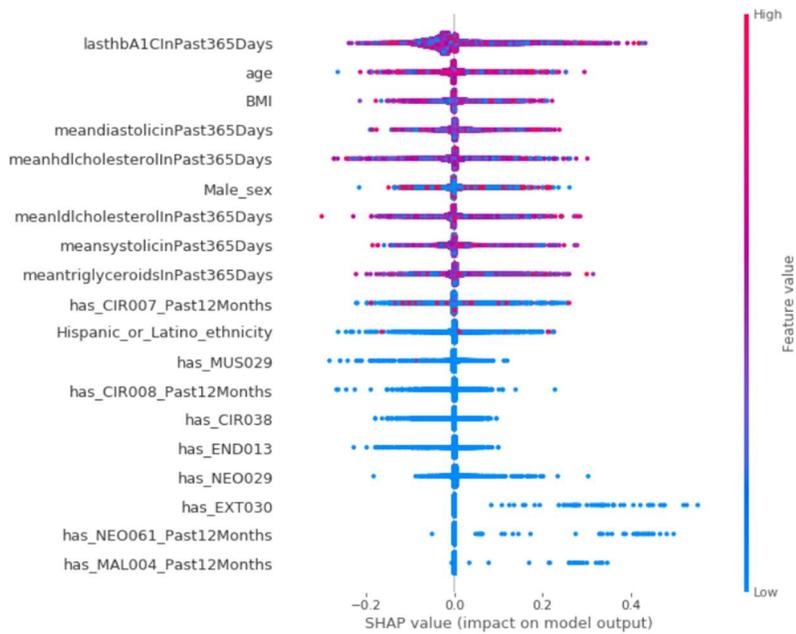

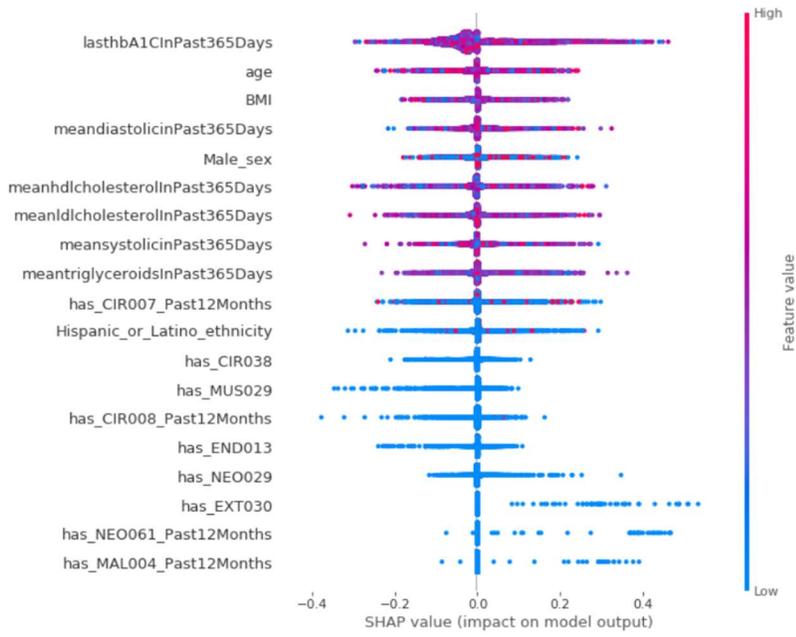

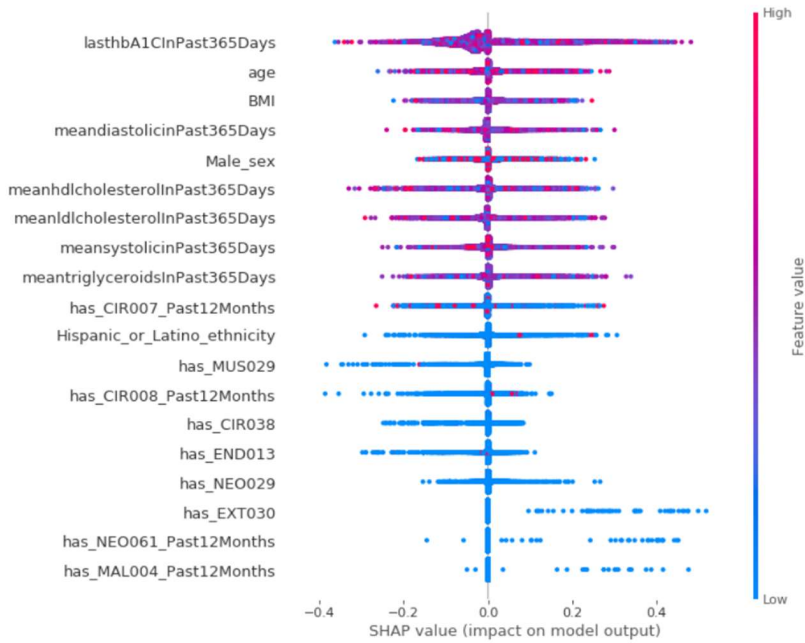

Local

SHAP values were also generated to explain the predictions of individual examples for each future time (3, 6, 9, 12, 18 and 24 months). A total of 3 examples were selected by sampling of risk scores at the 5[th], 50[th] and 95[th] percentile to represent instances at low, medium and high risks respectively.

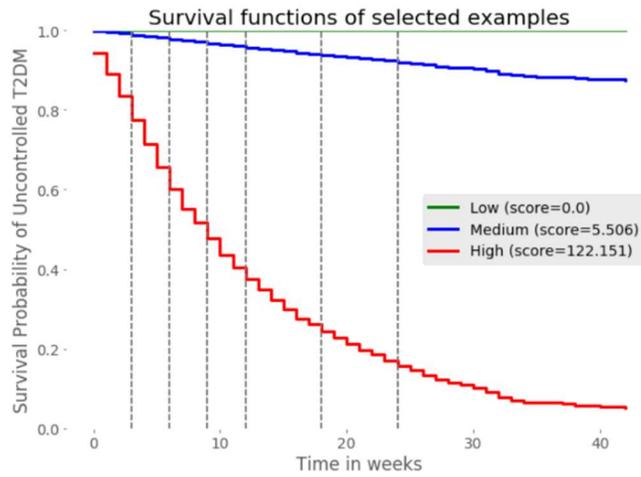

**Low Risk**
Risk Score: 0.0

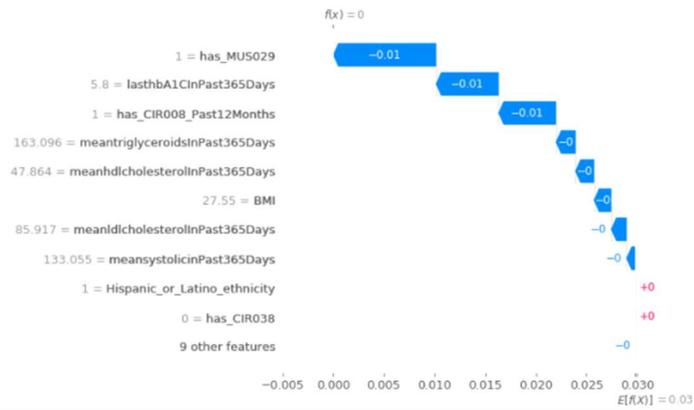

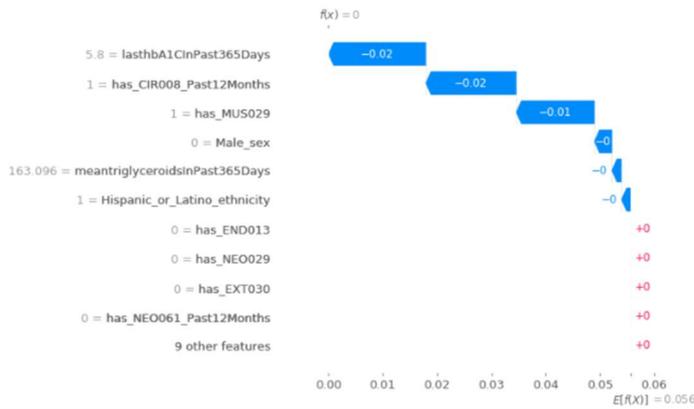

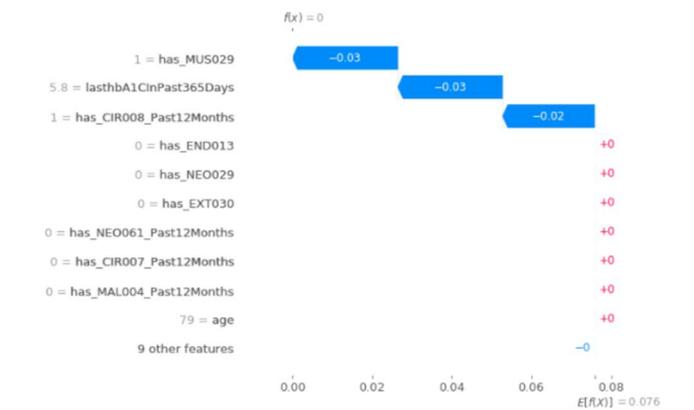

**Low Risk**
Risk Score: 0.0

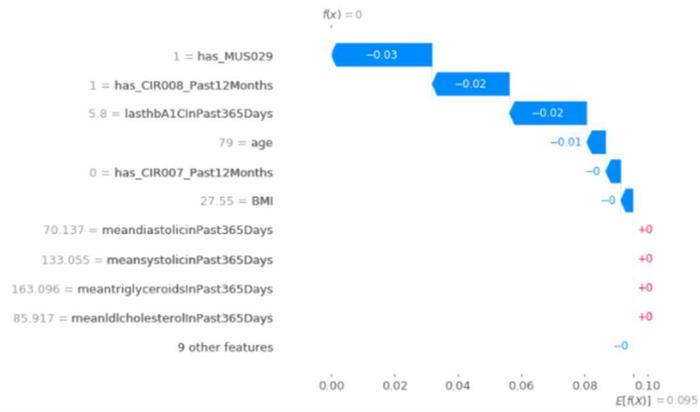

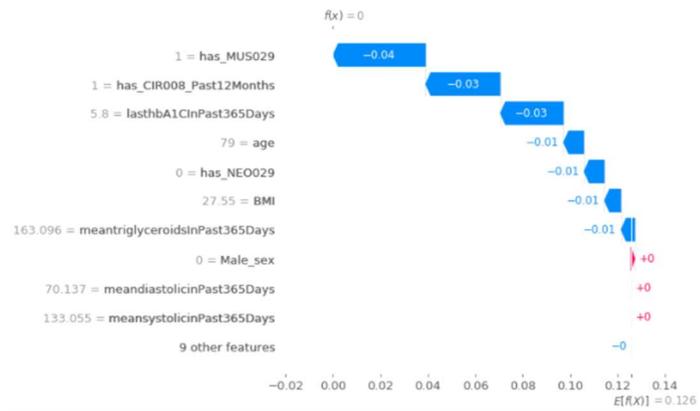

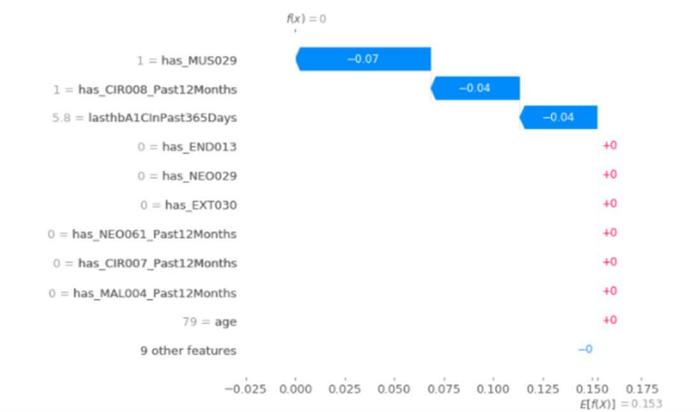

**Medium Risk**
Risk Score: 5.506

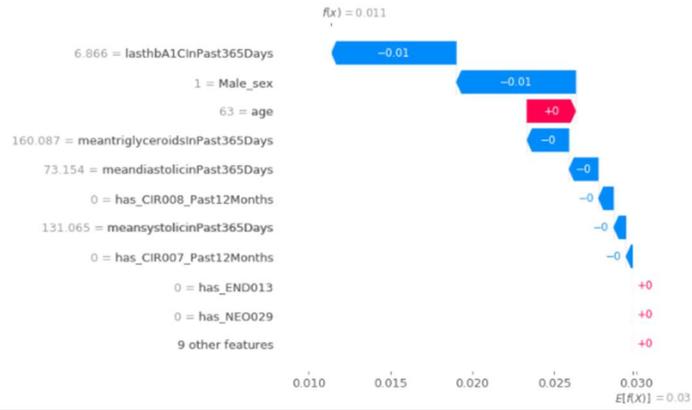

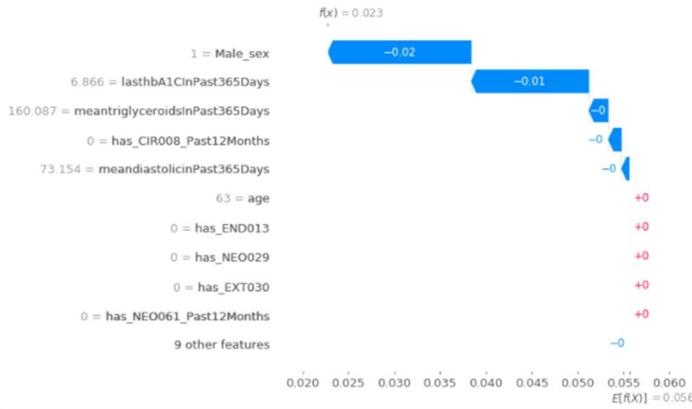

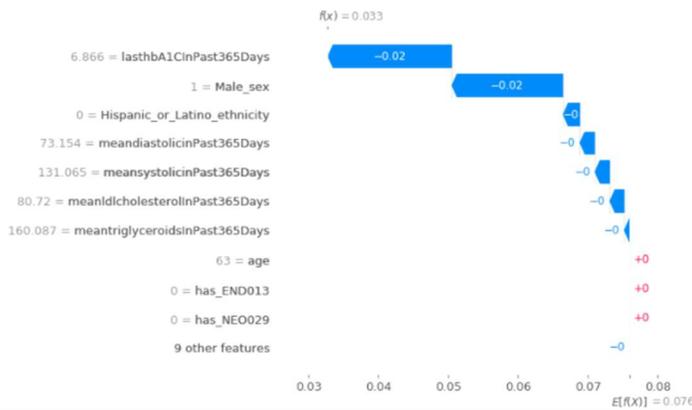

**Medium Risk**
Risk Score: 5.506

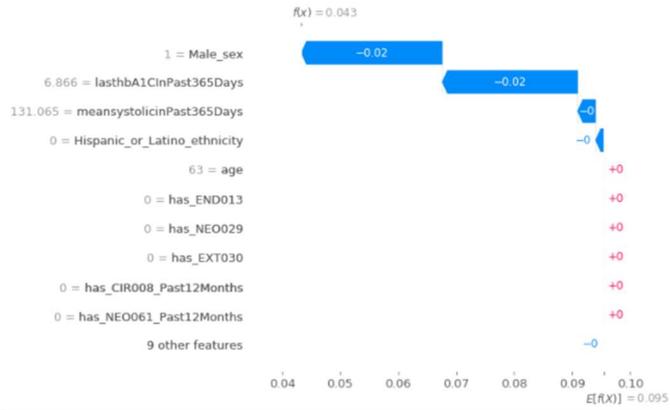

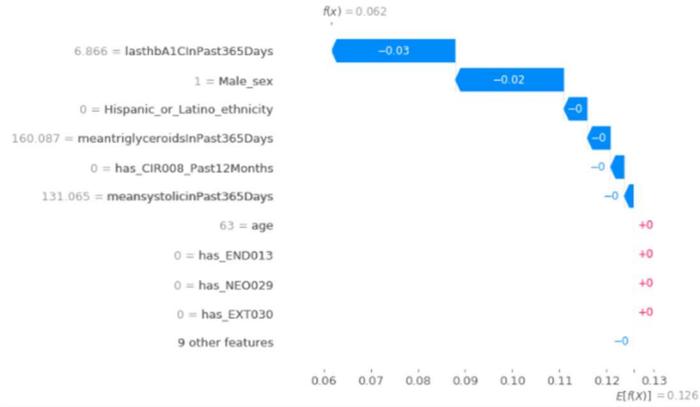

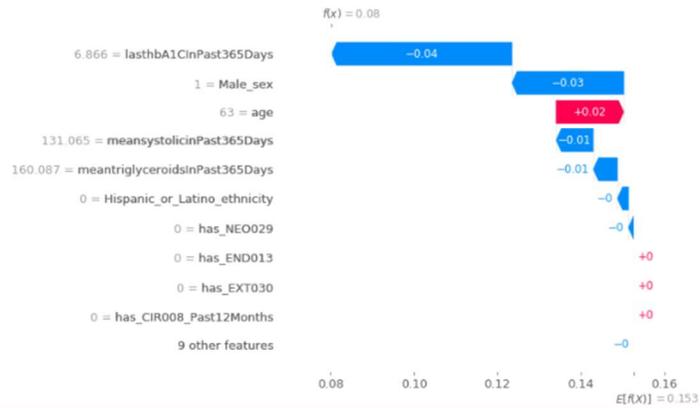

**High Risk**
Risk Score: 122.151

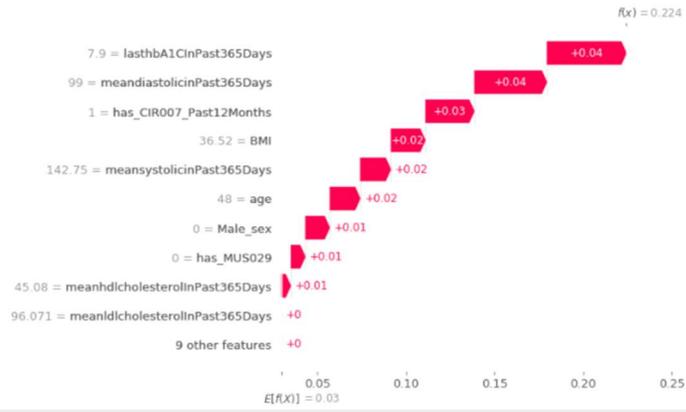

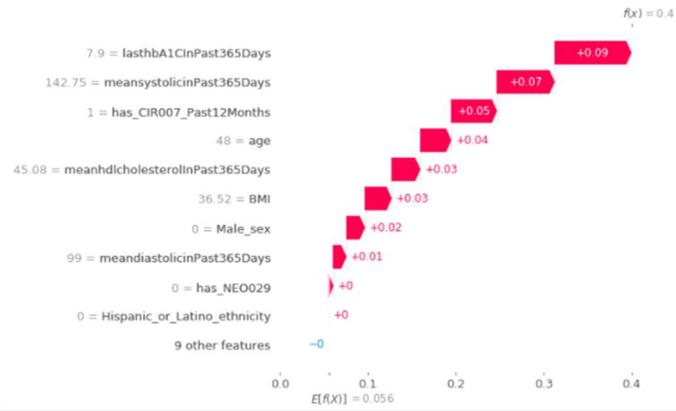

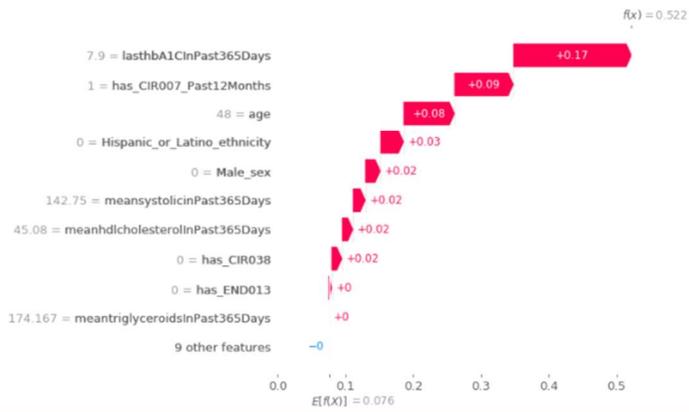

**High Risk**
Risk Score: 122.151

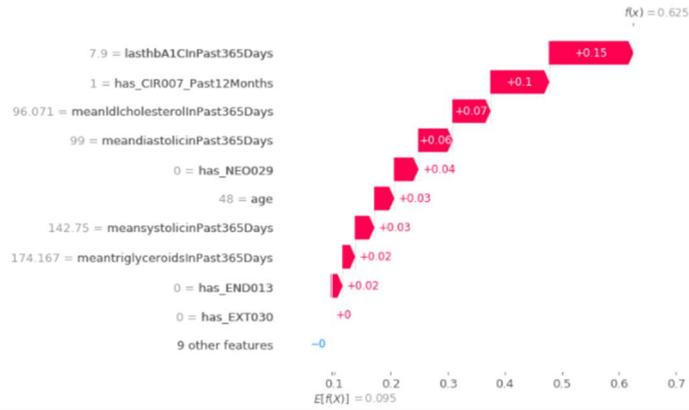

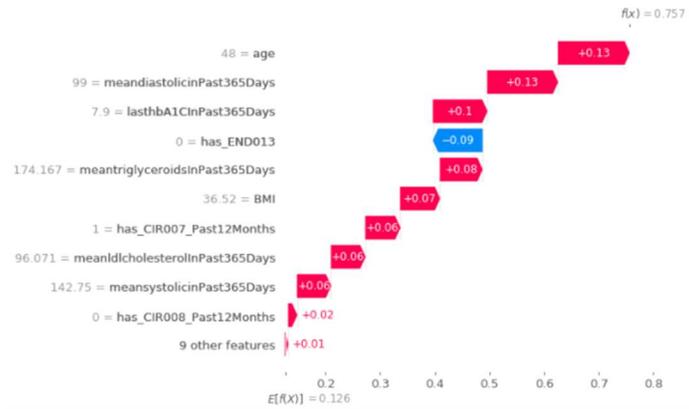

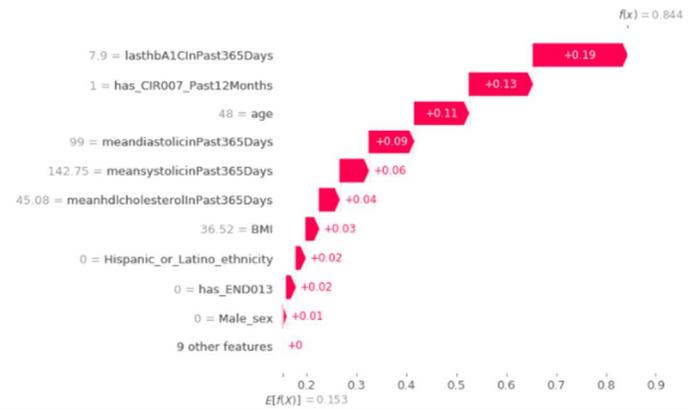

## Pre-DM to Uncontrolled DM Prediction

| Item | Specification |
|---|---|
| **Business Goal** | Enable care managers to identify the patients who are at risk of developing uncontrolled DM |
| **Usage Setting** | Outpatient |
| **ML Task** | Predict risk and/or time from DM to uncontrolled DM |
| **ML Class** | Survival |
| **Instances for Prediction** | Encounters |
| **Labels for Instances** | Binary indicator and time to event or censoring for uncontrolled DM |
| **Cohort Criteria** | • 2016-01-01 $\leq$ encounter date $\leq$ 2020-06-30 (available Epic data, excluding outliers)<br>• Encounter date is not within the first 90 days of when the patient entered the data set, to adjust for left-censoring<br>• 18 $\leq$ age $\leq$ 110 (adults without outliers)<br>• No T1DM diagnosis<br>• Not pregnant<br>• No "Do Not Resuscitate" diagnosis<br><br>• Pre-DM event before encounter date<br>• No DM event before encounter date<br>• No DM event up to 6 days after encounter date<br>• No uncontrolled DM event before encounter date<br>• No uncontrolled DM event up to 6 days after encounter date (encounters where lab results confirm event within the week) |
| **Input Features** | • Demographic<br>• Diagnosis, except:<br>    ○ hasDiabetes*<br>    ○ has_END002* (Diabetes mellitus without complication)<br>    ○ has_END003* (Diabetes mellitus with complication)<br>    ○ has_END004* (Diabetes mellitus, Type 1)<br>    ○ has_END005* (Diabetes mellitus, Type 2)<br>    ○ has_END006* (Diabetes mellitus, due to underlying condition, drug or chemical induced, or other specified type)<br>• Labs<br>• Utilization<br>• Vitals<br><br>See Appendix E for details. |
| **Evaluation Metrics** | • Concordance Index<br>• Integrated Brier Score |

### Data

The following charts summarize the key characteristics of the data after applying the cohort criteria stated above, along with selected features (see Model Signature below).

| Category | Variable | count | mean | stddev | min | 25% | 50% | 75% | max |
|---|---|---|---|---|---|---|---|---|---|
| Demographic | AgeBucket_18_to_39 | 258142.0 | 0.069903 | 0.254985 | 0.00 | 0.000 | 0.000 | 0.000 | 1.00 |
| | AgeBucket_40_to_59 | 258142.0 | 0.248332 | 0.432046 | 0.00 | 0.000 | 0.000 | 0.000 | 1.00 |
| | AgeBucket_60_to_79 | 258142.0 | 0.543267 | 0.498125 | 0.00 | 0.000 | 1.000 | 1.000 | 1.00 |
| | AgeBucket_80_to_109 | 258142.0 | 0.138497 | 0.345422 | 0.00 | 0.000 | 0.000 | 0.000 | 1.00 |
| | Sex_Female | 258142.0 | 0.627515 | 0.483467 | 0.00 | 0.000 | 1.000 | 1.000 | 1.00 |
| | Sex_Male | 258142.0 | 0.372485 | 0.483467 | 0.00 | 0.000 | 0.000 | 1.000 | 1.00 |
| | Ethnicity_Hispanic_or_Latino | 258142.0 | 0.054687 | 0.227369 | 0.00 | 0.000 | 0.000 | 0.000 | 1.00 |
| | Ethnicity_Not_Hispanic_or_Latino | 258142.0 | 0.945313 | 0.227369 | 0.00 | 1.000 | 1.000 | 1.000 | 1.00 |
| Encounter | EncounterType_Emergency | 258142.0 | 0.183232 | 0.386858 | 0.00 | 0.000 | 0.000 | 0.000 | 1.00 |
| | EncounterType_Inpatient | 258142.0 | 0.079848 | 0.271058 | 0.00 | 0.000 | 0.000 | 0.000 | 1.00 |
| | EncounterType_Outpatient | 258142.0 | 0.736920 | 0.440306 | 0.00 | 0.000 | 1.000 | 1.000 | 1.00 |
| Label | Time | 258142.0 | 16.275395 | 11.351408 | 0.00 | 7.000 | 14.000 | 25.000 | 50.00 |
| | Event | 258142.0 | 0.023673 | 0.152029 | 0.00 | 0.000 | 0.000 | 0.000 | 1.00 |
| Feature | age | 258142.0 | 64.481131 | 14.647176 | 18.00 | 56.000 | 66.000 | 75.000 | 107.00 |
| | Male_sex | 258142.0 | 0.372485 | 0.483467 | 0.00 | 0.000 | 0.000 | 1.000 | 1.00 |
| | Hispanic_or_Latino_ethnicity | 258142.0 | 0.054687 | 0.227369 | 0.00 | 0.000 | 0.000 | 0.000 | 1.00 |
| | lasthbA1CInPast365Days | 258142.0 | 6.034896 | 0.491028 | 2.20 | 5.800 | 6.022 | 6.103 | 8.90 |
| | meandiastolicinPast365Days | 258142.0 | 74.510641 | 7.465089 | 43.63 | 71.000 | 74.678 | 77.830 | 107.67 |
| | meansystolicinPast365Days | 258142.0 | 130.507721 | 11.237634 | 81.00 | 125.910 | 130.162 | 134.500 | 180.00 |
| | BMI | 258142.0 | 32.052144 | 6.993803 | 2.71 | 28.009 | 31.894 | 35.243 | 326.11 |
| | meantriglyceroidsInPast365Days | 258142.0 | 141.515197 | 48.161881 | 11.00 | 119.839 | 142.732 | 147.721 | 468.50 |
| | meanldlcholesterolInPast365Days | 258142.0 | 95.182333 | 23.631750 | 2.80 | 84.968 | 97.608 | 107.000 | 201.00 |
| | meanhdlcholesterolInPast365Days | 258142.0 | 47.882752 | 10.369947 | 6.00 | 42.348 | 47.000 | 52.436 | 98.00 |
| | has_BLD005_Past12Months | 258142.0 | 0.001189 | 0.034465 | 0.00 | 0.000 | 0.000 | 0.000 | 1.00 |
| | has_SKN005_Past12Months | 258142.0 | 0.004970 | 0.070324 | 0.00 | 0.000 | 0.000 | 0.000 | 1.00 |
| | has_INJ014_Past12Months | 258142.0 | 0.000686 | 0.026176 | 0.00 | 0.000 | 0.000 | 0.000 | 1.00 |
| | has_CIR007_Past12Months | 258142.0 | 0.165409 | 0.371550 | 0.00 | 0.000 | 0.000 | 0.000 | 1.00 |
| | has_NEO022_Past12Months | 258142.0 | 0.026927 | 0.161871 | 0.00 | 0.000 | 0.000 | 0.000 | 1.00 |
| | has_CIR008_Past12Months | 258142.0 | 0.009239 | 0.095675 | 0.00 | 0.000 | 0.000 | 0.000 | 1.00 |
| | has_NEO029_Past12Months | 258142.0 | 0.004269 | 0.065198 | 0.00 | 0.000 | 0.000 | 0.000 | 1.00 |
| | has_BLD005 | 258142.0 | 0.002119 | 0.045984 | 0.00 | 0.000 | 0.000 | 0.000 | 1.00 |
| | has_INJ007 | 258142.0 | 0.004846 | 0.069446 | 0.00 | 0.000 | 0.000 | 0.000 | 1.00 |
| | has_INJ054 | 258142.0 | 0.001522 | 0.038989 | 0.00 | 0.000 | 0.000 | 0.000 | 1.00 |
| | has_GEN018 | 258142.0 | 0.014829 | 0.120869 | 0.00 | 0.000 | 0.000 | 0.000 | 1.00 |
| | has_NEO044 | 258142.0 | 0.001120 | 0.033441 | 0.00 | 0.000 | 0.000 | 0.000 | 1.00 |
| | has_INJ035 | 258142.0 | 0.007031 | 0.083556 | 0.00 | 0.000 | 0.000 | 0.000 | 1.00 |
| | has_INJ049 | 258142.0 | 0.001941 | 0.044012 | 0.00 | 0.000 | 0.000 | 0.000 | 1.00 |
| | has_NEO012 | 258142.0 | 0.005745 | 0.075577 | 0.00 | 0.000 | 0.000 | 0.000 | 1.00 |

(Percentages for binary variables can be read from the "mean" column.)

## Encounters and Patients

| | Examples | Encounters | Patients |
|---|---|---|---|
| 1 | 258142 | 258142 | 45851 |

## Train and Test Sets

| | Set | No Event | No Event % | Event | Event % | Total | Total % |
|---|---|---|---|---|---|---|---|
| 1 | Train | 176915 | 70.2 | 4282 | 70.1 | 181197 | 70.2 |
| 2 | Test | 75116 | 29.8 | 1829 | 29.9 | 76945 | 29.8 |
| 3 | Total | 252031 | 100 | 6111 | 100 | 258142 | 100 |

### Encounters with Event (1) or Censoring (0)
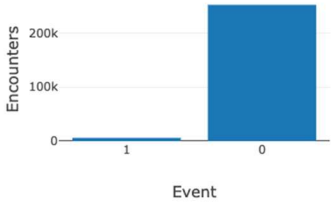

### Patients with Event (1) or Censoring (0)
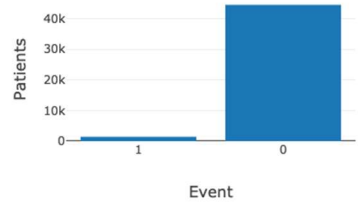

### Encounters by Age and Sex
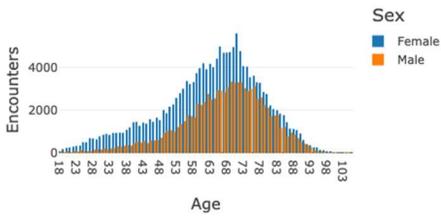

### Patients by Age and Sex
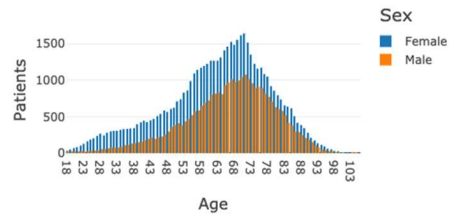

### Encounters by Ethnicity
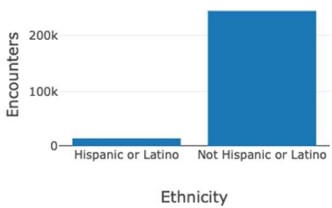

### Patients by Ethnicity
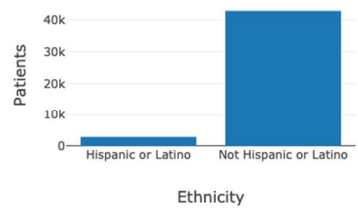

### Encounters by Encounter Type
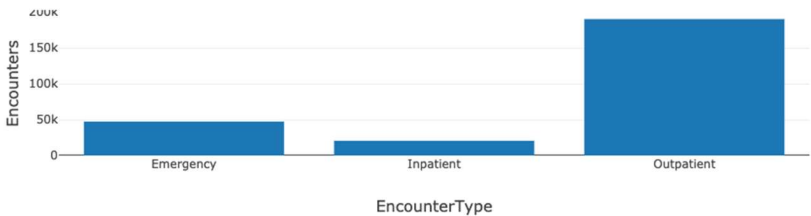

### Encounters by Time to Event (1) or Cen…
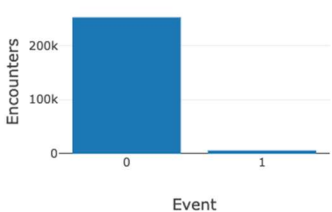

### Kaplan-Meier Estimate of Survival Functi…
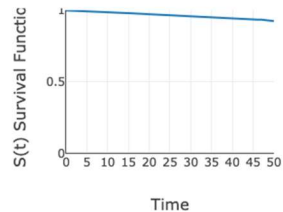

## Model Signature

The model signature has 25 features, comprising of 12 mandatory features and 13 other selected features. These are the selected features, in rank order (the last feature to be eliminated is ranked 1):

1. has_BLD005_Past12Months (Sickle cell trait/anemia)
2. has_INJ035 (Complication of internal orthopedic device or implant, initial encounter)
3. has_INJ014_Past12Months (Amputation of a limb, initial encounter)
4. has_INJ049 (Open wounds to limbs, subsequent encounter)
5. has_NEO022_Past12Months (Respiratory cancers)
6. has_INJ054 (Superficial injury; contusion, subsequent encounter)
7. has_NEO044 (Urinary system cancers - ureter and renal pelvis)
8. lasthbA1CInPast365Days
9. has_SKN005_Past12Months (Contact dermatitis)
10. has_NEO029_Past12Months (Breast cancer - ductal carcinoma in situ (DCIS))
11. has_BLD005 (Sickle cell trait/anemia)
12. has_NEO012 (Gastrointestinal cancers - esophagus)
13. has_INJ007 (Dislocations, initial encounter)
14. has_GEN018 (Inflammatory diseases of female pelvic organs)
15. Male_sex
16. has_CIR008_Past12Months (Hypertension with complications and secondary hypertension)
17. Hispanic_or_Latino_ethnicity
18. age
19. BMI
20. meanhdlcholesterolInPast365Days
21. has_CIR007_Past12Months (Essential hypertension)
22. meansystolicinPast365Days
23. meantriglyceroidsInPast365Days
24. meandiastolicinPast365Days
25. meanldlcholesterolInPast365Days

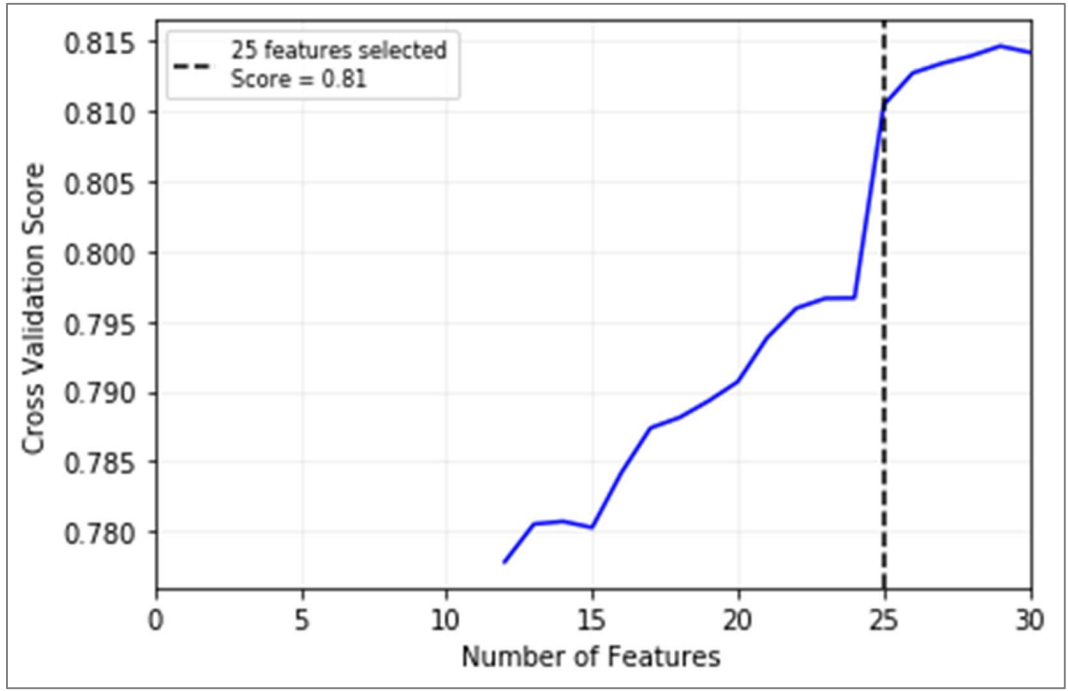

## Model Performance

The following table and chart summarize the performance of all candidate models on the test set for this prediction task in terms of the Concordance Index and the Integrated Brier Score.

| Model | No. of Parameter Combinations Successfully Tested | Concordance Index | Integrated Brier Score |
|---|---|---|---|
| **CoxPH** | 83 | 0.79 | **0.03** |
| **DeepSurv** | 81 | **0.81** | 0.03 |
| **RSF** | 1 | 0.75 | **0.03** |
| **CSF** | 5 | 0.75 | **0.03** |
| **EST** | 1 | 0.77 | **0.03** |

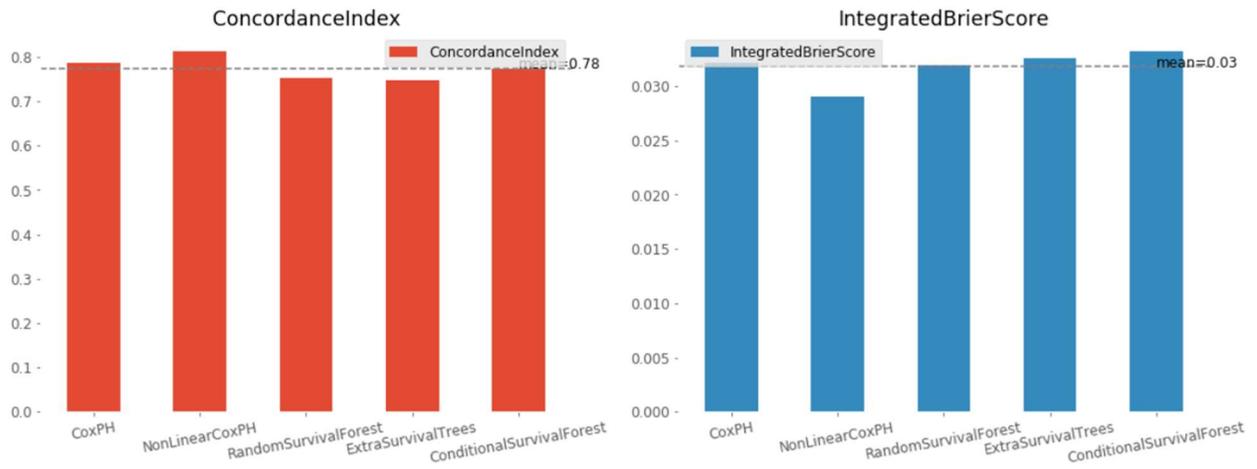

The following chart shows how the average survival function curves of the candidate models compare to the KM survival curve, the more similar their curves are to the KM survival curve the better.

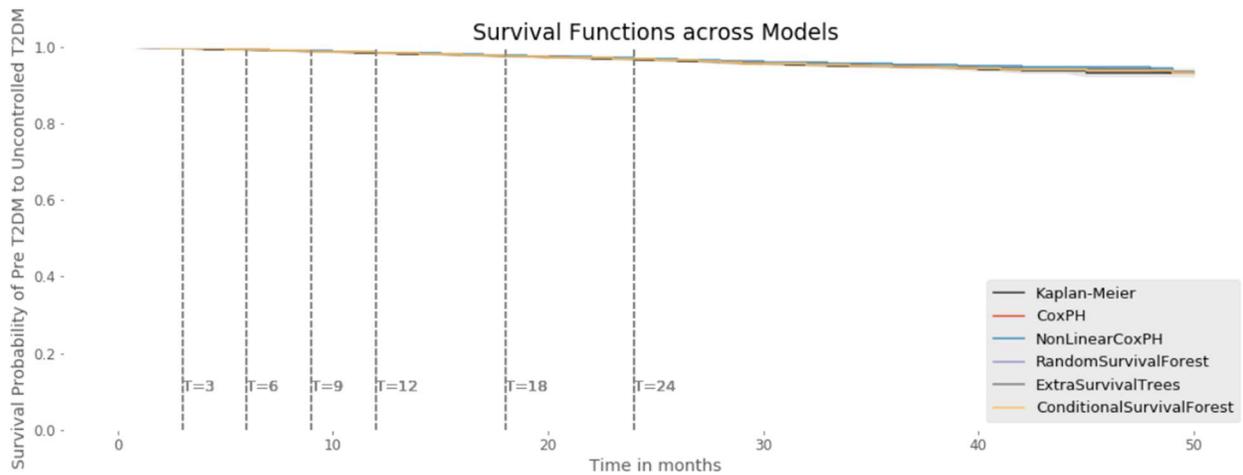

This chart shows the change in the Brier Score over time for all candidate models, the closer the scores are to 0 the better.

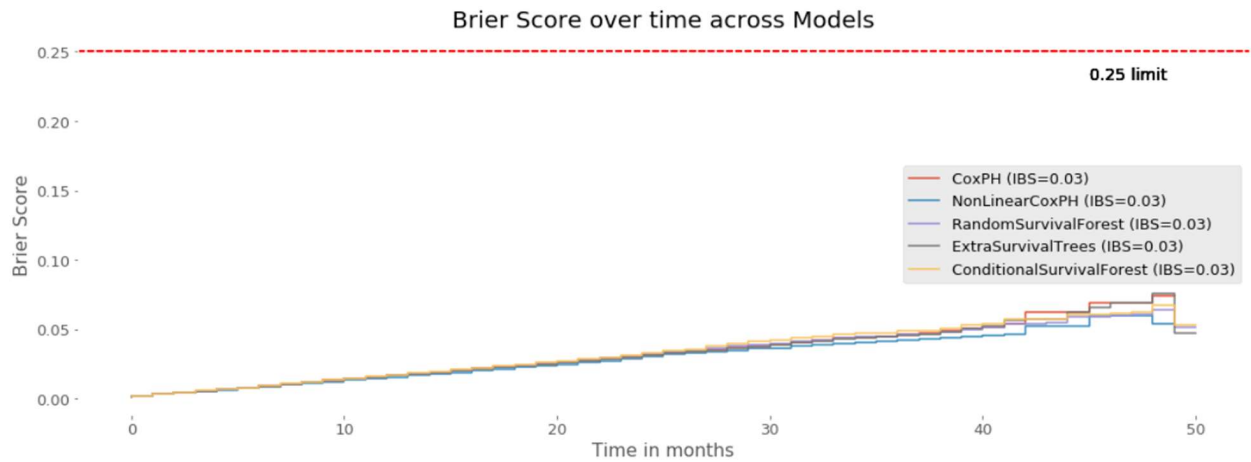

## Model Evaluation of Selected Model (DeepSurv)

Overall

This chart shows the change in the Brier Score over time for the selected model, the closer the scores are to 0 the better.

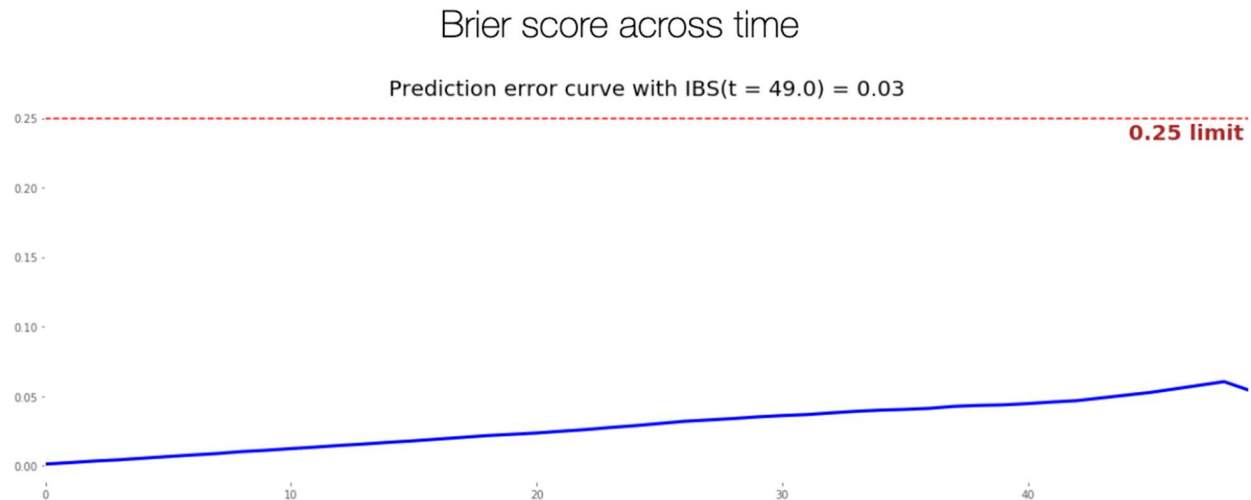

The following chart shows the actual vs. predicted density functions, i.e. number of instances that get the disease / complication at each time point and the RMSE, Median Absolute Error and Mean Absolute Error across the time points.

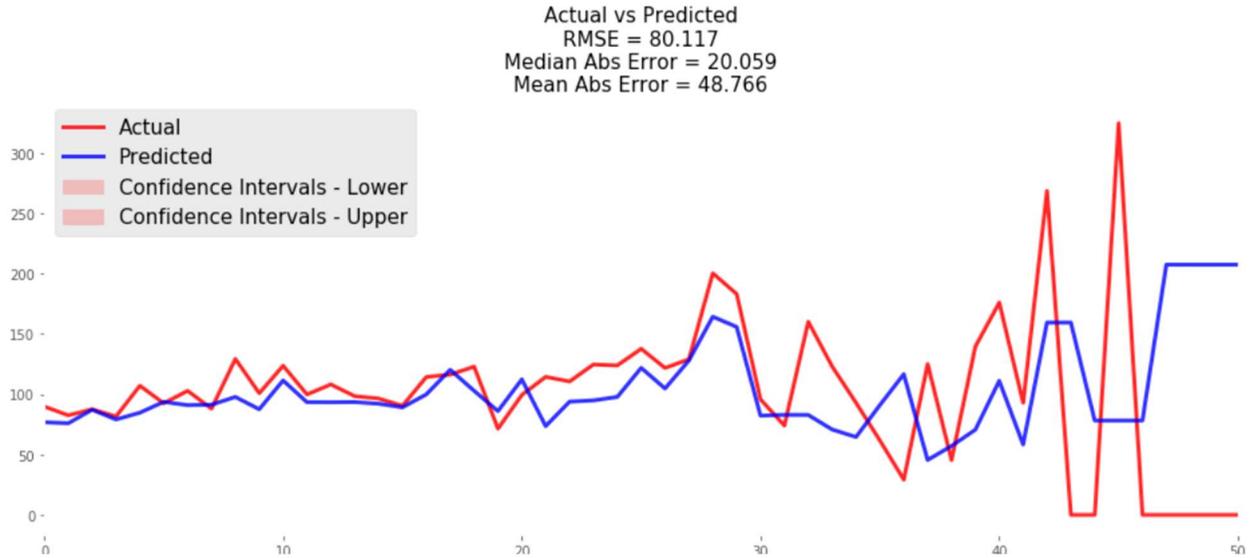

The following chart shows the actual vs. predicted survival functions, i.e. the number of instances that have not had the disease / complication by each time point and the RMSE, Median Absolute Error and Mean Absolute Error across the time points.

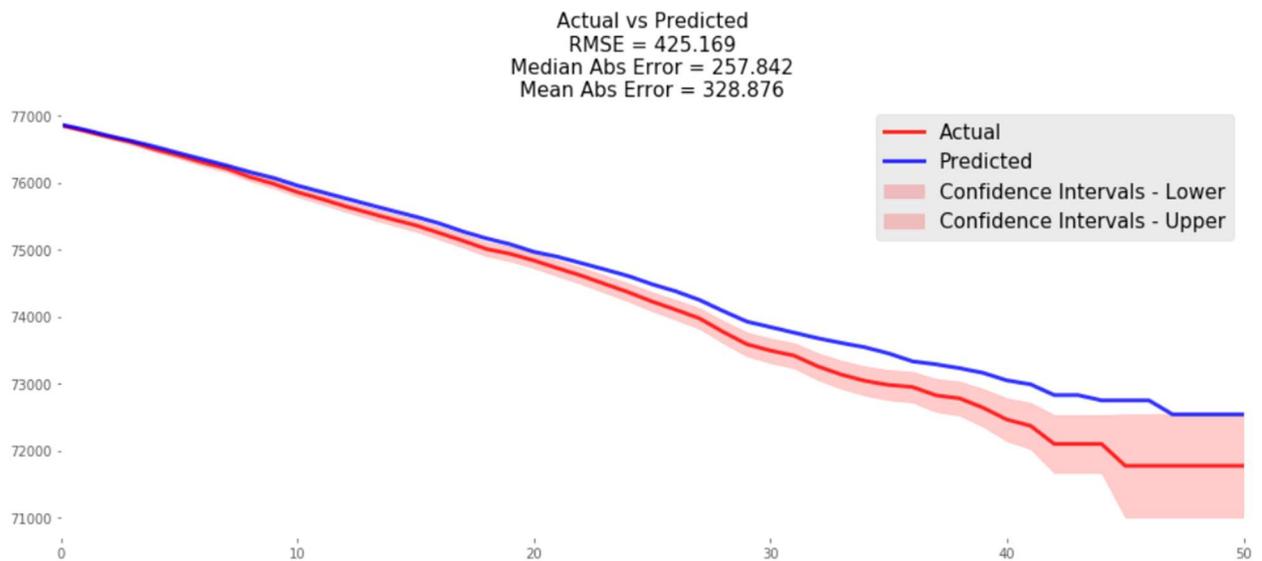

Risk Stratification

The low, medium and high risk groups are defined as examples with predicted risk scores belonging to the first quartile, second to third quartiles, and fourth quartile respectively.

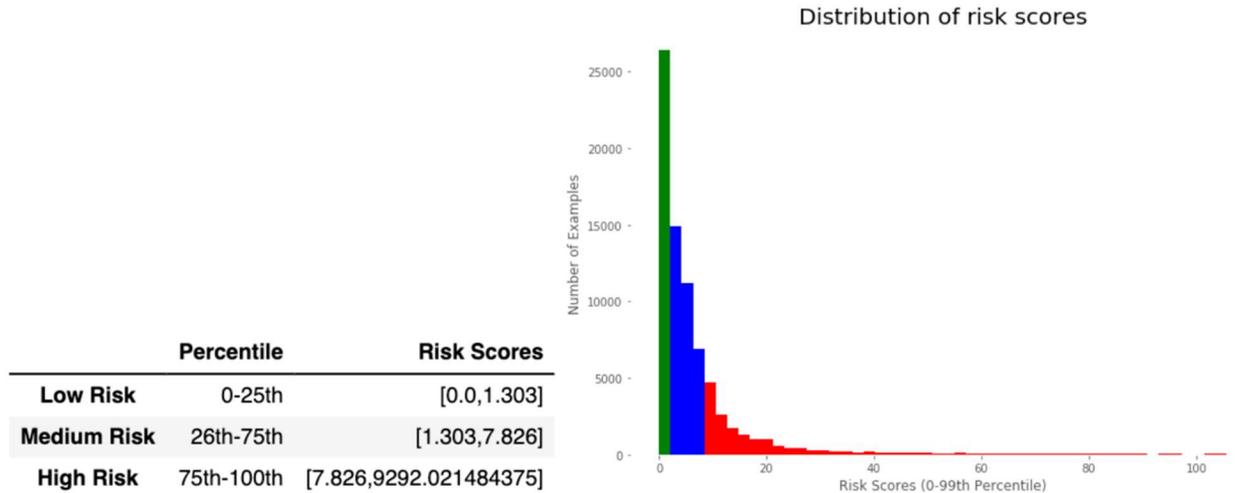

|  | Percentile | Risk Scores |
|---|---|---|
| **Low Risk** | 0-25th | [0.0,1.303] |
| **Medium Risk** | 26th-75th | [1.303,7.826] |
| **High Risk** | 75th-100th | [7.826,9292.021484375] |

Summary Metrics across Subgroups

The table below displays the summary metrics across subgroups of risk, age, sex, ethnicity and patient history.

| Category | Subgroup | Cohort Size | Concordance Index | Brier Score | Mean AUC | Mean Specificity | Mean Sensitivity | S(t), t=3 | S(t), t=6 | S(t), t=9 | S(t), t=12 | S(t), t=18 | S(t), t=24 |
|---|---|---|---|---|---|---|---|---|---|---|---|---|---|
| NaN | Overall | 76945.00 | 0.81 | 0.03 | 0.84 | 0.83 | 0.68 | 1.00 | 0.99 | 0.99 | 0.98 | 0.97 | 0.97 |
| Risk | Low | 19236.00 | 0.61 | 0.00 | 0.62 | 1.00 | 0.00 | 1.00 | 1.00 | 1.00 | 1.00 | 1.00 | 1.00 |
| Risk | Medium | 38472.00 | 0.60 | 0.02 | 0.64 | 1.00 | 0.00 | 1.00 | 1.00 | 0.99 | 0.99 | 0.99 | 0.98 |
| Risk | High | 19237.00 | 0.74 | 0.07 | 0.76 | 0.33 | 0.88 | 0.99 | 0.98 | 0.97 | 0.95 | 0.93 | 0.91 |
| Age Bucket | 18 to 39 | 5468.00 | 0.86 | 0.02 | 0.83 | 0.90 | 0.62 | 1.00 | 0.99 | 0.99 | 0.99 | 0.98 | 0.98 |
| Age Bucket | 40 to 59 | 19051.00 | 0.80 | 0.04 | 0.84 | 0.80 | 0.71 | 0.99 | 0.99 | 0.99 | 0.98 | 0.97 | 0.96 |
| Age Bucket | 60 to 79 | 41876.00 | 0.80 | 0.03 | 0.84 | 0.82 | 0.68 | 1.00 | 0.99 | 0.99 | 0.98 | 0.98 | 0.97 |
| Age Bucket | 80 to 109 | 10550.00 | 0.81 | 0.02 | 0.83 | 0.92 | 0.54 | 1.00 | 1.00 | 0.99 | 0.99 | 0.99 | 0.98 |
| Sex | Male | 28594.00 | 0.81 | 0.04 | 0.81 | 0.72 | 0.72 | 0.99 | 0.99 | 0.98 | 0.98 | 0.97 | 0.96 |
| Sex | Female | 48351.00 | 0.81 | 0.02 | 0.86 | 0.89 | 0.64 | 1.00 | 0.99 | 0.99 | 0.99 | 0.98 | 0.98 |
| Ethnicity | Hispanic or Latino | 4192.00 | 0.86 | 0.03 | 0.87 | 0.85 | 0.71 | 0.99 | 0.99 | 0.98 | 0.98 | 0.97 | 0.96 |
| Ethnicity | Not Hispanic or Latino | 72753.00 | 0.81 | 0.03 | 0.84 | 0.83 | 0.67 | 1.00 | 0.99 | 0.99 | 0.99 | 0.98 | 0.97 |
| History Bucket | <= 6 | 2437.00 | 0.90 | 0.02 | 0.88 | 0.81 | 0.77 | 1.00 | 0.99 | 0.99 | 0.98 | 0.98 | 0.97 |
| History Bucket | 7 to 12 | 6688.00 | 0.81 | 0.03 | 0.83 | 0.81 | 0.67 | 1.00 | 0.99 | 0.99 | 0.98 | 0.97 | 0.96 |
| History Bucket | 13 to 24 | 20985.00 | 0.81 | 0.03 | 0.83 | 0.83 | 0.65 | 1.00 | 0.99 | 0.99 | 0.98 | 0.97 | 0.97 |
| History Bucket | 25 to 36 | 22329.00 | 0.82 | 0.02 | 0.86 | 0.85 | 0.68 | 1.00 | 0.99 | 0.99 | 0.99 | 0.98 | 0.97 |
| History Bucket | 37 to 60 | 24506.00 | 0.85 | 0.01 | nan | nan | 0.70 | 1.00 | 0.99 | 0.99 | 0.99 | 0.98 | 0.97 |

Concordance Index & Integrated Brier Score

The following charts show how the Concordance Index and Integrated Brier Score varies among subgroups of risk, age, sex, ethnicity and patient history.

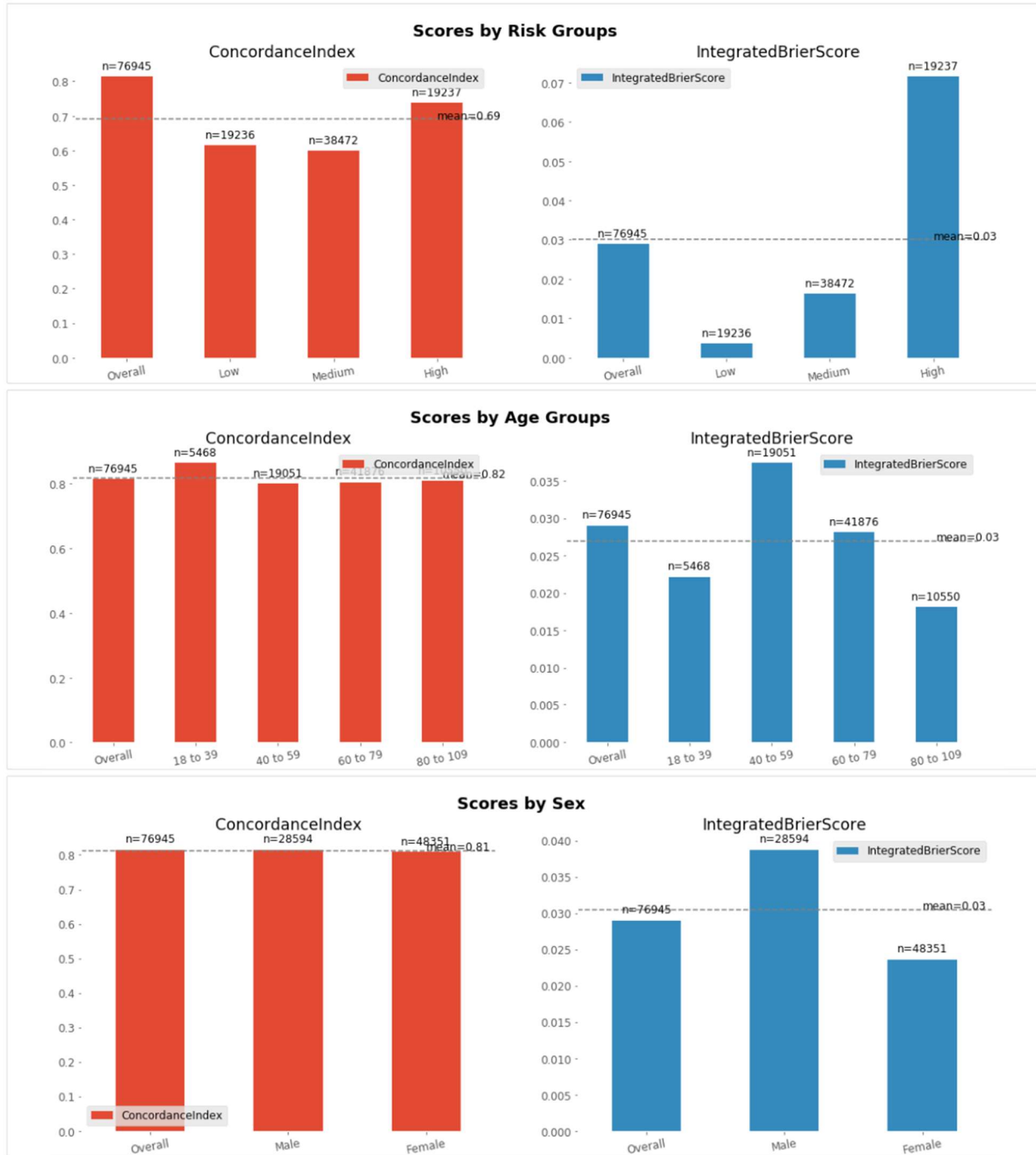

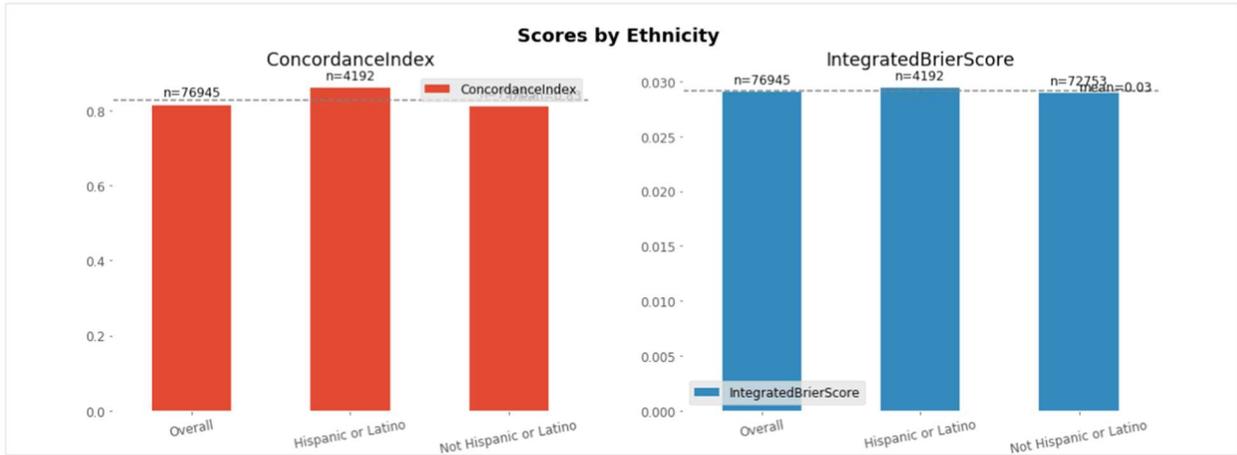
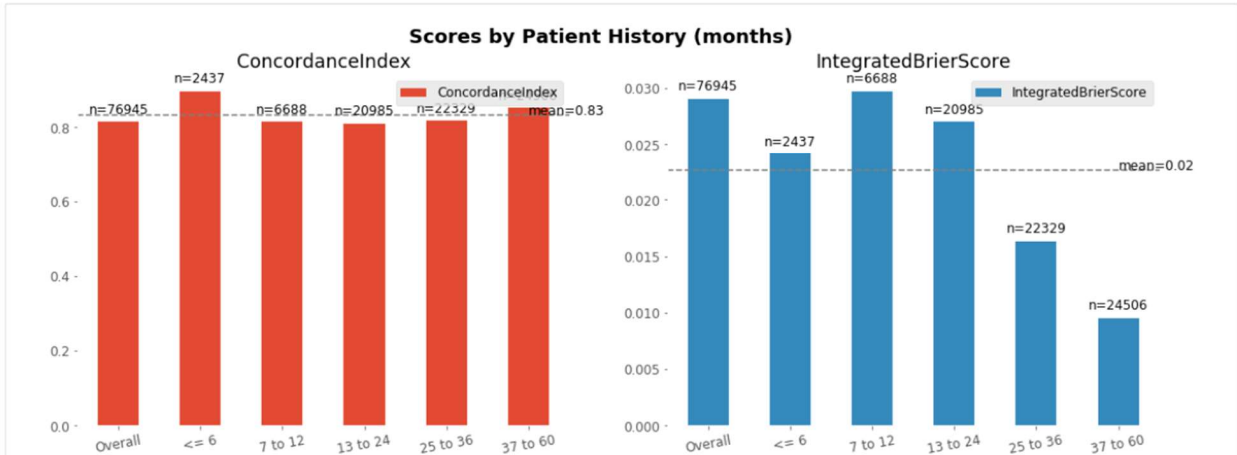

Average Survival Function Curves

The following charts show how the average survival function curve varies among subgroups of risk, age, sex, ethnicity and patient history.

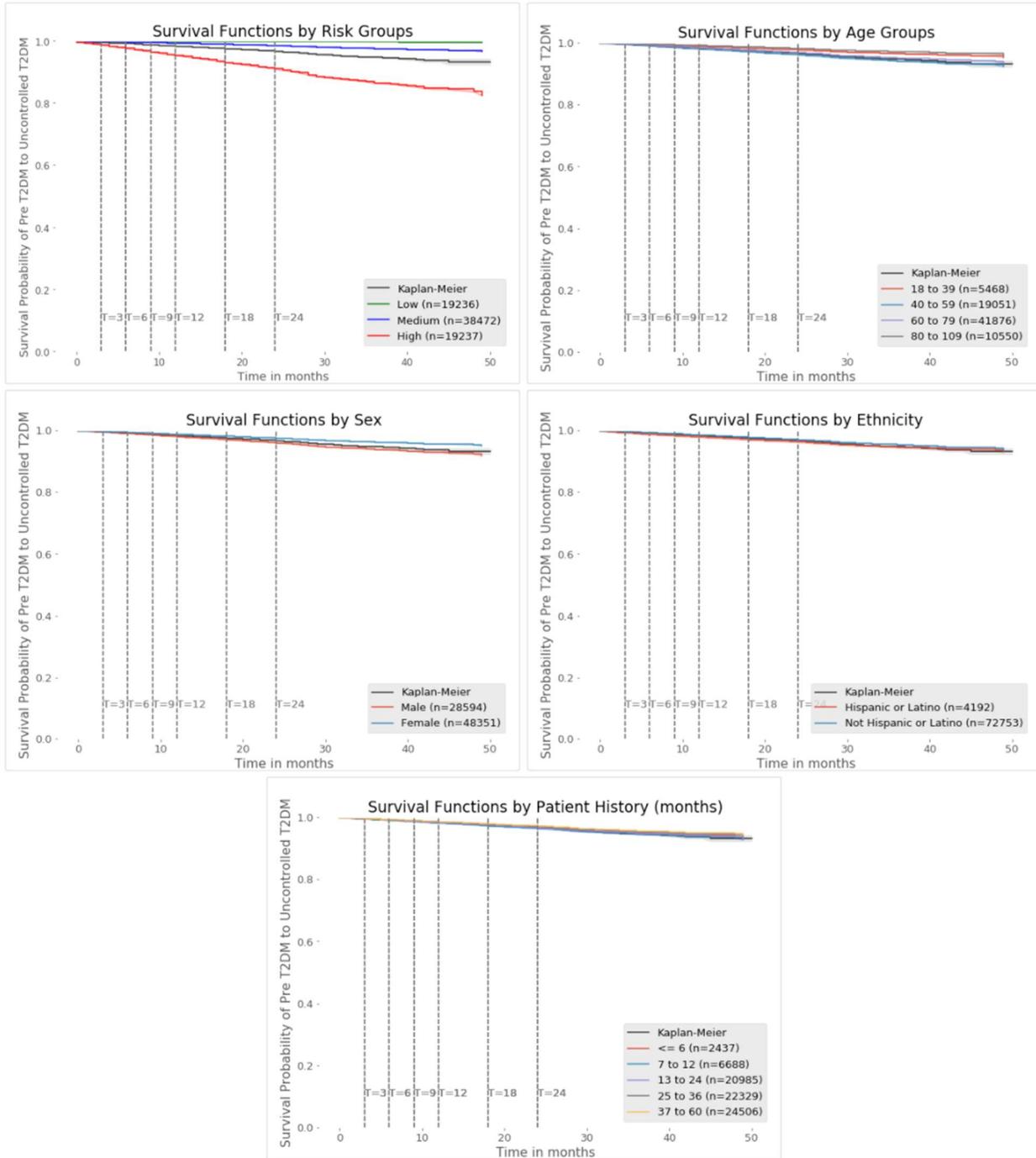

Time-dependent AUC

The following charts show how the AUC across time varies among subgroups of risk, age, sex, ethnicity and patient history.

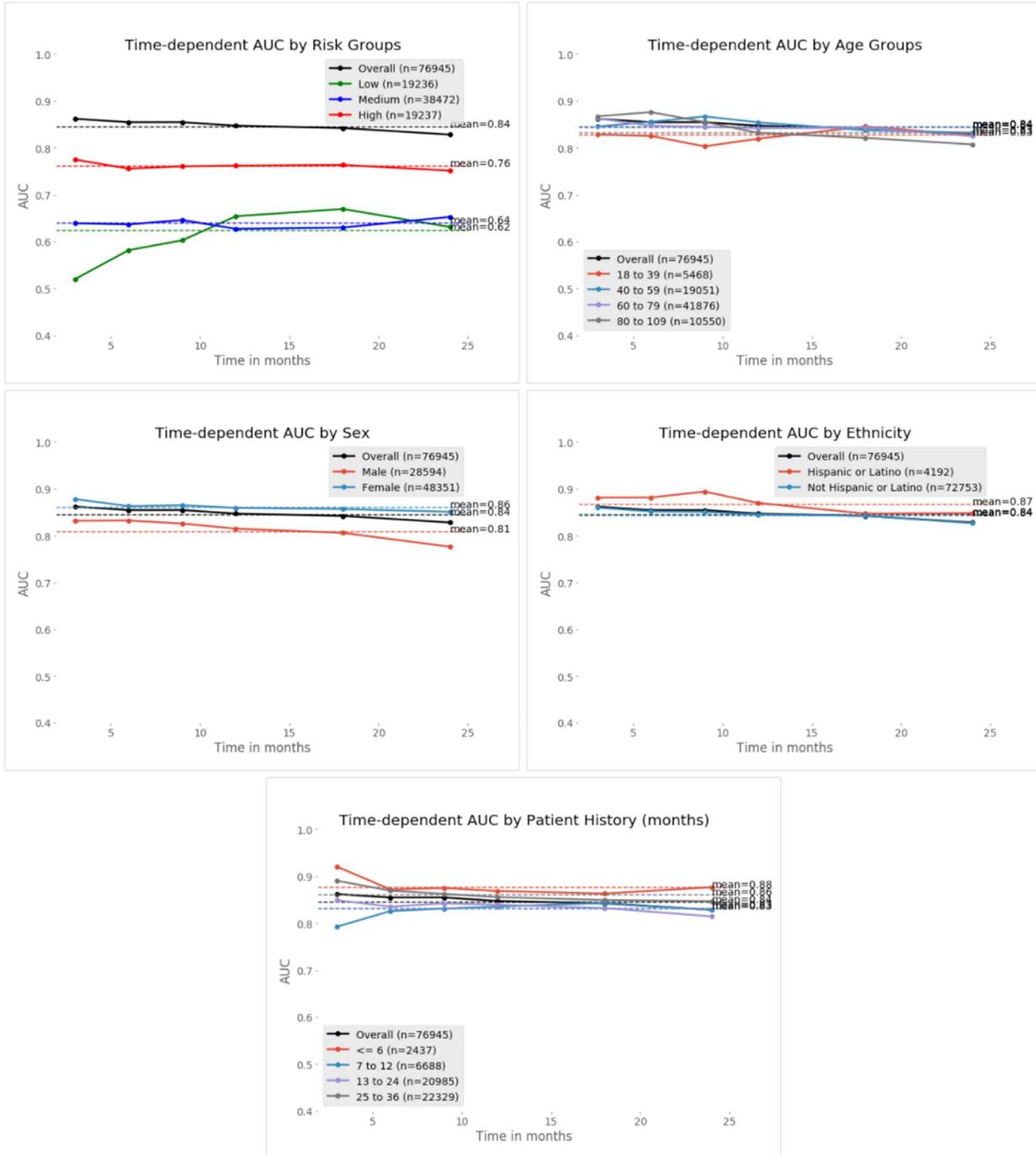

Time-dependent Specificity

The following charts show how the specificity across time varies among subgroups of risk, age, sex, ethnicity and patient history.

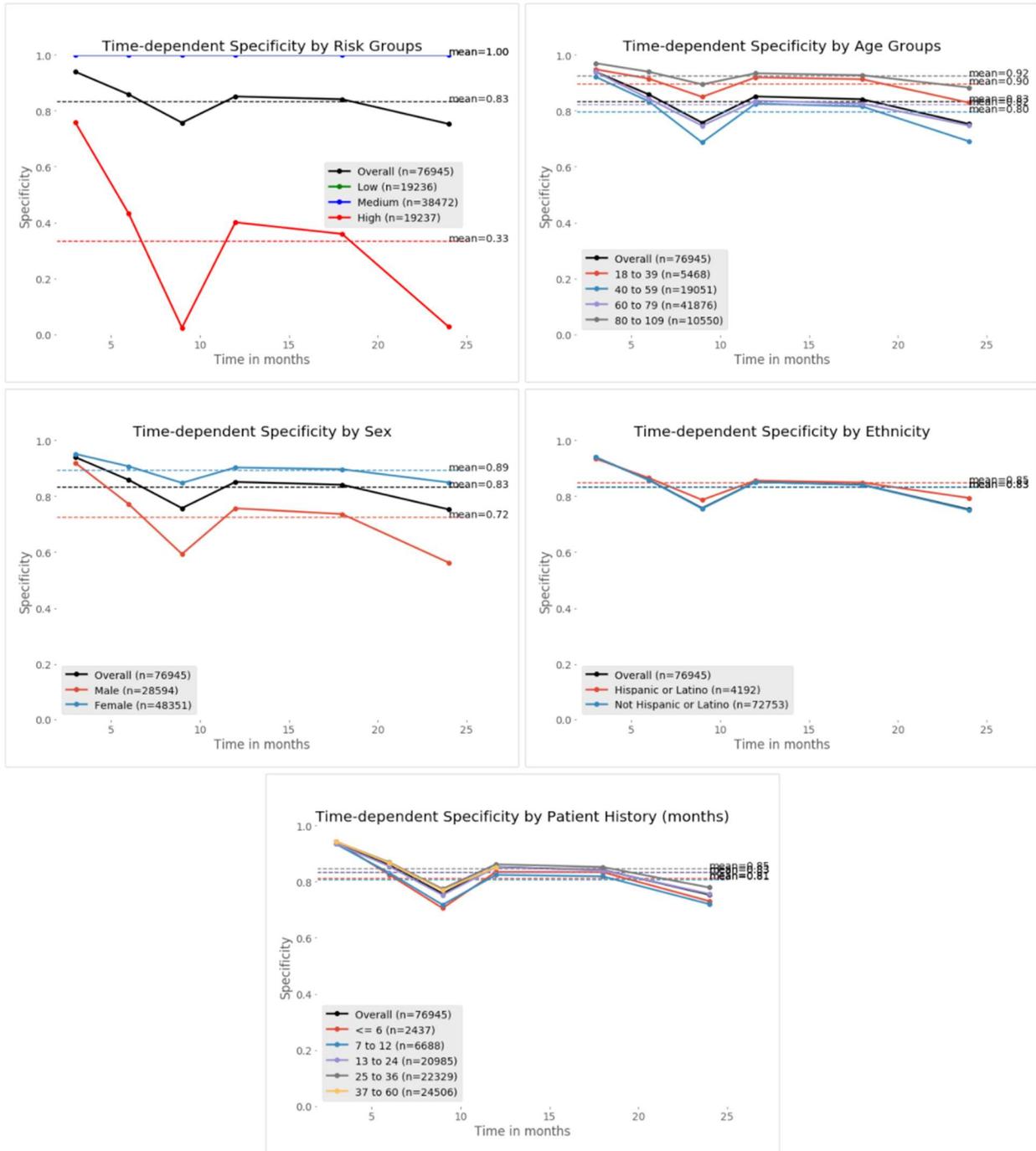

Time-dependent Sensitivity

The following charts show how the sensitivity across time varies among subgroups of risk, age, sex, ethnicity and patient history.

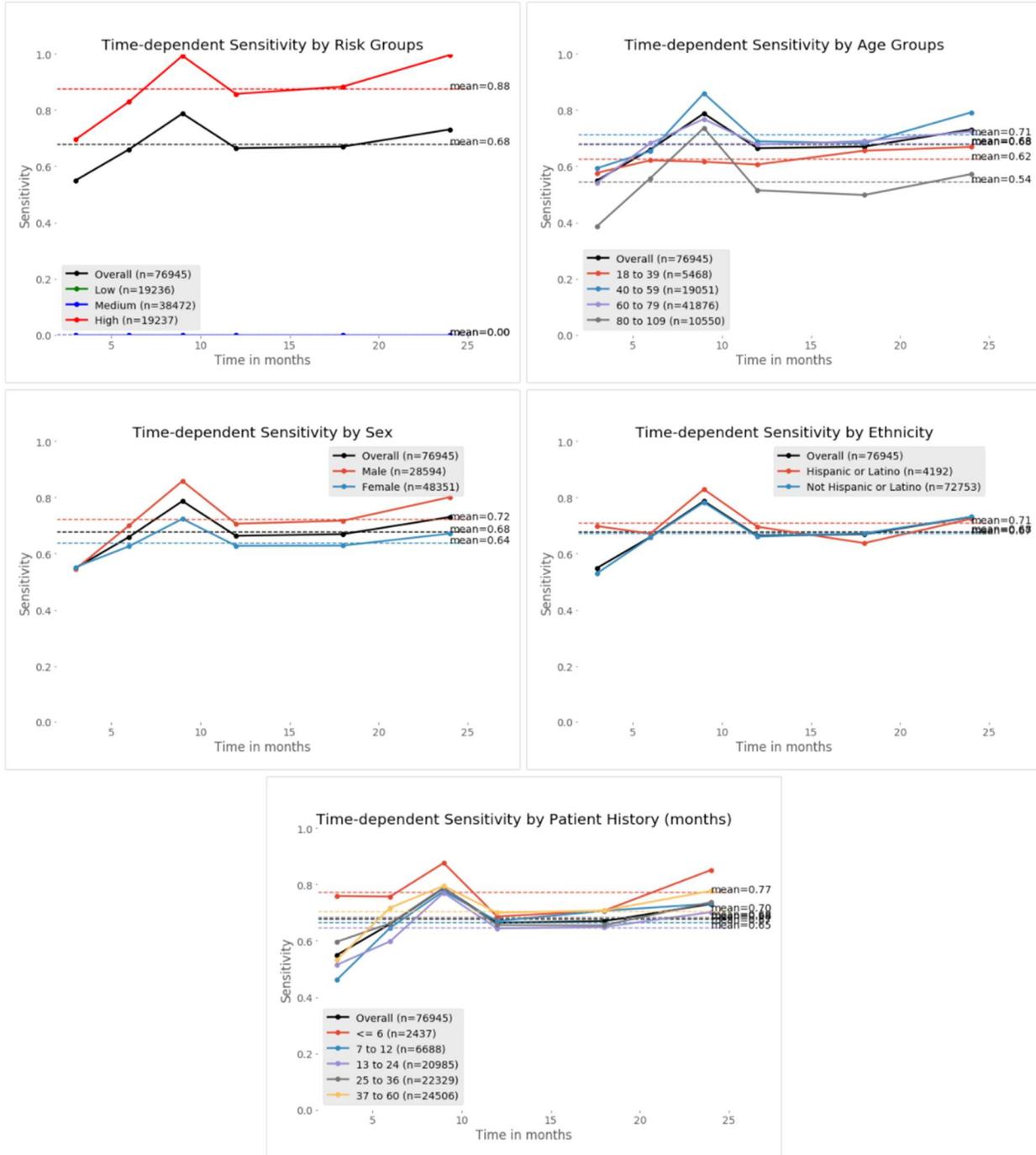

## Model Explanation (DeepSurv)

### Global

The following plots show the SHAP values of each instance in the training set for each future time (3, 6, 9, 12, 18 and 24 months). The features are sorted by the total magnitude of the SHAP values over all instances and the distribution of the effect that each feature has on the model's output can be observed.

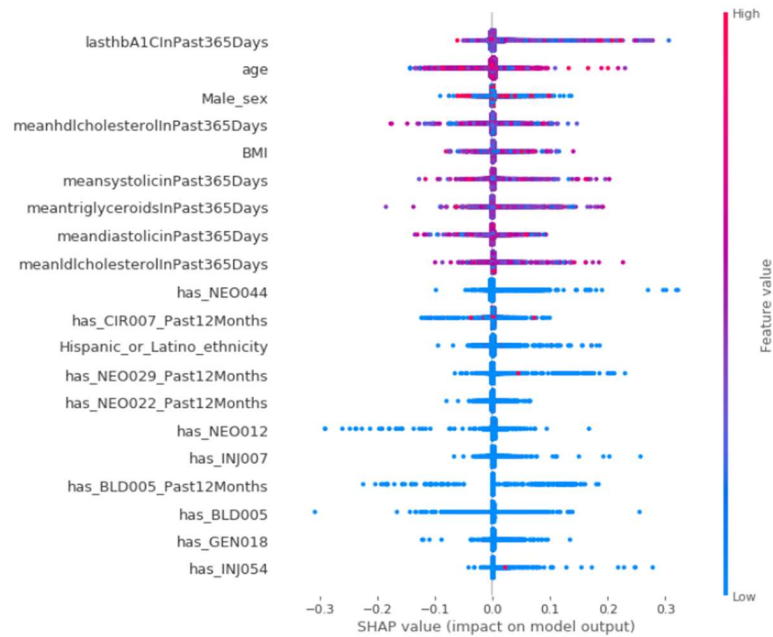

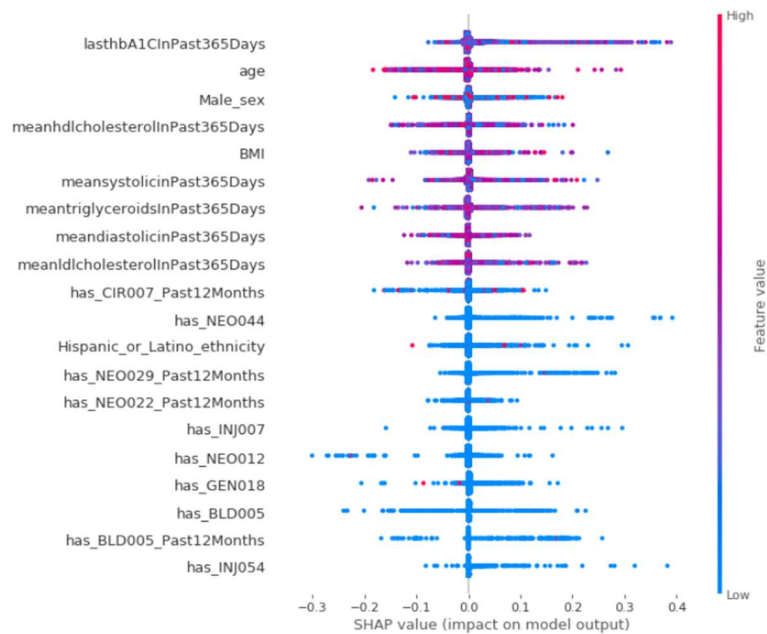

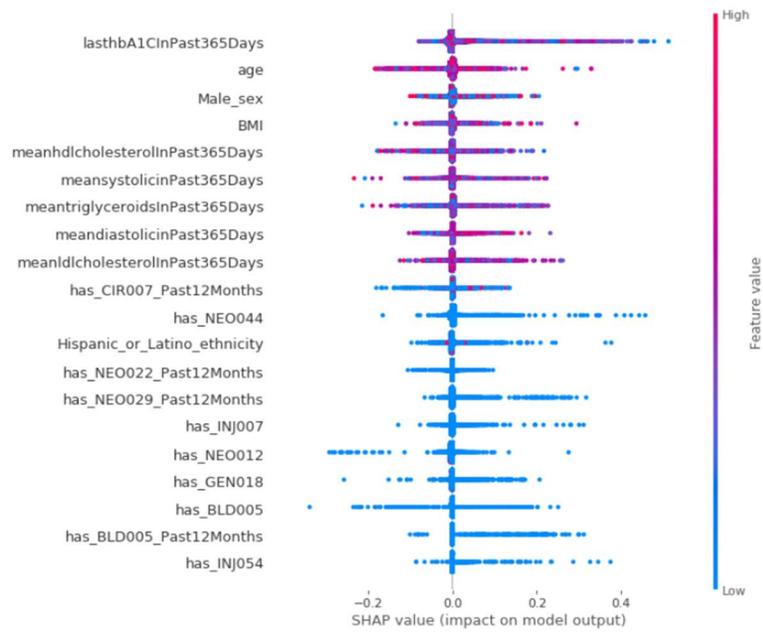

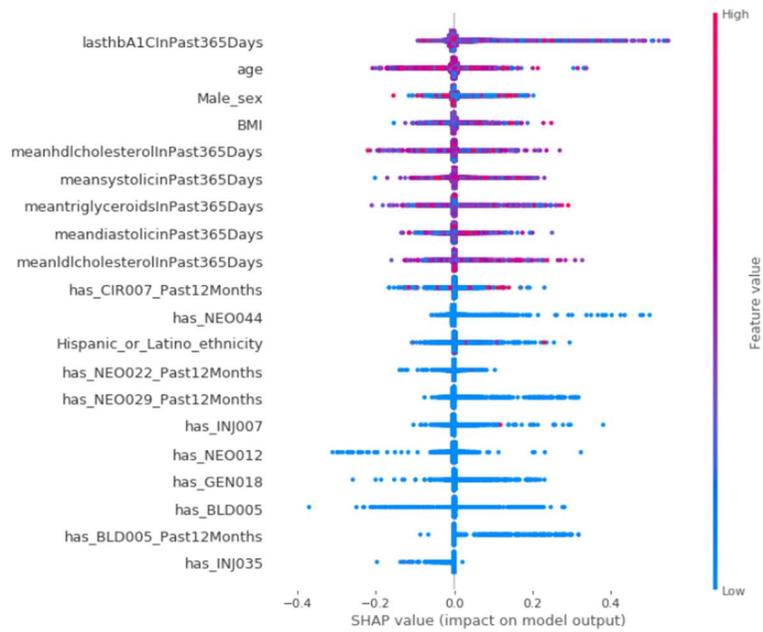

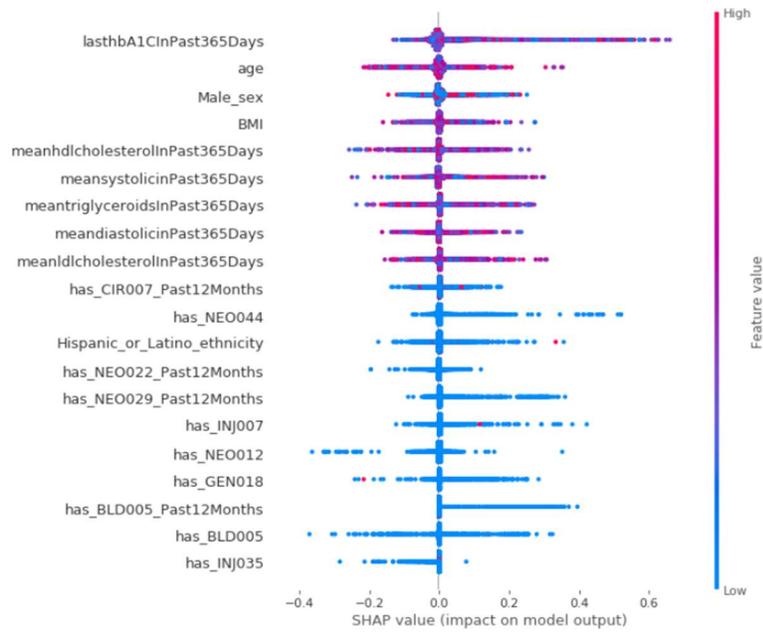

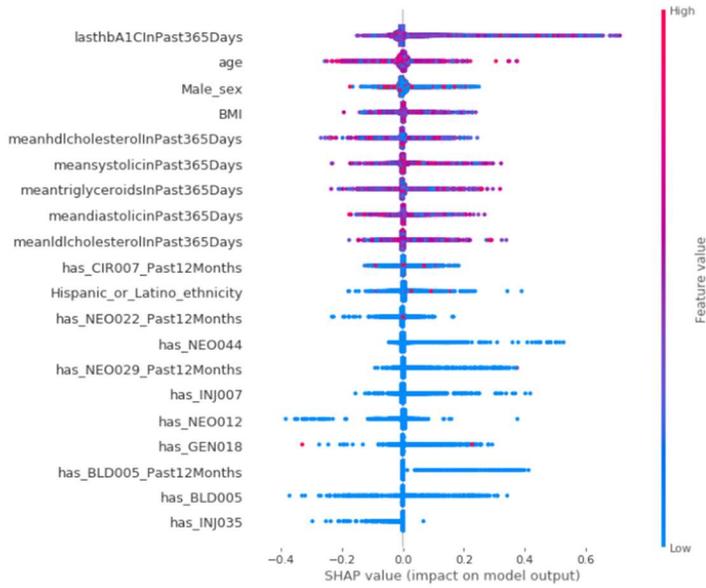

Local

SHAP values were also generated to explain the predictions of individual examples for each future time (3, 6, 9, 12, 18 and 24 months). A total of 3 examples were selected by sampling of risk scores at the 5th, 50th and 95th percentile to represent instances at low, medium and high risks respectively.

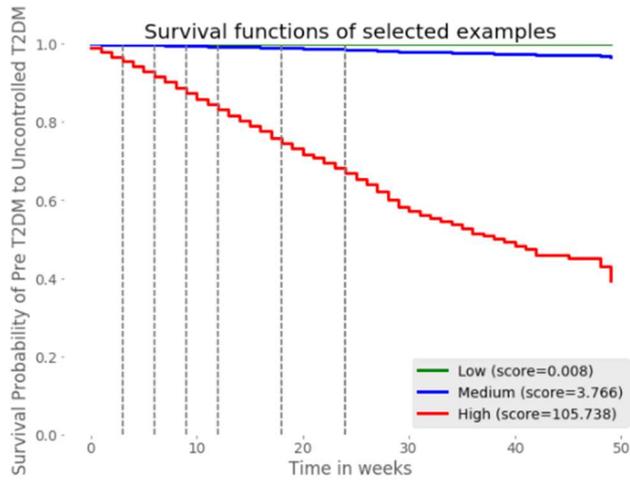

**Low Risk**
Risk Score: 0.008

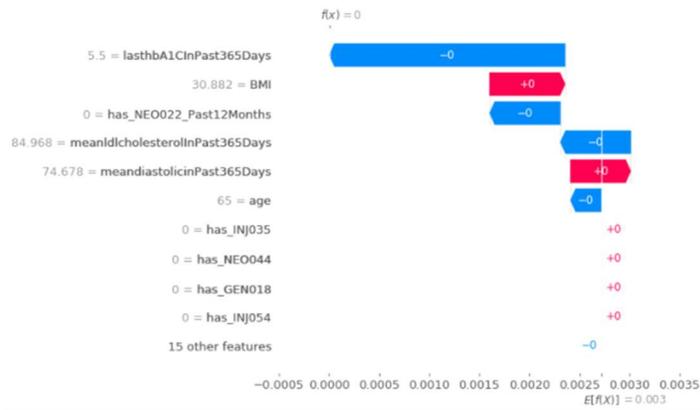

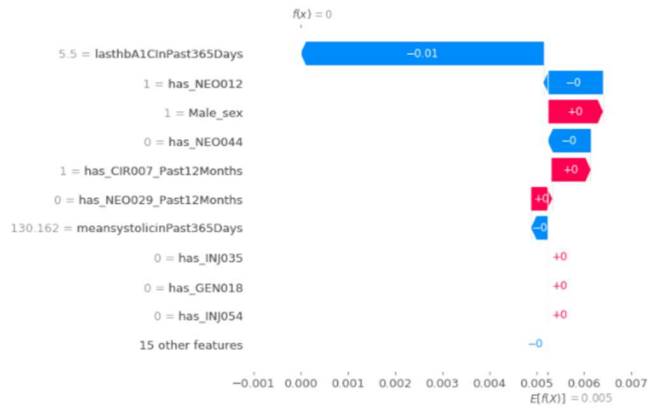

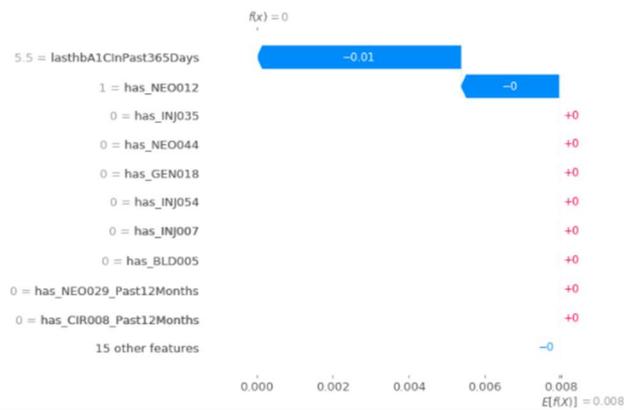

**Low Risk**
Risk Score: 0.008

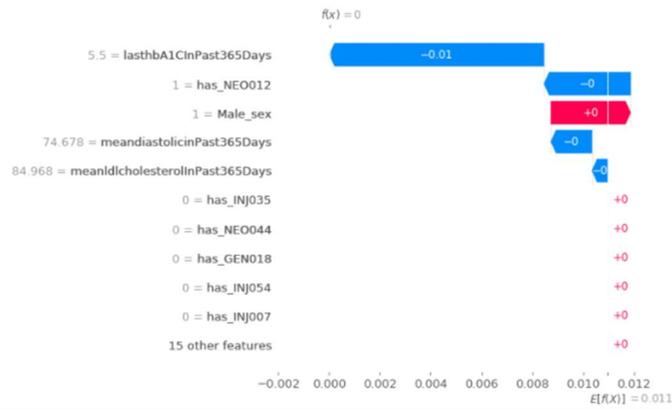

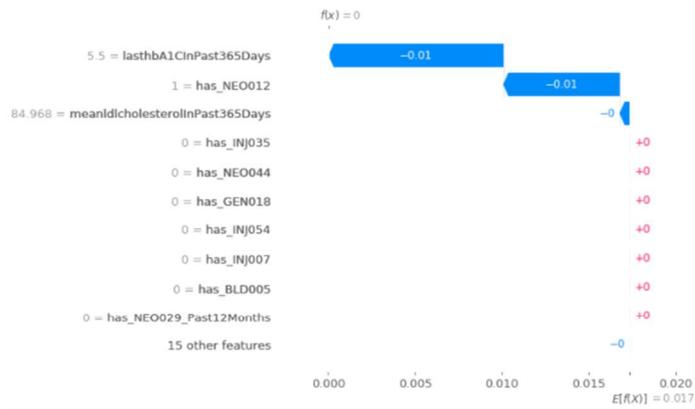

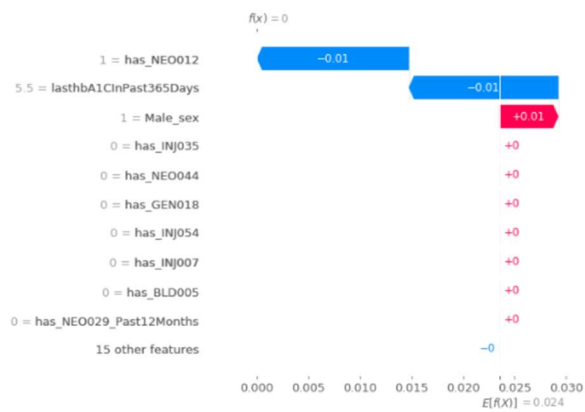

**Medium Risk**
Risk Score: 3.766

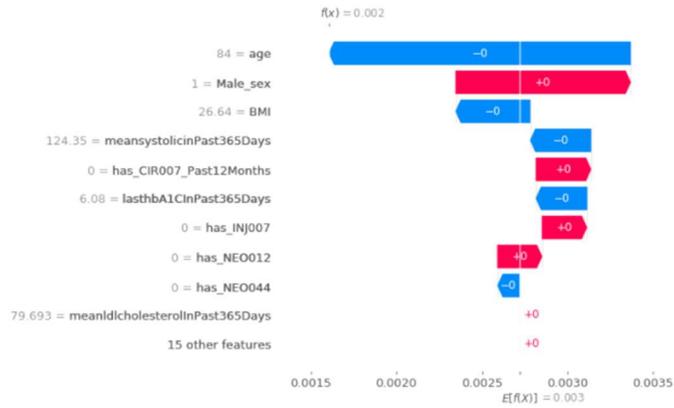

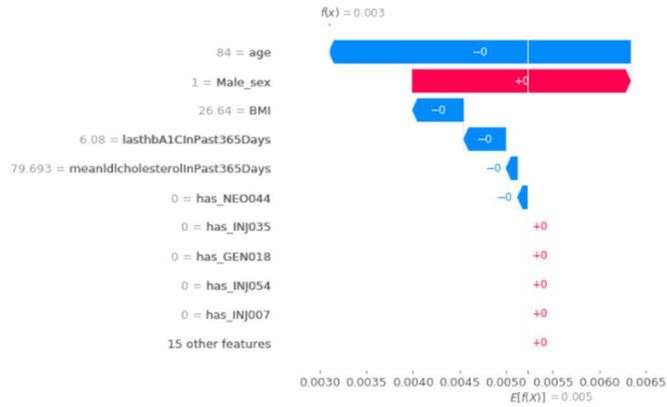

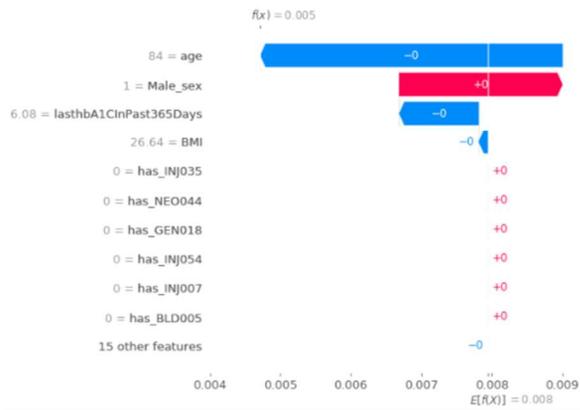

**Medium Risk**
Risk Score: 3.766

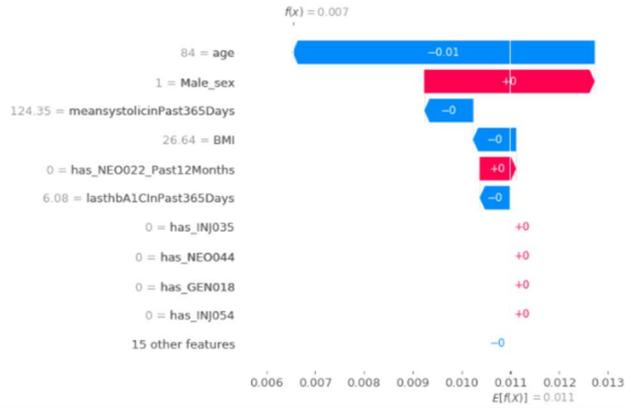

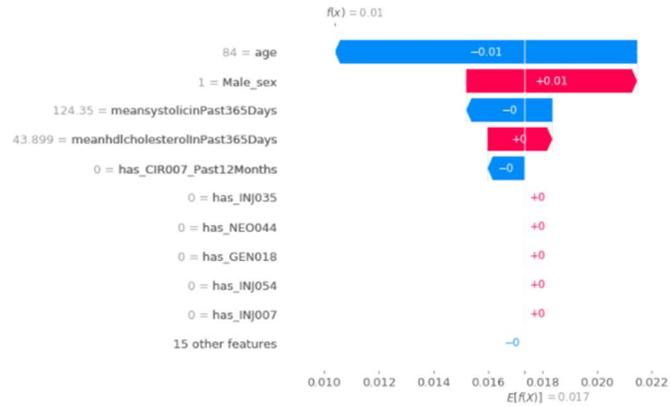

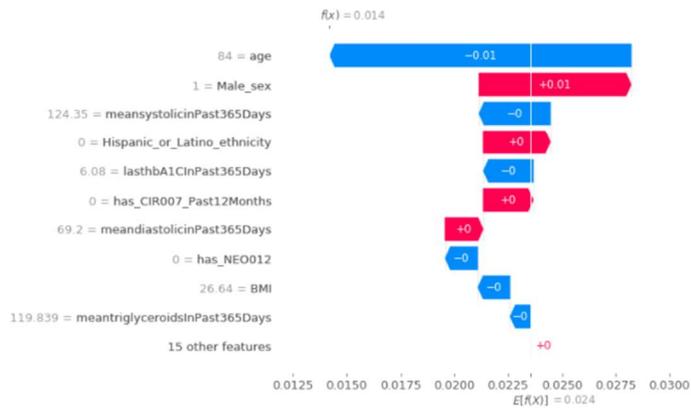

**High Risk**
Risk Score: 105.738

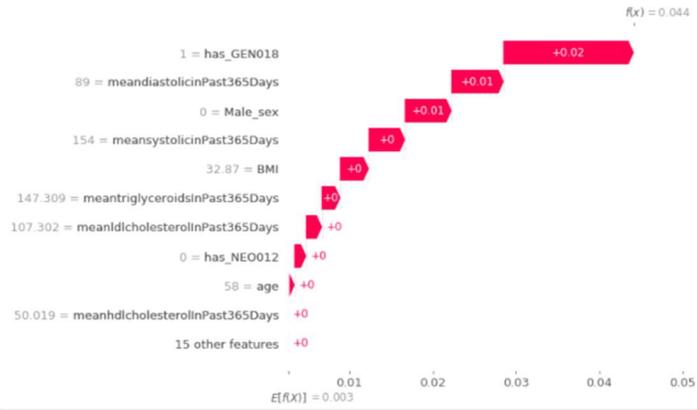

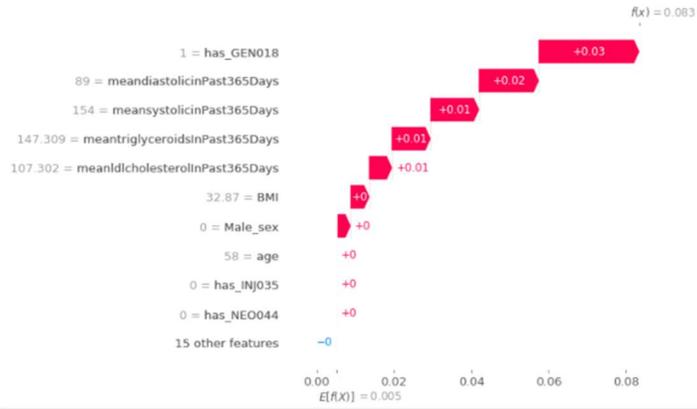

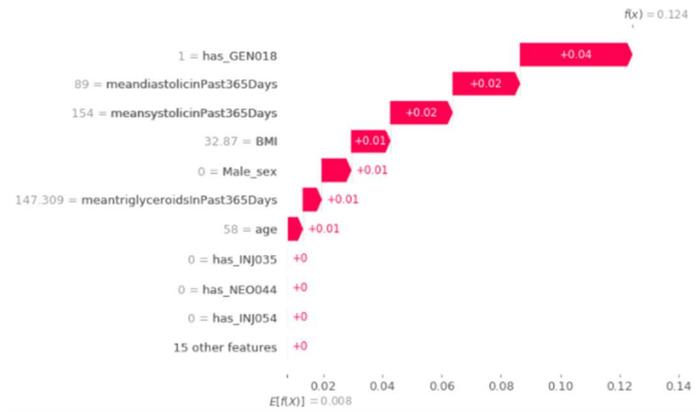

**High Risk**
Risk Score: 105.738

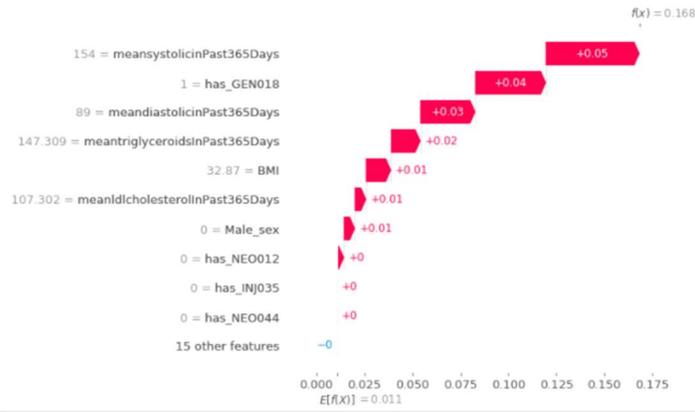

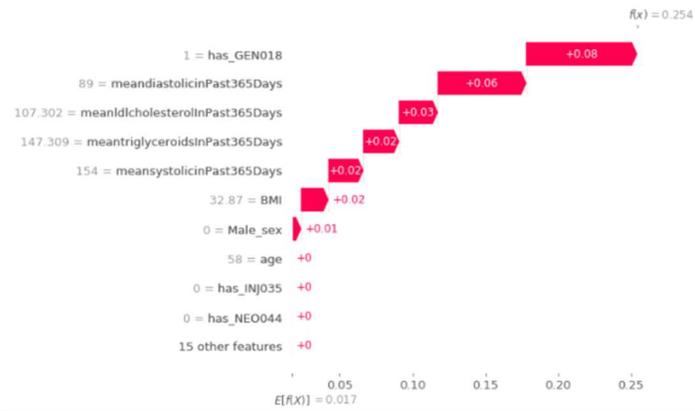

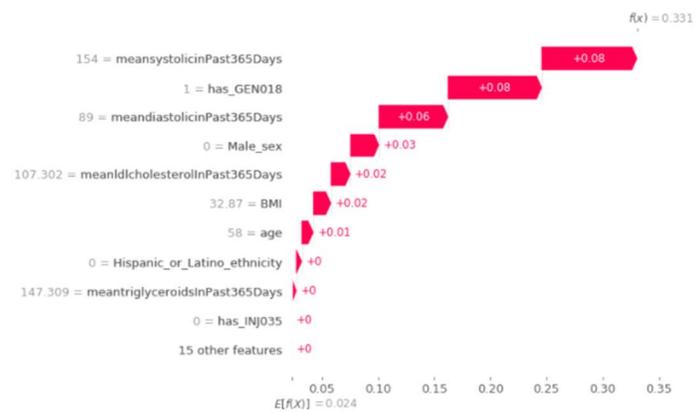

# DM to Diabetic Nephropathy Prediction

| Item | Specification |
| --- | --- |
| **Business Goal** | Enable care managers to identify the patients who are at risk of developing diabetic nephropathy |
| **Usage Setting** | Outpatient |
| **ML Task** | Predict risk and/or time from DM or uncontrolled DM to diabetic nephropathy |
| **ML Class** | Survival |
| **Instances for Prediction** | Encounters |
| **Labels for Instances** | Binary indicator and time to event or censoring for diabetic nephropathy |
| **Cohort Criteria** | <ul><li>2016-01-01 $\leq$ encounter date $\leq$ 2020-06-30 (available Epic data, excluding outliers)</li><li>Encounter date is not within the first 90 days of when the patient entered the data set, to adjust for left-censoring</li><li>18 $\leq$ age $\leq$ 110 (adults without outliers)</li><li>No T1DM diagnosis</li><li>Not pregnant</li><li>No "Do Not Resuscitate" diagnosis</li><li>DM or uncontrolled DM event before encounter date</li><li>No diabetic nephropathy event before encounter date</li><li>No diabetic nephropathy event up to 6 days after encounter date (encounters where diagnoses confirm event within the week)</li></ul> |
| **Input Features** | <ul><li>Demographic</li><li>Diagnosis, except:<ul><li>hasDiabetesNephropathy*</li><li>has_GEN001* (Nephritis; nephrosis; renal sclerosis)</li><li>has_GEN002* (Acute and unspecified renal failure)</li><li>has_GEN003* (Chronic kidney disease)</li><li>has_GEN006* (Other specified and unspecified diseases of kidney and ureters)</li></ul></li><li>Labs</li><li>Utilization</li><li>Vitals</li></ul>See Appendix E for details. |
| **Evaluation Metrics** | <ul><li>Concordance Index</li><li>Integrated Brier Score</li></ul> |

## Data

The following charts summarize the key characteristics of the data after applying the cohort criteria stated above, along with selected features (see Model Signature below).

| Category | Variable | count | mean | stddev | min | 25% | 50% | 75% | max |
|---|---|---|---|---|---|---|---|---|---|
| Demographic | AgeBucket_18_to_39 | 63765.0 | 0.049400 | 0.216704 | 0.00 | 0.000 | 0.000 | 0.000 | 1.00 |
| | AgeBucket_40_to_59 | 63765.0 | 0.299976 | 0.458251 | 0.00 | 0.000 | 0.000 | 1.000 | 1.00 |
| | AgeBucket_60_to_79 | 63765.0 | 0.552309 | 0.497260 | 0.00 | 0.000 | 1.000 | 1.000 | 1.00 |
| | AgeBucket_80_to_109 | 63765.0 | 0.098314 | 0.297741 | 0.00 | 0.000 | 0.000 | 0.000 | 1.00 |
| | Sex_Female | 63765.0 | 0.572367 | 0.494739 | 0.00 | 0.000 | 1.000 | 1.000 | 1.00 |
| | Sex_Male | 63765.0 | 0.427633 | 0.494739 | 0.00 | 0.000 | 0.000 | 1.000 | 1.00 |
| | Ethnicity_Hispanic_or_Latino | 63765.0 | 0.073332 | 0.260682 | 0.00 | 0.000 | 0.000 | 0.000 | 1.00 |
| | Ethnicity_Not_Hispanic_or_Latino | 63765.0 | 0.926668 | 0.260682 | 0.00 | 1.000 | 1.000 | 1.000 | 1.00 |
| Encounter | EncounterType_Emergency | 63765.0 | 0.224731 | 0.417409 | 0.00 | 0.000 | 0.000 | 0.000 | 1.00 |
| | EncounterType_Inpatient | 63765.0 | 0.091759 | 0.288687 | 0.00 | 0.000 | 0.000 | 0.000 | 1.00 |
| | EncounterType_Outpatient | 63765.0 | 0.683510 | 0.465110 | 0.00 | 0.000 | 1.000 | 1.000 | 1.00 |
| Label | Time | 63765.0 | 15.200580 | 11.223469 | 0.00 | 6.000 | 13.000 | 23.000 | 50.00 |
| | Event | 63765.0 | 0.116365 | 0.320664 | 0.00 | 0.000 | 0.000 | 0.000 | 1.00 |
| Feature | age | 63765.0 | 63.402556 | 13.069697 | 18.00 | 55.000 | 65.000 | 72.000 | 100.00 |
| | Male_sex | 63765.0 | 0.427633 | 0.494739 | 0.00 | 0.000 | 0.000 | 1.000 | 1.00 |
| | Hispanic_or_Latino_ethnicity | 63765.0 | 0.073332 | 0.260682 | 0.00 | 0.000 | 0.000 | 0.000 | 1.00 |
| | lasthbA1CInPast365Days | 63765.0 | 7.600242 | 1.215191 | 3.50 | 6.900 | 7.529 | 8.000 | 12.70 |
| | meandiastolicinPast365Days | 63765.0 | 74.097812 | 7.814768 | 44.00 | 70.050 | 74.041 | 79.110 | 107.56 |
| | meansystolicinPast365Days | 63765.0 | 131.832214 | 11.574736 | 82.00 | 127.597 | 131.574 | 135.440 | 180.00 |
| | BMI | 63765.0 | 33.819156 | 6.815868 | 13.43 | 29.840 | 33.718 | 37.340 | 85.68 |
| | meantriglyceroidsInPast365Days | 63765.0 | 169.037094 | 57.815160 | 19.00 | 141.000 | 167.131 | 182.483 | 460.50 |
| | meanldlcholesterolInPast365Days | 63765.0 | 89.617346 | 23.508239 | 8.00 | 81.282 | 89.776 | 97.772 | 201.00 |
| | meanhdlcholesterolInPast365Days | 63765.0 | 43.042894 | 9.097396 | 5.00 | 37.685 | 42.780 | 47.646 | 98.00 |
| | has_NEO024_Past12Months | 63765.0 | 0.001600 | 0.039964 | 0.00 | 0.000 | 0.000 | 0.000 | 1.00 |
| | has_RSP010_Past12Months | 63765.0 | 0.005003 | 0.070553 | 0.00 | 0.000 | 0.000 | 0.000 | 1.00 |
| | has_CIR007_Past12Months | 63765.0 | 0.191578 | 0.393546 | 0.00 | 0.000 | 0.000 | 0.000 | 1.00 |
| | has_CIR008_Past12Months | 63765.0 | 0.008829 | 0.093549 | 0.00 | 0.000 | 0.000 | 0.000 | 1.00 |
| | has_NVS018 | 63765.0 | 0.008970 | 0.094287 | 0.00 | 0.000 | 0.000 | 0.000 | 1.00 |
| | has_NEO050 | 63765.0 | 0.003936 | 0.062617 | 0.00 | 0.000 | 0.000 | 0.000 | 1.00 |
| | has_INJ026 | 63765.0 | 0.003607 | 0.059950 | 0.00 | 0.000 | 0.000 | 0.000 | 1.00 |
| | has_EXT017 | 63765.0 | 0.004077 | 0.063725 | 0.00 | 0.000 | 0.000 | 0.000 | 1.00 |
| | has_CIR034 | 63765.0 | 0.004768 | 0.068883 | 0.00 | 0.000 | 0.000 | 0.000 | 1.00 |
| | has_EYE012 | 63765.0 | 0.005018 | 0.070663 | 0.00 | 0.000 | 0.000 | 0.000 | 1.00 |
| | has_FAC023 | 63765.0 | 0.043770 | 0.204585 | 0.00 | 0.000 | 0.000 | 0.000 | 1.00 |
| | has_NEO008 | 63765.0 | 0.002729 | 0.052167 | 0.00 | 0.000 | 0.000 | 0.000 | 1.00 |
| | has_NVS005 | 63765.0 | 0.004642 | 0.067975 | 0.00 | 0.000 | 0.000 | 0.000 | 1.00 |
| | has_GEN010 | 63765.0 | 0.015306 | 0.122769 | 0.00 | 0.000 | 0.000 | 0.000 | 1.00 |
| | has_END011 | 63765.0 | 0.238767 | 0.426334 | 0.00 | 0.000 | 0.000 | 0.000 | 1.00 |
| | has_CIR008 | 63765.0 | 0.031459 | 0.174557 | 0.00 | 0.000 | 0.000 | 0.000 | 1.00 |

(Percentages for binary variables can be read from the "mean" column.)

### Encounters and Patients

| | Examples | Encounters | Patients |
|---|---|---|---|
| 1 | 63765 | 63765 | 11801 |

### Train and Test Sets

| | Set | No Event | No Event % | Event | Event % | Total | Total % |
|---|---|---|---|---|---|---|---|
| 1 | Train | 39492 | 70.1 | 5214 | 70.3 | 44706 | 70.1 |
| 2 | Test | 16853 | 29.9 | 2206 | 29.7 | 19059 | 29.9 |
| 3 | Total | 56345 | 100 | 7420 | 100 | 63765 | 100 |

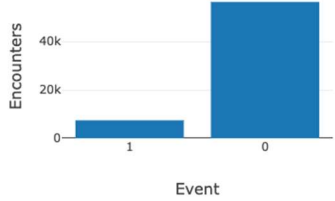
Encounters with Event (1) or Censoring (0)

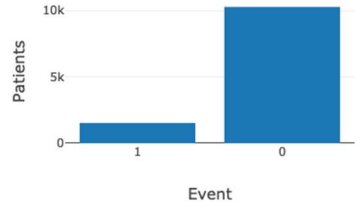
Patients with Event (1) or Censoring (0)

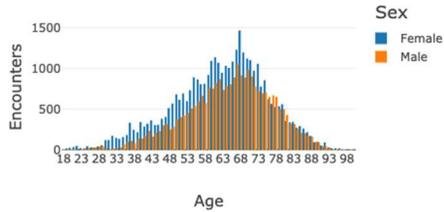
Encounters by Age and Sex

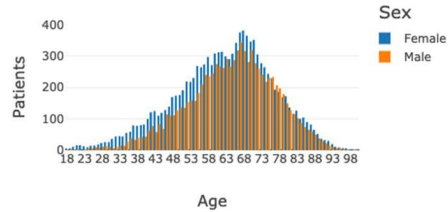
Patients by Age and Sex

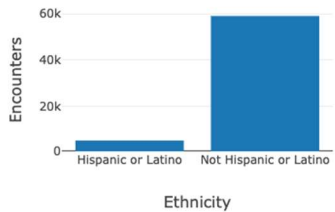
Encounters by Ethnicity

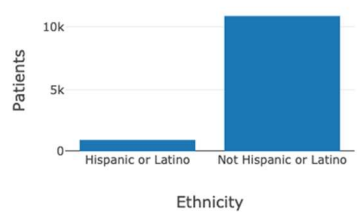
Patients by Ethnicity

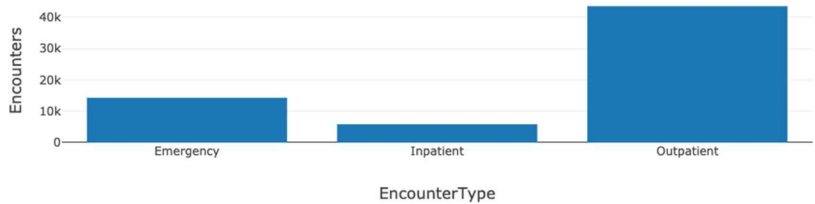
Encounters by Encounter Type

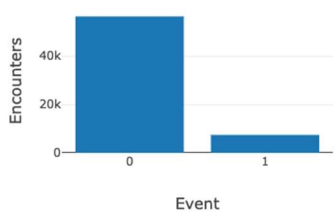
Encounters by Time to Event (1) or Cen…

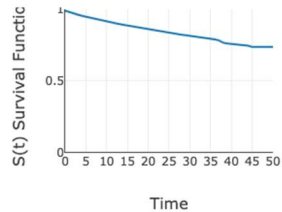
Kaplan-Meier Estimate of Survival Functi…

## Model Signature

The model signature has 26 features, comprising of 12 mandatory features and 14 other selected features. These are the selected features, in rank order (the last feature to be eliminated is ranked 1):

1. has_NEO008 (Head and neck cancers - laryngeal)
2. has_EXT017 (External cause codes: suffocation/inhalation; initial encounter)
3. has_NEO024_Past12Months (Sarcoma)
4. has_EYE012 (Other specified eye disorders)
5. has_CIR034 (Chronic phlebitis; thrombophlebitis and thromboembolism)
6. has_NVS005 (Multiple sclerosis)
7. has_INJ026 (Other specified injury)
8. has_CIR008 (Hypertension with complications and secondary hypertension)
9. has_RSP010_Past12Months (Aspiration pneumonitis)
10. has_GEN010 (Proteinuria)
11. has_END011 (Fluid and electrolyte disorders)
12. has_NVS018 (Myopathies)
13. has_NEO050 (Endocrine system cancers - thyroid)
14. has_FAC023 (Organ transplant status)
15. Hispanic_or_Latino_ethnicity
16. Male_sex
17. has_CIR007_Past12Months (Essential hypertension)
18. lasthbA1CInPast365Days
19. age
20. BMI
21. meandiastolicinPast365Days
22. meansystolicinPast365Days
23. meanhdlcholesterolInPast365Days
24. meanldlcholesterolInPast365Days
25. meantriglyceroidsInPast365Days
26. has_CIR008_Past12Months (Hypertension with complications and secondary hypertension)

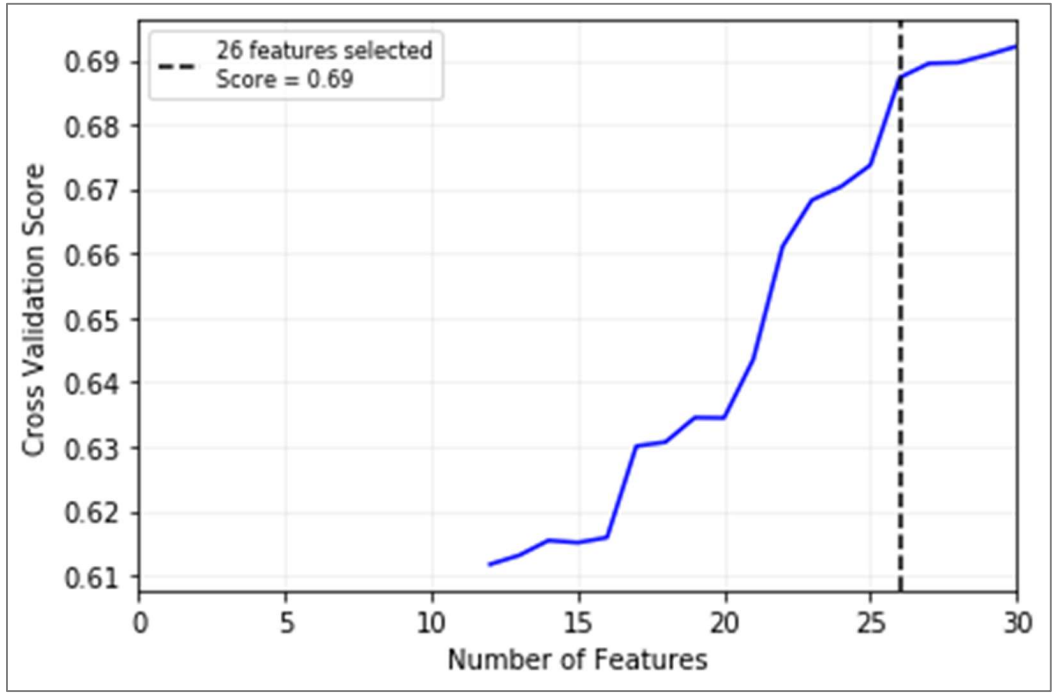

## Model Performance

The following table and chart summarize the performance of all candidate models on the test set for this prediction task in terms of the Concordance Index and the Integrated Brier Score.

| Model | No. of Parameter Combinations Successfully Tested | Concordance Index | Integrated Brier Score |
|---|---|---|---|
| **CoxPH** | 117 | 0.68 | 0.11 |
| **DeepSurv** | 175 | **0.80** | **0.09** |
| **RSF** | 35 | 0.71 | 0.12 |
| **CSF** | 35 | 0.69 | 0.12 |
| **EST** | 27 | 0.71 | 0.12 |

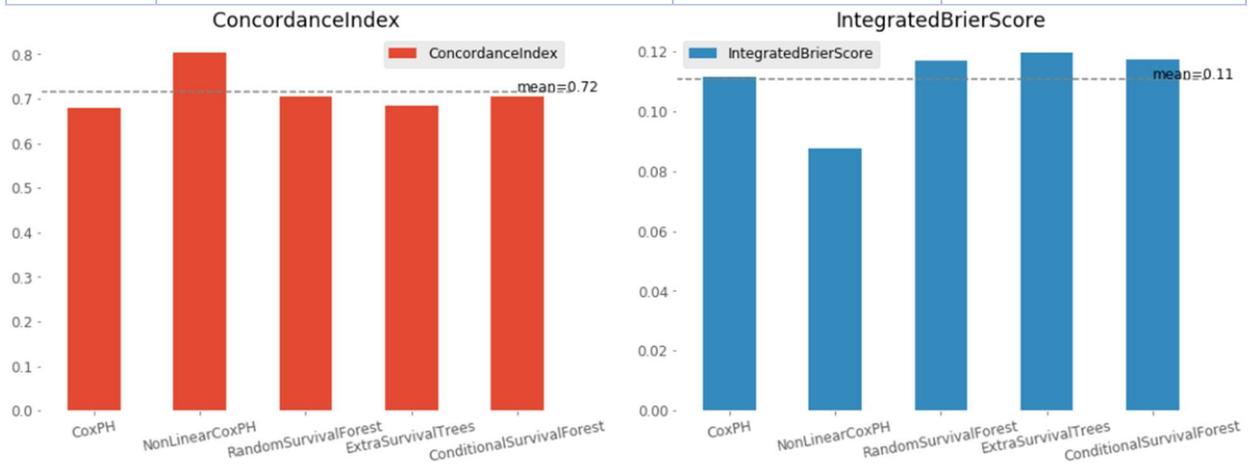

The following chart shows how the average survival function curves of the candidate models compare to the KM survival curve, the more similar their curves are to the KM survival curve the better.

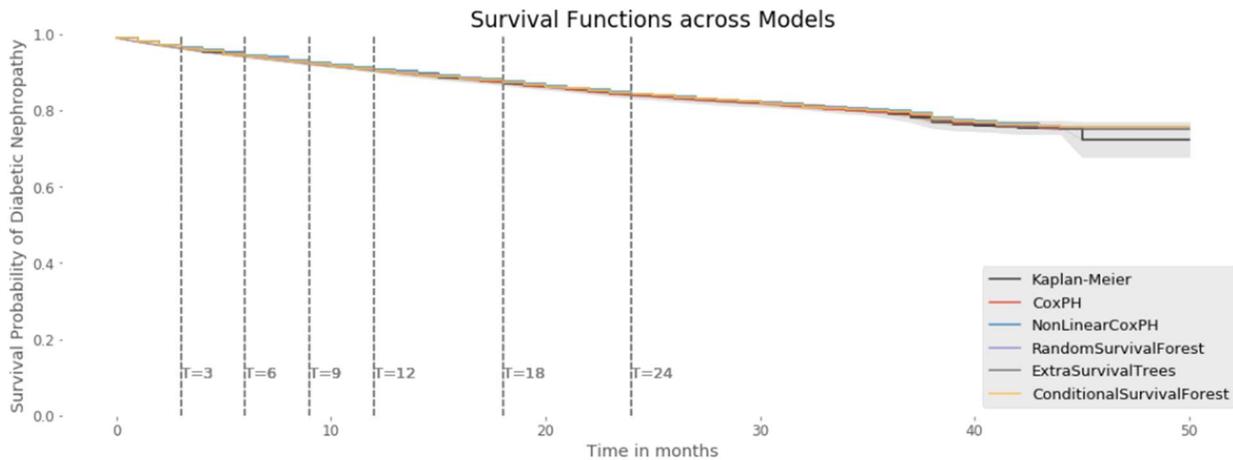

This chart shows the change in the Brier Score over time for all candidate models, the closer the scores are to 0 the better.

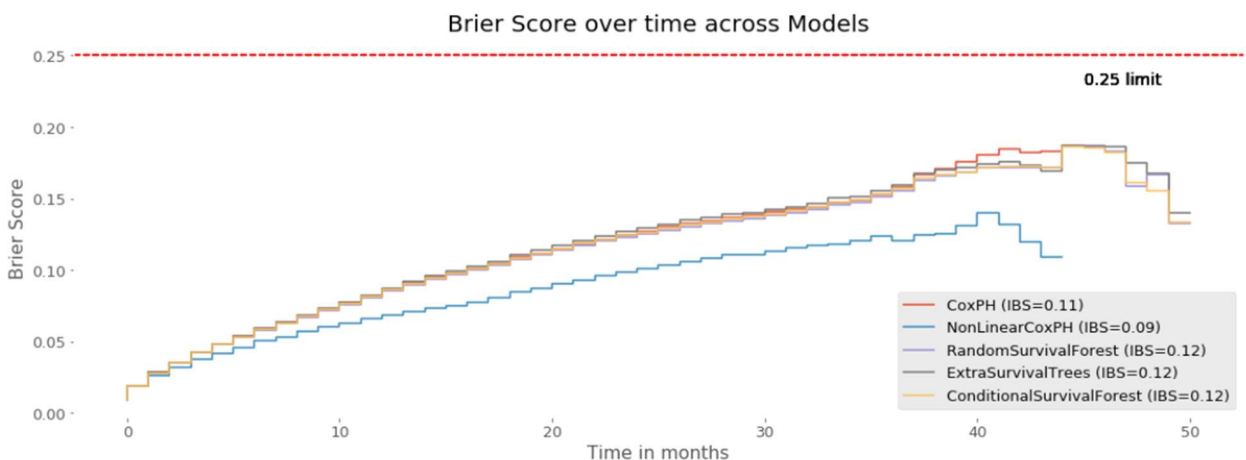

## Model Evaluation of Selected Model (DeepSurv)

### Overall

This chart shows the change in the Brier Score over time for the selected model, the closer the scores are to 0 the better.

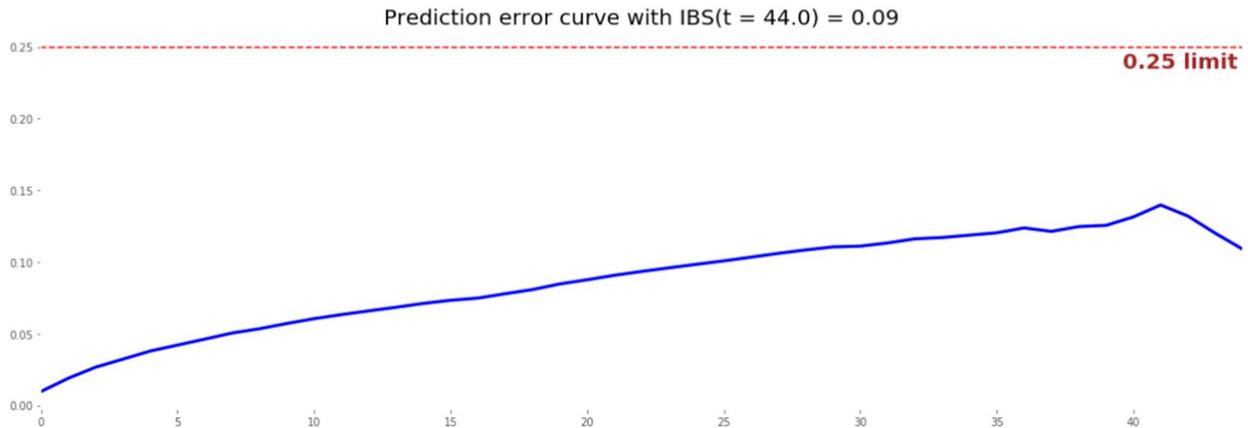

The following chart shows the actual vs. predicted density functions, i.e. number of instances that get the disease / complication at each time point and the RMSE, Median Absolute Error and Mean Absolute Error across the time points.

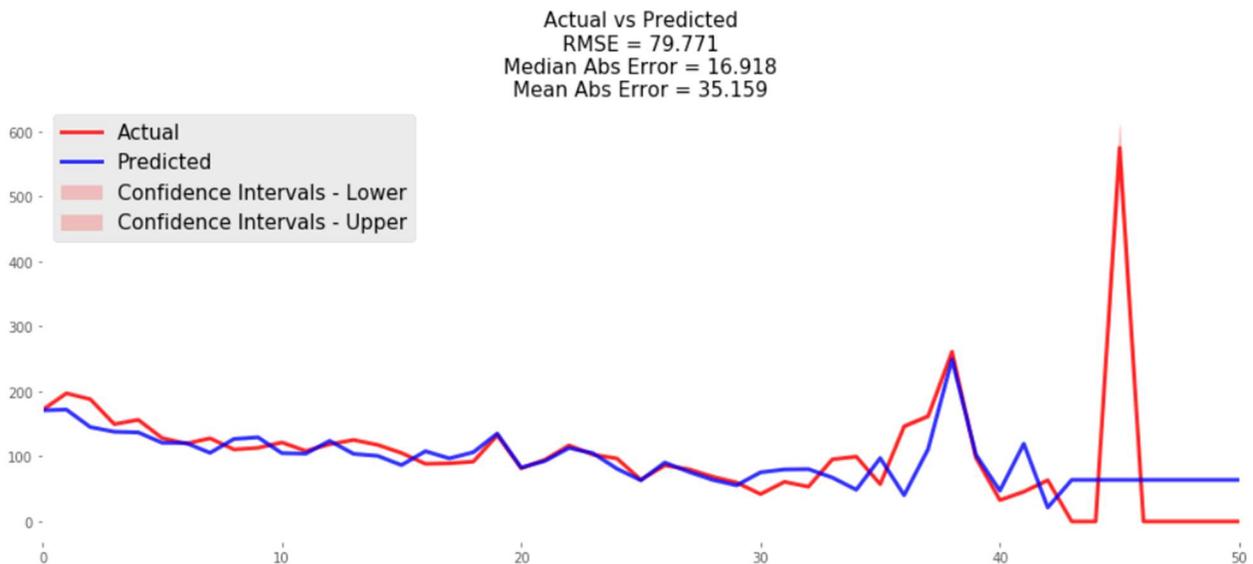

The following chart shows the actual vs. predicted survival functions, i.e. the number of instances that have not had the disease / complication by each time point and the RMSE, Median Absolute Error and Mean Absolute Error across the time points.

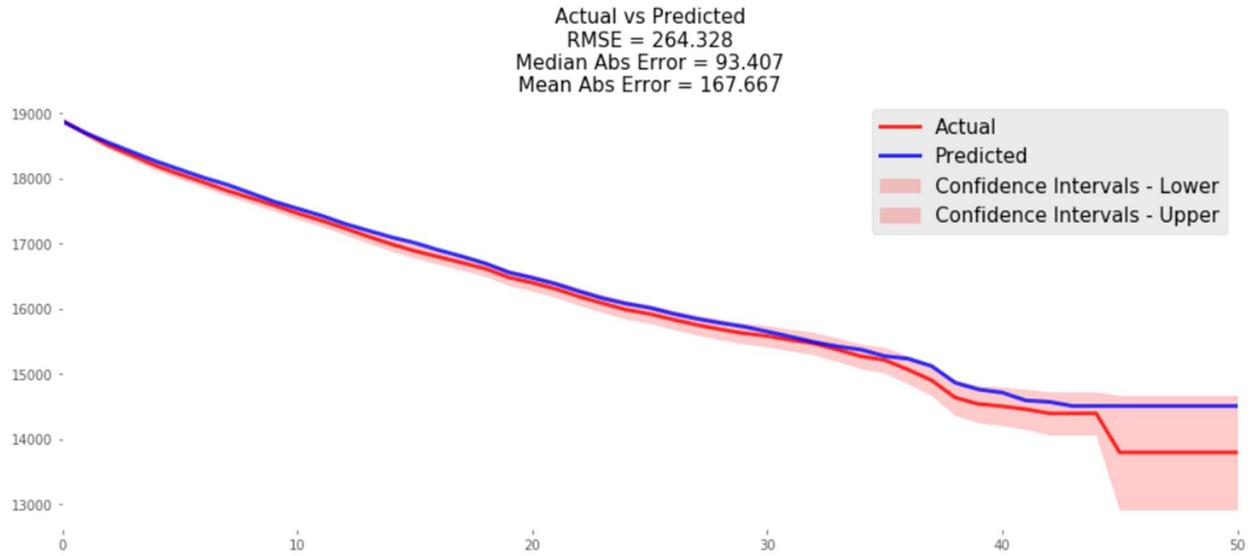

Risk Stratification

The low, medium and high risk groups are defined as examples with predicted risk scores belonging to the first quartile, second to third quartiles, and fourth quartile respectively.

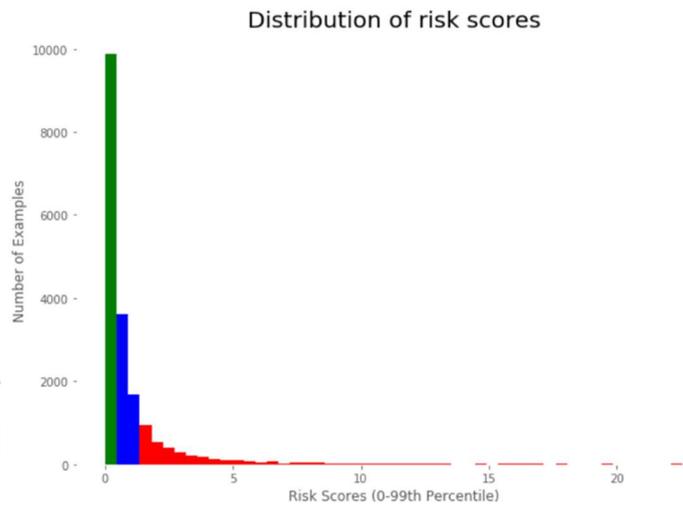

|  | Percentile | Risk Scores |
|---|---|---|
| Low Risk | 0-25th | [0.0,0.143] |
| Medium Risk | 26th-75th | [0.143,1.095] |
| High Risk | 75th-100th | [1.095,13426.0400390625] |

Summary Metrics across Subgroups

The table below displays the summary metrics across subgroups of risk, age, sex, ethnicity and patient history.

| Category | Subgroup | Cohort Size | Concordance Index | Brier Score | Mean AUC | Mean Specificity | Mean Sensitivity | S(t), t=3 | S(t), t=6 | S(t), t=9 | S(t), t=12 | S(t), t=18 | S(t), t=24 |
|---|---|---|---|---|---|---|---|---|---|---|---|---|---|
| NaN | Overall | 19059.00 | 0.80 | 0.09 | 0.83 | 0.76 | 0.73 | 0.96 | 0.94 | 0.92 | 0.90 | 0.87 | 0.84 |
| Risk | Low | 4765.00 | 0.65 | 0.02 | 0.66 | 1.00 | 0.00 | 1.00 | 1.00 | 1.00 | 1.00 | 0.99 | 0.99 |
| Risk | Medium | 9529.00 | 0.69 | 0.08 | 0.63 | 0.94 | 0.13 | 0.99 | 0.98 | 0.97 | 0.96 | 0.94 | 0.91 |
| Risk | High | 4765.00 | 0.70 | 0.16 | 0.75 | 0.00 | 1.00 | 0.89 | 0.83 | 0.77 | 0.72 | 0.64 | 0.56 |
| Age Bucket | 18 to 39 | 902.00 | 0.84 | 0.04 | 0.85 | 0.95 | 0.50 | 0.99 | 0.98 | 0.97 | 0.97 | 0.96 | 0.95 |
| Age Bucket | 40 to 59 | 5734.00 | 0.81 | 0.06 | 0.87 | 0.88 | 0.68 | 0.98 | 0.96 | 0.95 | 0.94 | 0.92 | 0.90 |
| Age Bucket | 60 to 79 | 10577.00 | 0.80 | 0.10 | 0.80 | 0.70 | 0.73 | 0.96 | 0.94 | 0.91 | 0.89 | 0.86 | 0.82 |
| Age Bucket | 80 to 109 | 1846.00 | 0.75 | 0.11 | 0.79 | 0.60 | 0.80 | 0.95 | 0.92 | 0.89 | 0.86 | 0.82 | 0.77 |
| Sex | Male | 8153.00 | 0.77 | 0.09 | 0.82 | 0.73 | 0.75 | 0.96 | 0.94 | 0.92 | 0.90 | 0.86 | 0.83 |
| Sex | Female | 10906.00 | 0.82 | 0.09 | 0.83 | 0.78 | 0.71 | 0.97 | 0.95 | 0.93 | 0.92 | 0.89 | 0.86 |
| Ethnicity | Hispanic or Latino | 1424.00 | 0.89 | 0.04 | 0.92 | 0.89 | 0.84 | 0.97 | 0.95 | 0.94 | 0.93 | 0.91 | 0.88 |
| Ethnicity | Not Hispanic or Latino | 17635.00 | 0.79 | 0.09 | 0.82 | 0.75 | 0.72 | 0.97 | 0.94 | 0.92 | 0.91 | 0.87 | 0.84 |
| History Bucket | <= 6 | 525.00 | 0.76 | 0.07 | 0.76 | 0.74 | 0.66 | 0.97 | 0.95 | 0.94 | 0.92 | 0.89 | 0.85 |
| History Bucket | 7 to 12 | 1856.00 | 0.82 | 0.09 | 0.86 | 0.76 | 0.81 | 0.96 | 0.94 | 0.92 | 0.90 | 0.87 | 0.84 |
| History Bucket | 13 to 24 | 5382.00 | 0.80 | 0.09 | 0.82 | 0.74 | 0.74 | 0.96 | 0.94 | 0.92 | 0.90 | 0.86 | 0.83 |
| History Bucket | 25 to 36 | 5444.00 | 0.83 | 0.06 | 0.84 | 0.78 | 0.72 | 0.96 | 0.94 | 0.92 | 0.91 | 0.88 | 0.84 |
| History Bucket | 37 to 60 | 5852.00 | 0.83 | 0.05 | nan | nan | 0.67 | 0.97 | 0.95 | 0.94 | 0.92 | 0.89 | 0.86 |

Concordance Index & Integrated Brier Score

The following charts show how the Concordance Index and Integrated Brier Score varies among subgroups of risk, age, sex, ethnicity and patient history.

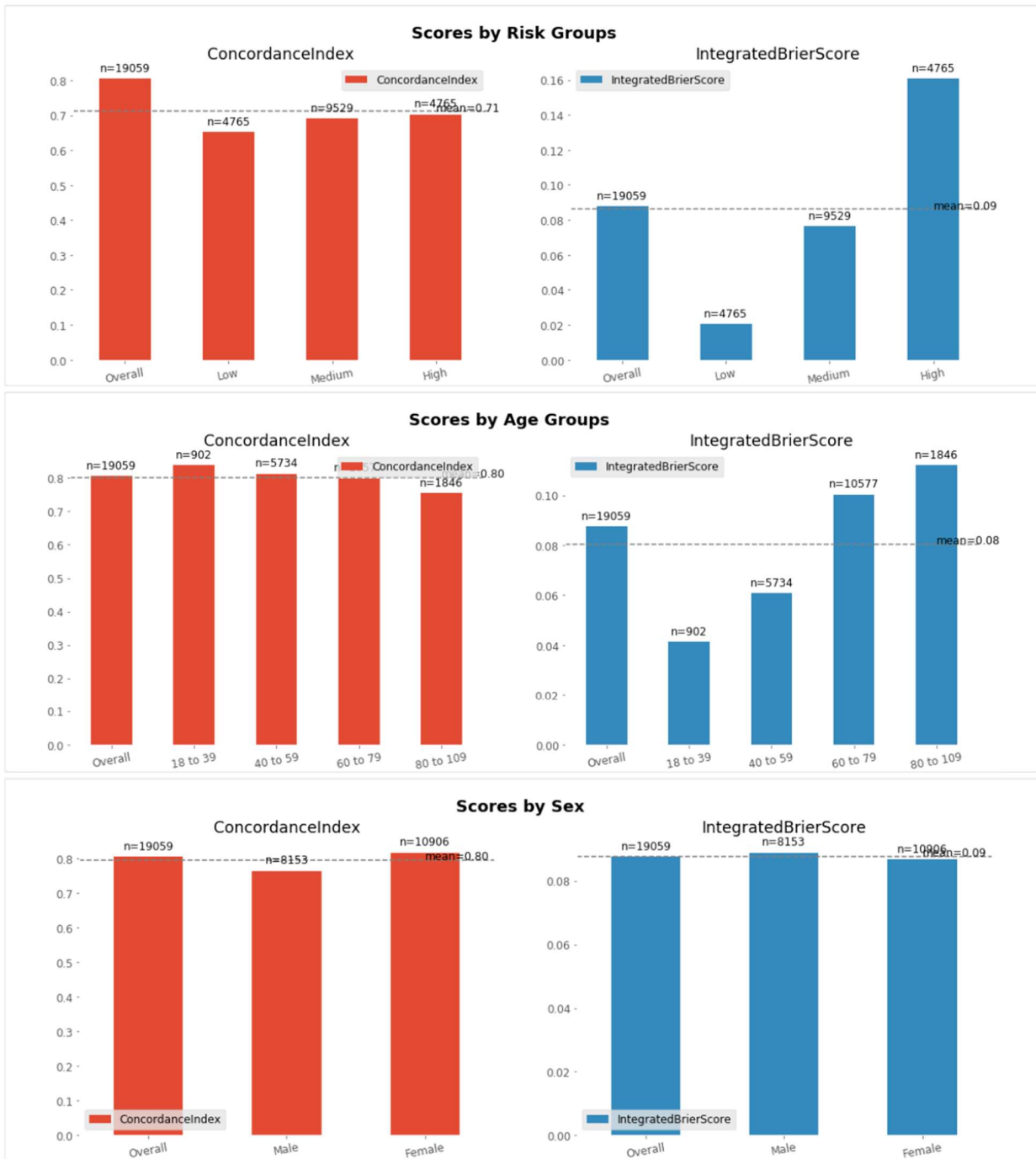

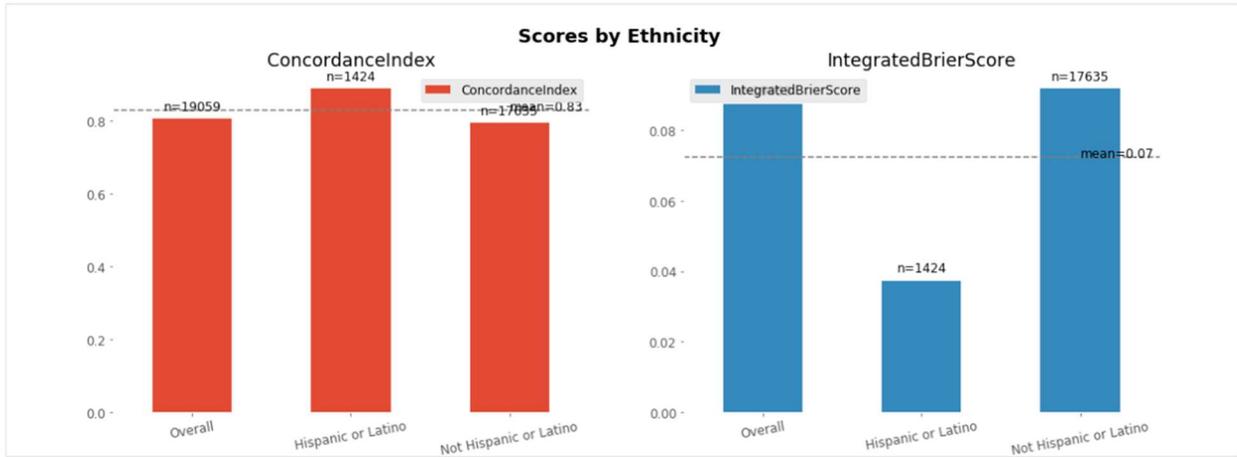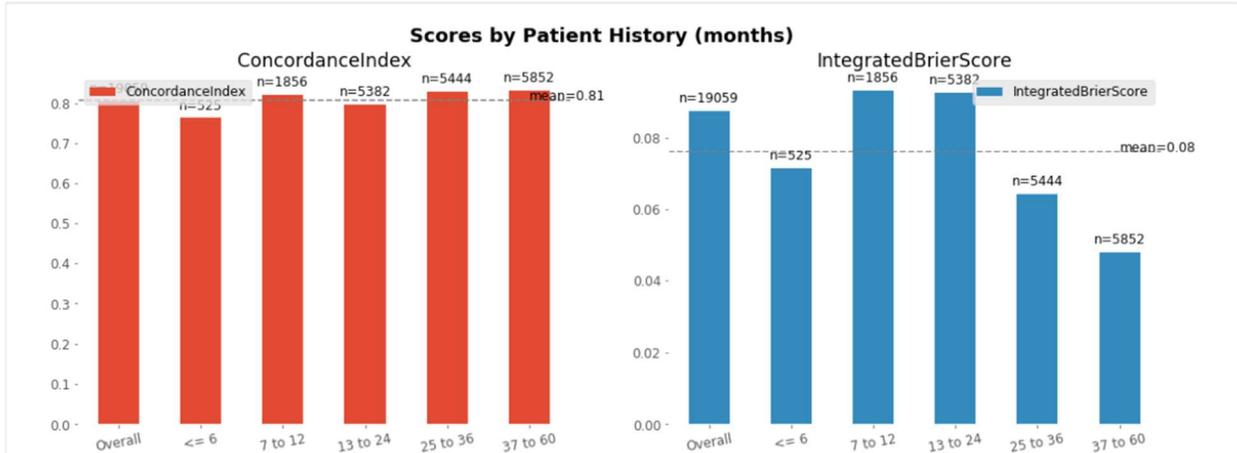

Average Survival Function Curves

The following charts show how the average survival function curve varies among subgroups of risk, age, sex, ethnicity and patient history.

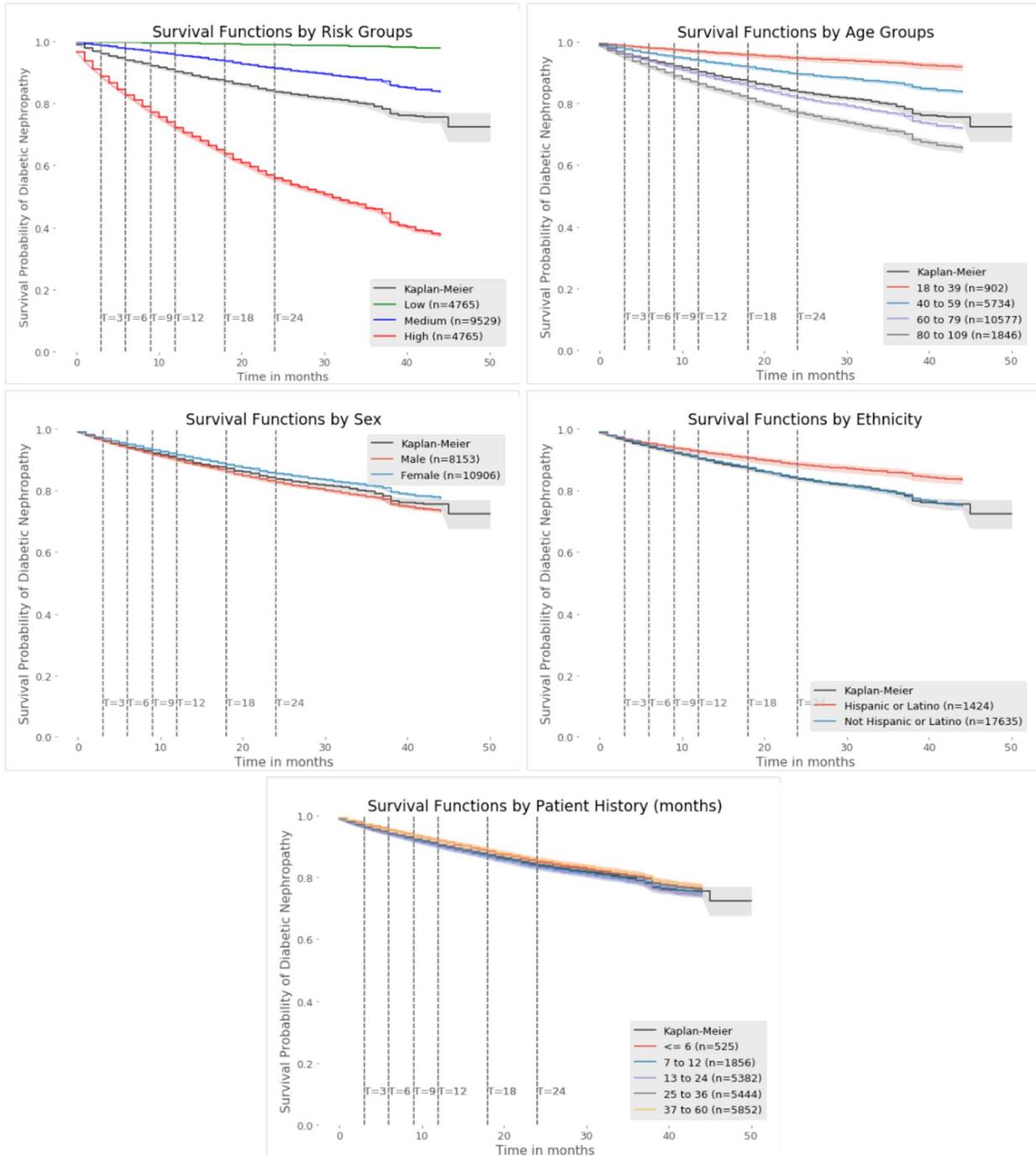

Time-dependent AUC

The following charts show how the AUC across time varies among subgroups of risk, age, sex, ethnicity and patient history.

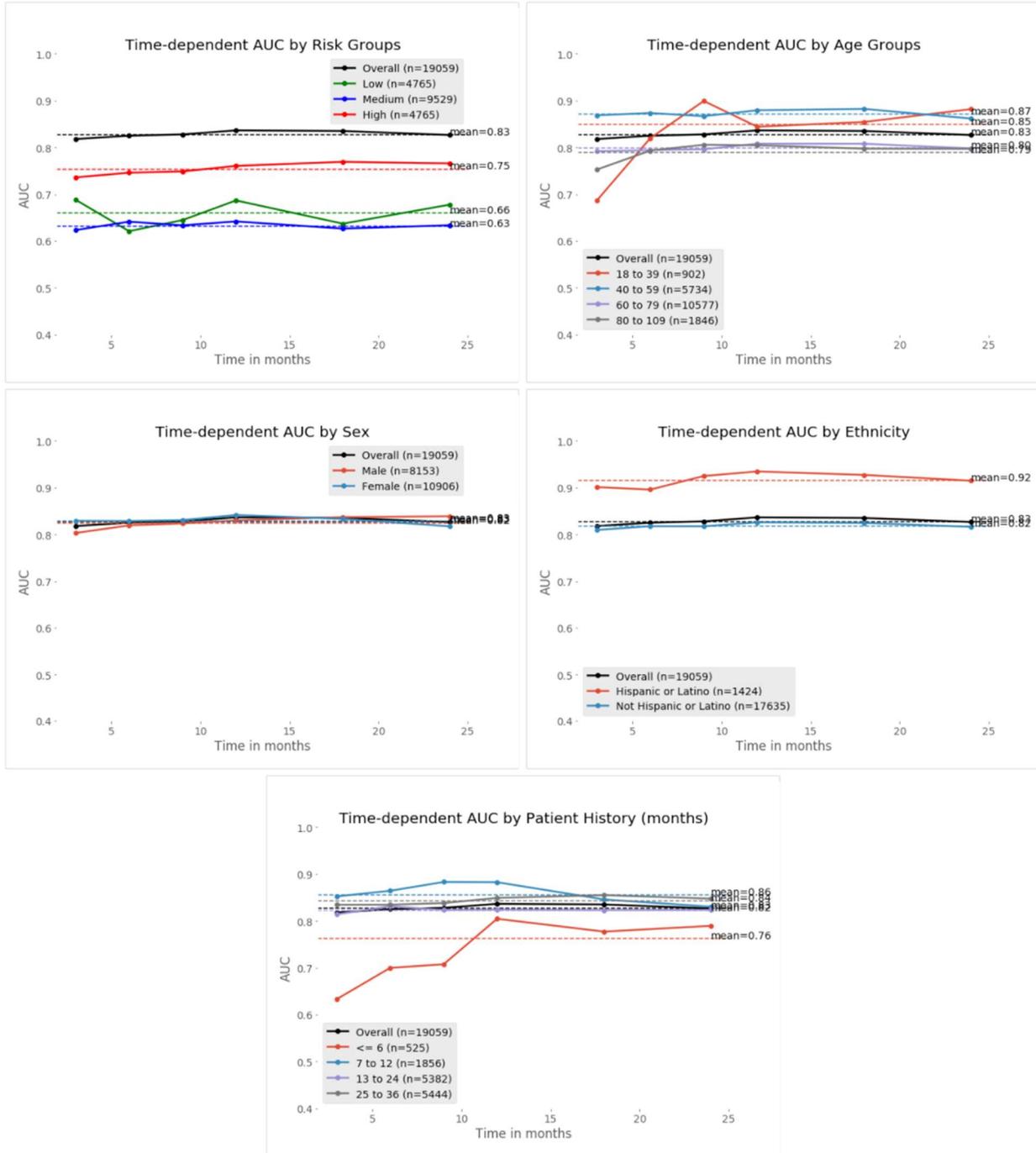

Time-dependent Specificity

The following charts show how the specificity across time varies among subgroups of risk, age, sex, ethnicity and patient history.

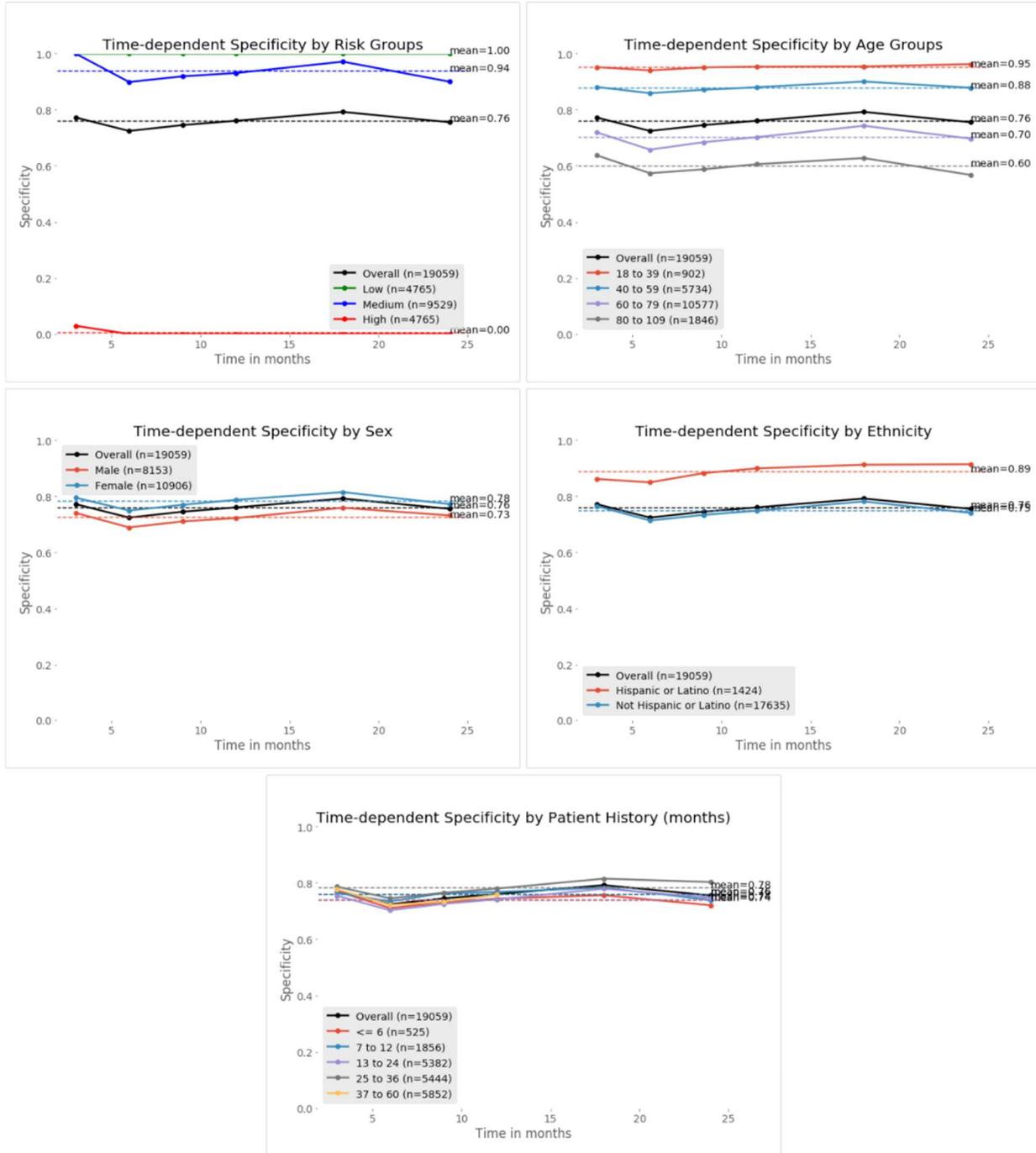

Time-dependent Sensitivity

The following charts show how the sensitivity across time varies among subgroups of risk, age, sex, ethnicity and patient history.

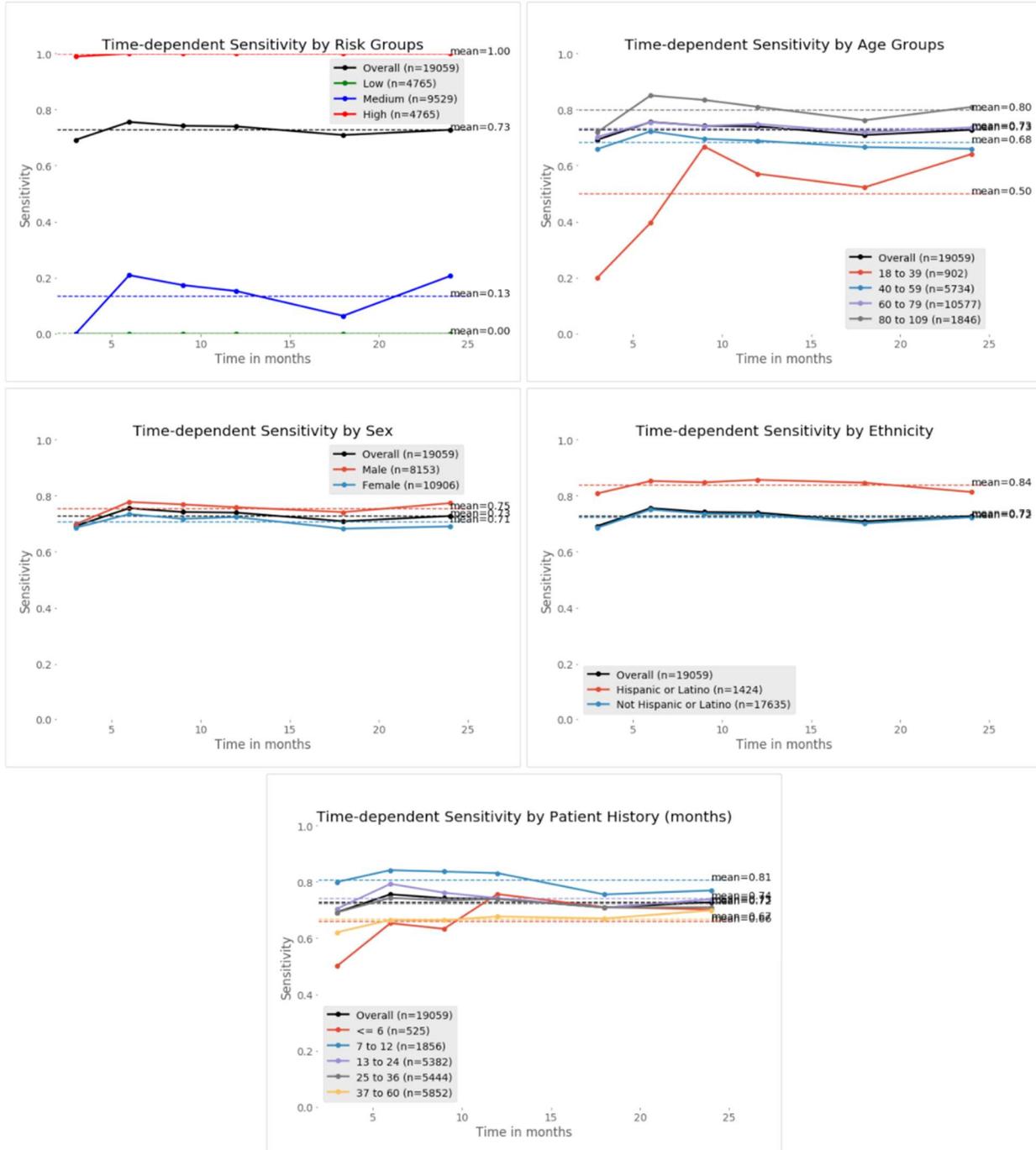

## Model Explanation (DeepSurv)
Global

The following plots show the SHAP values of each instance in the training set for each future time (3, 6, 9, 12, 18 and 24 months). The features are sorted by the total magnitude of the SHAP values over all instances and the distribution of the effect that each feature has on the model's output can be observed.

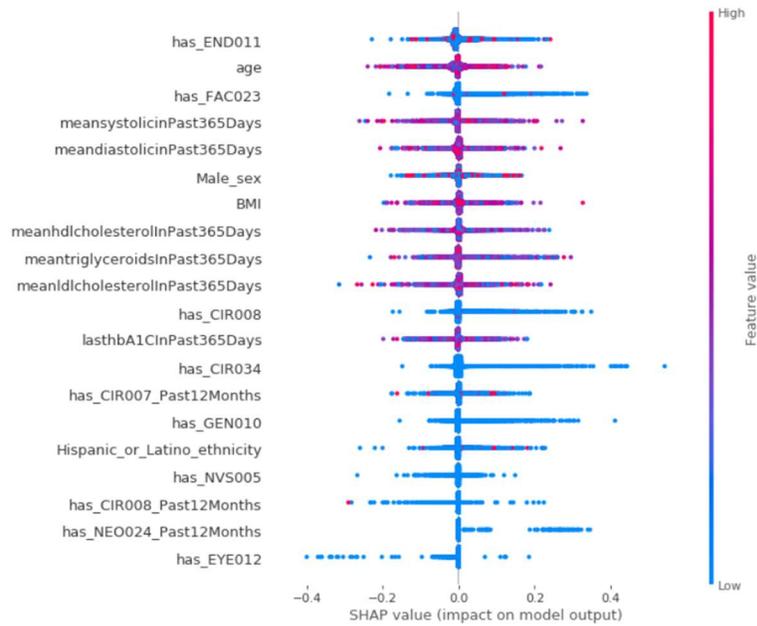

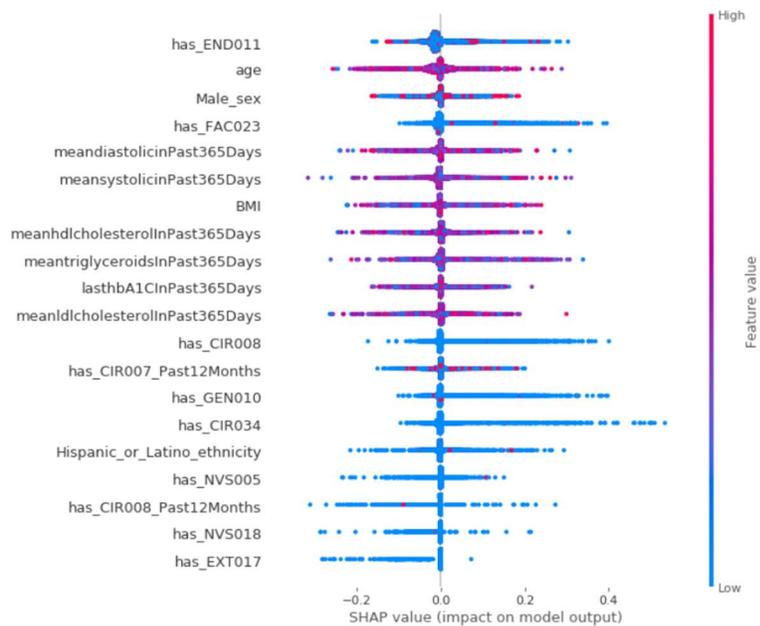

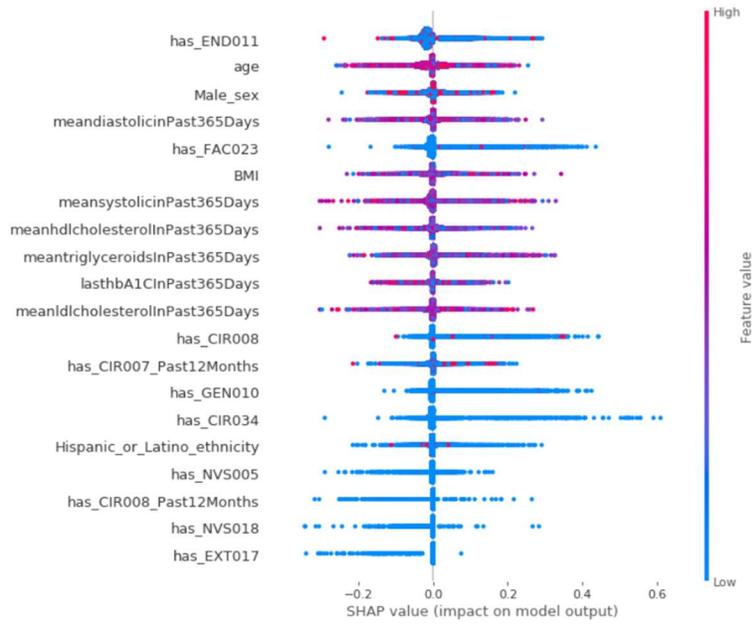

Prediction at 9 months

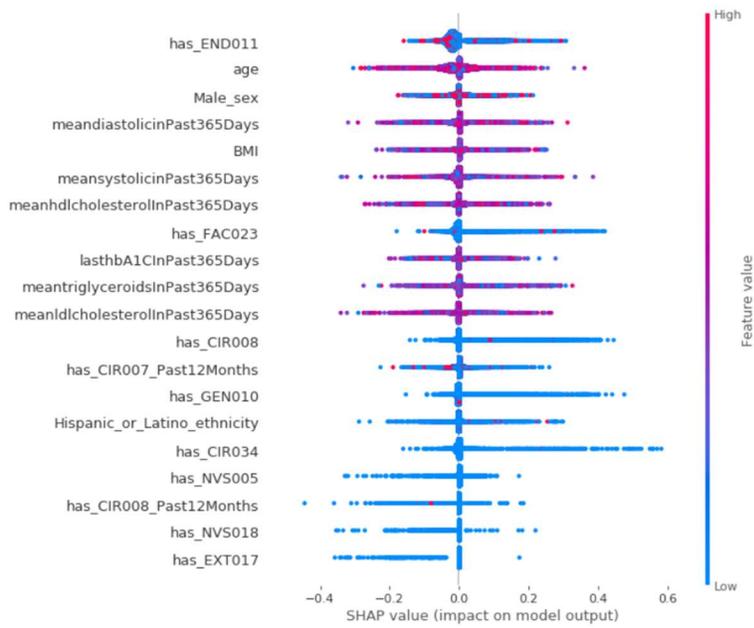

Prediction at 12 months

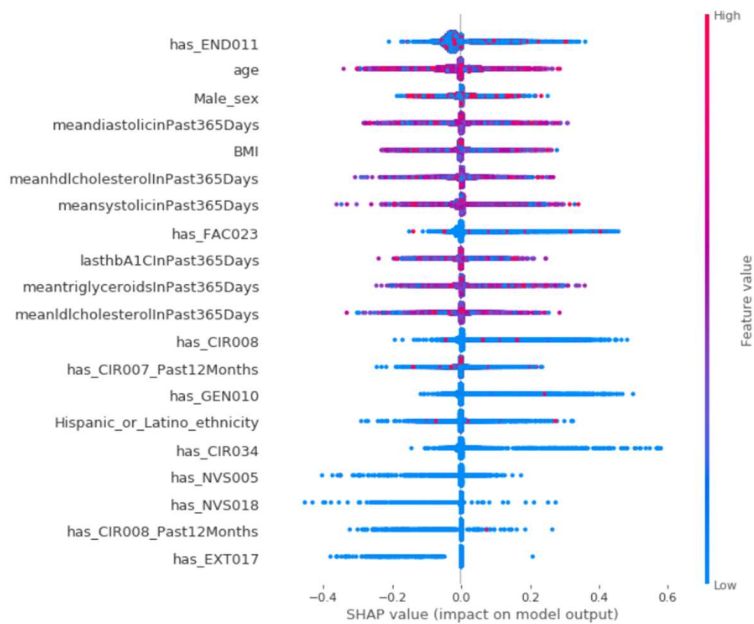

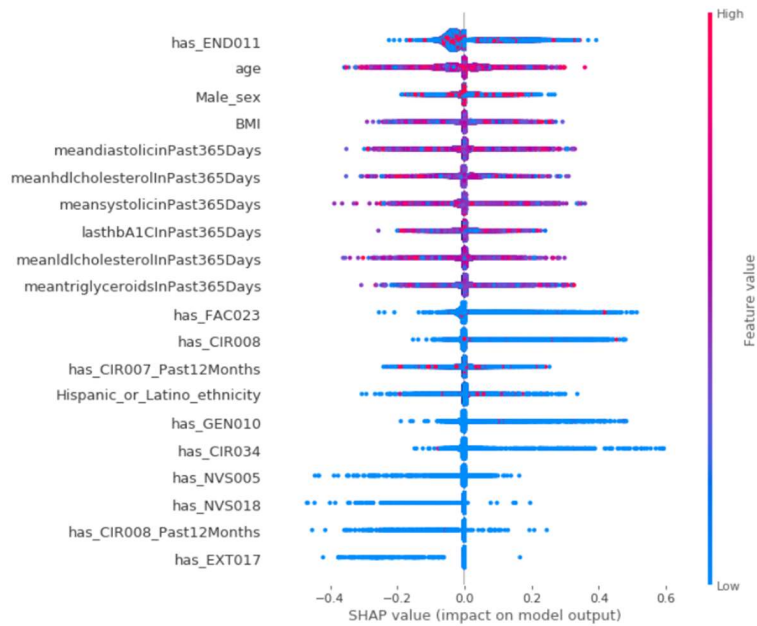

Local

SHAP values were also generated to explain the predictions of individual examples for each future time (3, 6, 9, 12, 18 and 24 months). A total of 3 examples were selected by sampling of risk scores at the 5[th], 50[th] and 95[th] percentile to represent instances at low, medium and high risks respectively.

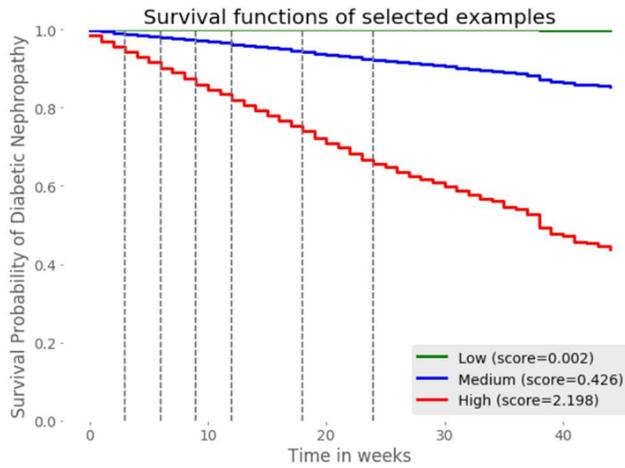

**Low Risk**
Risk Score: 0.002

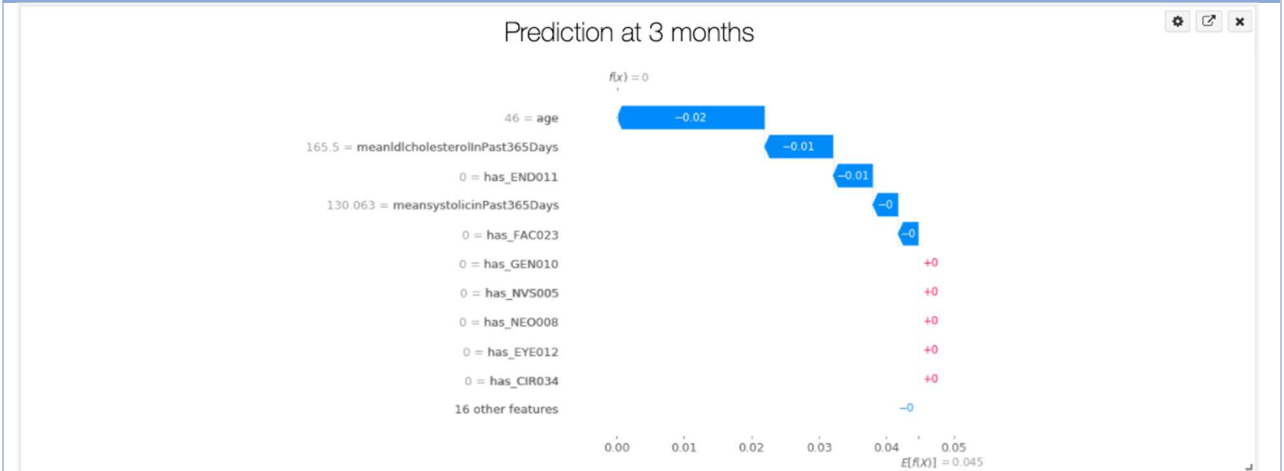

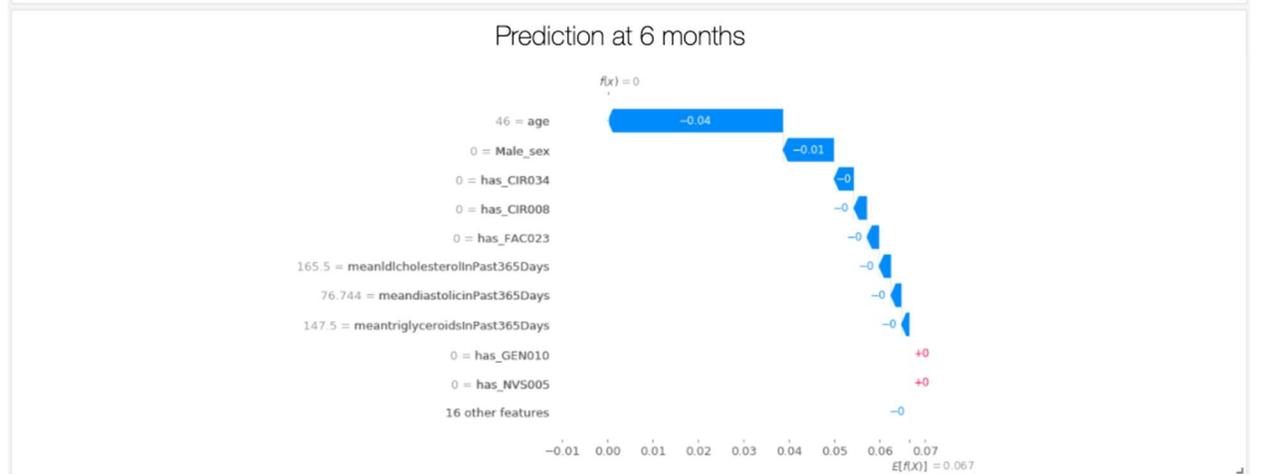

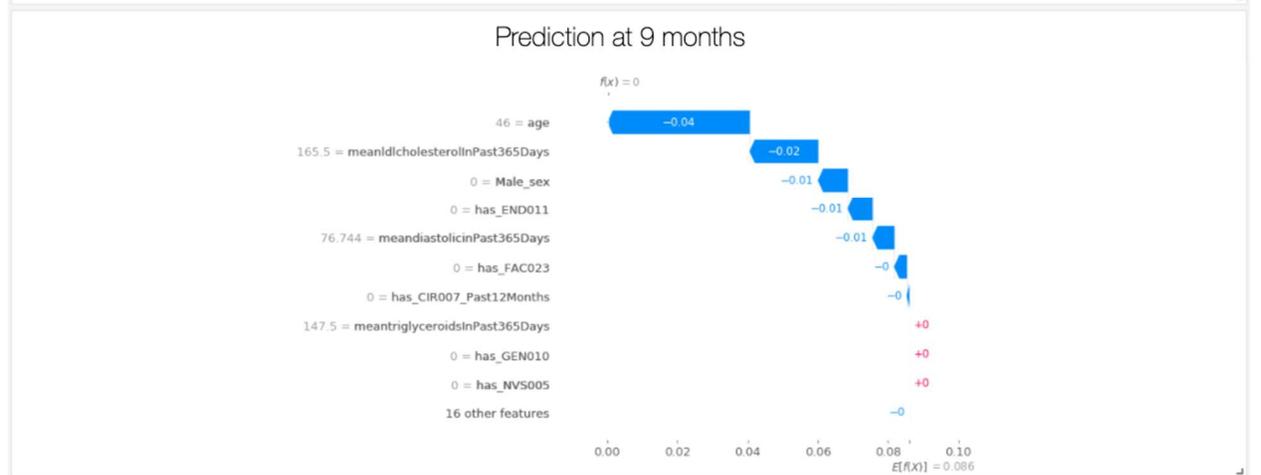

**Low Risk**
Risk Score: 0.002

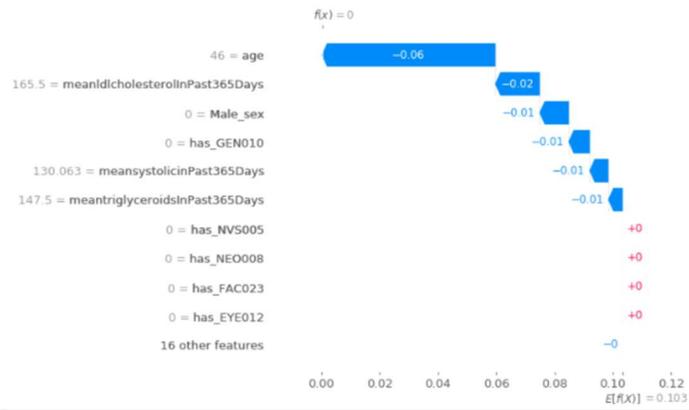

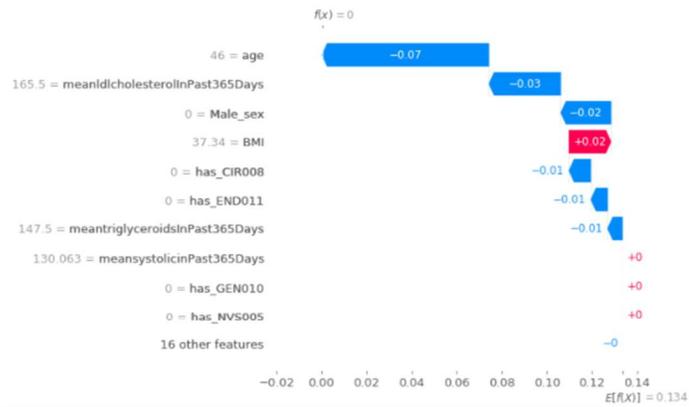

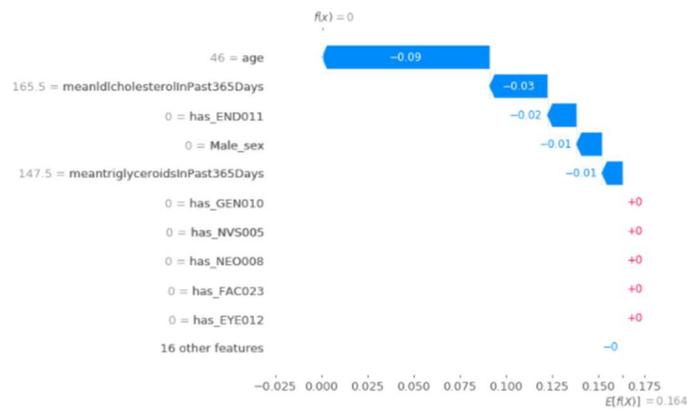

**Medium Risk**
Risk Score: 0.426

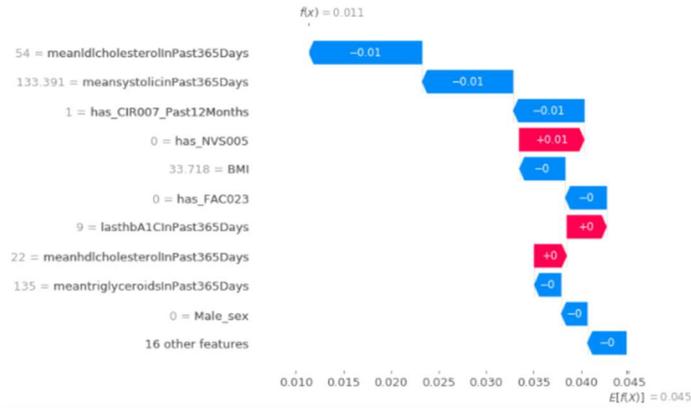

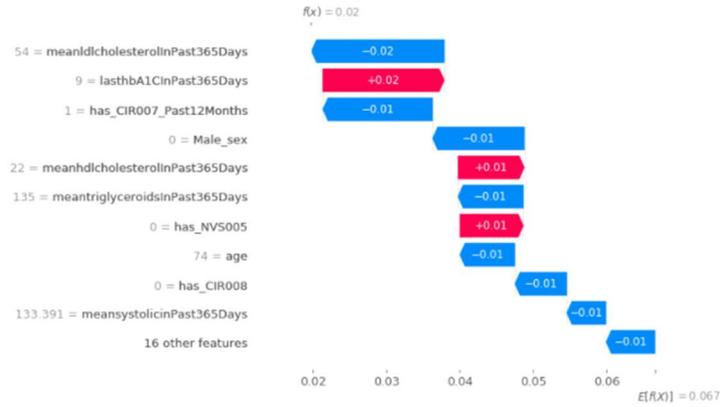

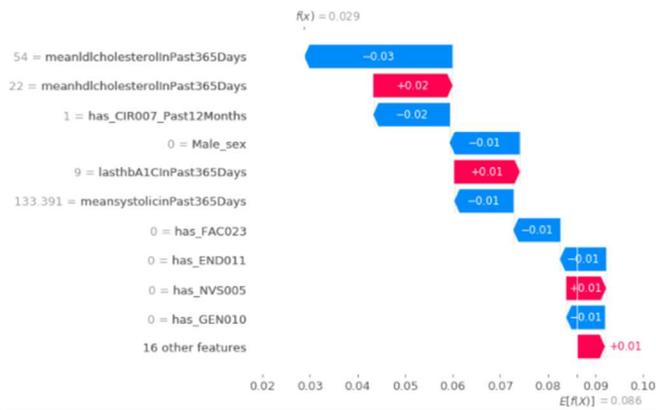

**Medium Risk**
Risk Score: 0.426

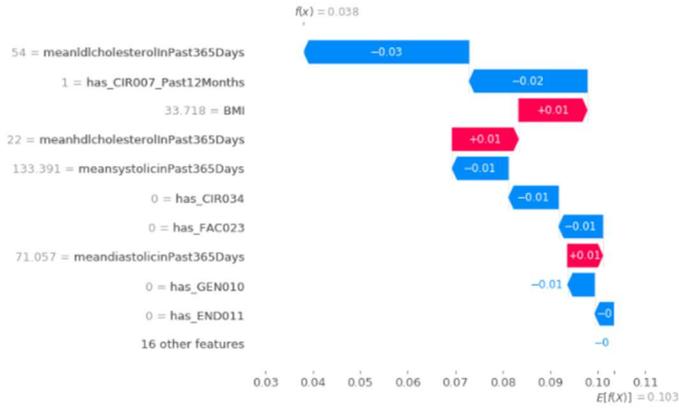

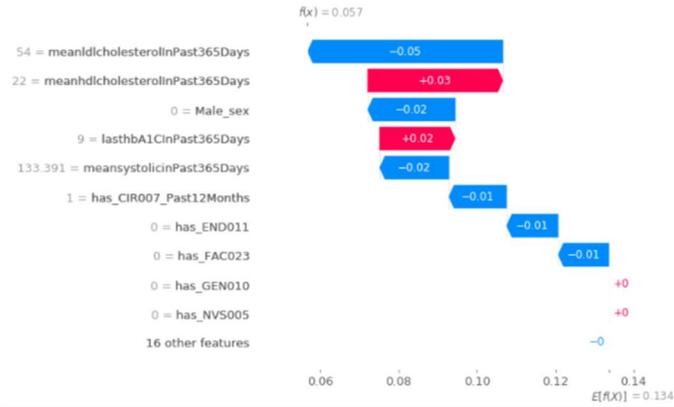

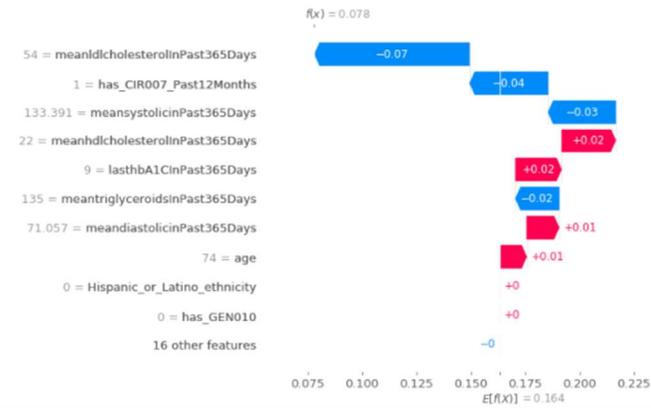

**High Risk**
Risk Score: 2.198

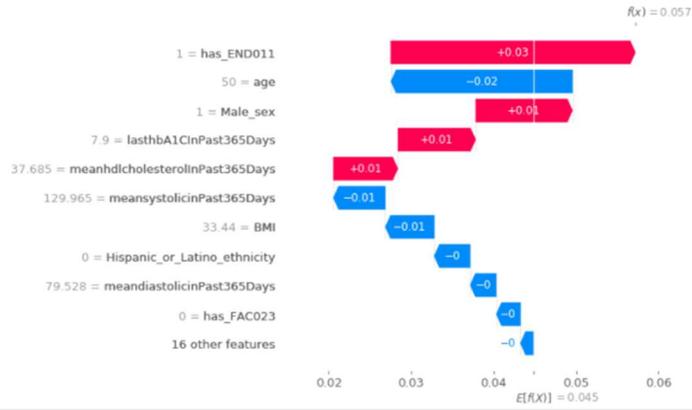

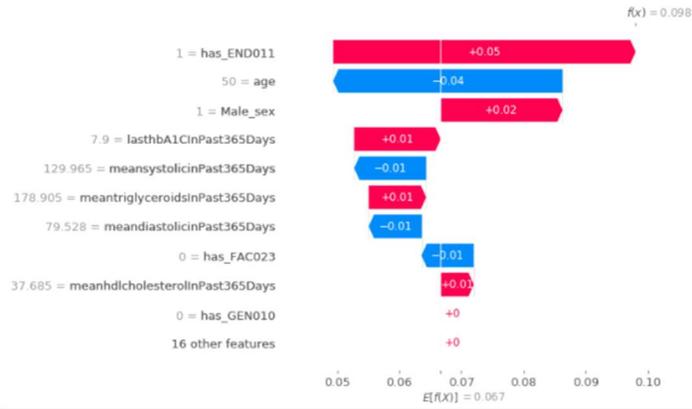

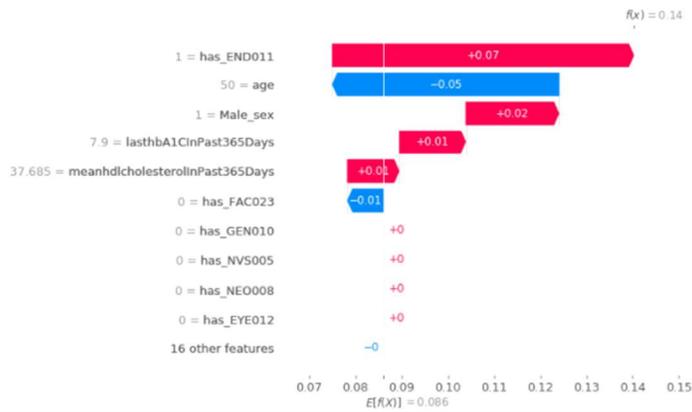

**High Risk**
Risk Score: 2.198

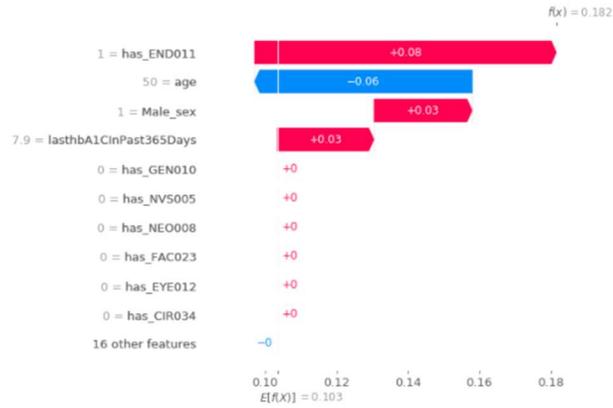

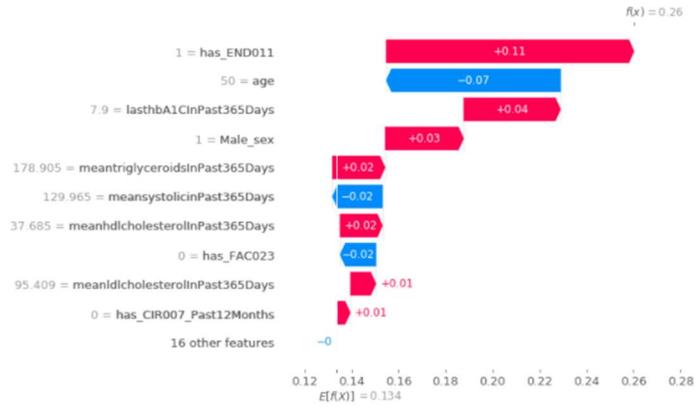

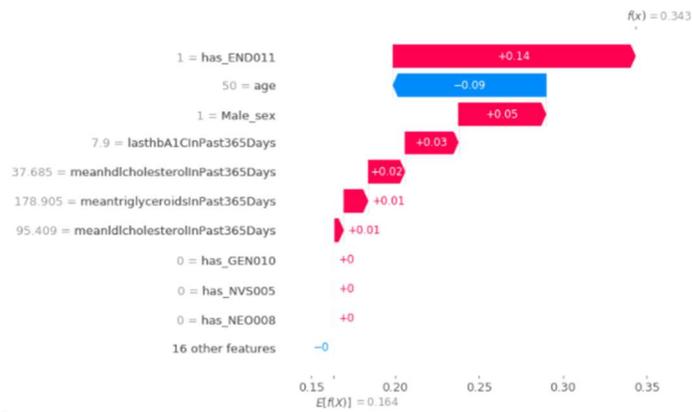

# DM to Diabetic Neuropathy Prediction

| Item | Specification |
| --- | --- |
| **Business Goal** | Enable care managers to identify the patients who are at risk of developing diabetic neuropathy |
| **Usage Setting** | Outpatient |
| **ML Task** | Predict risk and/or time from DM or uncontrolled DM to diabetic neuropathy |
| **ML Class** | Survival |
| **Instances for Prediction** | Encounters |
| **Labels for Instances** | Binary indicator and time to event or censoring for diabetic neuropathy |
| **Cohort Criteria** | <ul><li>2016-01-01 $\leq$ encounter date $\leq$ 2020-06-30 (available Epic data, excluding outliers)</li><li>Encounter date is not within the first 90 days of when the patient entered the data set, to adjust for left-censoring</li><li>18 $\leq$ age $\leq$ 110 (adults without outliers)</li><li>No T1DM diagnosis</li><li>Not pregnant</li><li>No "Do Not Resuscitate" diagnosis</li><li>DM or uncontrolled DM event before encounter date</li><li>No diabetic neuropathy event before encounter date</li><li>No diabetic neuropathy event up to 6 days after encounter date (encounters where diagnoses confirm event within the week)</li></ul> |
| **Input Features** | <ul><li>Demographic</li><li>Diagnosis, except:<ul><li>hasDiabetesNeuropathy*</li><li>has_NVS015* (Polyneuropathies)</li><li>has_NVS017* (Nerve and nerve root disorders)</li><li>has_NVS020* (Other specified nervous system disorders)</li></ul></li><li>Labs</li><li>Utilization</li><li>Vitals</li></ul> See Appendix E for details. |
| **Evaluation Metrics** | <ul><li>Concordance Index</li><li>Integrated Brier Score</li></ul> |

## Data

The following charts summarize the key characteristics of the data after applying the cohort criteria stated above, along with selected features (see Model Signature below).

| Category | Variable | count | mean | stddev | min | 25% | 50% | 75% | max |
|---|---|---|---|---|---|---|---|---|---|
| Demographic | AgeBucket_18_to_39 | 62735.0 | 0.046704 | 0.211007 | 0.00 | 0.000 | 0.000 | 0.000 | 1.00 |
| | AgeBucket_40_to_59 | 62735.0 | 0.277246 | 0.447642 | 0.00 | 0.000 | 0.000 | 1.000 | 1.00 |
| | AgeBucket_60_to_79 | 62735.0 | 0.561266 | 0.496236 | 0.00 | 0.000 | 1.000 | 1.000 | 1.00 |
| | AgeBucket_80_to_109 | 62735.0 | 0.114784 | 0.318764 | 0.00 | 0.000 | 0.000 | 0.000 | 1.00 |
| | Sex_Female | 62735.0 | 0.557201 | 0.496721 | 0.00 | 0.000 | 1.000 | 1.000 | 1.00 |
| | Sex_Male | 62735.0 | 0.442799 | 0.496721 | 0.00 | 0.000 | 0.000 | 1.000 | 1.00 |
| | Ethnicity_Hispanic_or_Latino | 62735.0 | 0.072113 | 0.258677 | 0.00 | 0.000 | 0.000 | 0.000 | 1.00 |
| | Ethnicity_Not_Hispanic_or_Latino | 62735.0 | 0.927887 | 0.258677 | 0.00 | 1.000 | 1.000 | 1.000 | 1.00 |
| Encounter | EncounterType_Emergency | 62735.0 | 0.217327 | 0.412430 | 0.00 | 0.000 | 0.000 | 0.000 | 1.00 |
| | EncounterType_Inpatient | 62735.0 | 0.105220 | 0.306840 | 0.00 | 0.000 | 0.000 | 0.000 | 1.00 |
| | EncounterType_Outpatient | 62735.0 | 0.677453 | 0.467455 | 0.00 | 0.000 | 1.000 | 1.000 | 1.00 |
| Label | Time | 62735.0 | 15.240121 | 11.253042 | 0.00 | 6.000 | 13.000 | 23.000 | 50.00 |
| | Event | 62735.0 | 0.083207 | 0.276197 | 0.00 | 0.000 | 0.000 | 0.000 | 1.00 |
| Feature | age | 62735.0 | 64.341325 | 13.274743 | 18.00 | 56.000 | 66.000 | 73.000 | 102.00 |
| | Male_sex | 62735.0 | 0.442799 | 0.496721 | 0.00 | 0.000 | 0.000 | 1.000 | 1.00 |
| | Hispanic_or_Latino_ethnicity | 62735.0 | 0.072113 | 0.258677 | 0.00 | 0.000 | 0.000 | 0.000 | 1.00 |
| | lasthbA1CInPast365Days | 62735.0 | 7.497466 | 1.188137 | 3.60 | 6.800 | 7.437 | 7.900 | 12.70 |
| | meandiastolicinPast365Days | 62735.0 | 73.863716 | 7.822996 | 43.62 | 69.760 | 73.991 | 78.750 | 107.56 |
| | meansystolicinPast365Days | 62735.0 | 132.057551 | 11.677764 | 82.00 | 127.478 | 131.921 | 135.900 | 180.00 |
| | BMI | 62735.0 | 33.435473 | 6.698827 | 10.37 | 29.360 | 33.670 | 37.040 | 157.25 |
| | meantriglyceroidsInPast365Days | 62735.0 | 165.787573 | 56.316318 | 19.00 | 138.750 | 164.470 | 180.836 | 460.50 |
| | meanldlcholesterolInPast365Days | 62735.0 | 88.837422 | 23.230243 | 8.00 | 81.915 | 89.011 | 98.902 | 201.00 |
| | meanhdlcholesterolInPast365Days | 62735.0 | 43.294861 | 9.027876 | 5.00 | 38.000 | 42.145 | 47.634 | 98.00 |
| | has_INJ004_Past12Months | 62735.0 | 0.006647 | 0.081258 | 0.00 | 0.000 | 0.000 | 0.000 | 1.00 |
| | has_NEO043_Past12Months | 62735.0 | 0.005149 | 0.071570 | 0.00 | 0.000 | 0.000 | 0.000 | 1.00 |
| | has_NEO012_Past12Months | 62735.0 | 0.003459 | 0.058712 | 0.00 | 0.000 | 0.000 | 0.000 | 1.00 |
| | has_NEO030_Past12Months | 62735.0 | 0.034670 | 0.182943 | 0.00 | 0.000 | 0.000 | 0.000 | 1.00 |
| | has_CIR007_Past12Months | 62735.0 | 0.192205 | 0.394037 | 0.00 | 0.000 | 0.000 | 0.000 | 1.00 |
| | has_MBD003_Past12Months | 62735.0 | 0.006344 | 0.079398 | 0.00 | 0.000 | 0.000 | 0.000 | 1.00 |
| | has_CIR008_Past12Months | 62735.0 | 0.009421 | 0.096602 | 0.00 | 0.000 | 0.000 | 0.000 | 1.00 |
| | has_FAC013 | 62735.0 | 0.008321 | 0.090838 | 0.00 | 0.000 | 0.000 | 0.000 | 1.00 |
| | has_DIG006 | 62735.0 | 0.006934 | 0.082982 | 0.00 | 0.000 | 0.000 | 0.000 | 1.00 |
| | has_NEO065 | 62735.0 | 0.004702 | 0.068413 | 0.00 | 0.000 | 0.000 | 0.000 | 1.00 |
| | has_FAC004 | 62735.0 | 0.001435 | 0.037849 | 0.00 | 0.000 | 0.000 | 0.000 | 1.00 |
| | has_GEN025 | 62735.0 | 0.041859 | 0.200268 | 0.00 | 0.000 | 0.000 | 0.000 | 1.00 |
| | has_RSP017 | 62735.0 | 0.011015 | 0.104372 | 0.00 | 0.000 | 0.000 | 0.000 | 1.00 |
| | has_NVS004 | 62735.0 | 0.009022 | 0.094556 | 0.00 | 0.000 | 0.000 | 0.000 | 1.00 |
| | has_MUS037 | 62735.0 | 0.011923 | 0.108541 | 0.00 | 0.000 | 0.000 | 0.000 | 1.00 |
| | has_CIR035 | 62735.0 | 0.009644 | 0.097729 | 0.00 | 0.000 | 0.000 | 0.000 | 1.00 |

(Percentages for binary variables can be read from the "mean" column.)

### Encounters and Patients

| | Examples | Encounters | Patients |
|---|---|---|---|
| 1 | 62735 | 62735 | 11539 |

### Train and Test Sets

| | Set | No Event | No Event % | Event | Event % | Total | Total % |
|---|---|---|---|---|---|---|---|
| 1 | Test | 17003 | 29.6 | 1540 | 29.5 | 18543 | 29.6 |
| 2 | Train | 40512 | 70.4 | 3680 | 70.5 | 44192 | 70.4 |
| 3 | Total | 57515 | 100 | 5220 | 100 | 62735 | 100 |

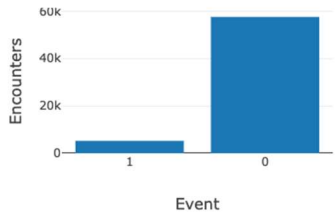
Encounters with Event (1) or Censoring (0)

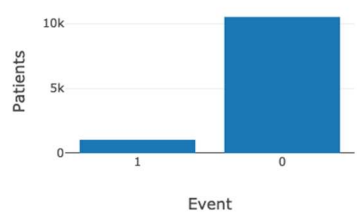
Patients with Event (1) or Censoring (0)

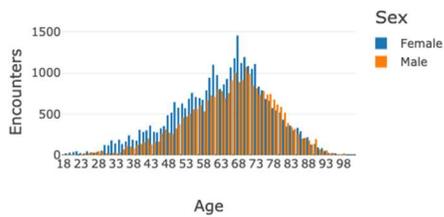
Encounters by Age and Sex

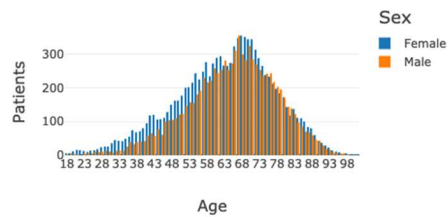
Patients by Age and Sex

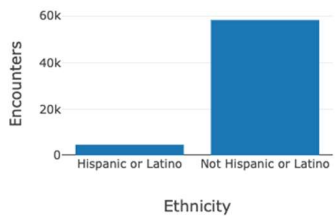
Encounters by Ethnicity

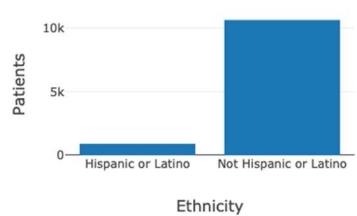
Patients by Ethnicity

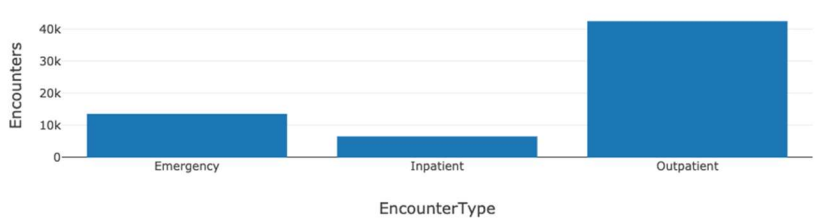
Encounters by Encounter Type

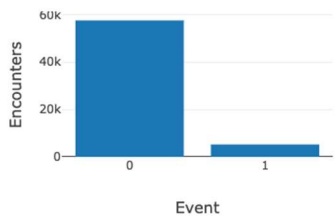
Encounters by Time to Event (1) or Cen…

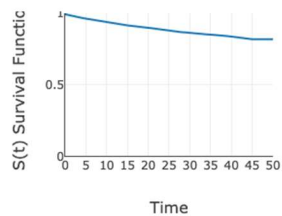
Kaplan-Meier Estimate of Survival Functi…

## Model Signature

The model signature has 26 features, comprising of 12 mandatory features and 14 other selected features. These are the selected features, in rank order (the last feature to be eliminated is ranked 1):

1. has_DIG006 (Gastrointestinal and biliary perforation)
2. has_FAC013 (Contraceptive and procreative management)
3. has_CIR035 (Varicose veins of lower extremity)
4. has_FAC004 (Encounter for prophylactic or other procedures)
5. has_NEO030_Past12Months (Breast cancer - all other types)
6. has_NEO012_Past12Months (Gastrointestinal cancers - esophagus)
7. has_MUS037 (Postprocedural or postoperative musculoskeletal system complication)
8. has_NEO043_Past12Months (Urinary system cancers - bladder)
9. has_NEO065 (Multiple myeloma)
10. has_INJ004_Past12Months (Fracture of the upper limb, initial encounter)
11. has_MBD003_Past12Months (Bipolar and related disorders)
12. has_RSP017 (Postprocedural or postoperative respiratory system complication)
13. has_NVS004 (Parkinson`s disease)
14. has_GEN025 (Other specified female genital disorders)
15. has_CIR008_Past12Months (Hypertension with complications and secondary hypertension)
16. has_CIR007_Past12Months (Essential hypertension)
17. Hispanic_or_Latino_ethnicity
18. Male_sex
19. lasthbA1CInPast365Days
20. meandiastolicinPast365Days
21. meansystolicinPast365Days
22. BMI
23. meanhdlcholesterolInPast365Days
24. meanldlcholesterolInPast365Days
25. age
26. meantriglyceroidsInPast365Days

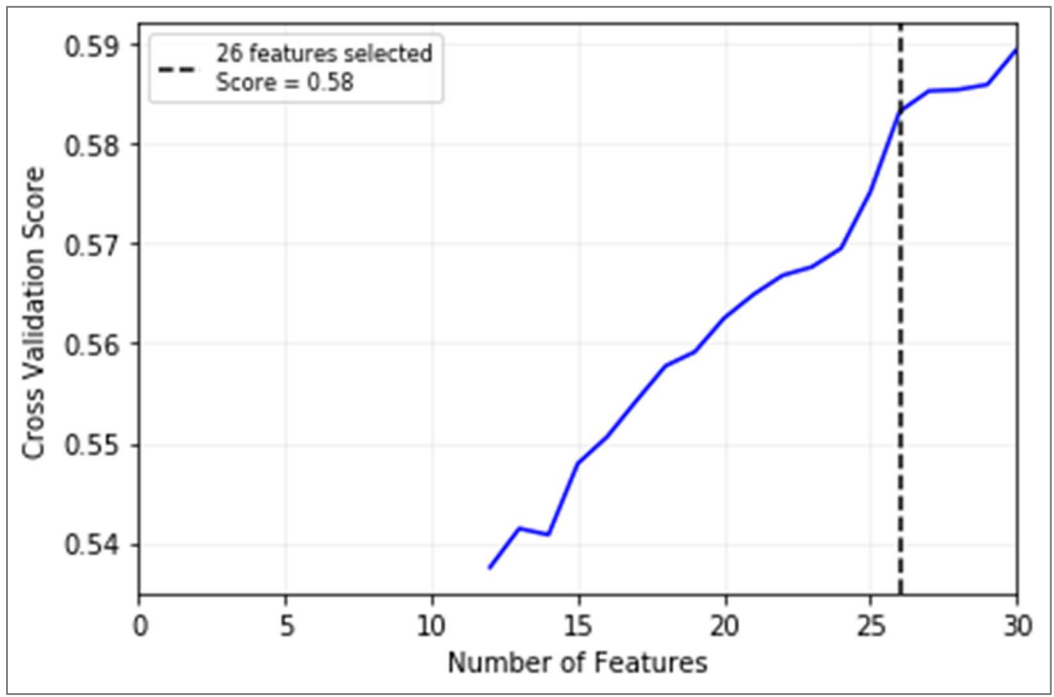

## Model Performance

The following table and chart summarize the performance of all candidate models on the test set for this prediction task in terms of the Concordance Index and the Integrated Brier Score.

| Model | No. of Parameter Combinations Successfully Tested | Concordance Index | Integrated Brier Score |
|---|---|---|---|
| **CoxPH** | 101 | 0.59 | 0.09 |
| **DeepSurv** | 168 | **0.72** | **0.08** |
| **RSF** | 67 | 0.64 | 0.09 |
| **CSF** | 36 | 0.63 | 0.09 |
| **EST** | 39 | 0.67 | 0.09 |

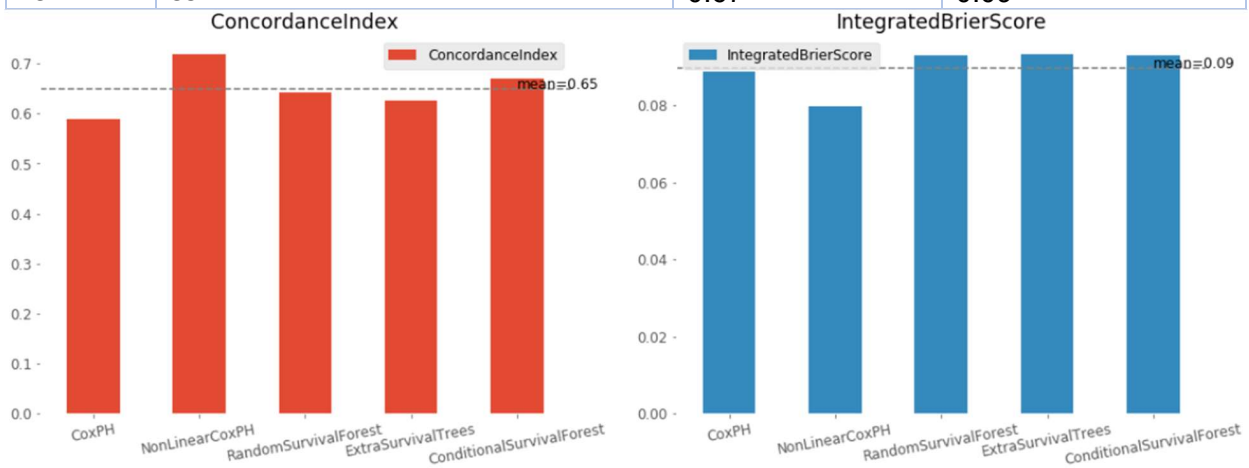

The following chart shows how the average survival function curves of the candidate models compare to the KM survival curve, the more similar their curves are to the KM survival curve the better.

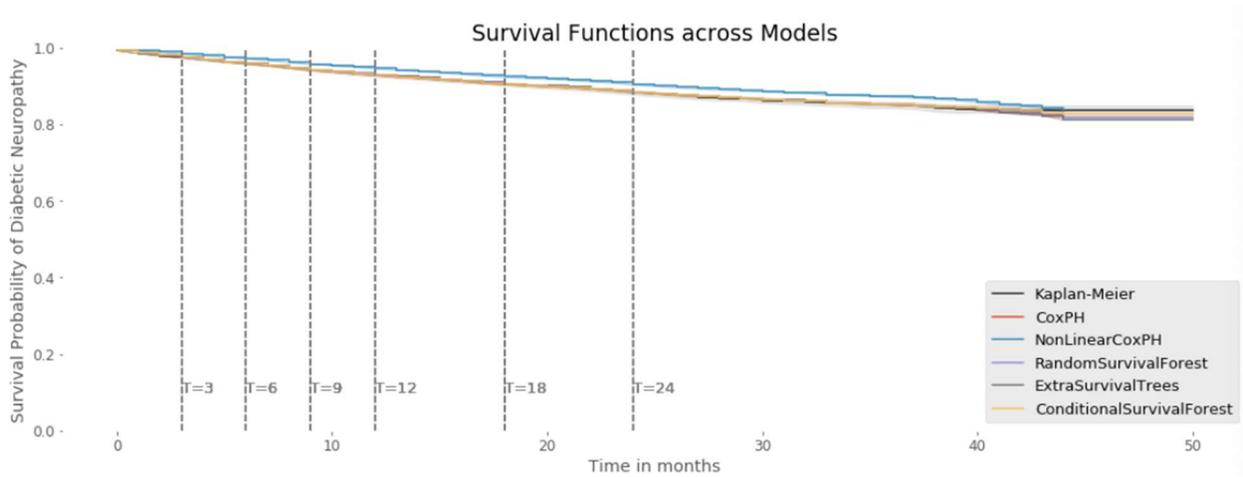

This chart shows the change in the Brier Score over time for all candidate models, the closer the scores are to 0 the better.

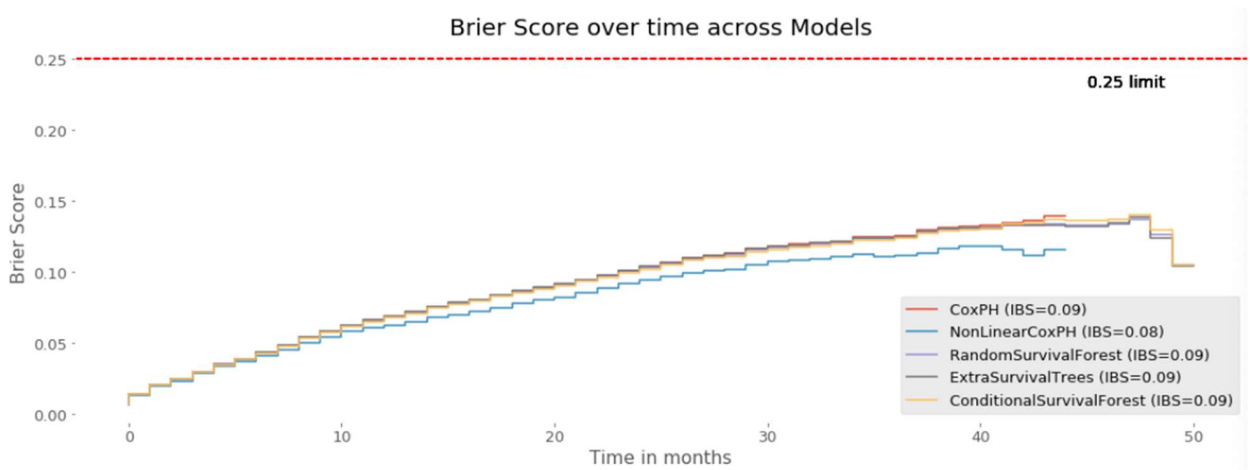

## Model Evaluation of Selected Model (DeepSurv)

### Overall

This chart shows the change in the Brier Score over time for the selected model, the closer the scores are to 0 the better.

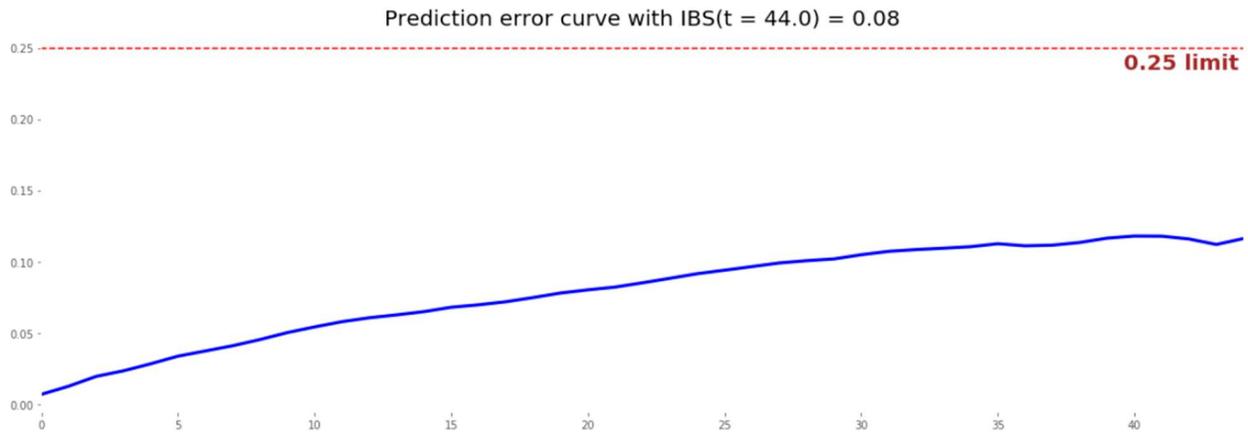

The following chart shows the actual vs. predicted density functions, i.e. number of instances that get the disease / complication at each time point and the RMSE, Median Absolute Error and Mean Absolute Error across the time points.

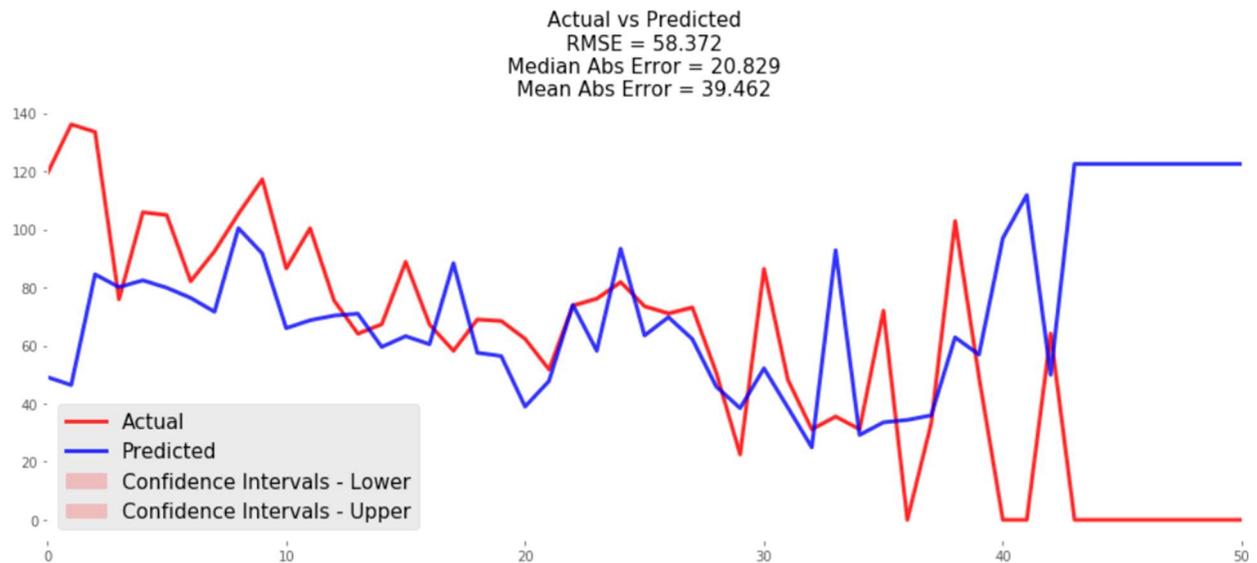

The following chart shows the actual vs. predicted survival functions, i.e. the number of instances that have not had the disease / complication by each time point and the RMSE, Median Absolute Error and Mean Absolute Error across the time points.

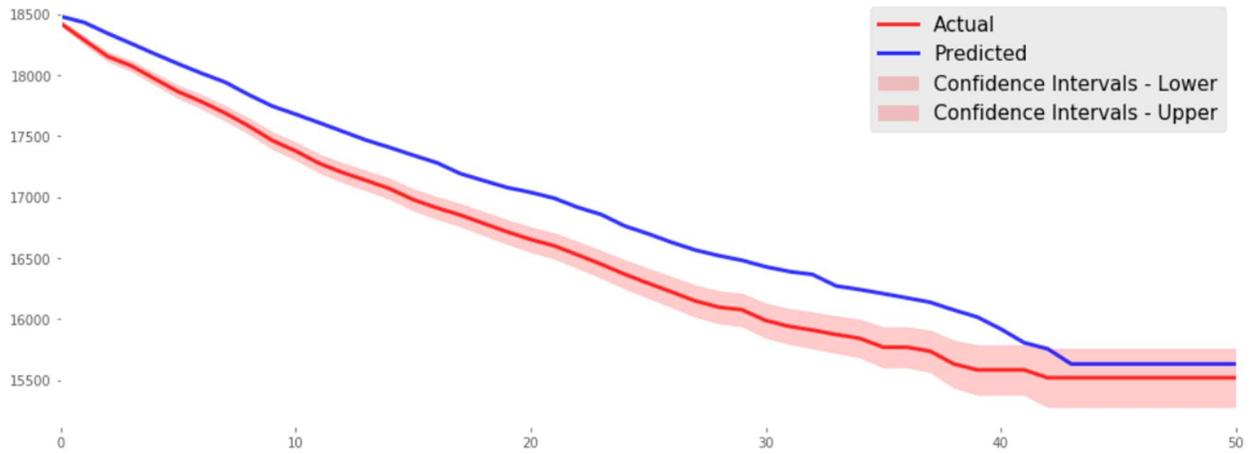

Risk Stratification

The low, medium and high risk groups are defined as examples with predicted risk scores belonging to the first quartile, second to third quartiles, and fourth quartile respectively.

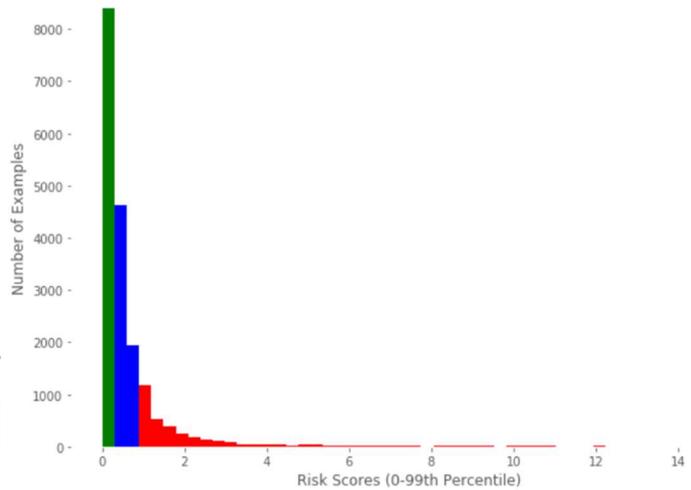

| | Percentile | Risk Scores |
|---|---|---|
| **Low Risk** | 0-25th | [0.0,0.11] |
| **Medium Risk** | 26th-75th | [0.11,0.718] |
| **High Risk** | 75th-100th | [0.718,433357.625] |

## Summary Metrics across Subgroups

The table below displays the summary metrics across subgroups of risk, age, sex, ethnicity and patient history.

| Category | Subgroup | Cohort Size | Concordance Index | Brier Score | Mean AUC | Mean Specificity | Mean Sensitivity | S(t), t=3 | S(t), t=6 | S(t), t=9 | S(t), t=12 | S(t), t=18 | S(t), t=24 |
|---|---|---|---|---|---|---|---|---|---|---|---|---|---|
| NaN | Overall | 18543.00 | 0.72 | 0.08 | 0.73 | 0.73 | 0.62 | 0.97 | 0.96 | 0.94 | 0.93 | 0.91 | 0.88 |
| Risk | Low | 4635.00 | 0.62 | 0.04 | 0.56 | 1.00 | 0.00 | 1.00 | 1.00 | 1.00 | 1.00 | 1.00 | 0.99 |
| Risk | Medium | 9272.00 | 0.56 | 0.07 | 0.59 | 0.89 | 0.18 | 0.99 | 0.99 | 0.98 | 0.97 | 0.96 | 0.94 |
| Risk | High | 4636.00 | 0.65 | 0.15 | 0.71 | 0.02 | 0.99 | 0.95 | 0.91 | 0.87 | 0.84 | 0.78 | 0.73 |
| Age Bucket | 18 to 39 | 825.00 | 0.80 | 0.04 | 0.81 | 0.84 | 0.65 | 0.99 | 0.97 | 0.96 | 0.95 | 0.93 | 0.91 |
| Age Bucket | 40 to 59 | 5142.00 | 0.76 | 0.07 | 0.74 | 0.70 | 0.64 | 0.99 | 0.97 | 0.96 | 0.95 | 0.93 | 0.91 |
| Age Bucket | 60 to 79 | 10436.00 | 0.69 | 0.09 | 0.71 | 0.73 | 0.59 | 0.99 | 0.97 | 0.96 | 0.95 | 0.93 | 0.91 |
| Age Bucket | 80 to 109 | 2140.00 | 0.75 | 0.08 | 0.80 | 0.70 | 0.71 | 0.98 | 0.96 | 0.94 | 0.93 | 0.91 | 0.88 |
| Sex | Male | 8244.00 | 0.69 | 0.08 | 0.71 | 0.73 | 0.58 | 0.98 | 0.97 | 0.95 | 0.94 | 0.92 | 0.90 |
| Sex | Female | 10299.00 | 0.74 | 0.08 | 0.75 | 0.72 | 0.65 | 0.99 | 0.97 | 0.96 | 0.95 | 0.93 | 0.90 |
| Ethnicity | Hispanic or Latino | 1360.00 | 0.85 | 0.07 | 0.83 | 0.60 | 0.82 | 0.97 | 0.95 | 0.92 | 0.90 | 0.87 | 0.84 |
| Ethnicity | Not Hispanic or Latino | 17183.00 | 0.71 | 0.08 | 0.73 | 0.74 | 0.61 | 0.99 | 0.97 | 0.96 | 0.95 | 0.93 | 0.91 |
| History Bucket | <= 6 | 549.00 | 0.73 | 0.06 | 0.62 | 0.71 | 0.52 | 0.99 | 0.98 | 0.96 | 0.95 | 0.93 | 0.91 |
| History Bucket | 7 to 12 | 1735.00 | 0.71 | 0.07 | 0.74 | 0.75 | 0.64 | 0.99 | 0.97 | 0.96 | 0.95 | 0.93 | 0.91 |
| History Bucket | 13 to 24 | 5206.00 | 0.70 | 0.09 | 0.71 | 0.72 | 0.60 | 0.98 | 0.96 | 0.95 | 0.94 | 0.91 | 0.89 |
| History Bucket | 25 to 36 | 5363.00 | 0.79 | 0.05 | 0.76 | 0.74 | 0.65 | 0.99 | 0.97 | 0.96 | 0.95 | 0.93 | 0.91 |
| History Bucket | 37 to 60 | 5690.00 | 0.71 | 0.04 | nan | nan | 0.61 | 0.99 | 0.98 | 0.96 | 0.95 | 0.93 | 0.91 |

Concordance Index & Integrated Brier Score

The following charts show how the Concordance Index and Integrated Brier Score varies among subgroups of risk, age, sex, ethnicity and patient history.

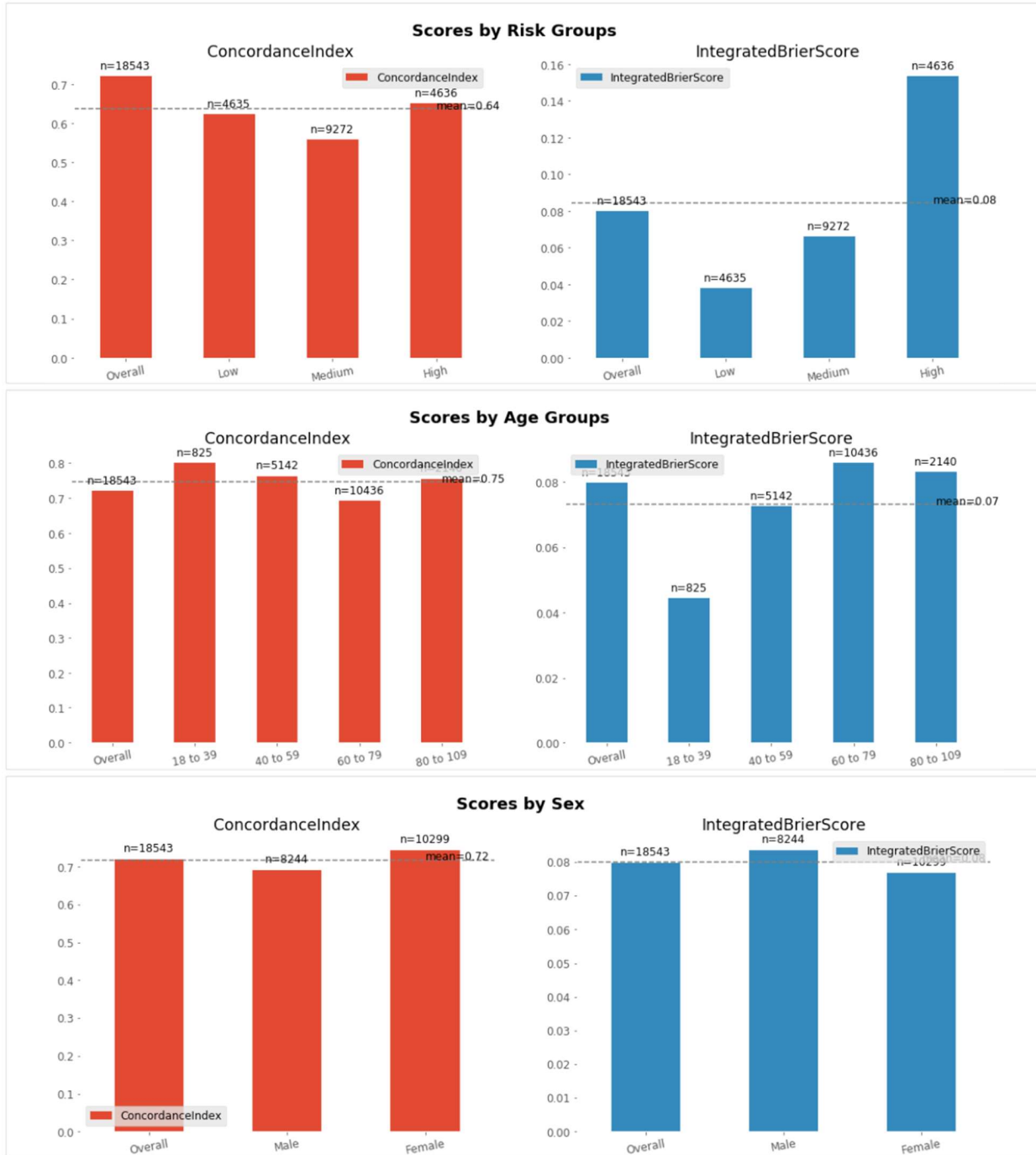

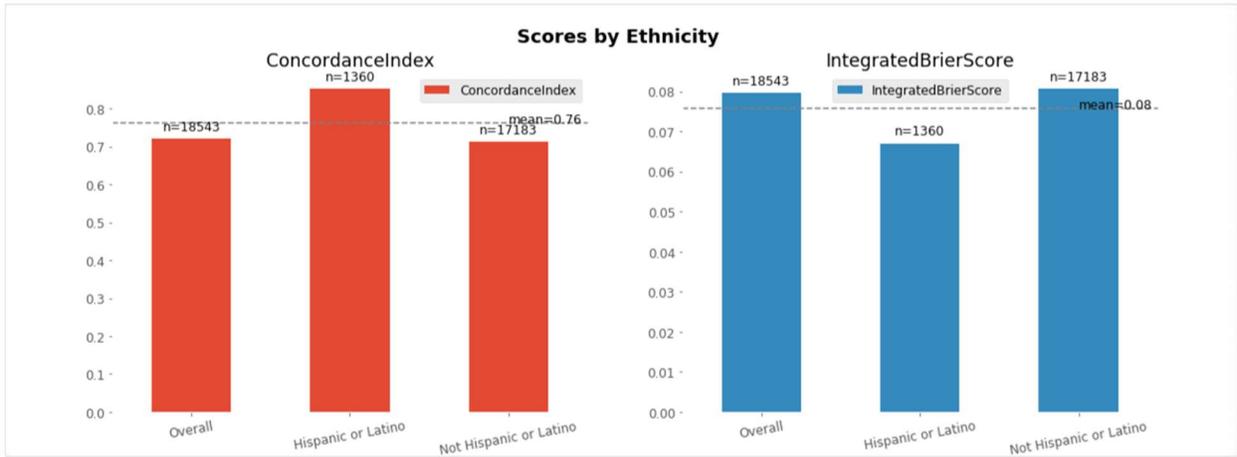
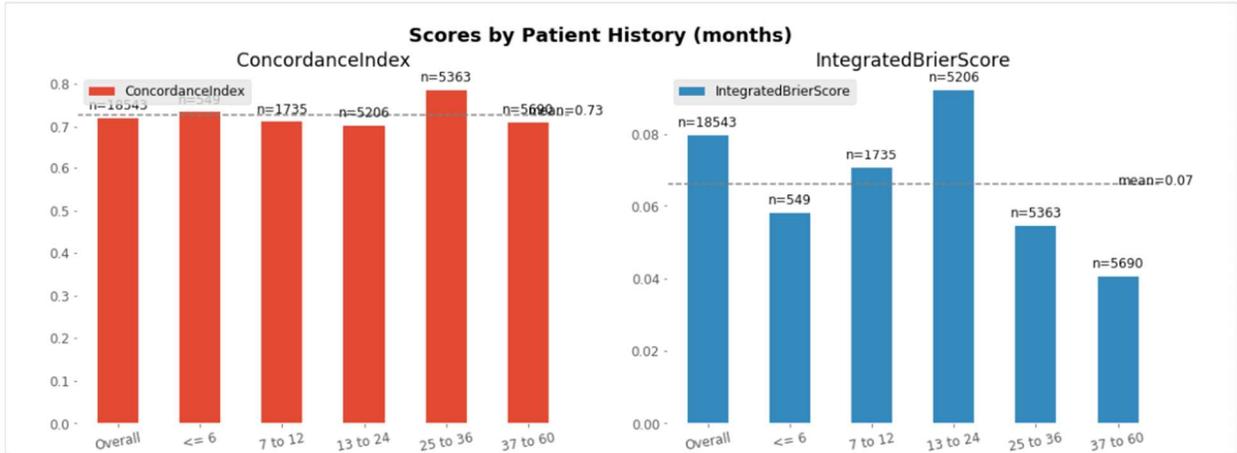

Average Survival Function Curves

The following charts show how the average survival function curve varies among subgroups of risk, age, sex, ethnicity and patient history.

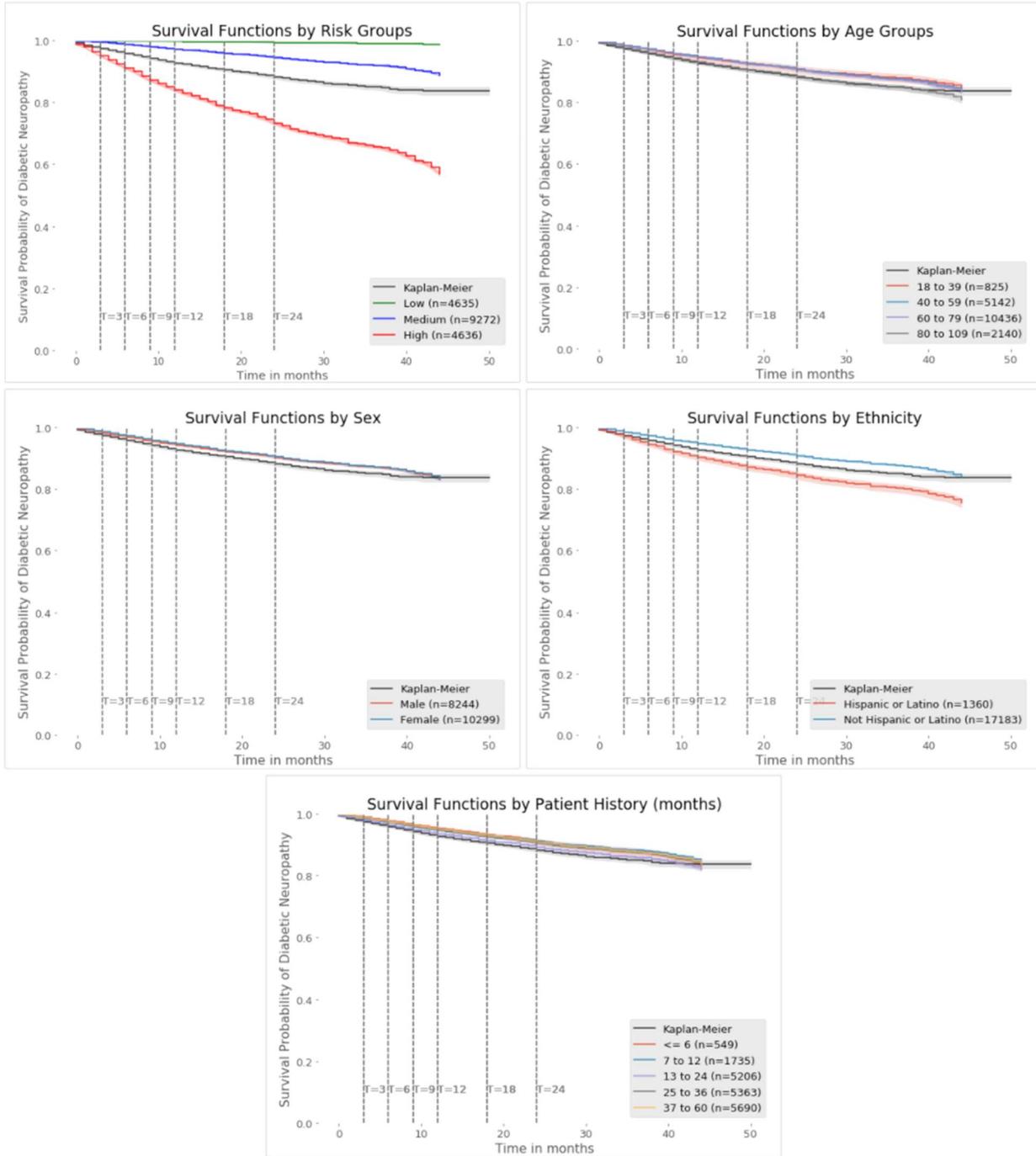

Time-dependent AUC

The following charts show how the AUC across time varies among subgroups of risk, age, sex, ethnicity and patient history.

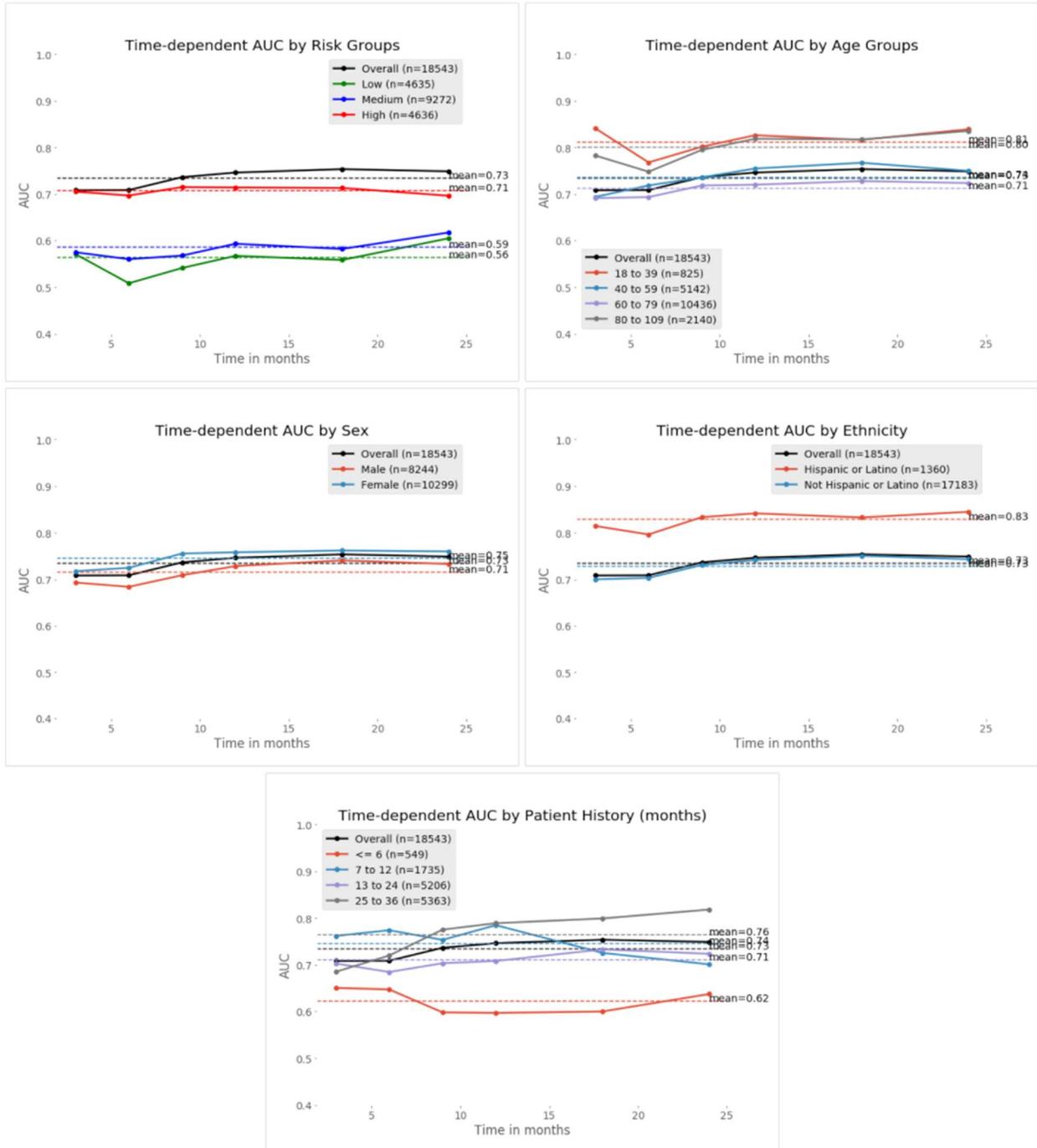

Time-dependent Specificity

The following charts show how the specificity across time varies among subgroups of risk, age, sex, ethnicity and patient history.

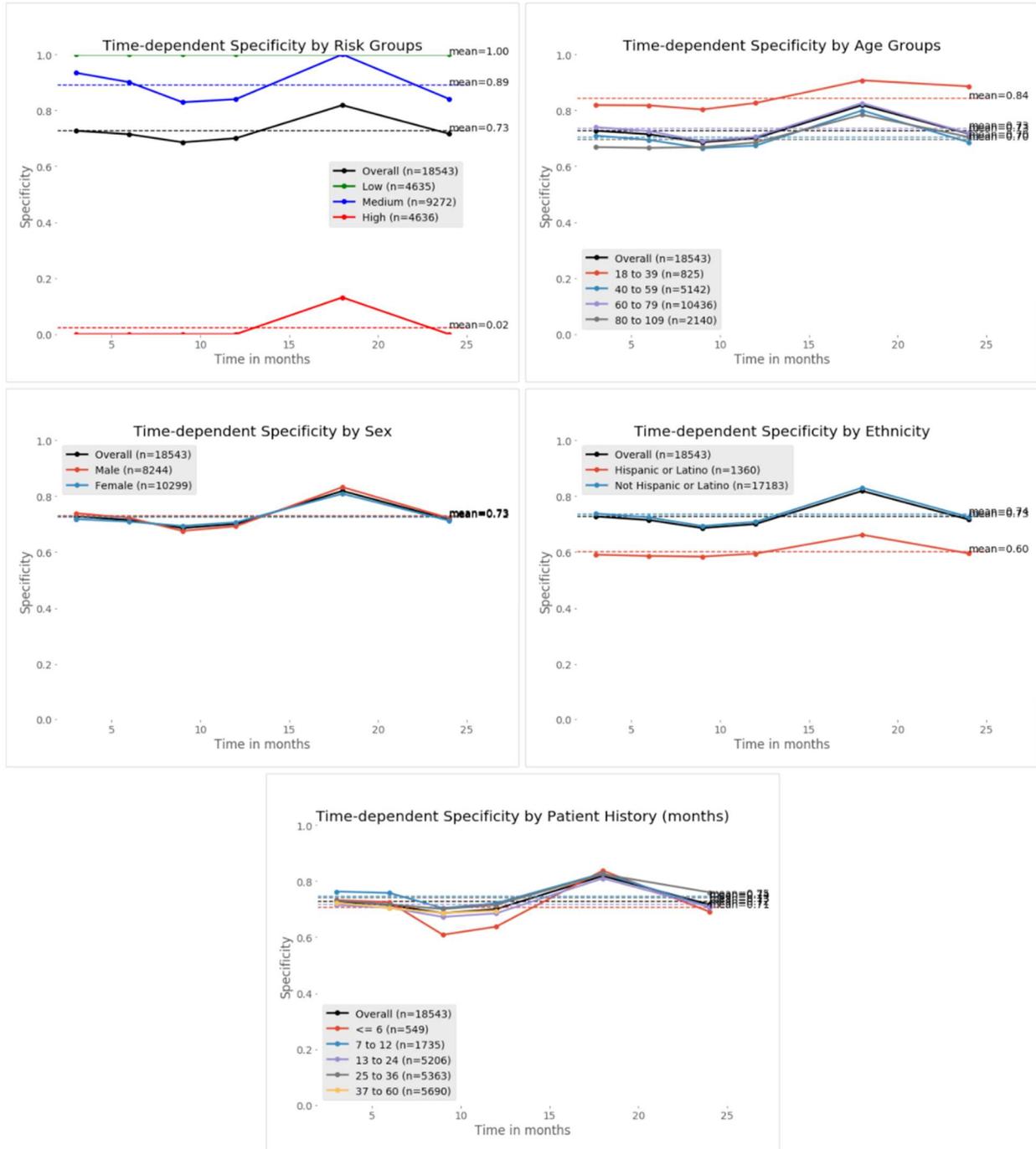

Time-dependent Sensitivity

The following charts show how the sensitivity across time varies among subgroups of risk, age, sex, ethnicity and patient history.

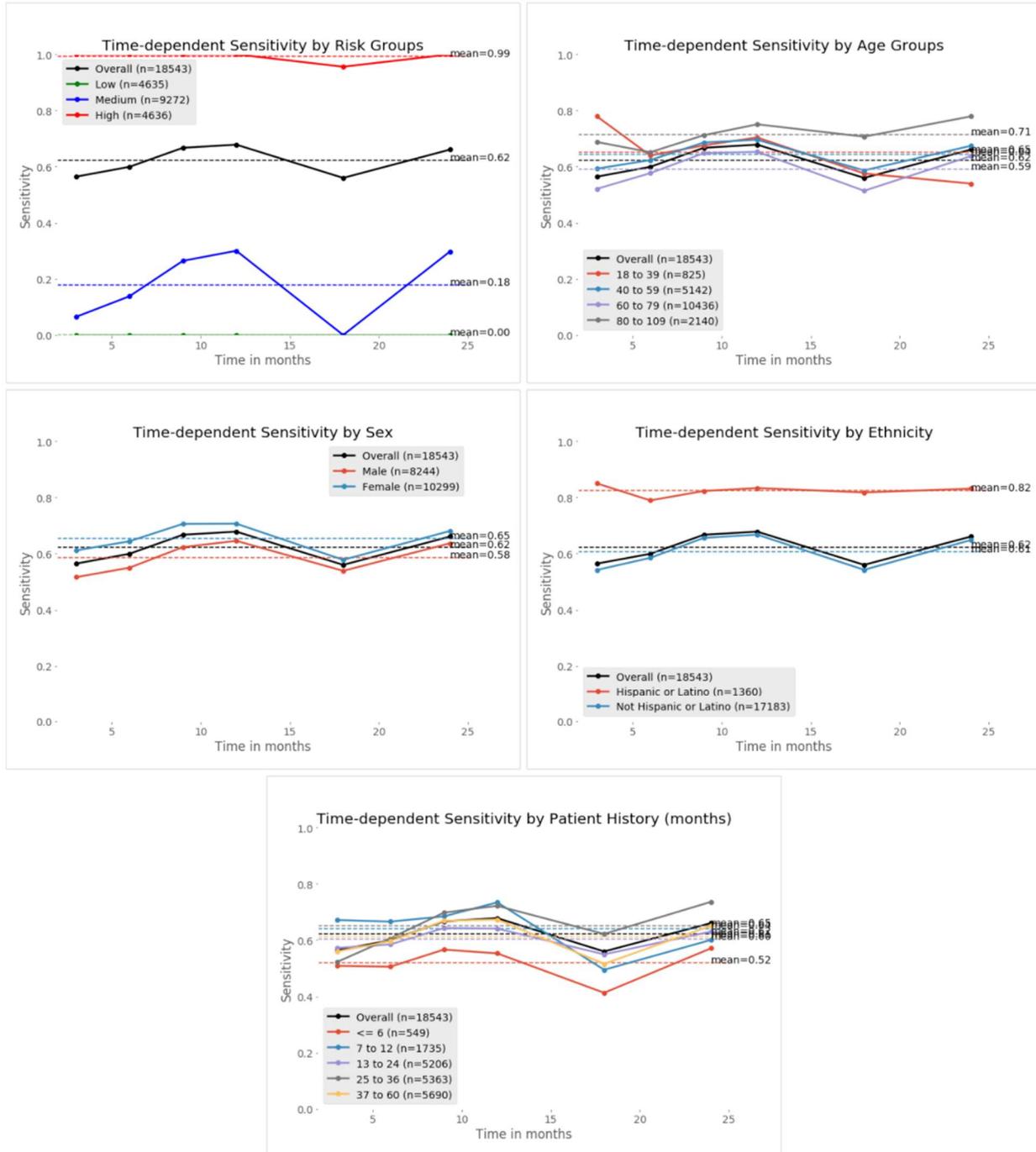

## Model Explanation (DeepSurv)

### Global

The following plots show the SHAP values of each instance in the training set for each future time (3, 6, 9, 12, 18 and 24 months). The features are sorted by the total magnitude of the SHAP values over all instances and the distribution of the effect that each feature has on the model's output can be observed.

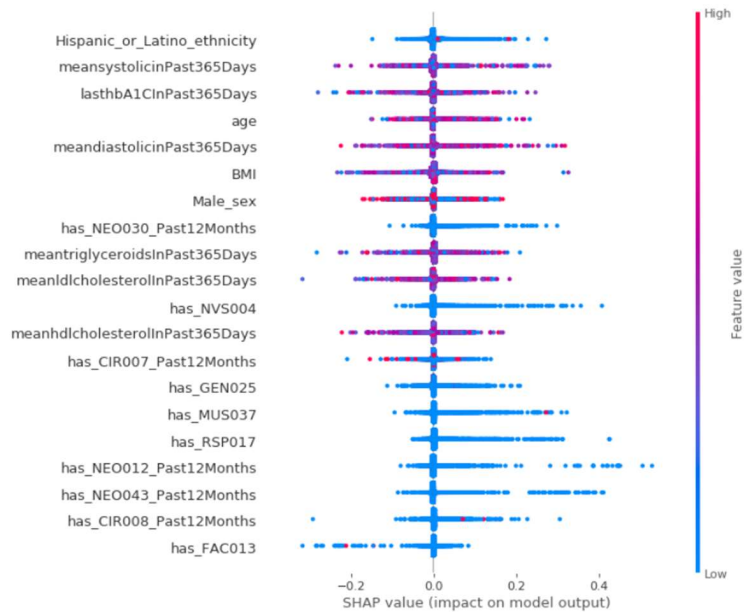

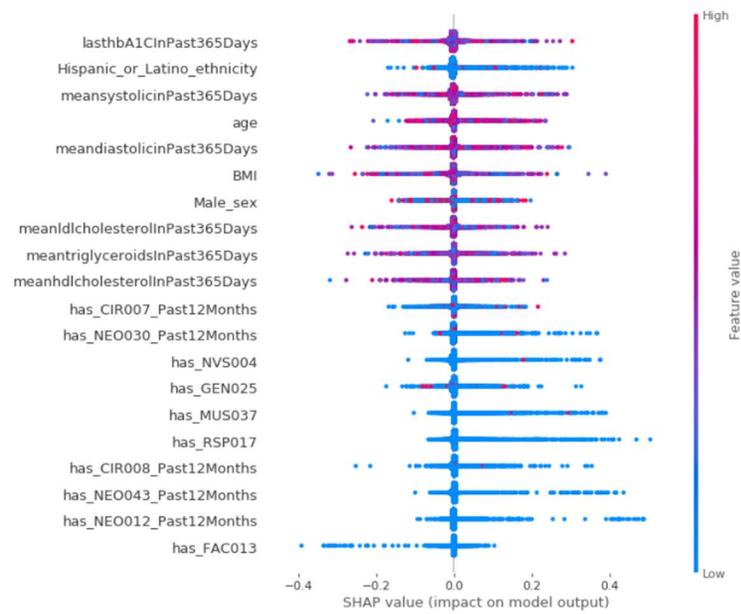

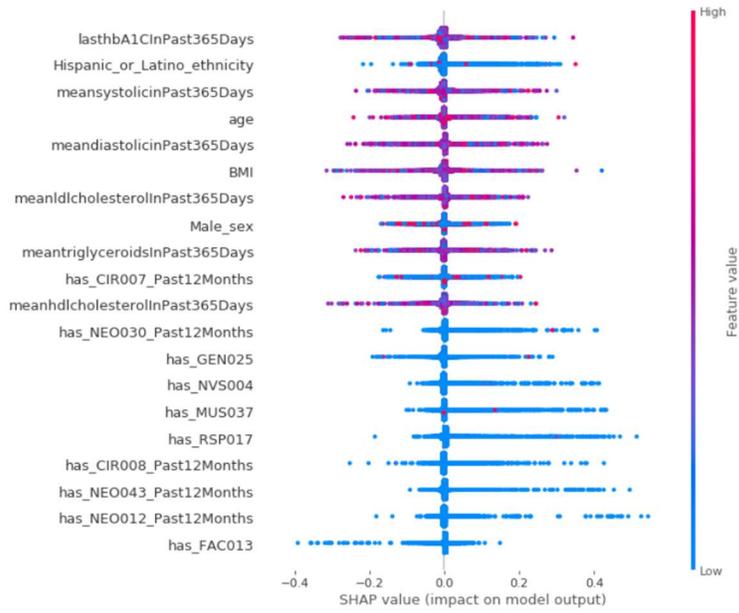

Prediction at 9 months

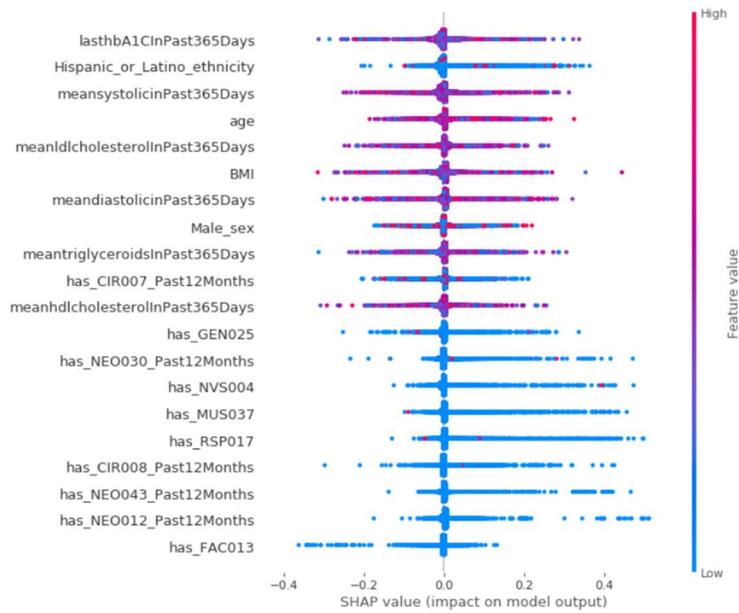

Prediction at 12 months

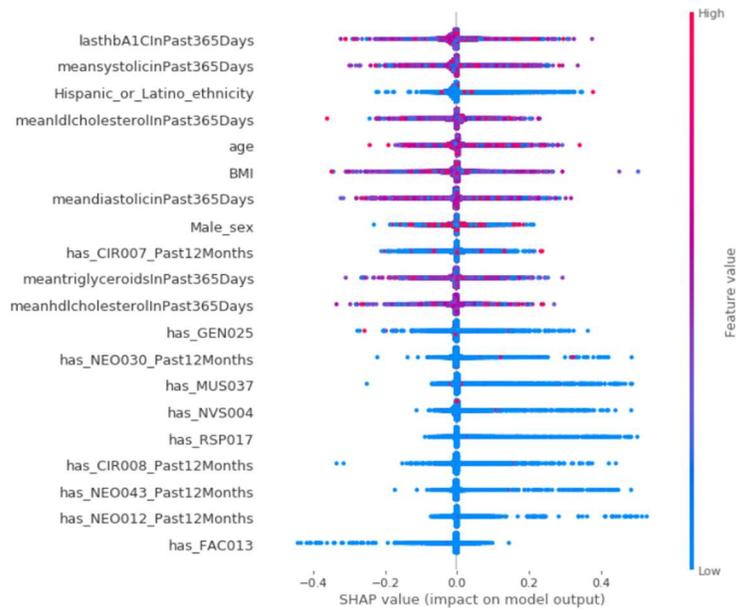

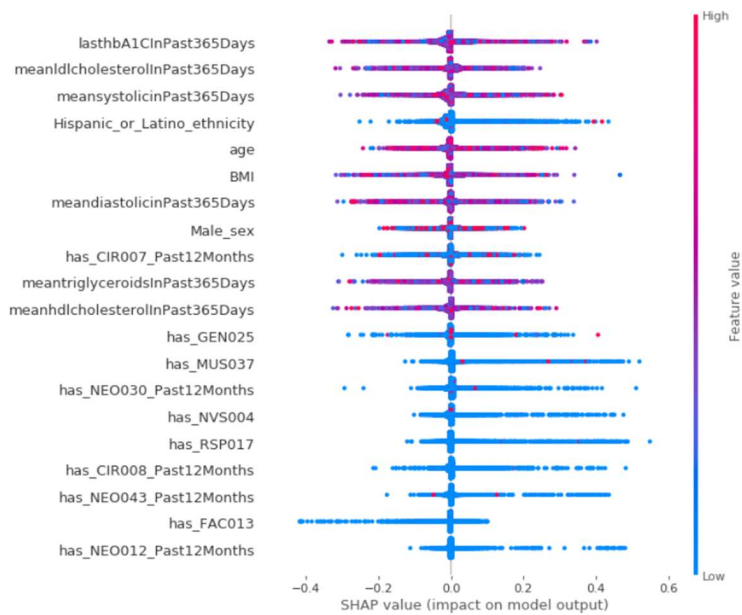

Local

SHAP values were also generated to explain the predictions of individual examples for each future time (3, 6, 9, 12, 18 and 24 months). A total of 3 examples were selected by sampling of risk scores at the 5th, 50th and 95th percentile to represent instances at low, medium and high risks respectively.

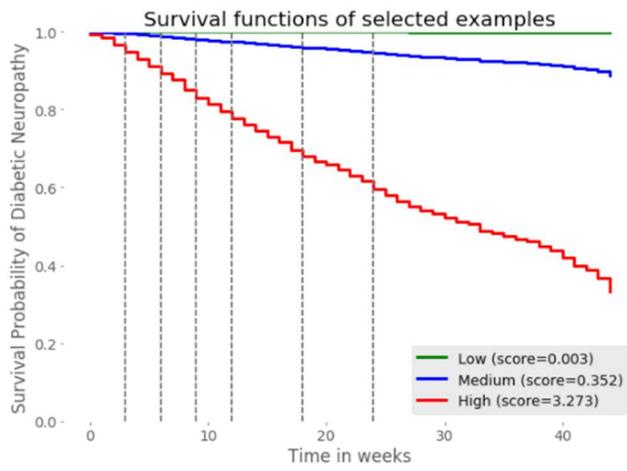

**Low Risk**
Risk Score: 0.003

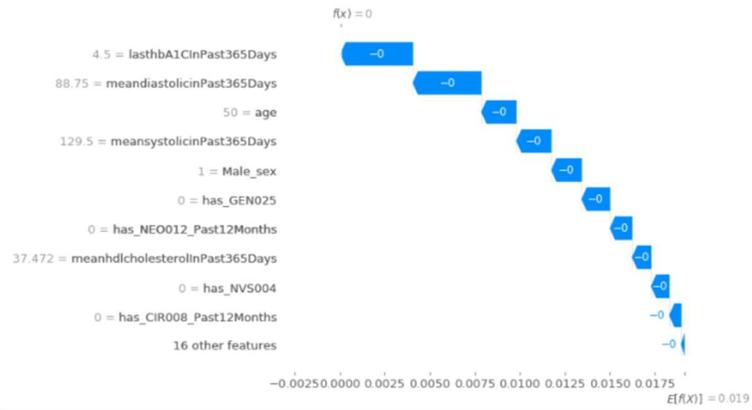

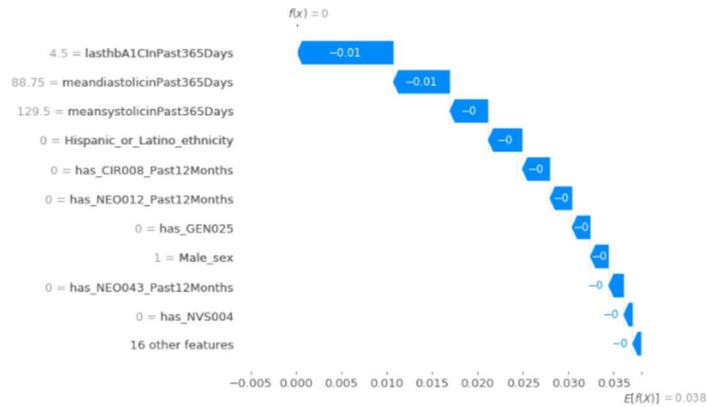

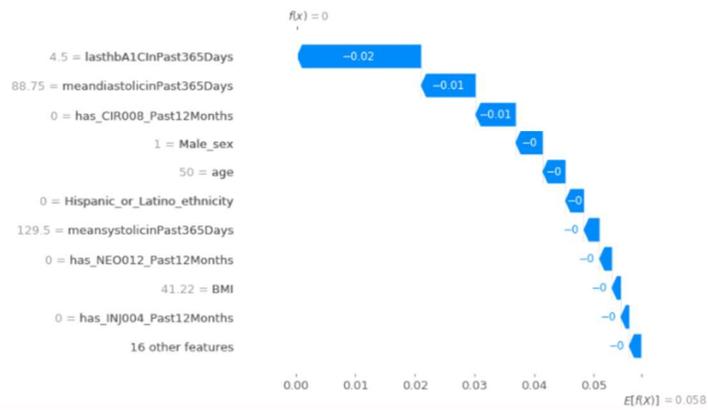

**Low Risk**
Risk Score: 0.003

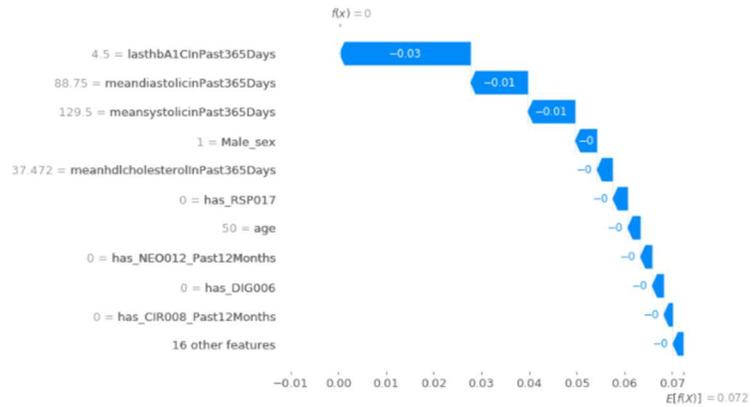

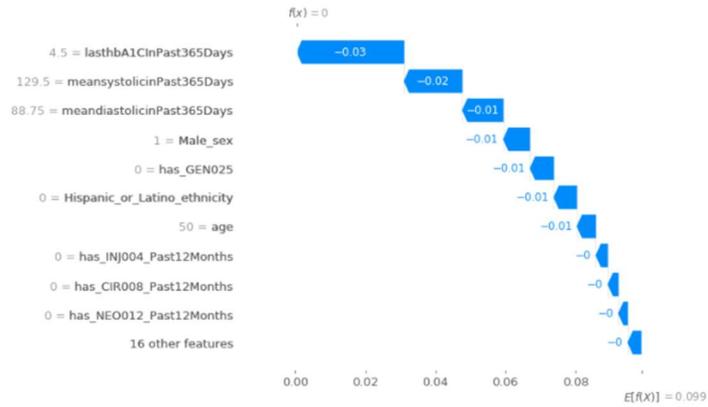

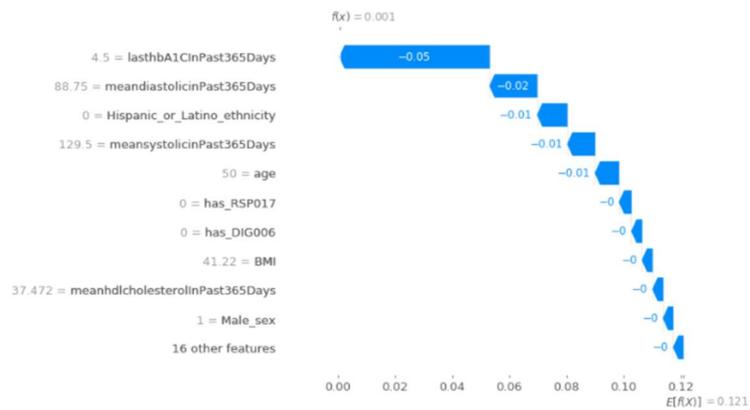

**Medium Risk**
Risk Score: 0.352

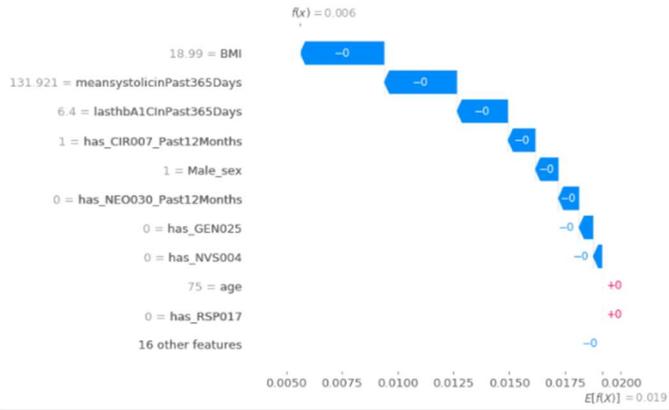

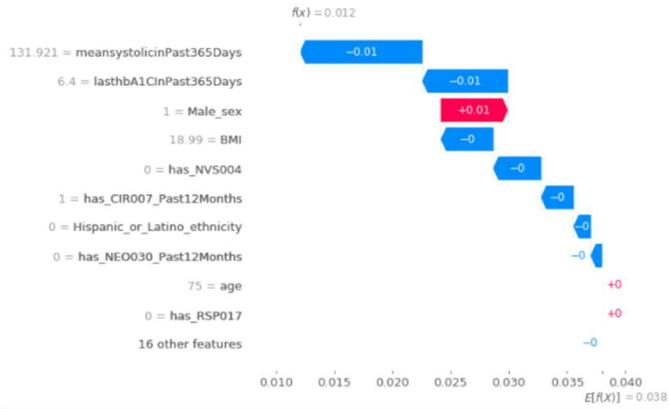

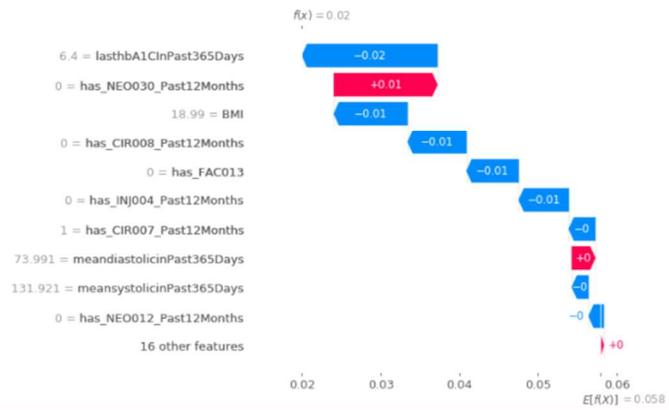

**Medium Risk**
Risk Score: 0.352

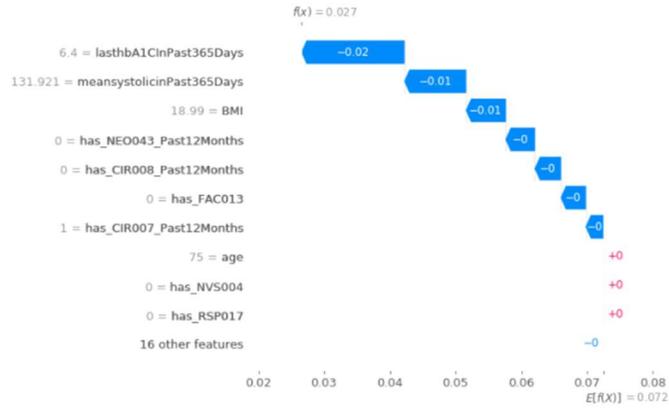

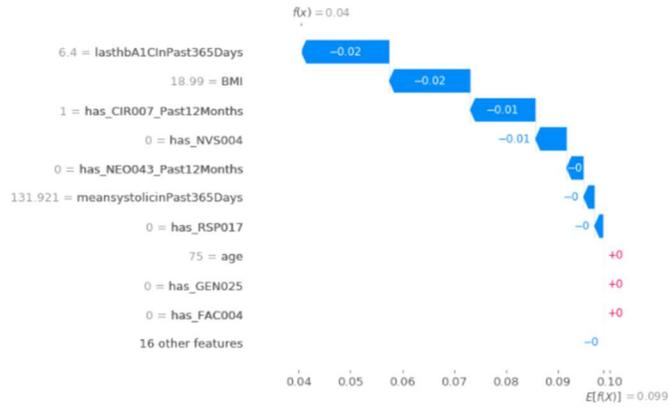

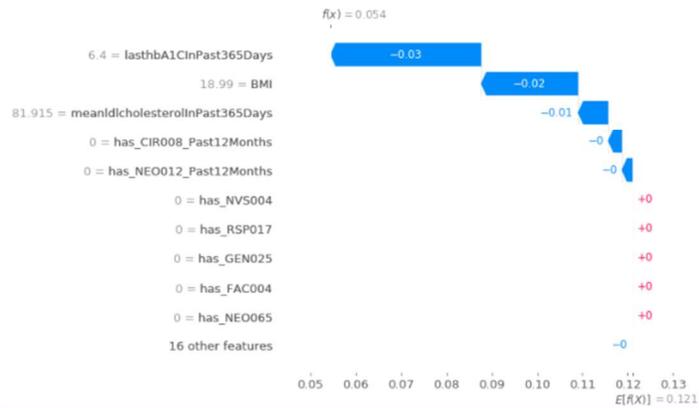

**High Risk**
Risk Score: 3.273

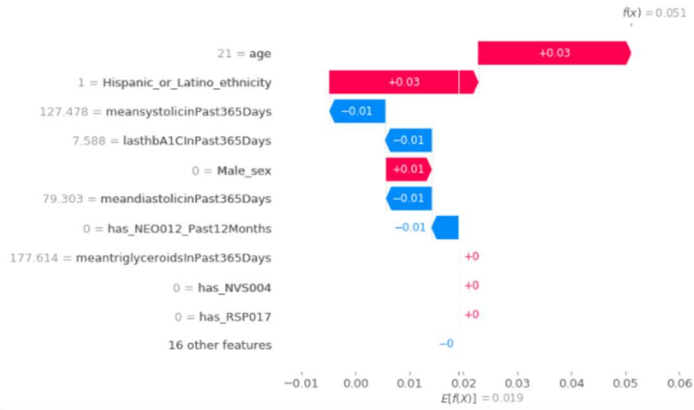

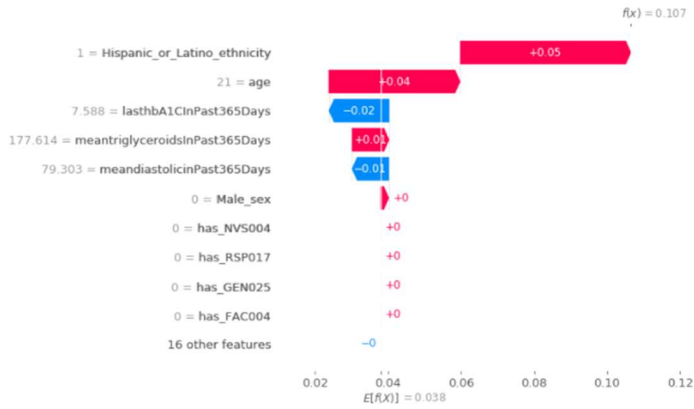

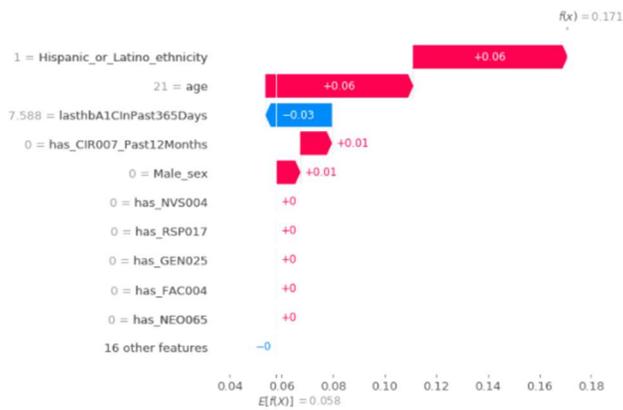

**High Risk**
Risk Score: 3.273

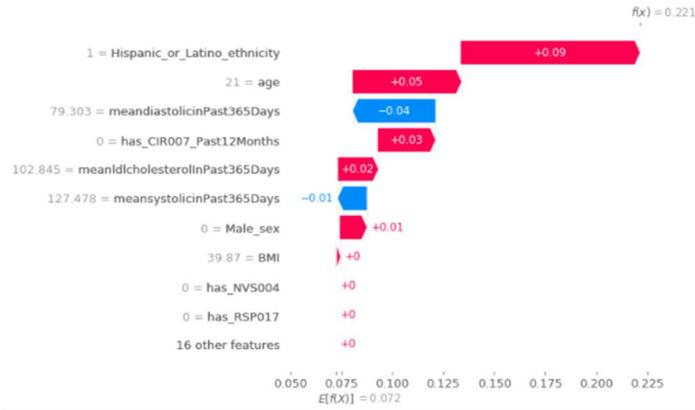

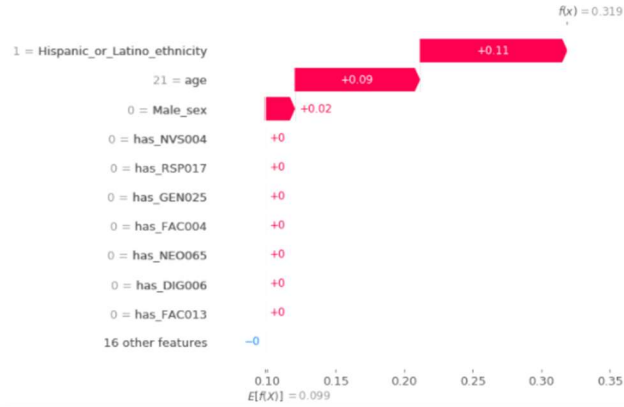

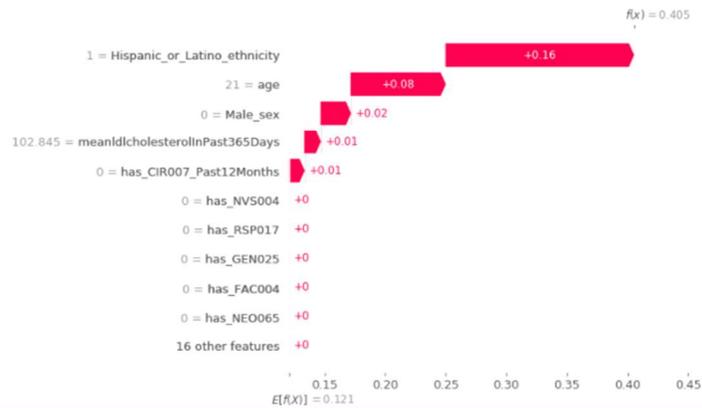

# DM to Diabetic Retinopathy Prediction

| Item | Specification |
|---|---|
| **Business Goal** | Enable care managers to identify the patients who are at risk of developing diabetic retinopathy |
| **Usage Setting** | Outpatient |
| **ML Task** | Predict risk and/or time from DM or uncontrolled DM to diabetic retinopathy |
| **ML Class** | Survival |
| **Instances for Prediction** | Encounters |
| **Labels for Instances** | Binary indicator and time to event or censoring for diabetic retinopathy |
| **Cohort Criteria** | • 2016-01-01 ≤ encounter date ≤ 2020-06-30 (available Epic data, excluding outliers)<br>• Encounter date is not within the first 90 days of when the patient entered the data set, to adjust for left-censoring<br>• 18 ≤ age ≤ 110 (adults without outliers)<br>• No T1DM diagnosis<br>• Not pregnant<br>• No "Do Not Resuscitate" diagnosis<br><br>• DM or uncontrolled DM event before encounter date<br>• No diabetic retinopathy event before encounter date<br>• No diabetic retinopathy event up to 6 days after encounter date (encounters where diagnoses confirm event within the week) |
| **Input Features** | • Demographic<br>• Diagnosis, except:<br>    o hasDiabetesRetinopathy*<br>    o has_EYE002* (cataract and other lens disorders)<br>    o has_EYE005* (retinal and vitreous conditions)<br>• Labs<br>• Utilization<br>• Vitals<br><br>See Appendix E for details. |
| **Evaluation Metrics** | • Concordance Index<br>• Integrated Brier Score |

## Data

The following charts summarize the key characteristics of the data after applying the cohort criteria stated above, along with selected features (see Model Signature below).

| Category | Variable | count | mean | stddev | min | 25% | 50% | 75% | max |
|---|---|---|---|---|---|---|---|---|---|
| Demographic | AgeBucket_18_to_39 | 80201.0 | 0.041994 | 0.200578 | 0.00 | 0.000 | 0.000 | 0.000 | 1.00 |
| | AgeBucket_40_to_59 | 80201.0 | 0.273924 | 0.445973 | 0.00 | 0.000 | 0.000 | 1.000 | 1.00 |
| | AgeBucket_60_to_79 | 80201.0 | 0.563297 | 0.495980 | 0.00 | 0.000 | 1.000 | 1.000 | 1.00 |
| | AgeBucket_80_to_109 | 80201.0 | 0.120784 | 0.325878 | 0.00 | 0.000 | 0.000 | 0.000 | 1.00 |
| | Sex_Female | 80201.0 | 0.554332 | 0.497042 | 0.00 | 0.000 | 1.000 | 1.000 | 1.00 |
| | Sex_Male | 80201.0 | 0.445668 | 0.497042 | 0.00 | 0.000 | 0.000 | 1.000 | 1.00 |
| | Ethnicity_Hispanic_or_Latino | 80201.0 | 0.068154 | 0.252011 | 0.00 | 0.000 | 0.000 | 0.000 | 1.00 |
| | Ethnicity_Not_Hispanic_or_Latino | 80201.0 | 0.931846 | 0.252011 | 0.00 | 1.000 | 1.000 | 1.000 | 1.00 |
| Encounter | EncounterType_Emergency | 80201.0 | 0.223027 | 0.416279 | 0.00 | 0.000 | 0.000 | 0.000 | 1.00 |
| | EncounterType_Inpatient | 80201.0 | 0.114201 | 0.318057 | 0.00 | 0.000 | 0.000 | 0.000 | 1.00 |
| | EncounterType_Outpatient | 80201.0 | 0.662772 | 0.472766 | 0.00 | 0.000 | 1.000 | 1.000 | 1.00 |
| Label | Time | 80201.0 | 15.808593 | 11.336733 | 0.00 | 6.000 | 14.000 | 24.000 | 50.00 |
| | Event | 80201.0 | 0.015436 | 0.123281 | 0.00 | 0.000 | 0.000 | 0.000 | 1.00 |
| Feature | age | 80201.0 | 64.692635 | 13.071276 | 18.00 | 56.000 | 66.000 | 74.000 | 102.00 |
| | Male_sex | 80201.0 | 0.445668 | 0.497042 | 0.00 | 0.000 | 0.000 | 1.000 | 1.00 |
| | Hispanic_or_Latino_ethnicity | 80201.0 | 0.068154 | 0.252011 | 0.00 | 0.000 | 0.000 | 0.000 | 1.00 |
| | lasthbA1CInPast365Days | 80201.0 | 7.550005 | 1.225381 | 3.50 | 6.800 | 7.468 | 8.000 | 12.70 |
| | meandiastolicinPast365Days | 80201.0 | 73.547908 | 7.848043 | 43.62 | 69.278 | 73.621 | 78.210 | 107.56 |
| | meansystolicinPast365Days | 80201.0 | 132.128908 | 11.565079 | 81.00 | 127.480 | 131.860 | 135.970 | 180.00 |
| | BMI | 80201.0 | 33.674120 | 6.790728 | 10.37 | 29.490 | 33.600 | 37.260 | 157.25 |
| | meantriglyceroidsInPast365Days | 80201.0 | 168.577678 | 57.720742 | 19.00 | 138.000 | 167.109 | 182.466 | 460.50 |
| | meanldlcholesterolInPast365Days | 80201.0 | 88.448366 | 23.579373 | 8.00 | 80.500 | 88.935 | 98.338 | 201.00 |
| | meanhdlcholesterolInPast365Days | 80201.0 | 42.829770 | 9.193400 | 5.00 | 37.500 | 42.000 | 47.190 | 98.00 |
| | has_GEN020_Past12Months | 80201.0 | 0.004377 | 0.066011 | 0.00 | 0.000 | 0.000 | 0.000 | 1.00 |
| | has_CIR009_Past12Months | 80201.0 | 0.018341 | 0.134184 | 0.00 | 0.000 | 0.000 | 0.000 | 1.00 |
| | has_BLD008_Past12Months | 80201.0 | 0.018105 | 0.133330 | 0.00 | 0.000 | 0.000 | 0.000 | 1.00 |
| | has_CIR007_Past12Months | 80201.0 | 0.190609 | 0.392784 | 0.00 | 0.000 | 0.000 | 0.000 | 1.00 |
| | has_NEO008_Past12Months | 80201.0 | 0.001820 | 0.042628 | 0.00 | 0.000 | 0.000 | 0.000 | 1.00 |
| | has_CIR008_Past12Months | 80201.0 | 0.012157 | 0.109587 | 0.00 | 0.000 | 0.000 | 0.000 | 1.00 |
| | has_MBD017 | 80201.0 | 0.022818 | 0.149323 | 0.00 | 0.000 | 0.000 | 0.000 | 1.00 |
| | has_INJ026 | 80201.0 | 0.003055 | 0.055186 | 0.00 | 0.000 | 0.000 | 0.000 | 1.00 |
| | has_GEN017 | 80201.0 | 0.035199 | 0.184284 | 0.00 | 0.000 | 0.000 | 0.000 | 1.00 |
| | has_MAL004 | 80201.0 | 0.001696 | 0.041145 | 0.00 | 0.000 | 0.000 | 0.000 | 1.00 |
| | has_NEO022 | 80201.0 | 0.037880 | 0.190907 | 0.00 | 0.000 | 0.000 | 0.000 | 1.00 |
| | has_INJ069 | 80201.0 | 0.001147 | 0.033850 | 0.00 | 0.000 | 0.000 | 0.000 | 1.00 |
| | has_RSP011 | 80201.0 | 0.063428 | 0.243733 | 0.00 | 0.000 | 0.000 | 0.000 | 1.00 |
| | has_INJ023 | 80201.0 | 0.002219 | 0.047059 | 0.00 | 0.000 | 0.000 | 0.000 | 1.00 |
| | has_MUS016 | 80201.0 | 0.002469 | 0.049626 | 0.00 | 0.000 | 0.000 | 0.000 | 1.00 |
| | has_NVS013 | 80201.0 | 0.019040 | 0.136665 | 0.00 | 0.000 | 0.000 | 0.000 | 1.00 |

(Percentages for binary variables can be read from the "mean" column.)

## Encounters and Patients

| | Examples | Encounters | Patients |
|---|---|---|---|
| 1 | 80201 | 80201 | 13624 |

## Train and Test Sets

| | Set | No Event | No Event % | Event | Event % | Total | Total % |
|---|---|---|---|---|---|---|---|
| 1 | Test | 23887 | 30.3 | 371 | 30 | 24258 | 30.2 |
| 2 | Train | 55076 | 69.7 | 867 | 70 | 55943 | 69.8 |
| 3 | Total | 78963 | 100 | 1238 | 100 | 80201 | 100 |

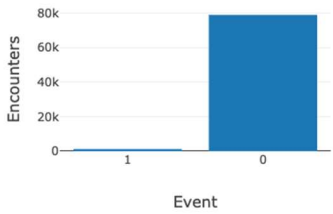

Encounters with Event (1) or Censoring (0)

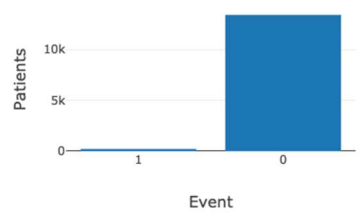

Patients with Event (1) or Censoring (0)

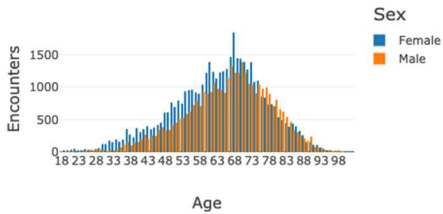

Encounters by Age and Sex

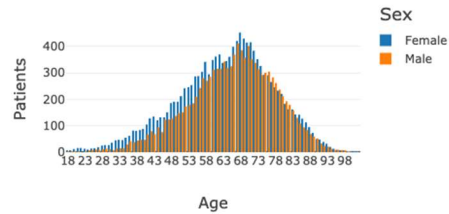

Patients by Age and Sex

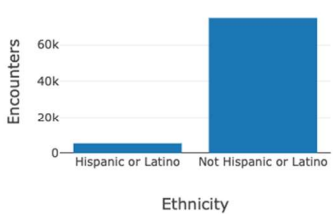

Encounters by Ethnicity

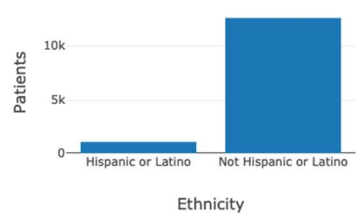

Patients by Ethnicity

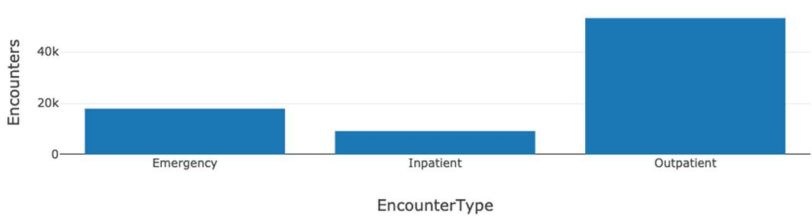

Encounters by Encounter Type

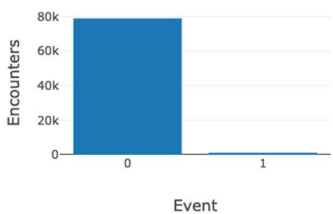

Encounters by Time to Event (1) or Cen…

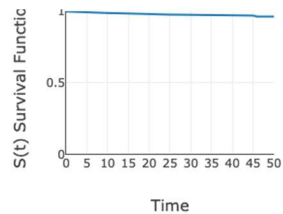

Kaplan-Meier Estimate of Survival Functi…

## Model Signature

The model signature has 26 features, comprising of 12 mandatory features and 14 other selected features. These are the selected features, in rank order (the last feature to be eliminated is ranked 1):

1. has_BLD008_Past12Months (Immunity disorders)
2. has_NEO022 (Respiratory cancers)
3. has_CIR009_Past12Months (Acute myocardial infarction)
4. has_INJ069 (Complication of cardiovascular device, implant or graft, subsequent encounter)
5. has_GEN020_Past12Months (Prolapse of female genital organs)
6. has_MUS016 (Stress fracture, initial encounter)
7. has_INJ023 (Toxic effects, initial encounter)
8. has_MAL004 (Nervous system congenital anomalies)
9. has_NVS013 (Coma; stupor; and brain damage)
10. has_GEN017 (Nonmalignant breast conditions)
11. has_INJ026 (Other specified injury)
12. has_NEO008_Past12Months (Head and neck cancers - laryngeal)
13. has_MBD017 (Alcohol-related disorders)
14. has_RSP011 (Pleurisy, pleural effusion and pulmonary collapse)
15. Hispanic_or_Latino_ethnicity
16. has_CIR008_Past12Months (Hypertension with complications and secondary hypertension)
17. lasthbA1CInPast365Days
18. Male_sex
19. has_CIR007_Past12Months (Essential hypertension)
20. meanhdlcholesterolInPast365Days
21. meandiastolicinPast365Days
22. age
23. meansystolicinPast365Days
24. BMI
25. meanldlcholesterolInPast365Days
26. meantriglyceroidsInPast365Days

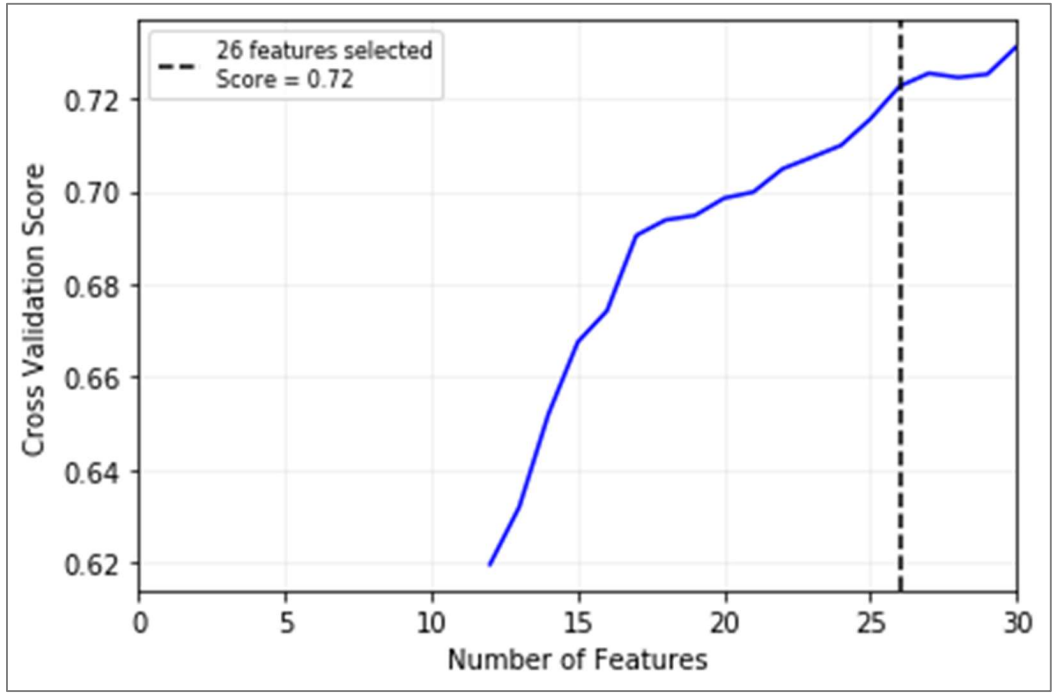

## Model Performance

The following table and chart summarize the performance of all candidate models on the test set for this prediction task in terms of the Concordance Index and the Integrated Brier Score.

| Model | No. of Parameter Combinations Successfully Tested | Concordance Index | Integrated Brier Score |
|---|---|---|---|
| **CoxPH** | 101 | 0.75 | 0.02 |
| **DeepSurv** | 169 | **0.87** | **0.01** |
| **RSF** | 31 | 0.73 | 0.02 |
| **CSF** | 22 | 0.72 | 0.02 |
| **EST** | 36 | 0.76 | 0.02 |

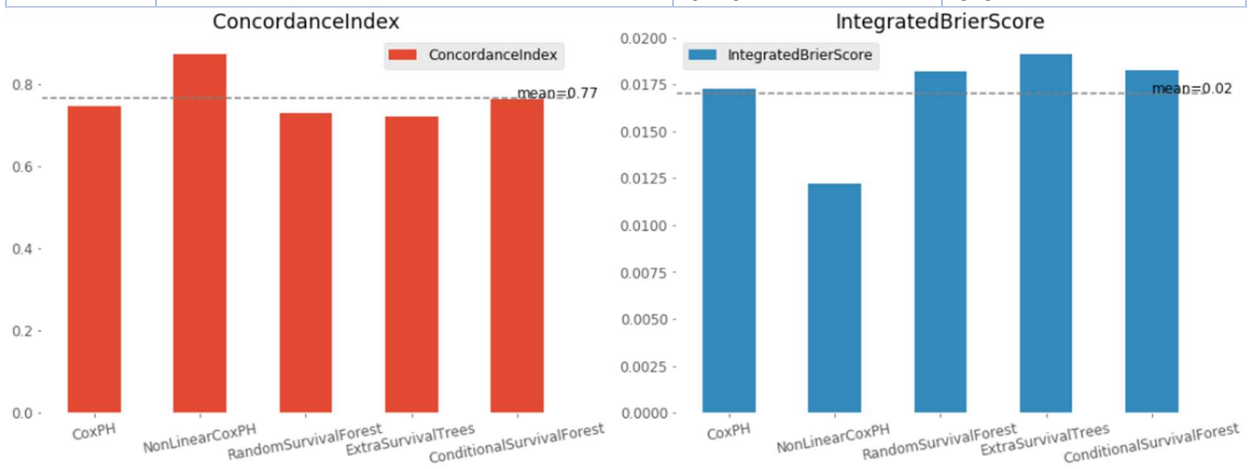

The following chart shows how the average survival function curves of the candidate models compare to the KM survival curve, the more similar their curves are to the KM survival curve the better.

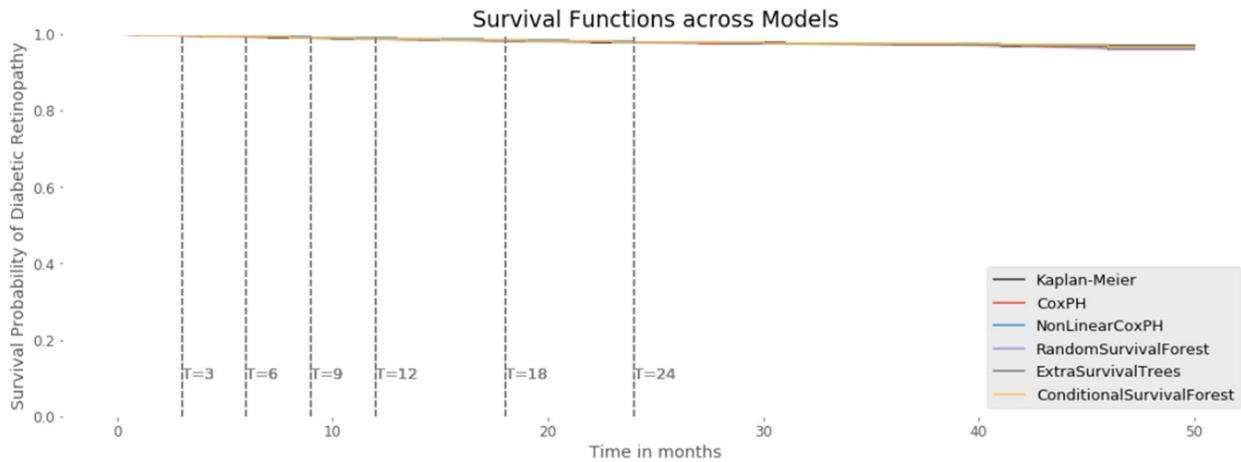

This chart shows the change in the Brier Score over time for all candidate models, the closer the scores are to 0 the better.

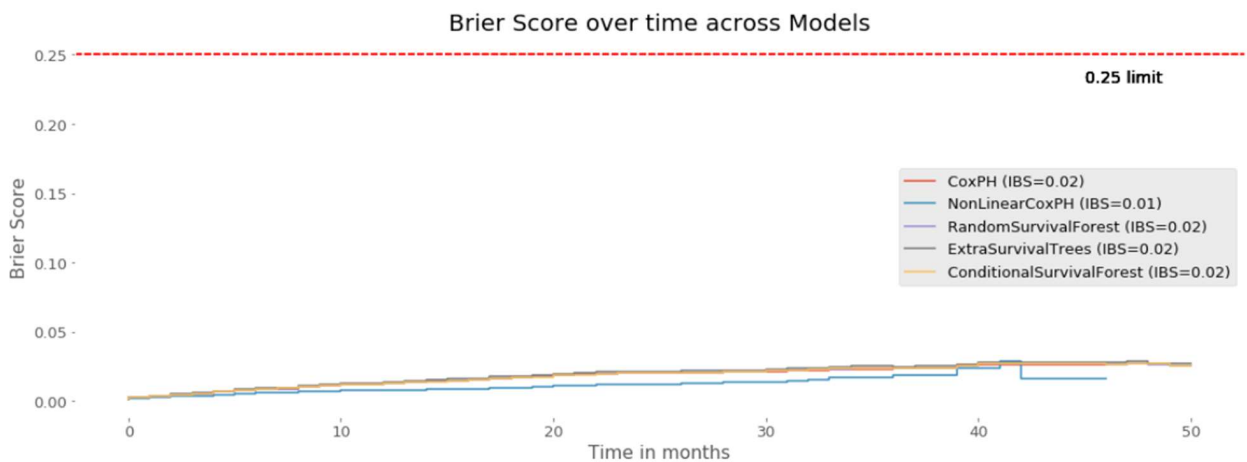

## Model Evaluation of Selected Model (DeepSurv)

### Overall

This chart shows the change in the Brier Score over time for the selected model, the closer the scores are to 0 the better.

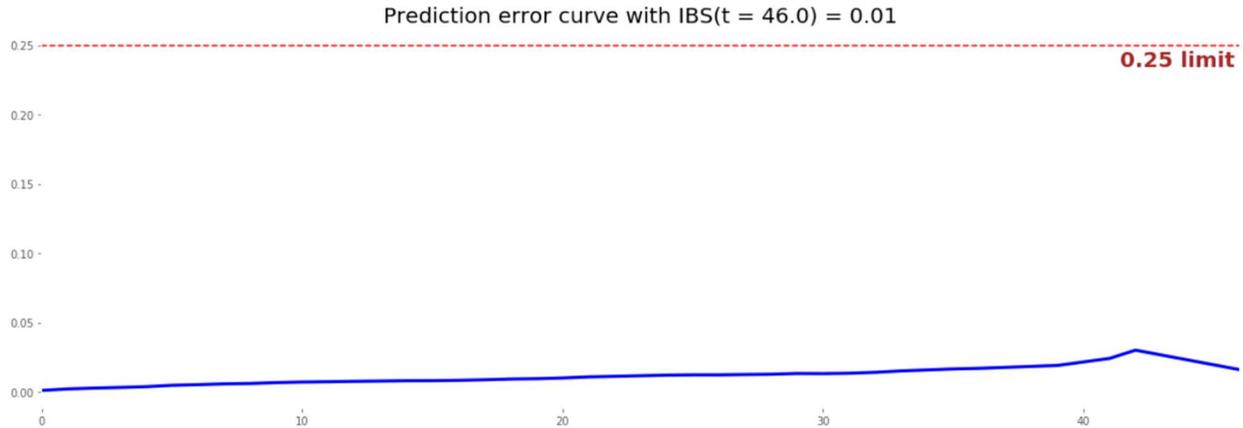

The following chart shows the actual vs. predicted density functions, i.e. number of instances that get the disease / complication at each time point and the RMSE, Median Absolute Error and Mean Absolute Error across the time points.

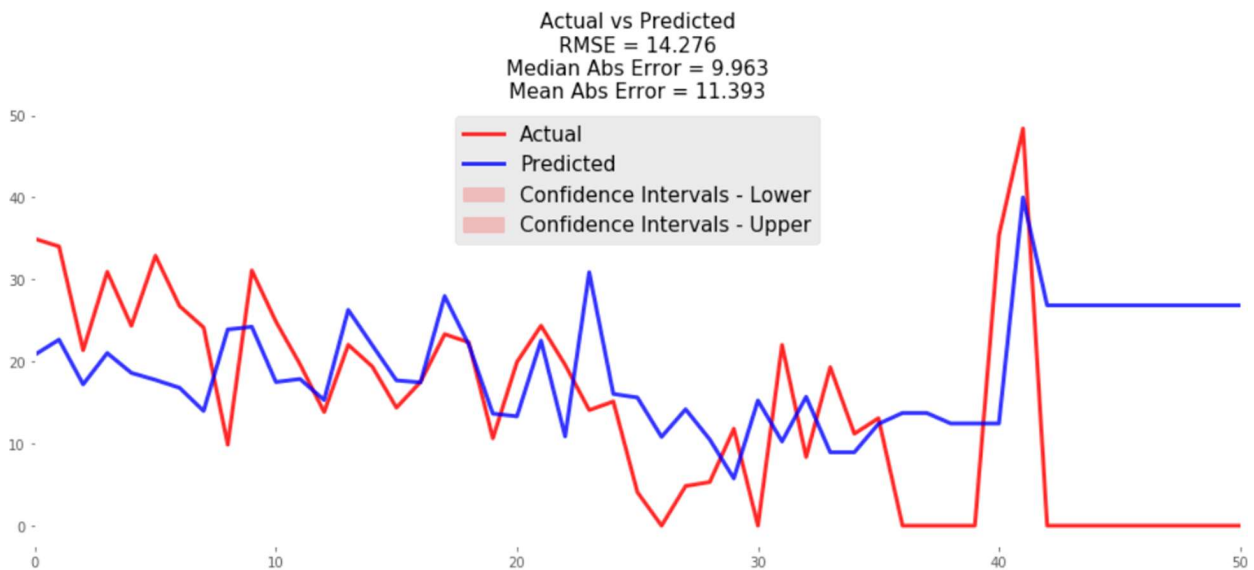

The following chart shows the actual vs. predicted survival functions, i.e. the number of instances that have not had the disease / complication by each time point and the RMSE, Median Absolute Error and Mean Absolute Error across the time points.

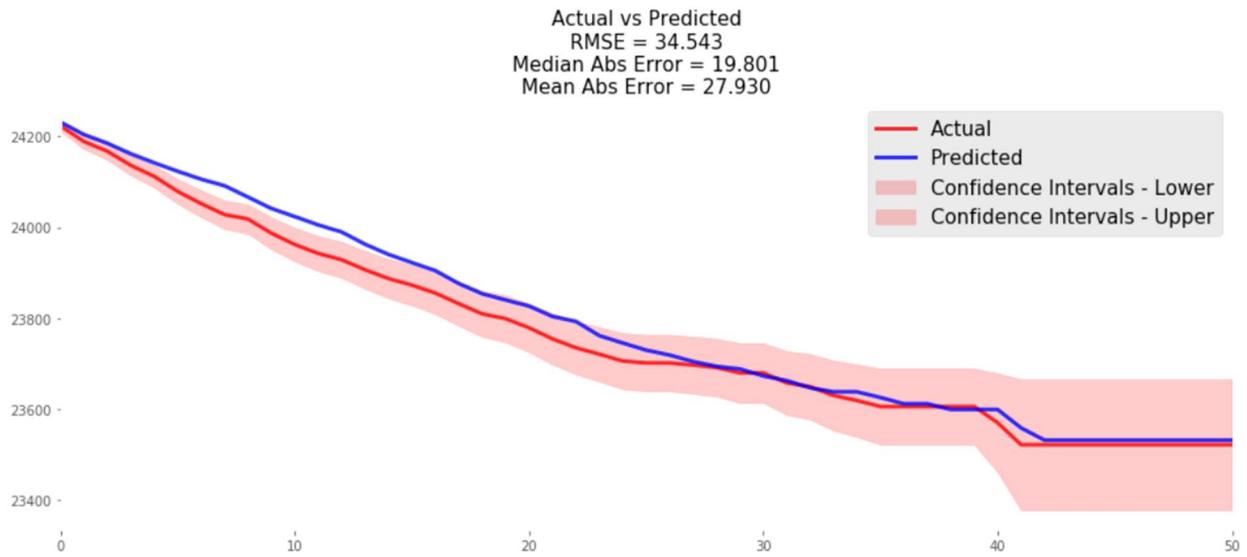

Risk Stratification

The low, medium and high risk groups are defined as examples with predicted risk scores belonging to the first quartile, second to third quartiles, and fourth quartile respectively.

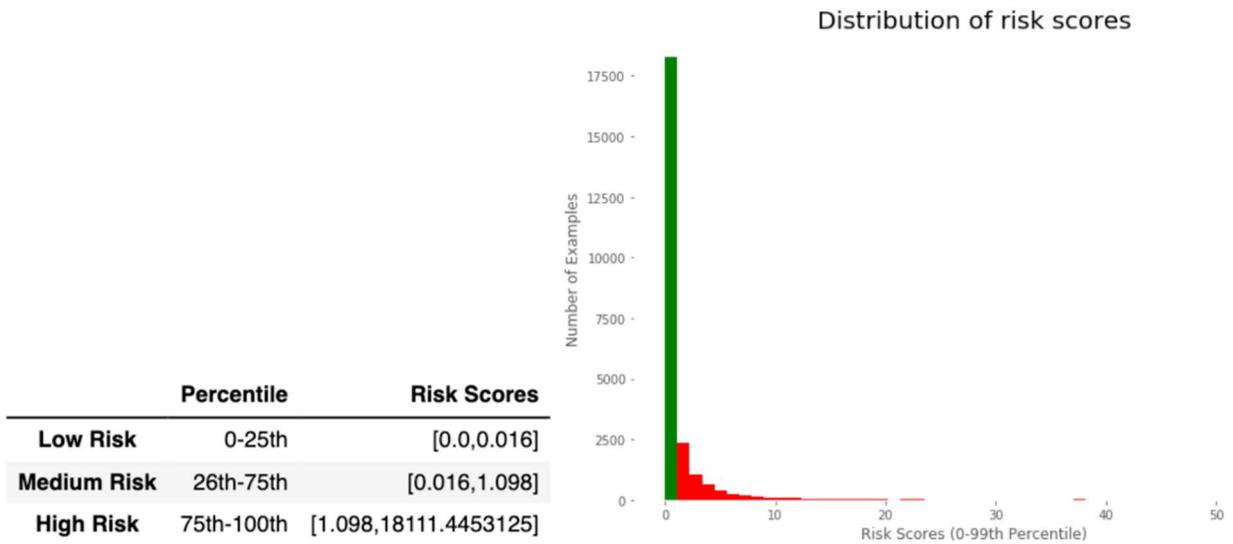

Summary Metrics across Subgroups

The table below displays the summary metrics across subgroups of risk, age, sex, ethnicity and patient history.

| Category | Subgroup | Cohort Size | Concordance Index | Brier Score | Mean AUC | Mean Specificity | Mean Sensitivity | S(t), t=3 | S(t), t=6 | S(t), t=9 | S(t), t=12 | S(t), t=18 | S(t), t=24 |
|---|---|---|---|---|---|---|---|---|---|---|---|---|---|
| NaN | Overall | 24258.00 | 0.87 | 0.01 | 0.90 | 0.97 | 0.68 | 0.99 | 0.99 | 0.99 | 0.99 | 0.98 | 0.98 |
| Risk | Low | 6065.00 | 0.84 | 0.00 | 0.86 | 1.00 | 0.00 | 1.00 | 1.00 | 1.00 | 1.00 | 1.00 | 1.00 |
| Risk | Medium | 12128.00 | 0.55 | 0.01 | 0.49 | 1.00 | 0.00 | 1.00 | 1.00 | 1.00 | 1.00 | 1.00 | 1.00 |
| Risk | High | 6065.00 | 0.81 | 0.04 | 0.87 | 0.88 | 0.80 | 0.98 | 0.98 | 0.97 | 0.96 | 0.94 | 0.92 |
| Age Bucket | 18 to 39 | 1035.00 | 0.99 | 0.01 | 0.99 | 0.99 | 0.75 | 1.00 | 1.00 | 1.00 | 0.99 | 0.99 | 0.99 |
| Age Bucket | 40 to 59 | 6588.00 | 0.89 | 0.01 | 0.93 | 0.96 | 0.77 | 0.99 | 0.99 | 0.99 | 0.98 | 0.98 | 0.97 |
| Age Bucket | 60 to 79 | 13713.00 | 0.86 | 0.01 | 0.89 | 0.97 | 0.64 | 1.00 | 0.99 | 0.99 | 0.99 | 0.98 | 0.98 |
| Age Bucket | 80 to 109 | 2922.00 | 0.75 | 0.00 | 0.72 | 0.98 | 0.36 | 1.00 | 1.00 | 1.00 | 0.99 | 0.99 | 0.99 |
| Sex | Male | 10669.00 | 0.83 | 0.01 | 0.85 | 0.97 | 0.60 | 1.00 | 0.99 | 0.99 | 0.99 | 0.99 | 0.98 |
| Sex | Female | 13589.00 | 0.89 | 0.01 | 0.92 | 0.97 | 0.73 | 1.00 | 0.99 | 0.99 | 0.99 | 0.98 | 0.98 |
| Ethnicity | Hispanic or Latino | 1674.00 | 0.94 | 0.02 | 0.88 | 0.95 | 0.67 | 0.99 | 0.99 | 0.98 | 0.98 | 0.97 | 0.96 |
| Ethnicity | Not Hispanic or Latino | 22584.00 | 0.86 | 0.01 | 0.90 | 0.97 | 0.68 | 1.00 | 0.99 | 0.99 | 0.99 | 0.98 | 0.98 |
| History Bucket | <= 6 | 607.00 | 0.92 | 0.01 | 0.92 | 0.98 | 0.59 | 1.00 | 1.00 | 0.99 | 0.99 | 0.99 | 0.98 |
| History Bucket | 7 to 12 | 2035.00 | 0.82 | 0.01 | 0.88 | 0.97 | 0.64 | 1.00 | 0.99 | 0.99 | 0.99 | 0.98 | 0.98 |
| History Bucket | 13 to 24 | 6532.00 | 0.84 | 0.01 | 0.87 | 0.97 | 0.58 | 1.00 | 0.99 | 0.99 | 0.99 | 0.98 | 0.98 |
| History Bucket | 25 to 36 | 7005.00 | 0.92 | 0.01 | 0.91 | 0.97 | 0.74 | 1.00 | 0.99 | 0.99 | 0.99 | 0.98 | 0.98 |
| History Bucket | 37 to 60 | 8079.00 | 0.90 | 0.01 | nan | nan | 0.75 | 0.99 | 0.99 | 0.99 | 0.99 | 0.98 | 0.98 |

Concordance Index & Integrated Brier Score

The following charts show how the Concordance Index and Integrated Brier Score varies among subgroups of risk, age, sex, ethnicity and patient history.

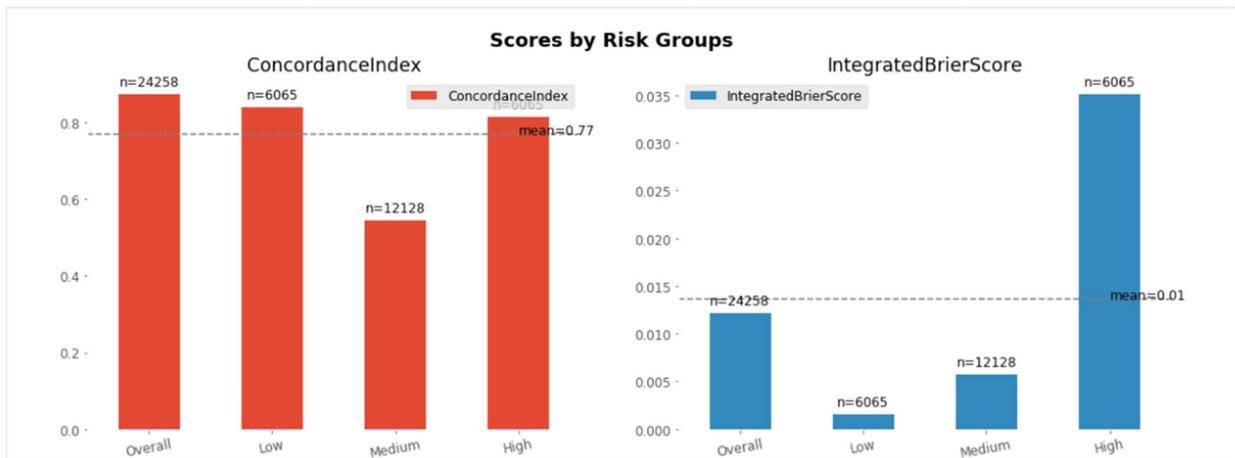

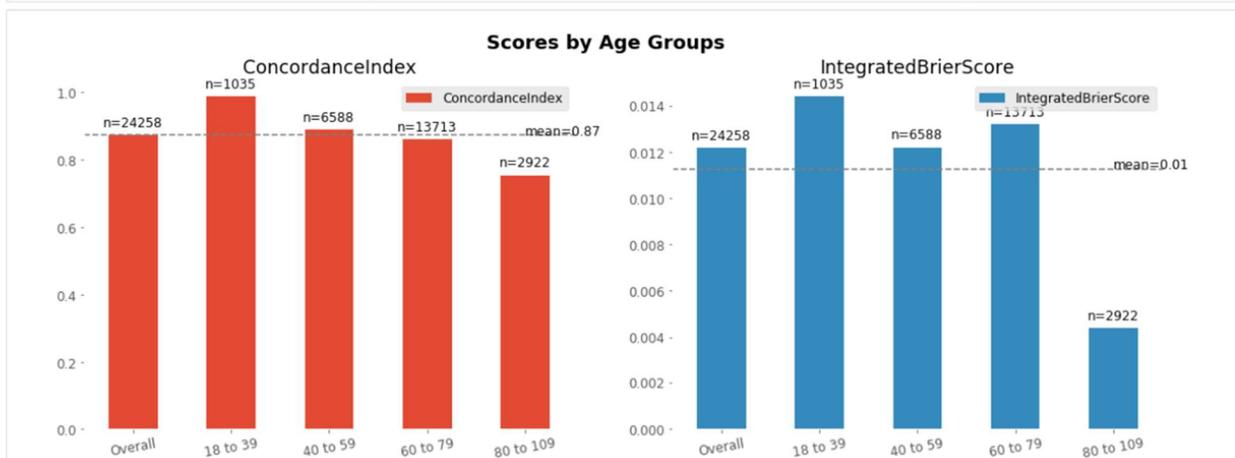

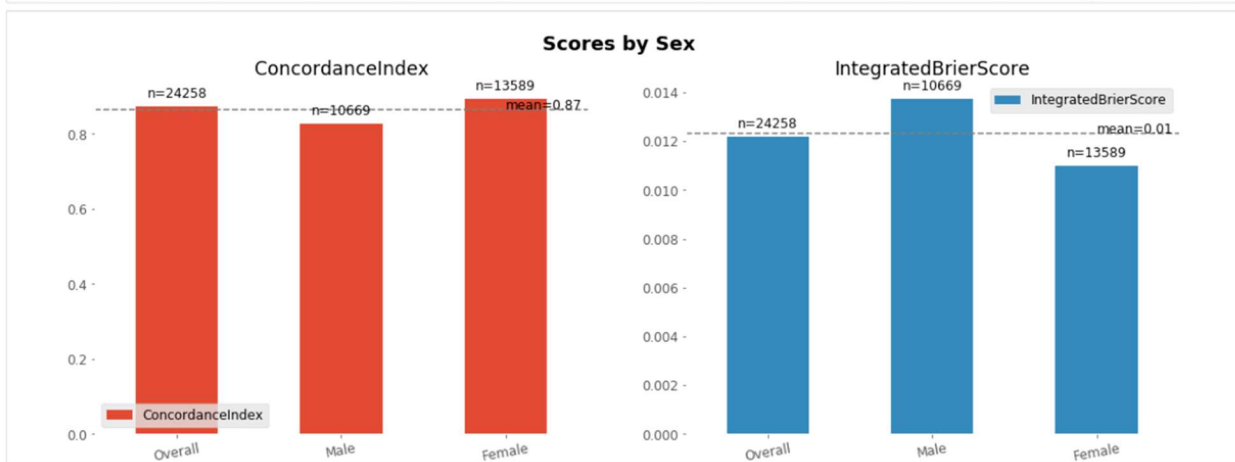

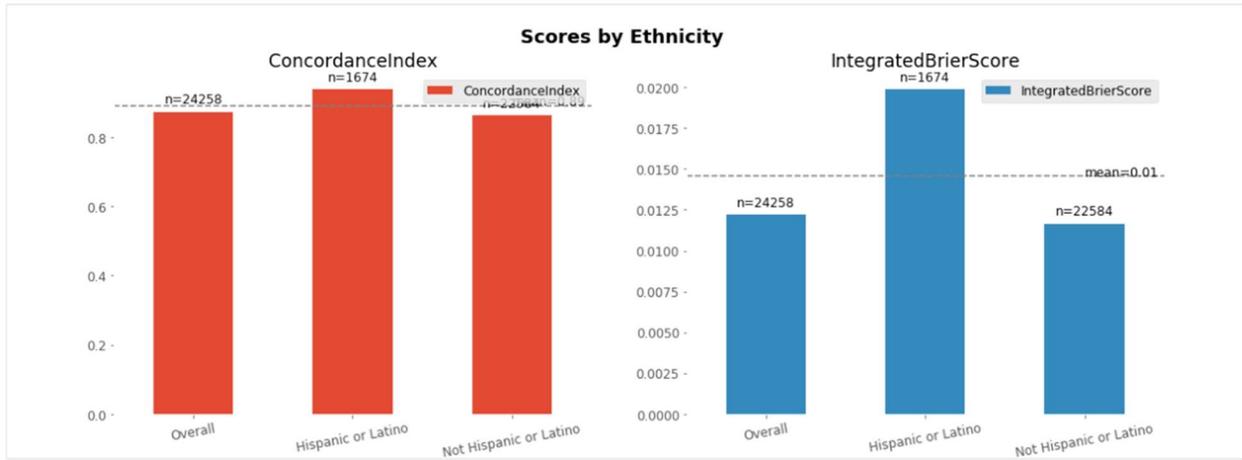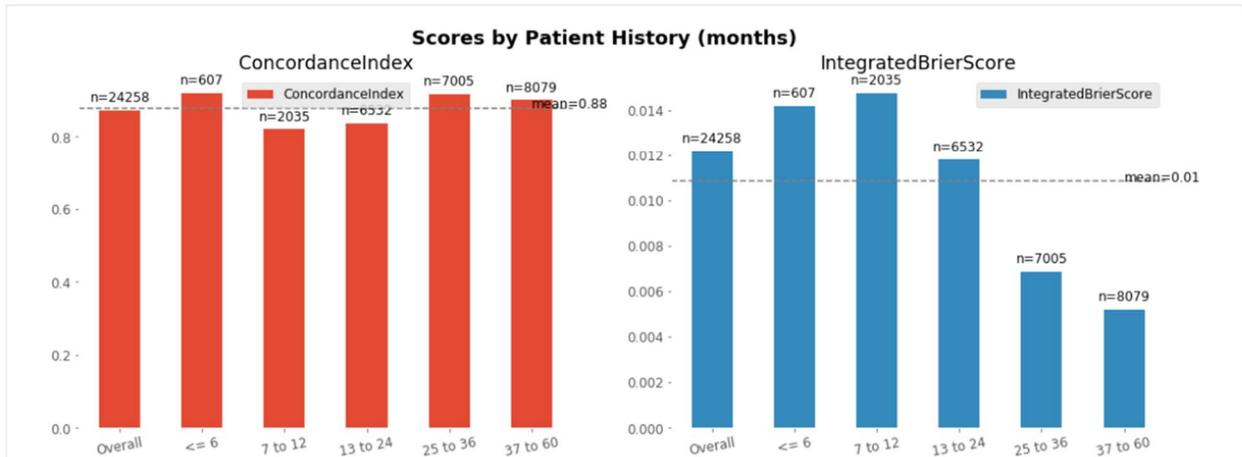

Average Survival Function Curves

The following charts show how the average survival function curve varies among subgroups of risk, age, sex, ethnicity and patient history.

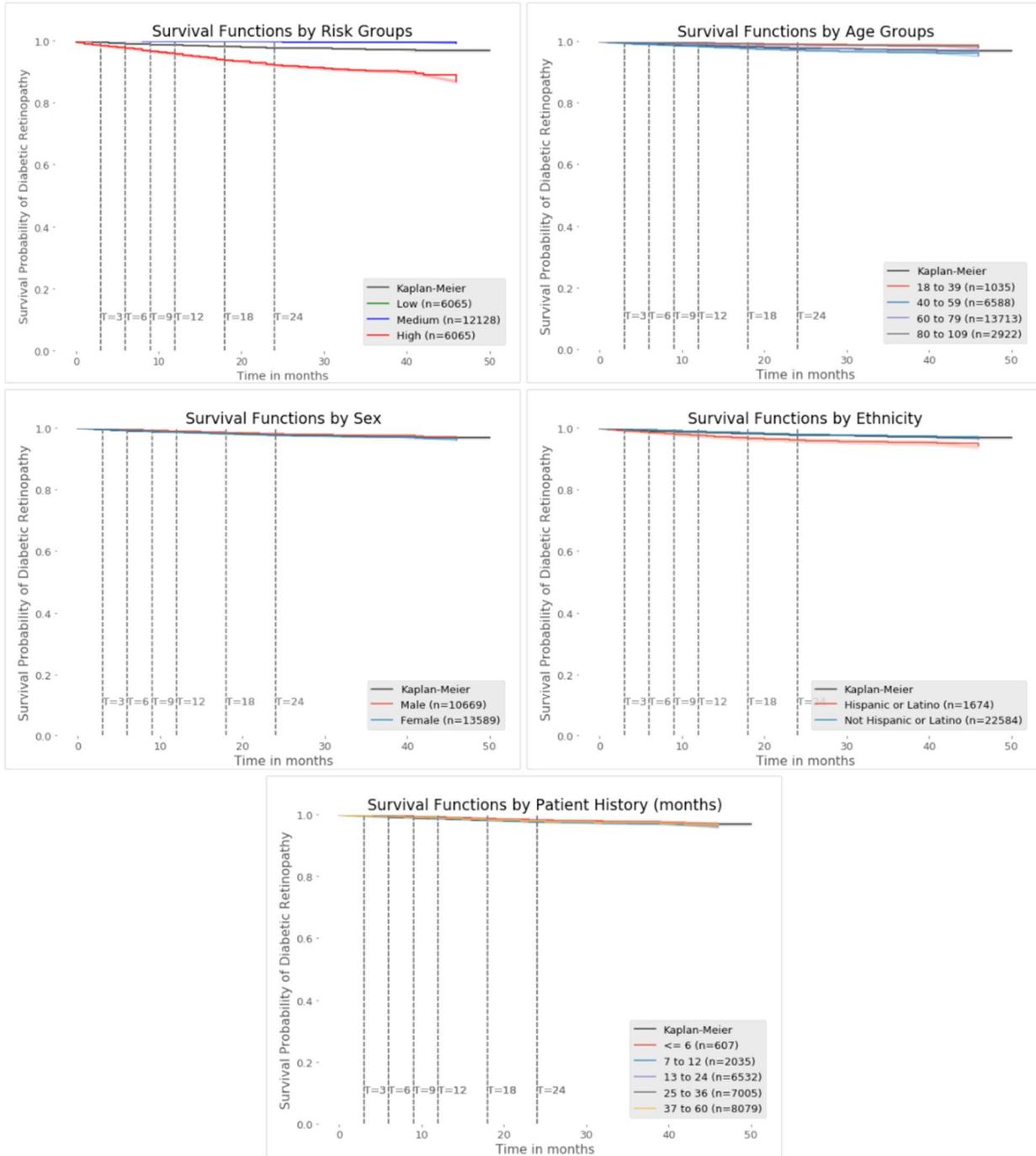

Time-dependent AUC

The following charts show how the AUC across time varies among subgroups of risk, age, sex, ethnicity and patient history.

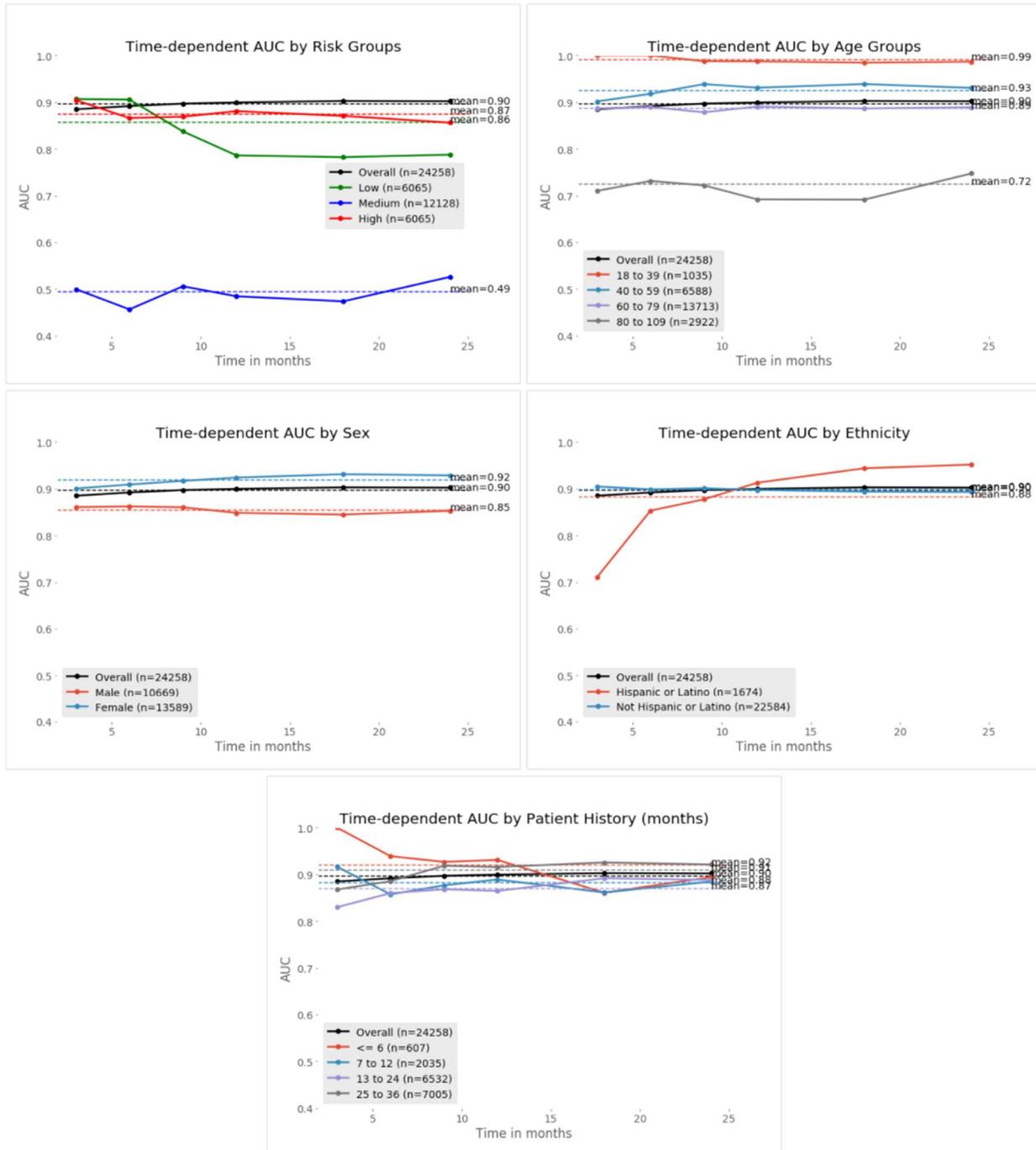

184184184

Time-dependent Specificity

The following charts show how the specificity across time varies among subgroups of risk, age, sex, ethnicity and patient history.

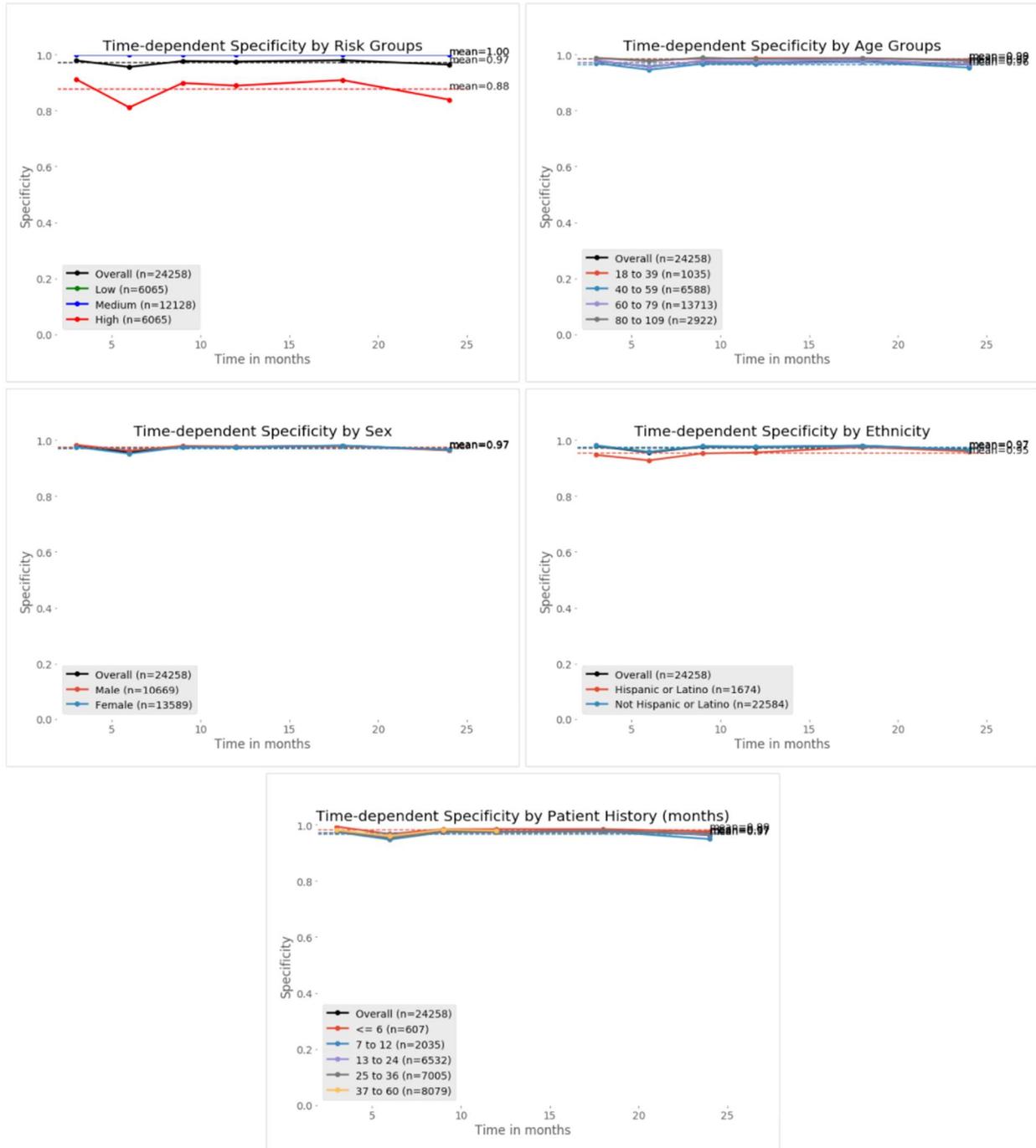

Time-dependent Sensitivity

The following charts show how the sensitivity across time varies among subgroups of risk, age, sex, ethnicity and patient history.

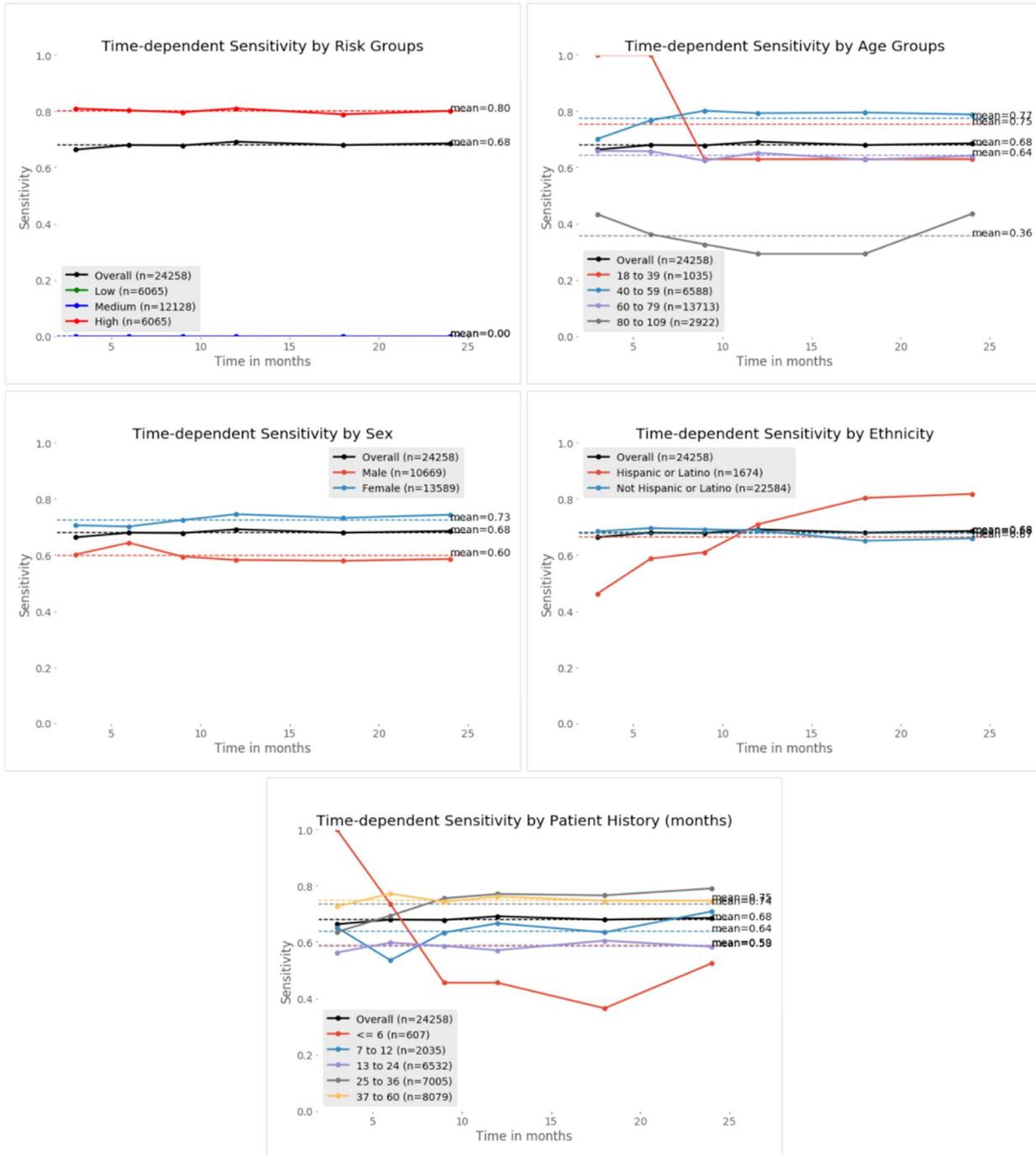

s

## Model Explanation (DeepSurv)

### Global

The following plots show the SHAP values of each instance in the training set for each future time (3, 6, 9, 12, 18 and 24 months). The features are sorted by the total magnitude of the SHAP values over all instances and the distribution of the effect that each feature has on the model's output can be observed.

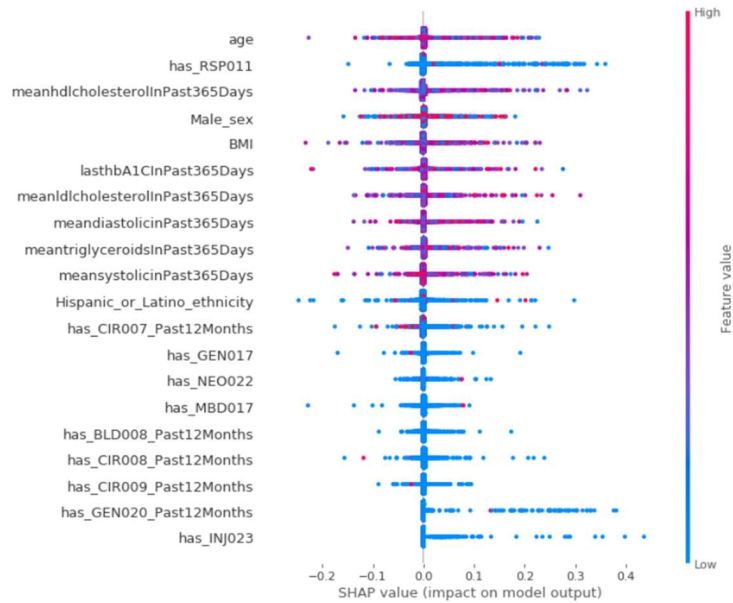

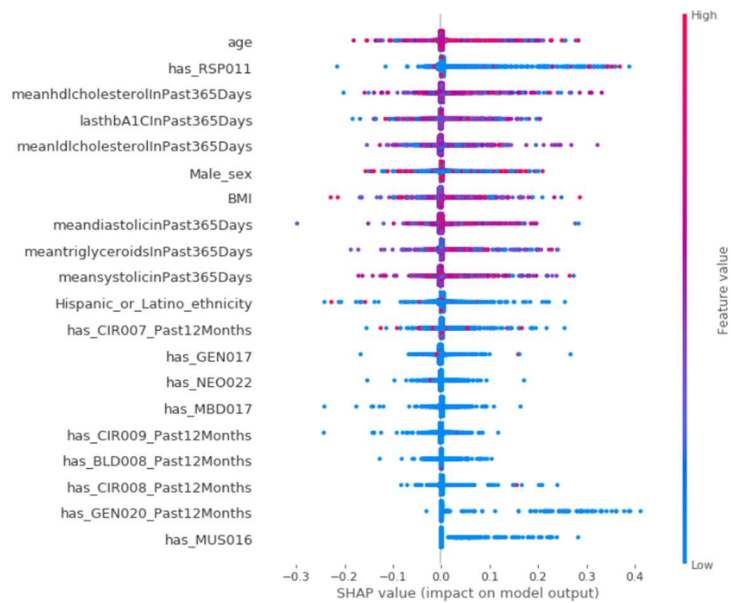

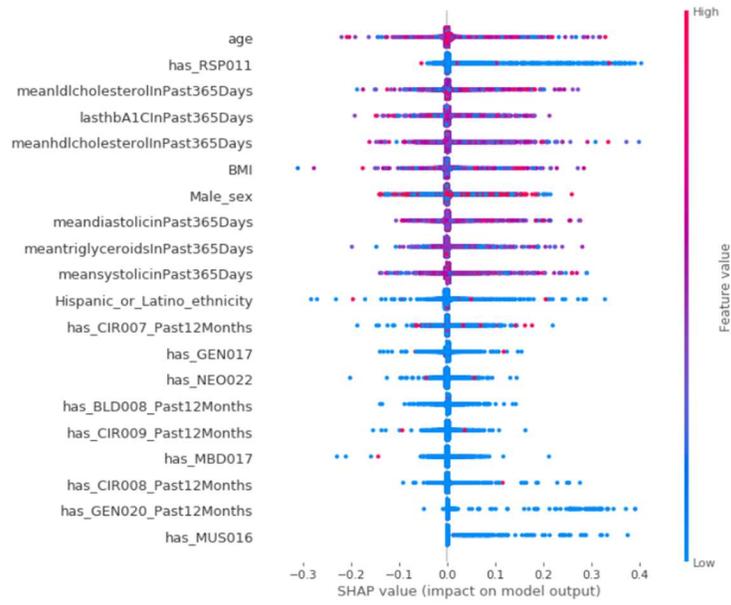
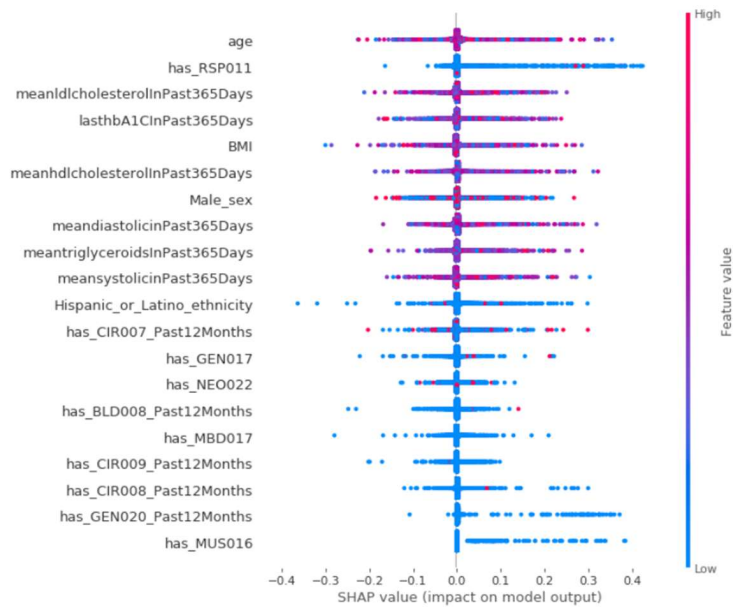

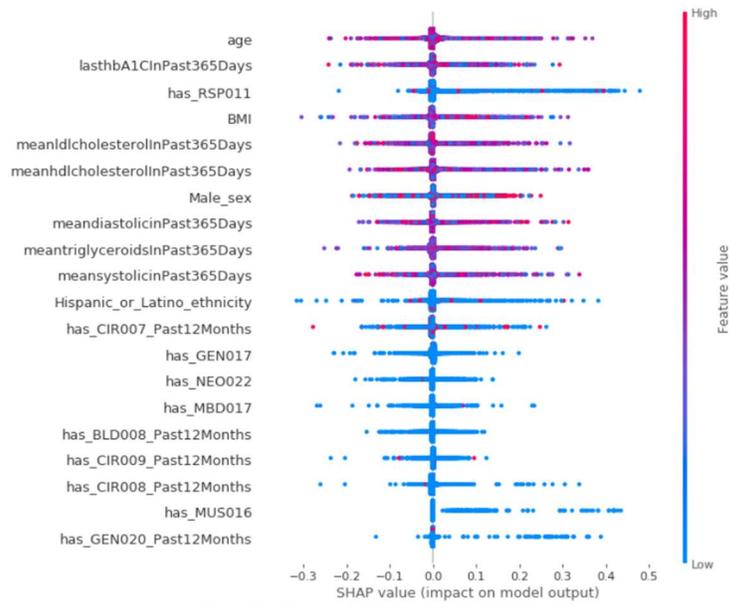
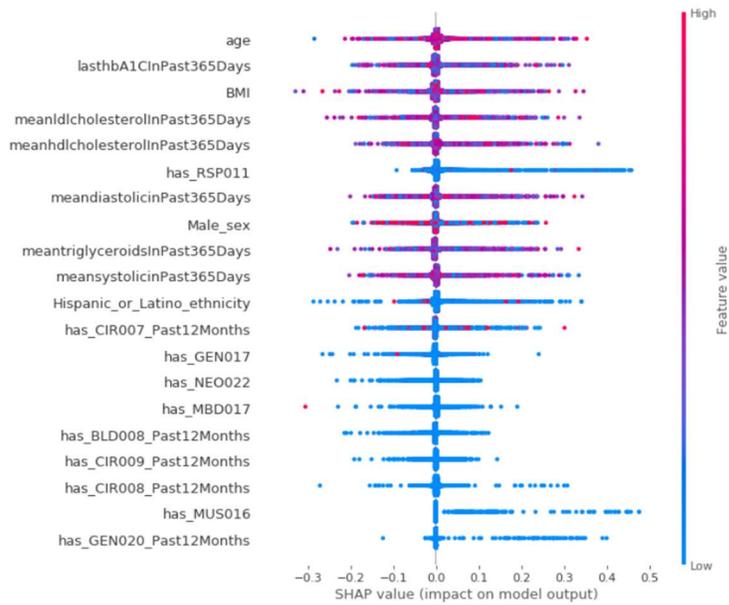

Local

SHAP values were also generated to explain the predictions of individual examples for each future time (3, 6, 9, 12, 18 and 24 months). A total of 3 examples were selected by sampling of risk scores at the 5[th], 50[th] and 95[th] percentile to represent instances at low, medium and high risks respectively.

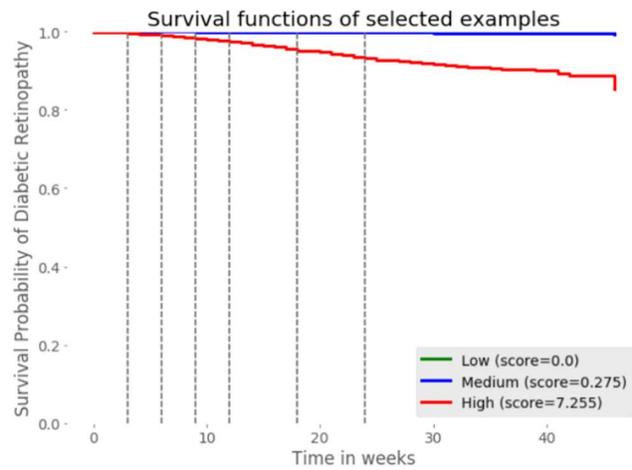

**Low Risk**
Risk Score: 0.0

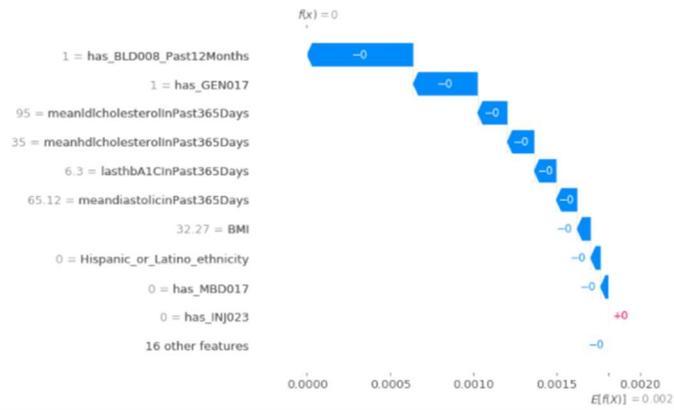

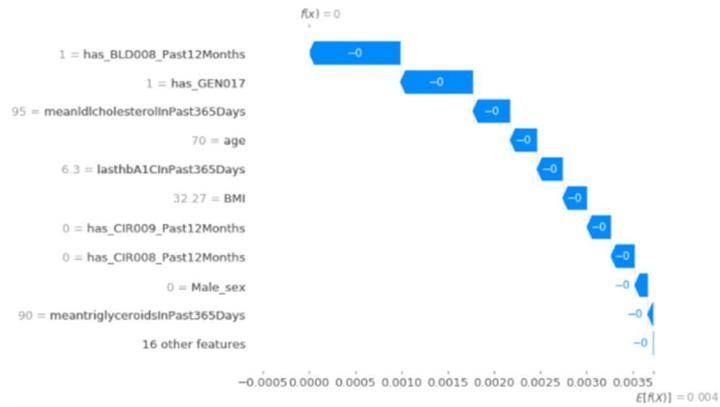

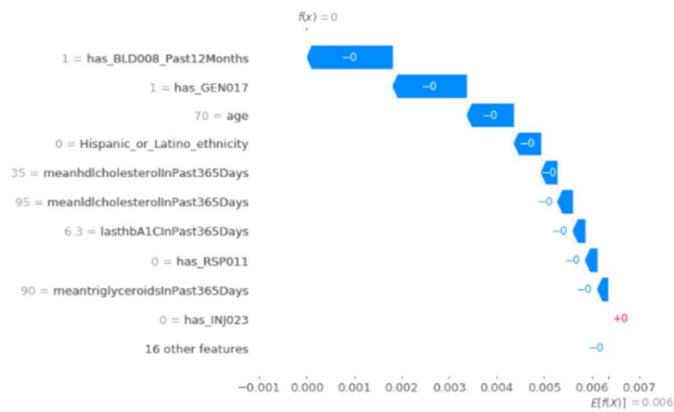

**Low Risk**
Risk Score: 0.0

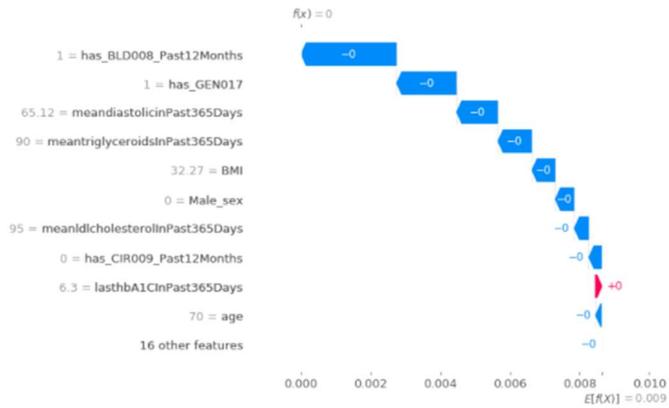

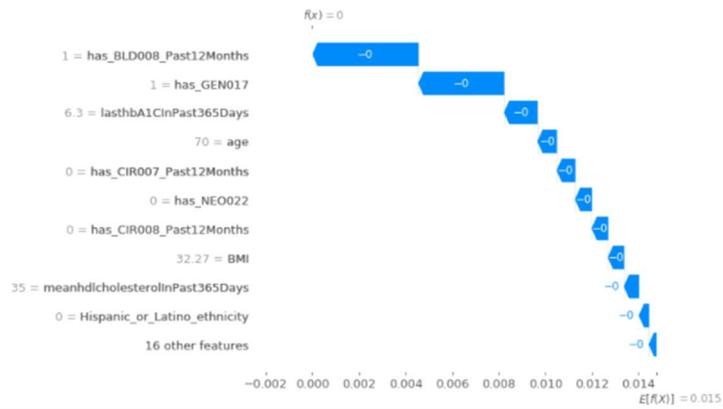

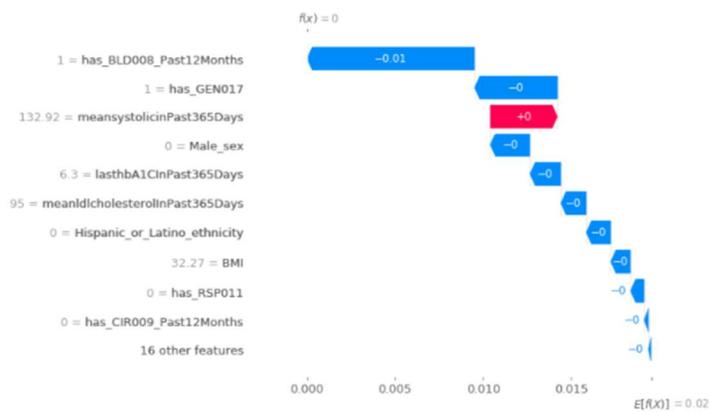

**Medium Risk**
Risk Score: 0.275

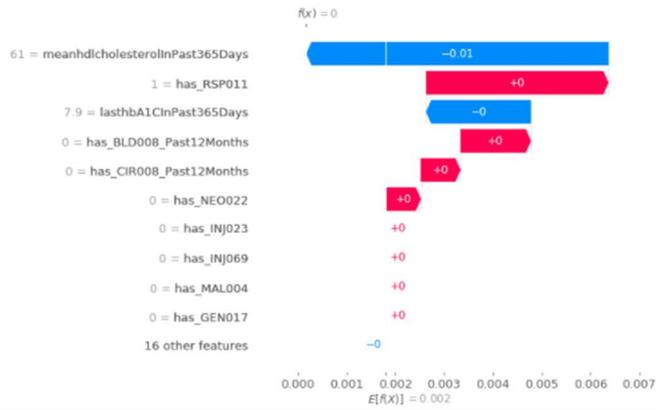

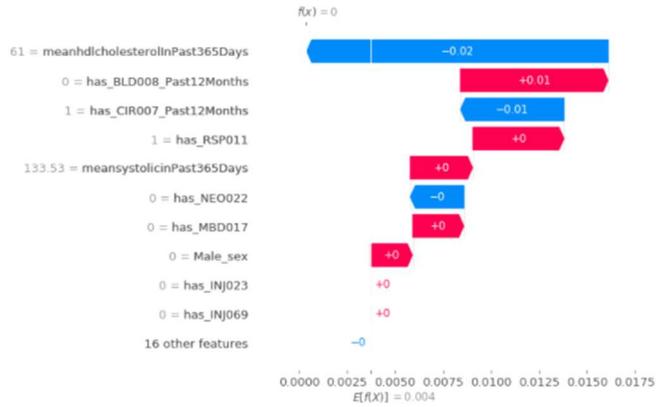

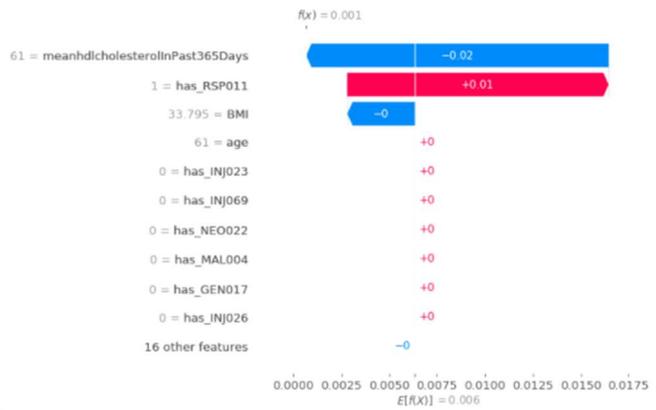

**Medium Risk**
Risk Score: 0.275

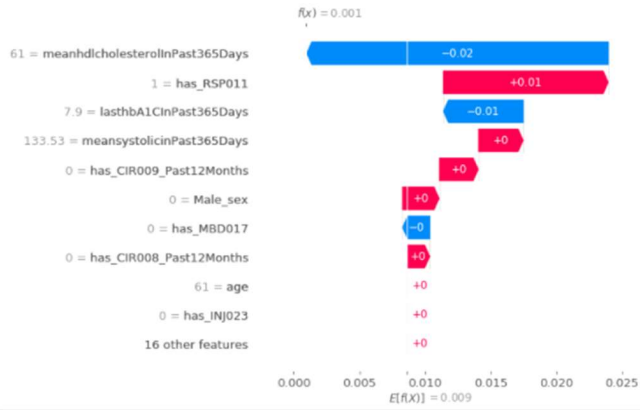

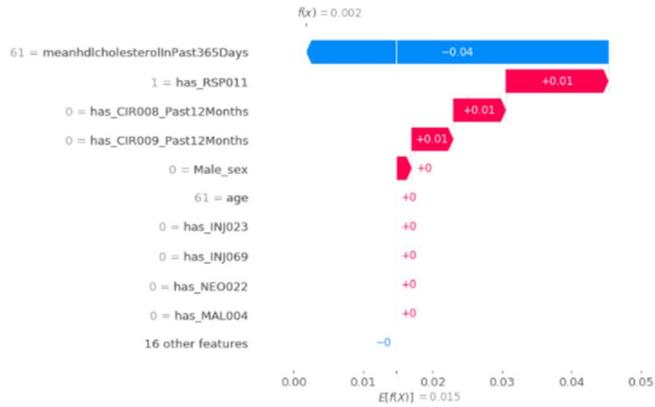

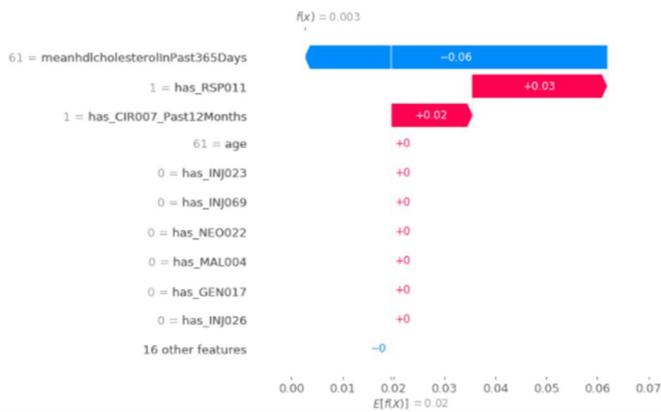

**High Risk**
Risk Score: 7.255

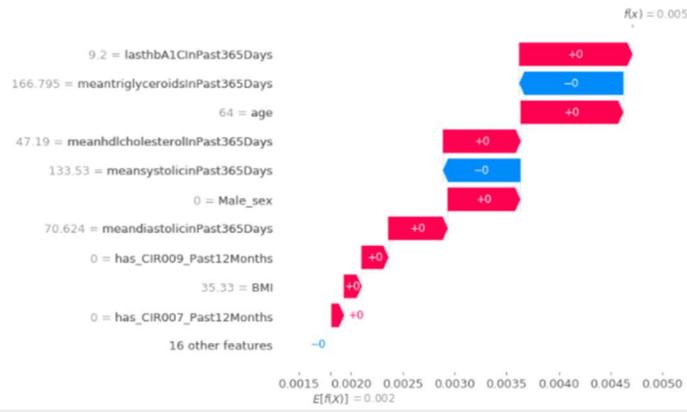

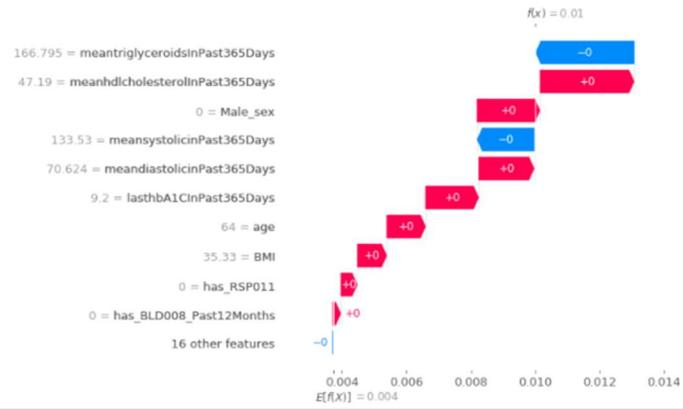

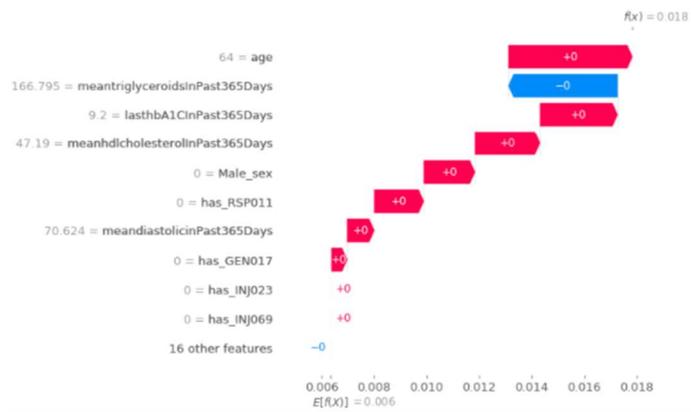

**High Risk**
Risk Score: 7.255

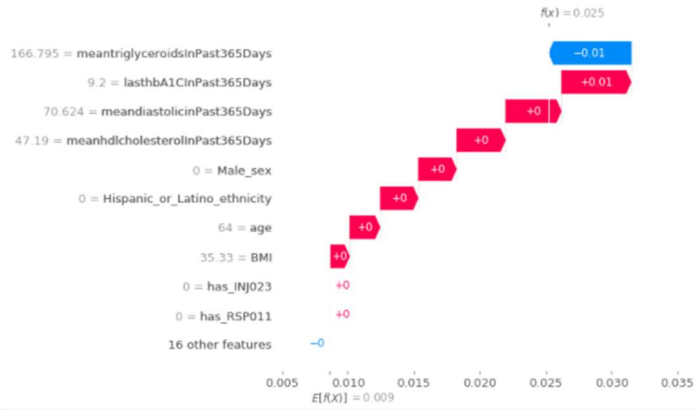

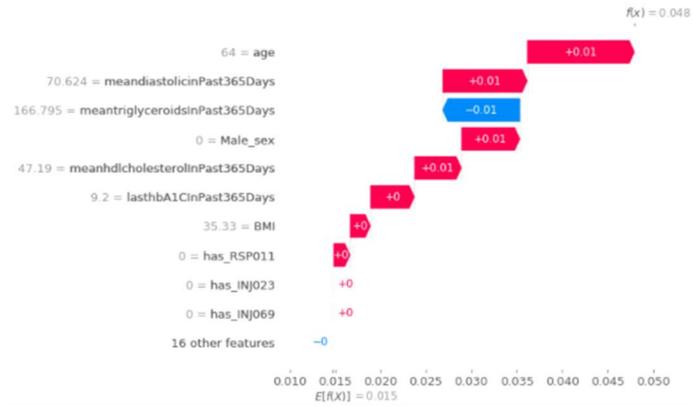

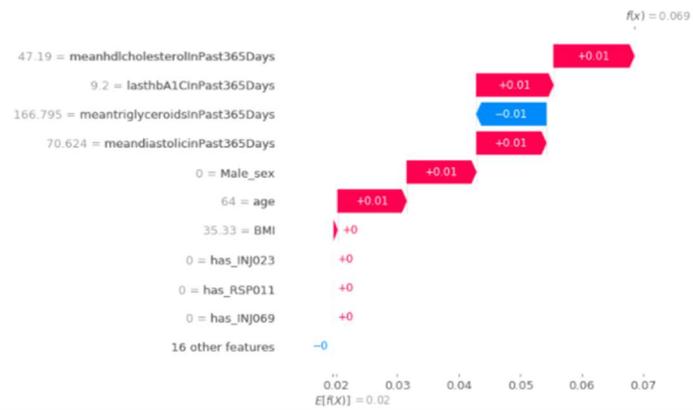

## Conclusion

In this document we presented an overview of problem settings for diabetes prediction and care management. We gave a detailed literature survey of various problems related to prediction of diabetes as well as management of diabetes. We note that often in literature the two are decoupled while in reality, prediction should always be mindful of disease management. We also specify the various ways to describe the problem and detailed descriptions of feature set that we later employ to build the prediction models. Additionally, we also provided detailed results of experiments for predicting diabetes complications, model calibration, optimization methods, baselines, model evaluation and model results. It is our hope that these results could be used as foundations and guidelines for model building for diabetes research in machine learning.

# Appendix A: Features for Model Training

The following table lists the candidate features for model training.

| Name | Category | Type | Description | Number of Features | Modifiable | Mandatory |
|---|---|---|---|---|---|---|
| **Age** | Demographic | Integer | In years, at time of encounter | 1 | No | Yes |
| **Age Above 65** | Demographic | Boolean | Whether the patient's age is above 65 | 1 | No | |
| **Ethnicity** | Demographic | Categorical | • Hispanic or Latino<br>• Not Hispanic or Latino | 1 | No | Yes |
| **Sex** | Demographic | Categorical | • Female<br>• Male | 1 | No | Yes |
| **CCSR Category (12 mth)** | Diagnosis | Boolean | Presence of CCSR Category in past 12 months | 540 | Depends | |
| **CCSR Category (Encounter)** | Diagnosis | Boolean | Presence of CCSR Category in the patient's whole history | 540 | Depends | |
| **Days Since Gestational Diabetes** | Diagnosis | Integer | Number of days since the patient's most recent diagnosis of gestational diabetes | 1 | No | |
| **Albumin** | Lab | Continuous | 365-day aggregates:<br>• Min<br>• Mean<br>• Median<br>• Max<br>• Last | 5 | | |
| **Alkaline Phosphatase** | Lab | Continuous | 365-day aggregates:<br>• Min<br>• Mean<br>• Median<br>• Max<br>• Last | 5 | | |
| **Bun** | Lab | Continuous | 365-day aggregates:<br>• Min<br>• Mean<br>• Median<br>• Max<br>• Last | 5 | | |
| **HBA1C** | Lab | Continuous | 365-day aggregates:<br>• Min<br>• Mean<br>• Median<br>• Max<br>• Last | 5 | | Mean |
| **Neutrophils** | Lab | Continuous | 365-day aggregates:<br>• Min<br>• Mean<br>• Median<br>• Max<br>• Last | 5 | | |

| Name | Category | Type | Description | Number of Features | Modifiable | Mandatory |
|---|---|---|---|---|---|---|
| **Potassium** | Lab | Continuous | 365-day aggregates:<br>• Min<br>• Mean<br>• Median<br>• Max<br>• Last | 5 | | |
| **Sodium** | Lab | Continuous | 365-day aggregates:<br>• Min<br>• Mean<br>• Median<br>• Max<br>• Last | 5 | | |
| **WBC** | Lab | Continuous | 365-day aggregates:<br>• Min<br>• Mean<br>• Median<br>• Max<br>• Last | 5 | | |
| **Days Since Last ED Visit** | Utilization | Integer | In days | 1 | | |
| **Days Since Last Inpatient Visit** | Utilization | Integer | In days | 1 | | |
| **Previous ED Visits Count** | Utilization | Integer | For days before current encounter:<br>• 90 days<br>• 180 days<br>• 365 days | 3 | | |
| **Previous Inpatient Encounters Count** | Utilization | Integer | For days before current encounter:<br>• 90 days<br>• 180 days<br>• 365 days | 3 | | |
| **ALT** | Vitals | Continuous | 365-day aggregates:<br>• Min<br>• Mean<br>• Median<br>• Max<br>• SD<br>• Last | 6 | Yes | Mean |
| **Diastolic** | Vitals | Continuous | 365-day aggregates:<br>• Min<br>• Mean<br>• Median<br>• Max<br>• SD<br>• Last | 6 | Yes | Mean |
| **Systolic** | Vitals | Continuous | 365-day aggregates:<br>• Min<br>• Mean<br>• Median<br>• Max<br>• SD<br>• Last | 6 | Yes | Mean |
| **LDL Cholesterol** | Vitals | Continuous | 365-day aggregates: | 6 | Yes | Mean |

| Name | Category | Type | Description | Number of Features | Modifiable | Mandatory |
|---|---|---|---|---|---|---|
| | | | • Min<br>• Mean<br>• Median<br>• Max<br>• SD<br>• Last | | | |
| **HDL Cholesterol** | Vitals | Continuous | 365-day aggregates:<br>• Min<br>• Mean<br>• Median<br>• Max<br>• SD<br>• Last | 6 | Yes | Mean |
| **Creatine** | Vitals | Continuous | 365-day aggregates:<br>• Min<br>• Mean<br>• Median<br>• Max<br>• SD<br>• Last | 6 | Yes | Mean |
| **Triglyceroids** | Vitals | Continuous | 365-day aggregates:<br>• Min<br>• Mean<br>• Median<br>• Max<br>• SD<br>• Last | 6 | Yes | Mean |

# Appendix B: Baseline Comparison using the Last A1C Value

In this section, we compare the performance of our models against a baseline. The last A1C value of the patients were used as a baseline to match the method which care managers have been using to identify high risk patients. The following metrics were compared using both the risk scores generated by our models and the last A1C value of the patients:

- Concordance Index and
- Time-dependent AUC

The table below shows the percentage of test instances with a last A1C value for each task which also represents the proportion of the test cohort which we will be carrying out the comparison on.

|  | % of instances with last A1C value |
| --- | --- |
| T2DM | 55.55 |
| Uncontrolled T2DM | 66.76 |
| Pre T2DM to Uncontrolled T2DM | 55.30 |
| Diabetic Nephropathy | 70.81 |
| Diabetic Neuropathy | 69.49 |
| Diabetic Retinopathy | 73.00 |

For all tasks except T2DM and Diabetic Neuropathy, the use of Risk Scores resulted in a higher Concordance Index, indicating that most of our models were able to beat the baseline in terms of ranking the instances correctly according to their risk levels.

|  | Concordance Index (Risk Scores) | Concordance Index (A1C Values) |
| --- | --- | --- |
| T2DM | 0.825329 | 0.868744 |
| Uncontrolled T2DM | 0.804390 | 0.782126 |
| Pre T2DM to Uncontrolled T2DM | 0.832033 | 0.825503 |
| Diabetic Nephropathy | 0.809212 | 0.767832 |
| Diabetic Neuropathy | 0.737294 | 0.759065 |
| Diabetic Retinopathy | 0.868380 | 0.737511 |

Using AUC as the metric gives a different perspective. The charts below show that in general, across different prediction tasks and points in time, the survival risks predicted by the ML models achieve higher AUC than the baseline of last A1C value.

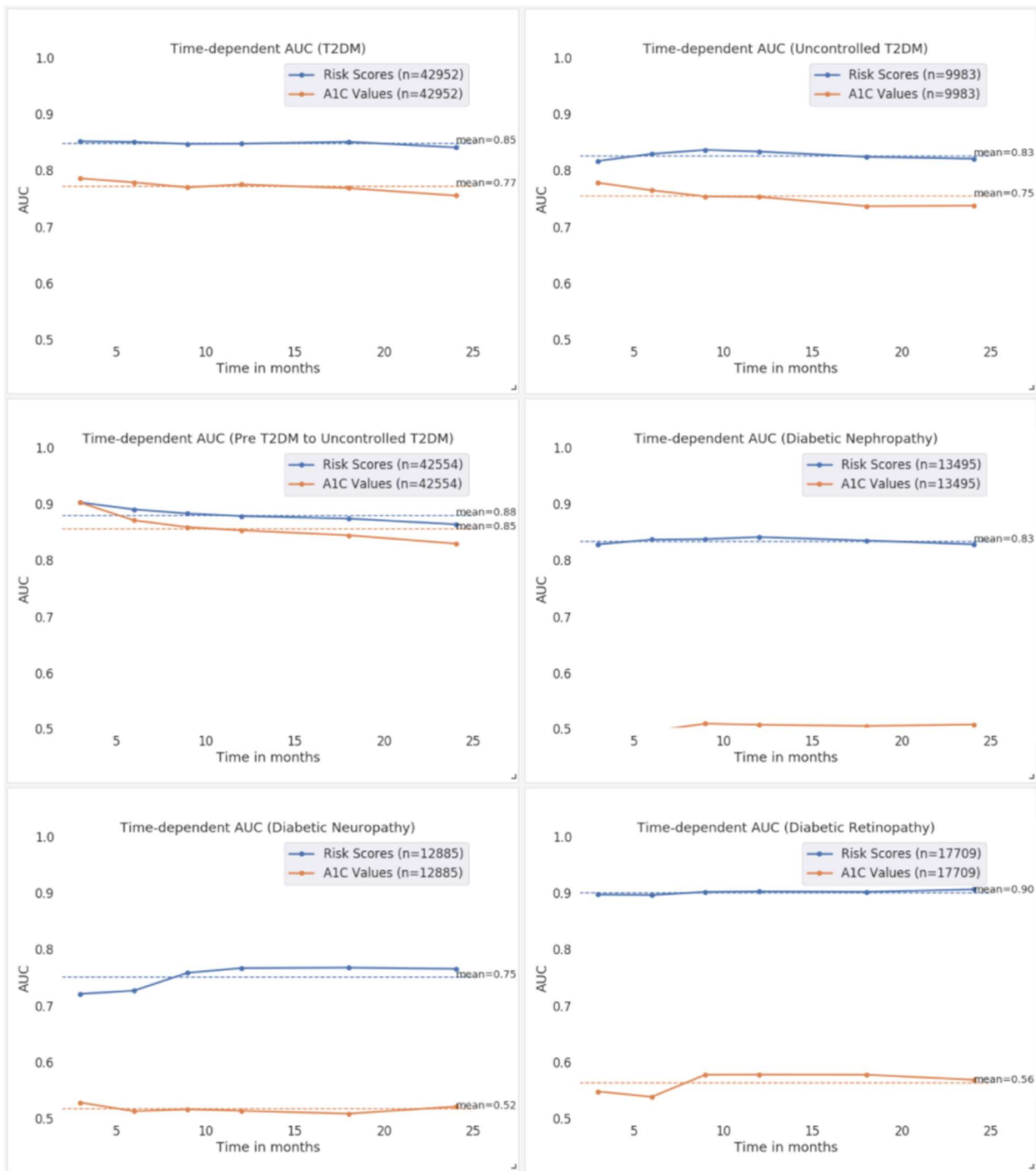

# Appendix B2: Baseline Comparison using Last A1C Values within 3 Months

In this section, we compare the performance of our models against a baseline. The last A1C value of the patients within 3 months from their encounter date were used as a baseline to match the method which care managers have been using to identify high risk patients. The following metrics were compared using both the risk scores generated by our models and the last A1C value of the patients:

- Concordance Index and
- Time-dependent AUC

The table below shows the percentage of test instances with missing A1C value (or A1C value > 20), last A1C value greater than 3 months from the encounter date and last A1C value within months of the encounter date which also represents the proportion of the test cohort which we will be carrying out the comparison on.

|  | % with missing A1C value | % with A1C value > 3 months | % with A1C value <= 3 months |
| --- | --- | --- | --- |
| T2DM | 27.76 | 44.09 | 28.15 |
| Uncontrolled T2DM | 18.02 | 43.68 | 38.30 |
| Pre T2DM to Uncontrolled T2DM | 27.88 | 44.39 | 27.73 |
| Diabetic Nephropathy | 13.86 | 43.36 | 42.78 |
| Diabetic Neuropathy | 14.44 | 43.55 | 42.01 |
| Diabetic Retinopathy | 12.07 | 43.25 | 44.68 |

For all tasks except Diabetic Retinopathy, the use of the Risk Scores resulted in a lower Concordance Index, indicating that most of our models were not able to beat the baseline in terms of ranking the instances correctly according to their risk levels.

|  | Concordance Index (Risk Scores) | Concordance Index (A1C Values) |
| --- | --- | --- |
| T2DM | 0.808911 | 0.959292 |
| Uncontrolled T2DM | 0.802203 | 0.919095 |
| Pre T2DM to Uncontrolled T2DM | 0.801713 | 0.968062 |
| Diabetic Nephropathy | 0.804889 | 0.910083 |
| Diabetic Neuropathy | 0.729231 | 0.910516 |
| Diabetic Retinopathy | 0.859881 | 0.857972 |

Using AUC as the metric gives a different perspective. The charts below show that in general, across different prediction tasks and points in time, the survival risks predicted by the ML models achieve higher AUC than the baseline of last A1C value.

This indicates that an A1C reading within the last 3 months provides a good indication of *relative* risks for DM and its complications between patients, i.e. who is likely to get a disease first, but the ML predicted risk scores are a more accurate predictor of *absolute* risk at specific points in time, i.e. who is likely to get a disease by a certain time.

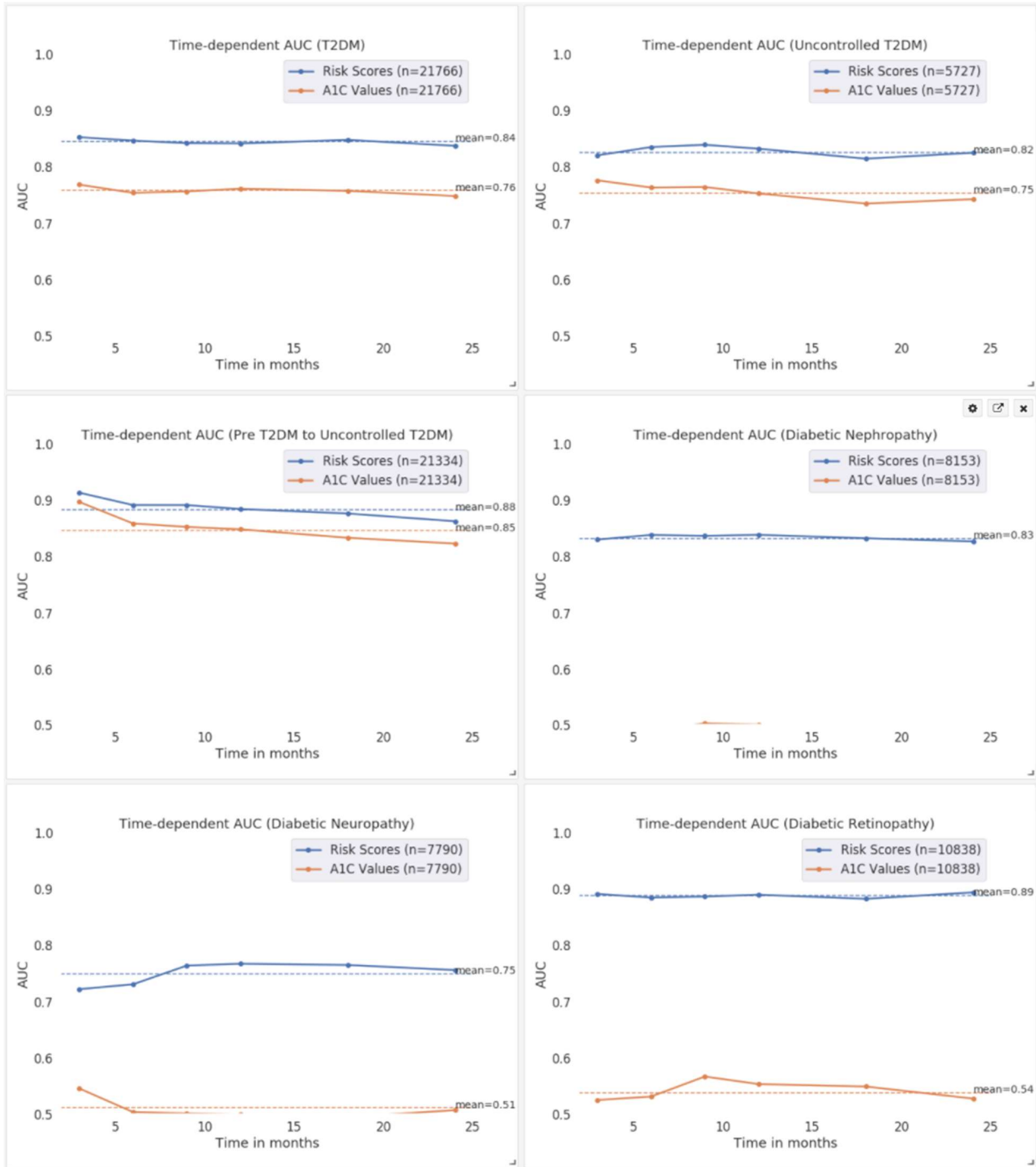

# Appendix C: Threshold Optimization for each Time Point

This section shows how the thresholds for the probability of experiencing the event were determined. The thresholds were optimized by maximizing the balanced accuracy score, which is the average of sensitivity and specificity. The table and charts below display the optimal threshold at each time point and for each task.

|  | 3 | 6 | 9 | 12 | 18 | 24 |
|---|---|---|---|---|---|---|
| T2DM | 0.02 | 0.04 | 0.05 | 0.05 | 0.08 | 0.10 |
| Uncontrolled T2DM | 0.02 | 0.04 | 0.06 | 0.07 | 0.11 | 0.15 |
| Pre T2DM to Uncontrolled T2DM | 0.01 | 0.01 | 0.01 | 0.02 | 0.03 | 0.03 |
| Diabetic Nephropathy | 0.03 | 0.04 | 0.06 | 0.08 | 0.13 | 0.15 |
| Diabetic Neuropathy | 0.01 | 0.02 | 0.03 | 0.04 | 0.09 | 0.08 |
| Diabetic Retinopathy | 0.01 | 0.01 | 0.03 | 0.04 | 0.08 | 0.07 |

Pre-DM to DM Prediction

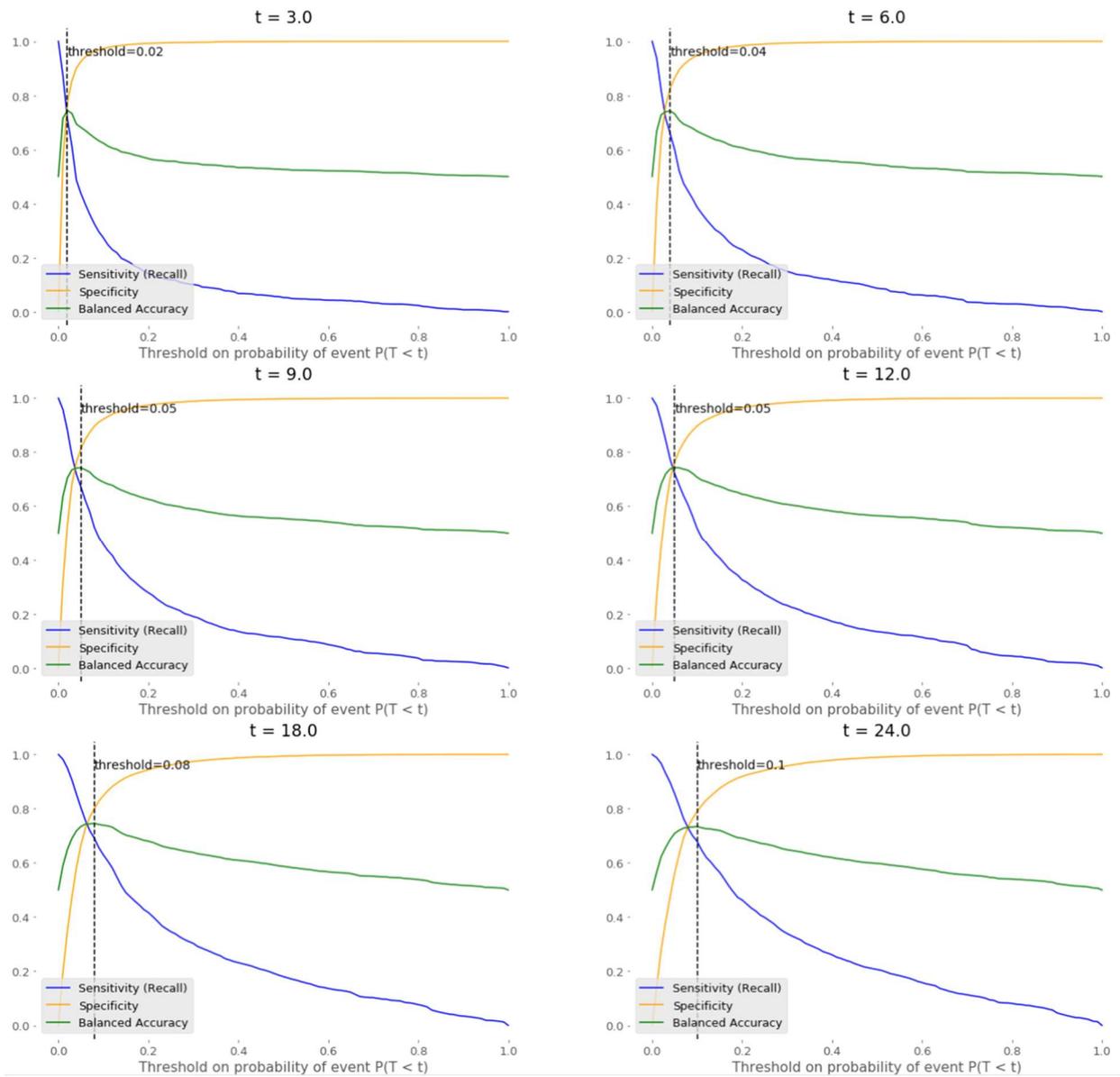

DM to Uncontrolled DM Prediction

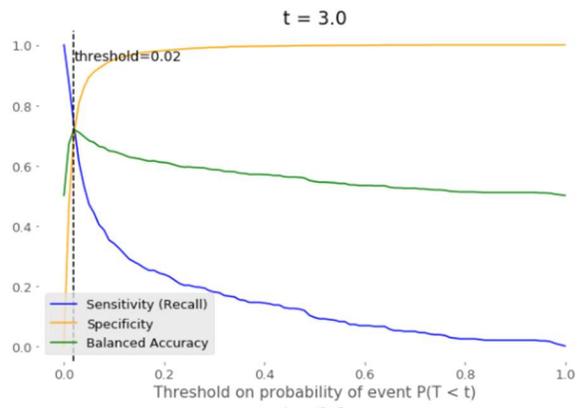
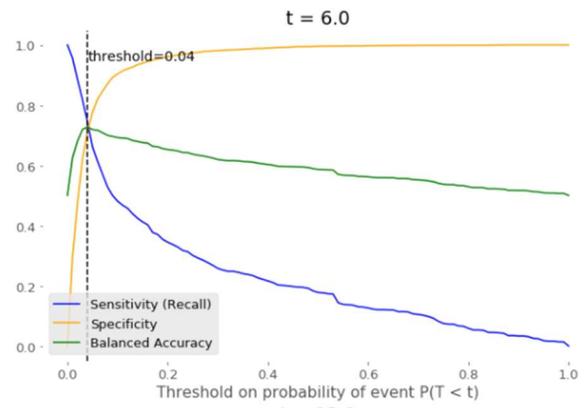
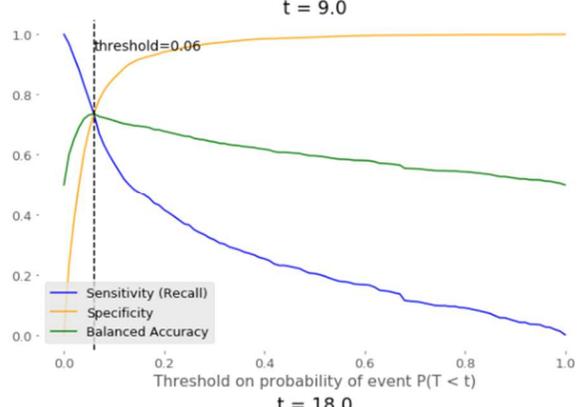
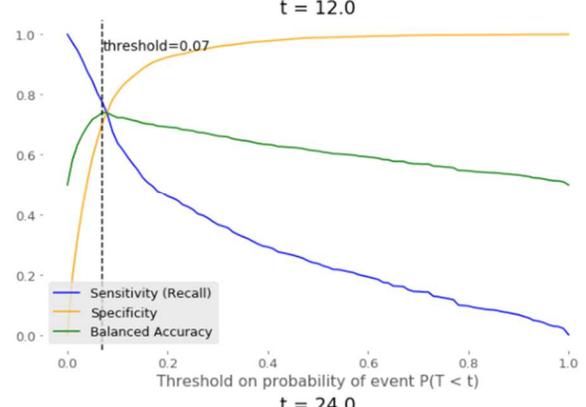
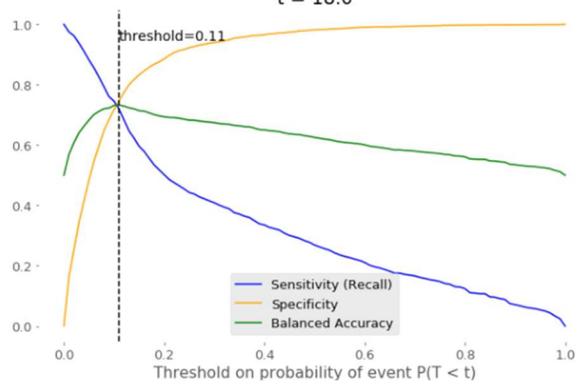
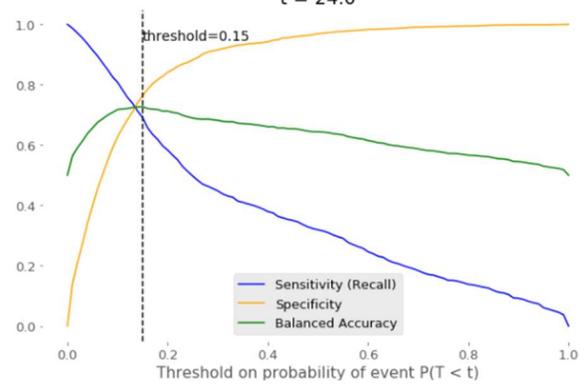

## Pre-DM to Uncontrolled DM Prediction

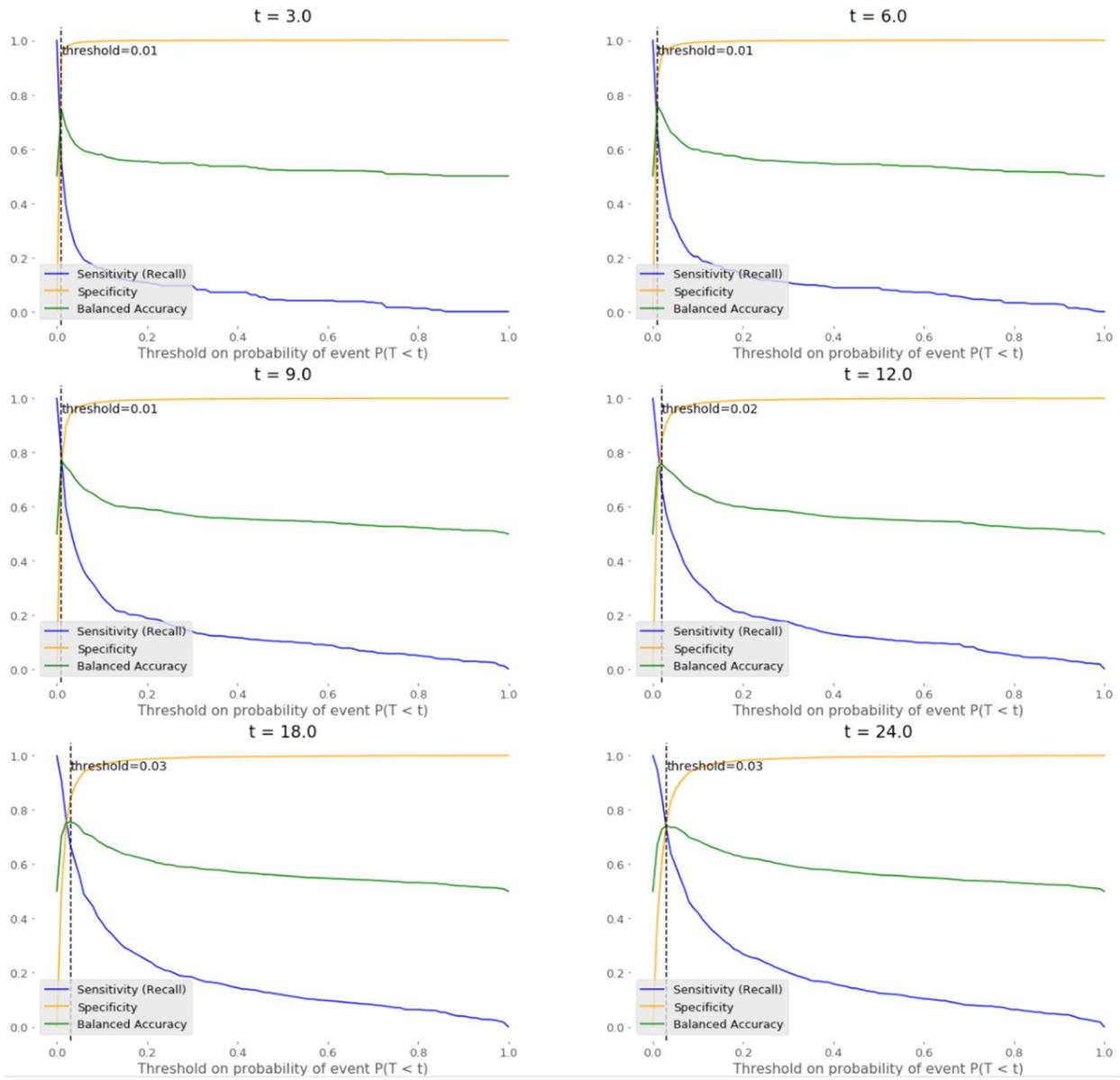

## DM to Diabetic Nephropathy Prediction

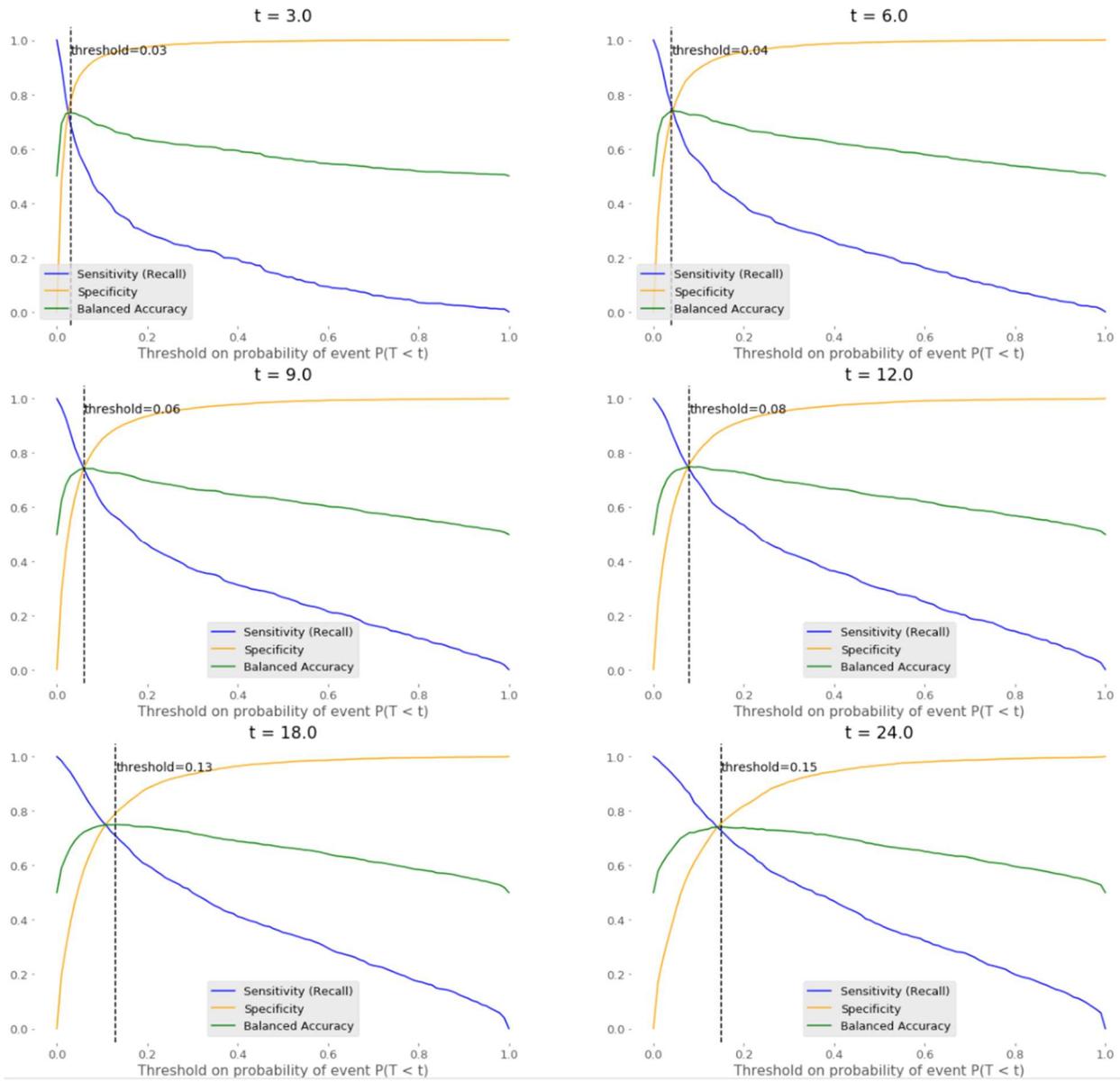

DM to Diabetic Neuropathy Prediction

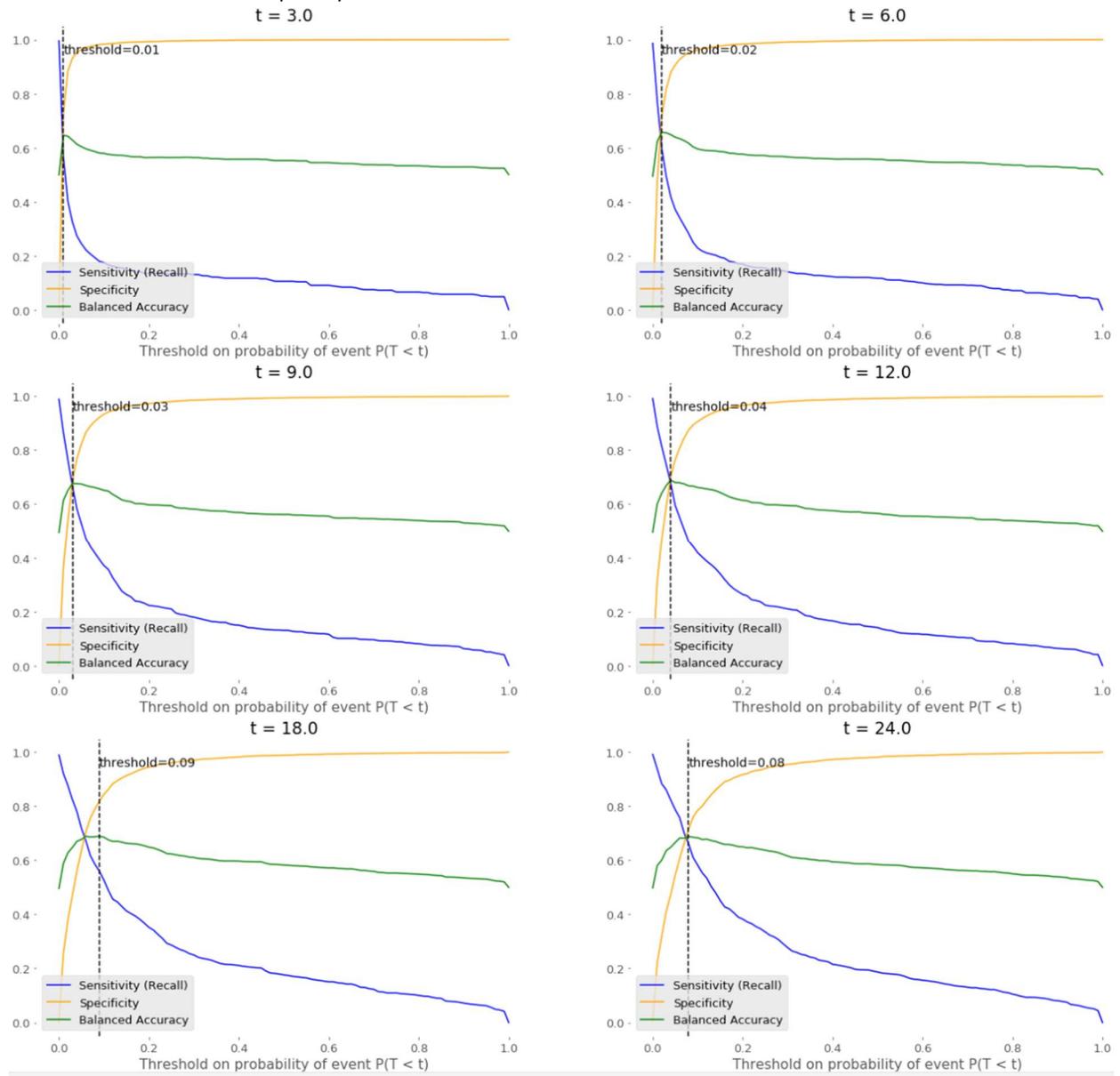

DM to Diabetic Retinopathy Prediction

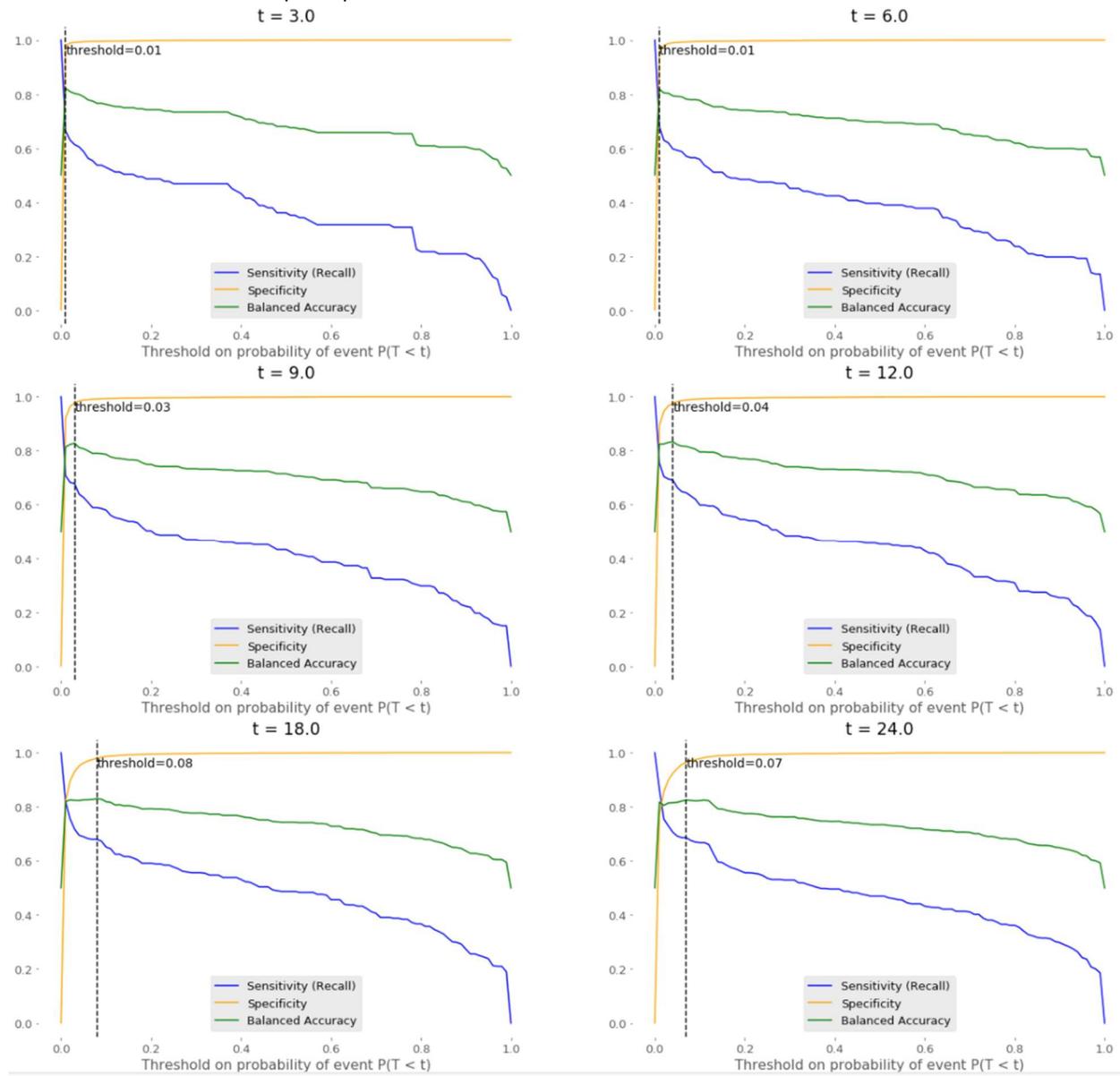

# Appendix D: Model Calibration Experiment

This model calibration approach was proposed by [15] and adopted by [16], [17], where a set of models predicting diabetes complications was trained on one population (U.S., 1980s), and calibrated for a different population (U.K., 2000s). This approach may be useful for the diabetes models in this document, where models pre-trained by KenSci can be quickly deployed at multiple customers, and calibrated for each target patient population.

## Methodology

1. Split data into in-sample (2016-01-01 to 2019-06-30; 2.5 years) and out-of-sample (2019-07-01 to 2020-06-30; 1 year).
2. Further split in-sample data into a 70% training set and 30% in-sample test set.
3. Train model on in-sample training set based on selected features and hyperparameters from Iteration 3.
4. Test model on in-sample test set.
5. Test model on out-of-sample test set, without calibration.
6. Calibrate model using the method in [1].
7. Re-test model on out-of-sample test set, after calibration.

## Summary of Results

Model metrics on the out-of-sample test sets, before and after calibration:

| Task | Concordance Index[2] | Integrated Brier Score | | Median Absolute Error of Survival Curves | | Calibration Coefficient |
|---|---|---|---|---|---|---|
| | | Before | After | Before | After | |
| **Pre-DM to DM** | 0.62 | 0.033 | 0.030 | 1,815 | 5.8 | 0.032 |
| **DM to U-DM** | 0.60 | 0.052 | 0.047 | 837 | 5.3 | 0.053 |
| **Pre-DM to U-DM** | 0.71 | 0.0099 | 0.0088 | 144 | 0.2 | -0.090 |
| **D. Nephropathy** | 0.69 | 0.051 | 0.044 | 266 | 3.2 | -0.012 |
| **D. Neuropathy** | 0.54 | 0.040 | 0.029 | 43 | 0.9 | -0.957 |

---

[2] Concordance index is not affected by calibration, because no covariates were added to the calibration model in this experiment, thus the rank order of predictions between instances remains the same.

| D. Retinopathy | 0.67 | 0.0099 | 0.0075 | 152 | 0.2 | 0.015 |

The following sections report the detailed results of model calibration for each task. Although the models are trained on the same features and using the same parameters, they are trained using a different sub-set of data (limited by date range), so the reported metrics will not be the same as those in the main document.

Key Observations & Learnings
1. Calibration reduced Integrated Brier Score and Median Absolute Error for all tasks, and closed the gap between the predicted and actual survival functions in the target population. This can be clearly seen in the charts comparing the predicted survival curve before and after calibration versus the actual survival curve. Thus, the calibrated models can provide more accurate estimates of a patient's survival (or disease) probability over time. Thus, calibration is useful where it is important to know the survival (or disease) probability (e.g. probability of a patient getting diabetic retinopathy in 12 months) for a specific patient.
2. On the other hand, when no additional covariates are added to the calibration model (as in this experiment), calibration has no effect on the rank order of the predictions, nor the Concordance Index. Without additional covariates, the calibration model is a linear function of the risk predicted by the original model. Therefore, while the calibration model may predict different absolute risks for each instance, the relative predicted risks between instances does not change. By extension, Concordance Index is identical for both models. Hence, calibration is not necessary if (a) there are no new covariates, and (b) the primary use case is to rank patients from high to low risk but absolute probabilities are not important.
3. In order to further optimize Concordance Index (i.e. correct rank order of predicted risks for the target population), additional covariates must be added to the calibration model – these can be covariates existing in the original model (effectively calibrating the coefficients of selected covariates), or new covariates that become available (adding new information).
4. The amount of data used for calibration should be at least as long as the intended prediction horizon. For example, if the model is to be used to predict risks up to 18 months in the future, then at least 18 months of data will be needed. This is because the calibration model needs to be able to estimate the baseline hazard function for the target population at least up to that time period.

## Pre-DM to DM

Number of instances and proportion of class labels in training, validation and test sets:

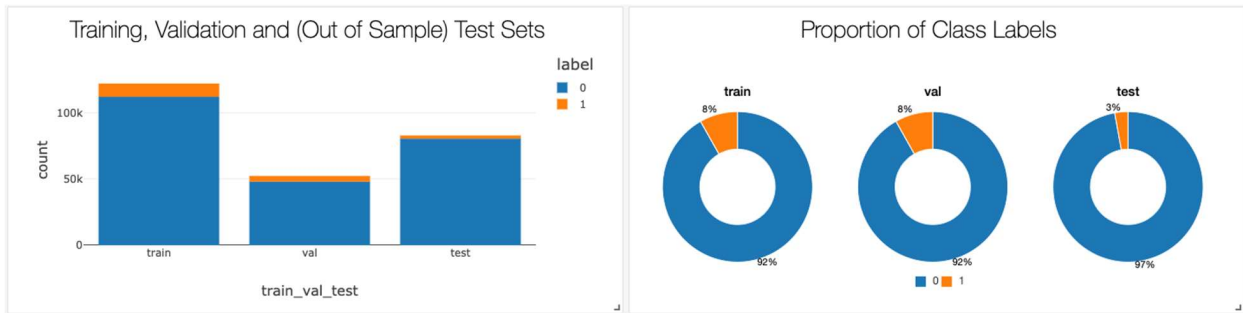

Distribution of features in training set:

|  | count | mean | std | min | 25% | 50% | 75% | max |
|---|---|---|---|---|---|---|---|---|
| age | 122501.0 | 64.264839 | 14.672148 | 18.00 | 56.000 | 66.000 | 75.000 | 107.00 |
| Male_sex | 122501.0 | 0.375336 | 0.484211 | 0.00 | 0.000 | 0.000 | 1.000 | 1.00 |
| Hispanic_or_Latino_ethnicity | 122501.0 | 0.056302 | 0.230504 | 0.00 | 0.000 | 0.000 | 0.000 | 1.00 |
| maxhbA1CInPast365Days | 122501.0 | 6.145194 | 0.542504 | 3.80 | 5.900 | 6.121 | 6.228 | 9.80 |
| lasthbA1CInPast365Days | 122501.0 | 6.032503 | 0.494381 | 3.80 | 5.800 | 6.022 | 6.103 | 8.90 |
| meandiastolicinPast365Days | 122501.0 | 74.578806 | 8.462522 | 43.63 | 69.900 | 74.600 | 79.696 | 107.50 |
| meansystolicinPast365Days | 122501.0 | 130.583874 | 12.925398 | 81.00 | 123.100 | 130.162 | 136.930 | 180.00 |
| BMI | 122501.0 | 32.062367 | 6.950208 | 11.66 | 28.009 | 31.894 | 35.243 | 273.52 |
| meantriglyceroidsInPast365Days | 122501.0 | 142.243926 | 48.492521 | 22.00 | 119.839 | 142.732 | 148.000 | 466.00 |
| meanldlcholesterolInPast365Days | 122501.0 | 95.220383 | 23.718540 | 3.78 | 84.968 | 97.608 | 107.000 | 201.00 |
| meanhdlcholesterolInPast365Days | 122501.0 | 47.528001 | 10.470784 | 6.00 | 42.000 | 46.000 | 52.436 | 98.00 |
| has_BLD005_Past12Months | 122501.0 | 0.001673 | 0.040874 | 0.00 | 0.000 | 0.000 | 0.000 | 1.00 |
| has_CIR007_Past12Months | 122501.0 | 0.195484 | 0.396575 | 0.00 | 0.000 | 0.000 | 0.000 | 1.00 |
| has_CIR008_Past12Months | 122501.0 | 0.008980 | 0.094334 | 0.00 | 0.000 | 0.000 | 0.000 | 1.00 |
| has_NEO016_Past12Months | 122501.0 | 0.002498 | 0.049917 | 0.00 | 0.000 | 0.000 | 0.000 | 1.00 |
| has_PRG029 | 122501.0 | 0.015151 | 0.122153 | 0.00 | 0.000 | 0.000 | 0.000 | 1.00 |
| has_FAC006 | 122501.0 | 0.040906 | 0.198073 | 0.00 | 0.000 | 0.000 | 0.000 | 1.00 |
| has_DIG018 | 122501.0 | 0.023894 | 0.152719 | 0.00 | 0.000 | 0.000 | 0.000 | 1.00 |
| has_SKN003 | 122501.0 | 0.055763 | 0.229464 | 0.00 | 0.000 | 0.000 | 0.000 | 1.00 |
| has_RSP015 | 122501.0 | 0.002327 | 0.048178 | 0.00 | 0.000 | 0.000 | 0.000 | 1.00 |
| has_NEO017 | 122501.0 | 0.012457 | 0.110914 | 0.00 | 0.000 | 0.000 | 0.000 | 1.00 |
| has_INJ057 | 122501.0 | 0.001747 | 0.041760 | 0.00 | 0.000 | 0.000 | 0.000 | 1.00 |

Distribution of features in out-of-sample test set:

|  | count | mean | std | min | 25% | 50% | 75% | max |
|---|---|---|---|---|---|---|---|---|
| age | 83169.0 | 64.928050 | 14.577695 | 18.00 | 57.000 | 67.000 | 75.000 | 103.00 |
| Male_sex | 83169.0 | 0.371004 | 0.483076 | 0.00 | 0.000 | 0.000 | 1.000 | 1.00 |
| Hispanic_or_Latino_ethnicity | 83169.0 | 0.050848 | 0.219689 | 0.00 | 0.000 | 0.000 | 0.000 | 1.00 |
| maxhbA1CInPast365Days | 83169.0 | 6.144681 | 0.521625 | 3.00 | 5.900 | 6.121 | 6.228 | 8.90 |
| lasthbA1CInPast365Days | 83169.0 | 6.040098 | 0.482872 | 2.20 | 5.800 | 6.022 | 6.103 | 8.90 |
| meandiastolicinPast365Days | 83169.0 | 74.433770 | 4.714079 | 44.00 | 72.425 | 74.678 | 77.667 | 107.00 |
| meansystolicinPast365Days | 83169.0 | 130.417804 | 6.406752 | 82.00 | 127.945 | 130.162 | 131.494 | 180.00 |
| BMI | 83169.0 | 32.058567 | 7.273502 | 2.71 | 27.876 | 31.690 | 35.243 | 326.11 |
| meantriglyceroidsInPast365Days | 83169.0 | 139.865553 | 46.986997 | 11.00 | 119.839 | 142.732 | 147.721 | 468.50 |
| meanldlcholesterolInPast365Days | 83169.0 | 95.183594 | 23.392690 | 2.80 | 84.968 | 97.608 | 107.302 | 201.00 |
| meanhdlcholesterolInPast365Days | 83169.0 | 48.551459 | 10.051132 | 8.00 | 42.348 | 49.000 | 52.436 | 98.00 |
| has_BLD005_Past12Months | 83169.0 | 0.000385 | 0.019612 | 0.00 | 0.000 | 0.000 | 0.000 | 1.00 |
| has_CIR007_Past12Months | 83169.0 | 0.101781 | 0.302362 | 0.00 | 0.000 | 0.000 | 0.000 | 1.00 |
| has_CIR008_Past12Months | 83169.0 | 0.010028 | 0.099636 | 0.00 | 0.000 | 0.000 | 0.000 | 1.00 |
| has_NEO016_Past12Months | 83169.0 | 0.004365 | 0.065921 | 0.00 | 0.000 | 0.000 | 0.000 | 1.00 |
| has_PRG029 | 83169.0 | 0.016845 | 0.128692 | 0.00 | 0.000 | 0.000 | 0.000 | 1.00 |
| has_FAC006 | 83169.0 | 0.040075 | 0.196136 | 0.00 | 0.000 | 0.000 | 0.000 | 1.00 |
| has_DIG018 | 83169.0 | 0.018853 | 0.136007 | 0.00 | 0.000 | 0.000 | 0.000 | 1.00 |
| has_SKN003 | 83169.0 | 0.045365 | 0.208106 | 0.00 | 0.000 | 0.000 | 0.000 | 1.00 |
| has_RSP015 | 83169.0 | 0.003547 | 0.059451 | 0.00 | 0.000 | 0.000 | 0.000 | 1.00 |
| has_NEO017 | 83169.0 | 0.010677 | 0.102777 | 0.00 | 0.000 | 0.000 | 0.000 | 1.00 |
| has_INJ057 | 83169.0 | 0.001515 | 0.038894 | 0.00 | 0.000 | 0.000 | 0.000 | 1.00 |

Calibration coefficient $\beta$ (confidence interval): 0.032 (0.031, 0.032)

Model metrics:

| Set | Concordance Index | Integrated Brier Score |
|---|---|---|
| **Train** | 0.56 | 0.079 |
| **Validation** | 0.59 | 0.079 |
| **Test (before calibration)** | 0.63 | 0.033 |
| **Test (after calibration)** | 0.63 | 0.030 |

Predicted vs Actual survival curves for test set, before and after calibration:

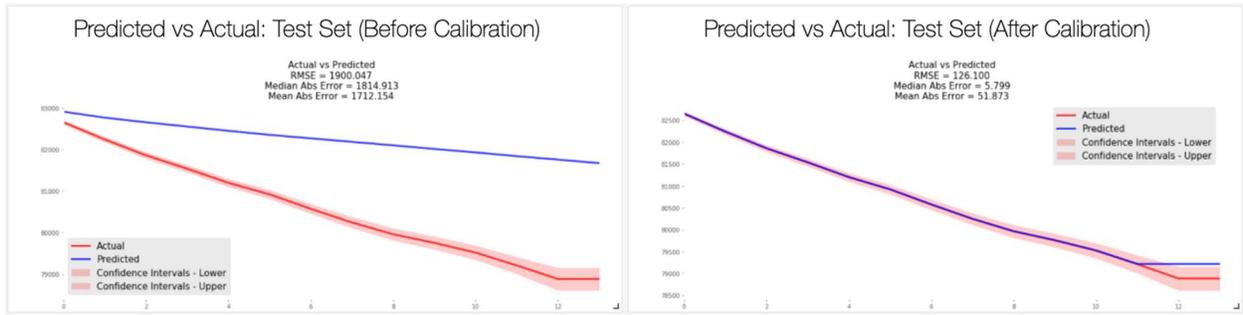

DM to Uncontrolled DM

Number of instances and proportion of class labels in training, validation and test sets:

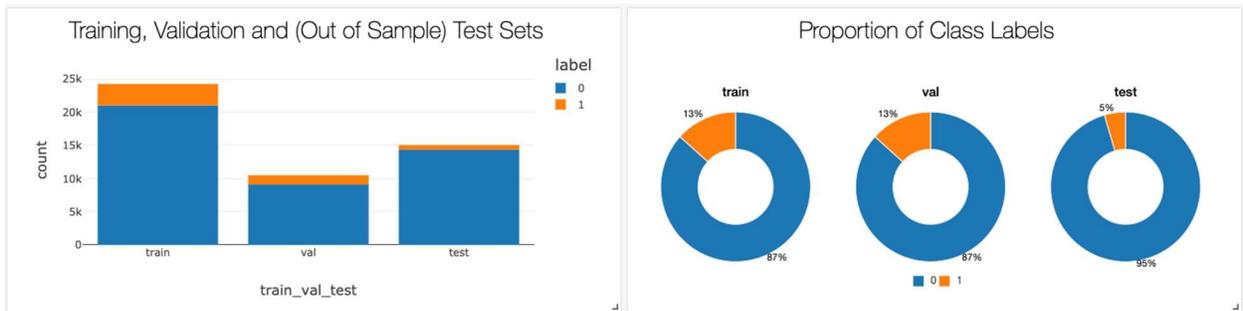

Distribution of features in training set:

|  | count | mean | std | min | 25% | 50% | 75% | max |
|---|---|---|---|---|---|---|---|---|
| age | 24277.0 | 65.975697 | 12.953298 | 18.00 | 58.00 | 68.000 | 75.000 | 102.00 |
| Male_sex | 24277.0 | 0.453969 | 0.497887 | 0.00 | 0.00 | 0.000 | 1.000 | 1.00 |
| Hispanic_or_Latino_ethnicity | 24277.0 | 0.057585 | 0.232963 | 0.00 | 0.00 | 0.000 | 0.000 | 1.00 |
| lasthbA1CInPast365Days | 24277.0 | 6.868240 | 0.775740 | 4.00 | 6.50 | 6.866 | 7.200 | 8.90 |
| meandiastolicinPast365Days | 24277.0 | 72.817306 | 8.881113 | 44.00 | 67.77 | 73.000 | 78.400 | 107.56 |
| meansystolicinPast365Days | 24277.0 | 131.733696 | 13.430075 | 83.00 | 123.82 | 131.065 | 138.500 | 180.00 |
| BMI | 24277.0 | 32.698618 | 6.376725 | 10.37 | 29.40 | 32.400 | 35.320 | 72.56 |
| meantriglyceroidsInPast365Days | 24277.0 | 161.643379 | 52.089780 | 20.00 | 140.50 | 163.096 | 170.874 | 456.50 |
| meanldlcholesterolInPast365Days | 24277.0 | 85.460050 | 21.733265 | 8.00 | 80.72 | 85.917 | 95.000 | 200.00 |
| meanhdlcholesterolInPast365Days | 24277.0 | 43.260053 | 9.102808 | 5.00 | 39.00 | 42.000 | 47.864 | 97.00 |
| has_MAL004_Past12Months | 24277.0 | 0.000206 | 0.014350 | 0.00 | 0.00 | 0.000 | 0.000 | 1.00 |
| has_CIR007_Past12Months | 24277.0 | 0.228611 | 0.419947 | 0.00 | 0.00 | 0.000 | 0.000 | 1.00 |
| has_NEO061_Past12Months | 24277.0 | 0.001442 | 0.037943 | 0.00 | 0.00 | 0.000 | 0.000 | 1.00 |
| has_CIR008_Past12Months | 24277.0 | 0.009309 | 0.096036 | 0.00 | 0.00 | 0.000 | 0.000 | 1.00 |
| has_EXT030 | 24277.0 | 0.001524 | 0.039010 | 0.00 | 0.00 | 0.000 | 0.000 | 1.00 |
| has_NEO029 | 24277.0 | 0.003336 | 0.057667 | 0.00 | 0.00 | 0.000 | 0.000 | 1.00 |
| has_END013 | 24277.0 | 0.004613 | 0.067767 | 0.00 | 0.00 | 0.000 | 0.000 | 1.00 |
| has_CIR038 | 24277.0 | 0.007414 | 0.085789 | 0.00 | 0.00 | 0.000 | 0.000 | 1.00 |
| has_MUS029 | 24277.0 | 0.007291 | 0.085076 | 0.00 | 0.00 | 0.000 | 0.000 | 1.00 |

Distribution of features in out-of-sample test set:

|  | count | mean | std | min | 25% | 50% | 75% | max |
|---|---|---|---|---|---|---|---|---|
| age | 15060.0 | 67.176228 | 13.144581 | 19.000 | 59.000 | 69.000 | 77.0000 | 100.00 |
| Male_sex | 15060.0 | 0.451328 | 0.497642 | 0.000 | 0.000 | 0.000 | 1.0000 | 1.00 |
| Hispanic_or_Latino_ethnicity | 15060.0 | 0.072776 | 0.259776 | 0.000 | 0.000 | 0.000 | 0.0000 | 1.00 |
| lasthbA1CInPast365Days | 15060.0 | 6.882857 | 0.749252 | 3.600 | 6.600 | 6.866 | 7.2000 | 8.90 |
| meandiastolicinPast365Days | 15060.0 | 72.492741 | 5.145955 | 43.620 | 70.137 | 73.154 | 75.9140 | 104.21 |
| meansystolicinPast365Days | 15060.0 | 131.365176 | 6.472913 | 90.000 | 129.358 | 131.065 | 133.0550 | 179.42 |
| BMI | 15060.0 | 32.707763 | 6.722531 | 13.430 | 28.978 | 32.420 | 36.0825 | 80.03 |
| meantriglyceroidsInPast365Days | 15060.0 | 159.501333 | 52.165235 | 27.000 | 132.000 | 160.500 | 170.8740 | 458.00 |
| meanldlcholesterolInPast365Days | 15060.0 | 85.987783 | 21.523274 | 11.000 | 80.000 | 85.917 | 96.0710 | 200.00 |
| meanhdlcholesterolInPast365Days | 15060.0 | 44.188717 | 8.854655 | 10.333 | 39.813 | 44.000 | 47.8640 | 97.00 |
| has_MAL004_Past12Months | 15060.0 | 0.001594 | 0.039890 | 0.000 | 0.000 | 0.000 | 0.0000 | 1.00 |
| has_CIR007_Past12Months | 15060.0 | 0.121846 | 0.327119 | 0.000 | 0.000 | 0.000 | 0.0000 | 1.00 |
| has_NEO061_Past12Months | 15060.0 | 0.000000 | 0.000000 | 0.000 | 0.000 | 0.000 | 0.0000 | 0.00 |
| has_CIR008_Past12Months | 15060.0 | 0.015405 | 0.123161 | 0.000 | 0.000 | 0.000 | 0.0000 | 1.00 |
| has_EXT030 | 15060.0 | 0.000531 | 0.023043 | 0.000 | 0.000 | 0.000 | 0.0000 | 1.00 |
| has_NEO029 | 15060.0 | 0.007769 | 0.087801 | 0.000 | 0.000 | 0.000 | 0.0000 | 1.00 |
| has_END013 | 15060.0 | 0.005378 | 0.073143 | 0.000 | 0.000 | 0.000 | 0.0000 | 1.00 |
| has_CIR038 | 15060.0 | 0.003652 | 0.060324 | 0.000 | 0.000 | 0.000 | 0.0000 | 1.00 |
| has_MUS029 | 15060.0 | 0.007437 | 0.085919 | 0.000 | 0.000 | 0.000 | 0.0000 | 1.00 |

Calibration coefficient $\beta$ (confidence interval): 0.053 (0.050, 0.056)

Model metrics:

| Set | Concordance Index | Integrated Brier Score |
|---|---|---|
| **Train** | 0.55 | 0.128 |
| **Validation** | 0.55 | 0.127 |
| **Test (before calibration)** | 0.60 | 0.052 |
| **Test (after calibration)** | 0.60 | 0.047 |

Predicted vs Actual survival curves for test set, before and after calibration:

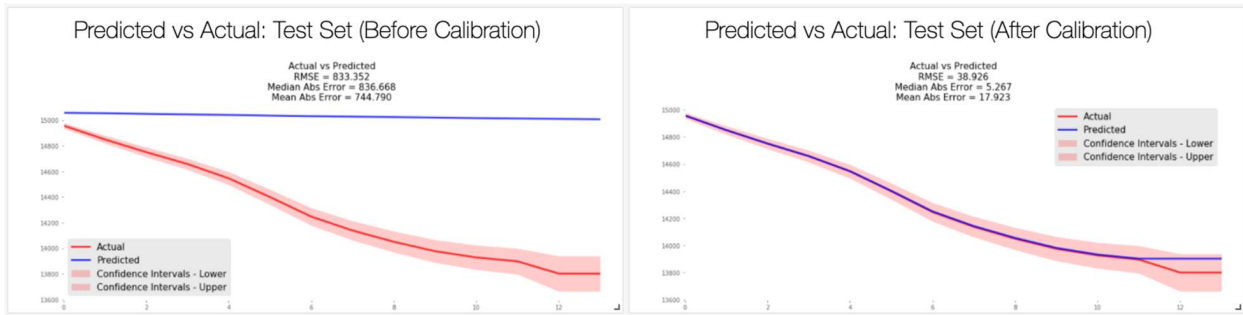

## Pre-DM to Uncontrolled DM
Number of instances and proportion of class labels in training, validation and test sets:

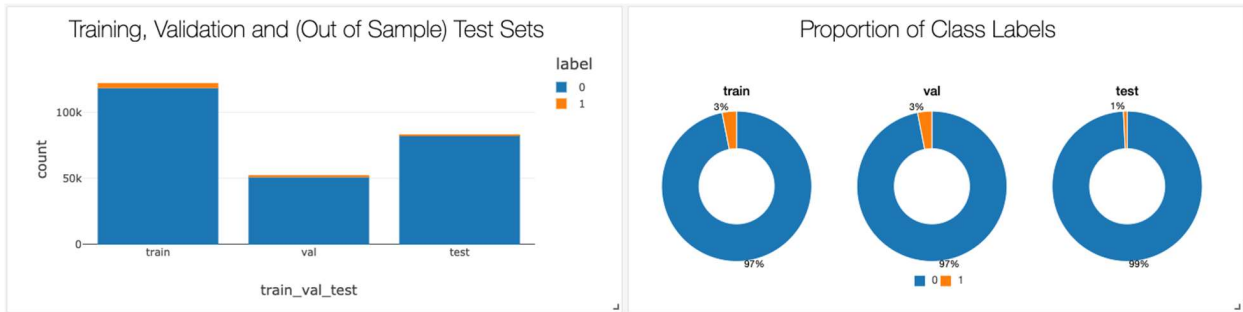

Distribution of features in training set:

| | count | mean | std | min | 25% | 50% | 75% | max |
|---|---|---|---|---|---|---|---|---|
| age | 122342.0 | 64.313506 | 14.659463 | 18.00 | 56.000 | 66.000 | 75.000 | 107.00 |
| Male_sex | 122342.0 | 0.373453 | 0.483723 | 0.00 | 0.000 | 0.000 | 1.000 | 1.00 |
| Hispanic_or_Latino_ethnicity | 122342.0 | 0.055991 | 0.229904 | 0.00 | 0.000 | 0.000 | 0.000 | 1.00 |
| lasthbA1CInPast365Days | 122342.0 | 6.032703 | 0.494929 | 3.70 | 5.800 | 6.022 | 6.103 | 8.90 |
| meandiastolicinPast365Days | 122342.0 | 74.523529 | 8.456022 | 43.63 | 69.860 | 74.520 | 79.620 | 107.67 |
| meansystolicinPast365Days | 122342.0 | 130.536081 | 12.899920 | 81.00 | 123.000 | 130.162 | 136.820 | 180.00 |
| BMI | 122342.0 | 32.047383 | 6.861005 | 11.66 | 28.009 | 31.894 | 35.243 | 273.52 |
| meantriglyceroidsInPast365Days | 122342.0 | 142.364649 | 49.023127 | 22.00 | 119.839 | 142.732 | 148.000 | 466.00 |
| meanldlcholesterolInPast365Days | 122342.0 | 95.223014 | 23.739237 | 3.78 | 84.968 | 97.608 | 107.000 | 201.00 |
| meanhdlcholesterolInPast365Days | 122342.0 | 47.592181 | 10.548001 | 6.00 | 42.000 | 46.333 | 52.436 | 98.00 |
| has_BLD005_Past12Months | 122342.0 | 0.001594 | 0.039892 | 0.00 | 0.000 | 0.000 | 0.000 | 1.00 |
| has_SKN005_Past12Months | 122342.0 | 0.005354 | 0.072974 | 0.00 | 0.000 | 0.000 | 0.000 | 1.00 |
| has_INJ014_Past12Months | 122342.0 | 0.000490 | 0.022140 | 0.00 | 0.000 | 0.000 | 0.000 | 1.00 |
| has_CIR007_Past12Months | 122342.0 | 0.196081 | 0.397033 | 0.00 | 0.000 | 0.000 | 0.000 | 1.00 |
| has_NEO022_Past12Months | 122342.0 | 0.028012 | 0.165007 | 0.00 | 0.000 | 0.000 | 0.000 | 1.00 |
| has_CIR008_Past12Months | 122342.0 | 0.008926 | 0.094054 | 0.00 | 0.000 | 0.000 | 0.000 | 1.00 |
| has_NEO029_Past12Months | 122342.0 | 0.004406 | 0.066229 | 0.00 | 0.000 | 0.000 | 0.000 | 1.00 |
| has_BLD005 | 122342.0 | 0.002501 | 0.049949 | 0.00 | 0.000 | 0.000 | 0.000 | 1.00 |
| has_INJ007 | 122342.0 | 0.004962 | 0.070263 | 0.00 | 0.000 | 0.000 | 0.000 | 1.00 |
| has_INJ054 | 122342.0 | 0.001782 | 0.042175 | 0.00 | 0.000 | 0.000 | 0.000 | 1.00 |
| has_GEN018 | 122342.0 | 0.015056 | 0.121777 | 0.00 | 0.000 | 0.000 | 0.000 | 1.00 |
| has_NEO044 | 122342.0 | 0.001177 | 0.034288 | 0.00 | 0.000 | 0.000 | 0.000 | 1.00 |
| has_INJ035 | 122342.0 | 0.007340 | 0.085360 | 0.00 | 0.000 | 0.000 | 0.000 | 1.00 |
| has_INJ049 | 122342.0 | 0.001954 | 0.044156 | 0.00 | 0.000 | 0.000 | 0.000 | 1.00 |
| has_NEO012 | 122342.0 | 0.006261 | 0.078880 | 0.00 | 0.000 | 0.000 | 0.000 | 1.00 |

Distribution of features in out-of-sample test set:

|  | count | mean | std | min | 25% | 50% | 75% | max |
|---|---|---|---|---|---|---|---|---|
| age | 83169.0 | 64.928050 | 14.577695 | 18.00 | 57.000 | 67.000 | 75.000 | 103.00 |
| Male_sex | 83169.0 | 0.371004 | 0.483076 | 0.00 | 0.000 | 0.000 | 1.000 | 1.00 |
| Hispanic_or_Latino_ethnicity | 83169.0 | 0.050848 | 0.219689 | 0.00 | 0.000 | 0.000 | 0.000 | 1.00 |
| lasthbA1CInPast365Days | 83169.0 | 6.040098 | 0.482872 | 2.20 | 5.800 | 6.022 | 6.103 | 8.90 |
| meandiastolicinPast365Days | 83169.0 | 74.433770 | 4.714079 | 44.00 | 72.425 | 74.678 | 77.667 | 107.00 |
| meansystolicinPast365Days | 83169.0 | 130.417804 | 6.406752 | 82.00 | 127.945 | 130.162 | 131.494 | 180.00 |
| BMI | 83169.0 | 32.058567 | 7.273502 | 2.71 | 27.876 | 31.690 | 35.243 | 326.11 |
| meantriglyceroidsInPast365Days | 83169.0 | 139.865553 | 46.986997 | 11.00 | 119.839 | 142.732 | 147.721 | 468.50 |
| meanldlcholesterolInPast365Days | 83169.0 | 95.183594 | 23.392690 | 2.80 | 84.968 | 97.608 | 107.302 | 201.00 |
| meanhdlcholesterolInPast365Days | 83169.0 | 48.551459 | 10.051132 | 8.00 | 42.348 | 49.000 | 52.436 | 98.00 |
| has_BLD005_Past12Months | 83169.0 | 0.000385 | 0.019612 | 0.00 | 0.000 | 0.000 | 0.000 | 1.00 |
| has_SKN005_Past12Months | 83169.0 | 0.004870 | 0.069613 | 0.00 | 0.000 | 0.000 | 0.000 | 1.00 |
| has_INJ014_Past12Months | 83169.0 | 0.000902 | 0.030016 | 0.00 | 0.000 | 0.000 | 0.000 | 1.00 |
| has_CIR007_Past12Months | 83169.0 | 0.101781 | 0.302362 | 0.00 | 0.000 | 0.000 | 0.000 | 1.00 |
| has_NEO022_Past12Months | 83169.0 | 0.023831 | 0.152523 | 0.00 | 0.000 | 0.000 | 0.000 | 1.00 |
| has_CIR008_Past12Months | 83169.0 | 0.010028 | 0.099636 | 0.00 | 0.000 | 0.000 | 0.000 | 1.00 |
| has_NEO029_Past12Months | 83169.0 | 0.004292 | 0.065377 | 0.00 | 0.000 | 0.000 | 0.000 | 1.00 |
| has_BLD005 | 83169.0 | 0.001335 | 0.036508 | 0.00 | 0.000 | 0.000 | 0.000 | 1.00 |
| has_INJ007 | 83169.0 | 0.004377 | 0.066012 | 0.00 | 0.000 | 0.000 | 0.000 | 1.00 |
| has_INJ054 | 83169.0 | 0.001238 | 0.035170 | 0.00 | 0.000 | 0.000 | 0.000 | 1.00 |
| has_GEN018 | 83169.0 | 0.014597 | 0.119933 | 0.00 | 0.000 | 0.000 | 0.000 | 1.00 |
| has_NEO044 | 83169.0 | 0.001130 | 0.033600 | 0.00 | 0.000 | 0.000 | 0.000 | 1.00 |
| has_INJ035 | 83169.0 | 0.006529 | 0.080538 | 0.00 | 0.000 | 0.000 | 0.000 | 1.00 |
| has_INJ049 | 83169.0 | 0.001816 | 0.042571 | 0.00 | 0.000 | 0.000 | 0.000 | 1.00 |
| has_NEO012 | 83169.0 | 0.004244 | 0.065011 | 0.00 | 0.000 | 0.000 | 0.000 | 1.00 |

Calibration coefficient $\beta$ (confidence interval): -0.09 (-0.096, -0.083)

Model metrics:

| Set | Concordance Index | Integrated Brier Score |
|---|---|---|
| Train | 0.67 | 0.0359 |
| Validation | 0.58 | 0.0355 |
| Test (before calibration) | 0.71 | 0.0099 |
| Test (after calibration) | 0.71 | 0.0088 |

Predicted vs Actual survival curves for test set, before and after calibration:

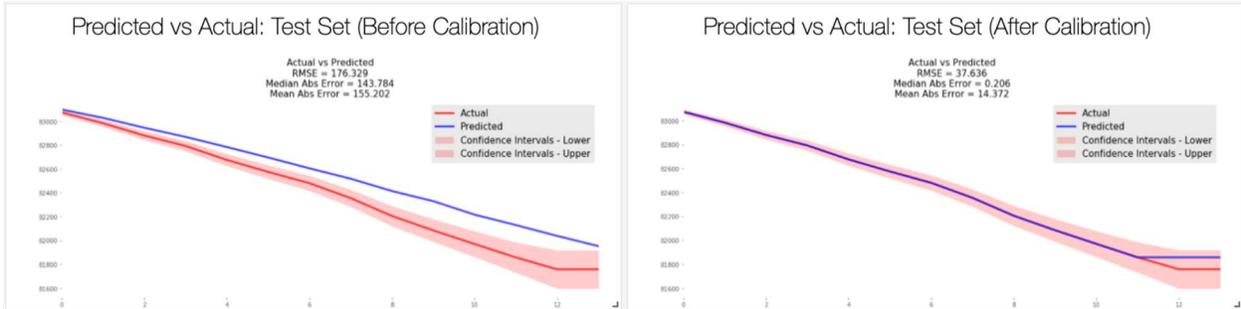

## Diabetic Nephropathy

Number of instances and proportion of class labels in training, validation and test sets:

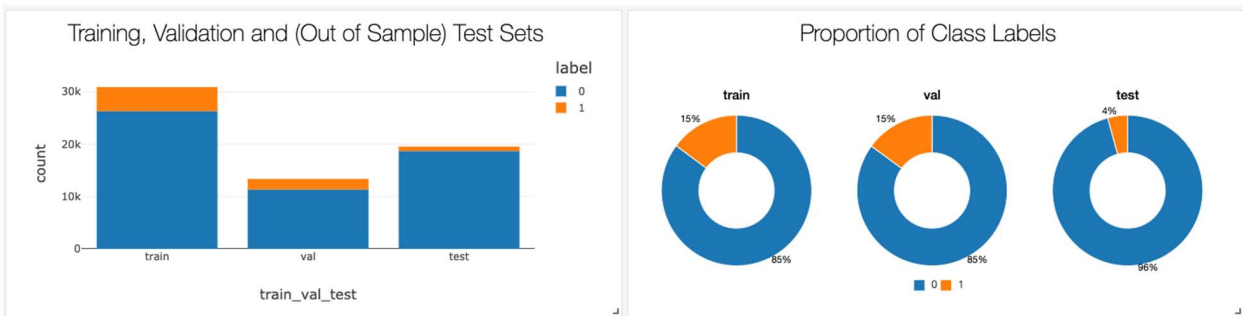

Distribution of features in training set:

| | count | mean | std | min | 25% | 50% | 75% | max |
|---|---|---|---|---|---|---|---|---|
| age | 30891.0 | 63.458192 | 12.980580 | 18.00 | 55.000 | 65.000 | 72.000 | 99.00 |
| Male_sex | 30891.0 | 0.432003 | 0.495363 | 0.00 | 0.000 | 0.000 | 1.000 | 1.00 |
| Hispanic_or_Latino_ethnicity | 30891.0 | 0.072125 | 0.258698 | 0.00 | 0.000 | 0.000 | 0.000 | 1.00 |
| lasthbA1CInPast365Days | 30891.0 | 7.591038 | 1.215052 | 3.90 | 6.900 | 7.529 | 8.000 | 12.70 |
| meandiastolicinPast365Days | 30891.0 | 74.101690 | 8.748834 | 44.00 | 69.110 | 74.041 | 79.528 | 107.00 |
| meansystolicinPast365Days | 30891.0 | 131.945299 | 13.249303 | 84.33 | 124.000 | 131.574 | 138.420 | 180.00 |
| BMI | 30891.0 | 33.759723 | 6.744802 | 13.49 | 29.960 | 33.718 | 37.340 | 76.98 |
| meantriglyceroidsInPast365Days | 30891.0 | 169.731001 | 57.312733 | 19.00 | 144.500 | 167.131 | 182.483 | 457.00 |
| meanldlcholesterolInPast365Days | 30891.0 | 89.362962 | 23.555869 | 8.00 | 81.282 | 89.776 | 97.772 | 201.00 |
| meanhdlcholesterolInPast365Days | 30891.0 | 42.784989 | 9.134578 | 5.00 | 37.685 | 42.780 | 47.646 | 97.00 |
| has_NEO024_Past12Months | 30891.0 | 0.001780 | 0.042159 | 0.00 | 0.000 | 0.000 | 0.000 | 1.00 |
| has_RSP010_Past12Months | 30891.0 | 0.005762 | 0.075691 | 0.00 | 0.000 | 0.000 | 0.000 | 1.00 |
| has_CIR007_Past12Months | 30891.0 | 0.221294 | 0.415125 | 0.00 | 0.000 | 0.000 | 0.000 | 1.00 |
| has_CIR008_Past12Months | 30891.0 | 0.006928 | 0.082945 | 0.00 | 0.000 | 0.000 | 0.000 | 1.00 |
| has_NVS018 | 30891.0 | 0.010456 | 0.101721 | 0.00 | 0.000 | 0.000 | 0.000 | 1.00 |
| has_NEO050 | 30891.0 | 0.004241 | 0.064984 | 0.00 | 0.000 | 0.000 | 0.000 | 1.00 |
| has_INJ026 | 30891.0 | 0.003788 | 0.061427 | 0.00 | 0.000 | 0.000 | 0.000 | 1.00 |
| has_EXT017 | 30891.0 | 0.004111 | 0.063988 | 0.00 | 0.000 | 0.000 | 0.000 | 1.00 |
| has_CIR034 | 30891.0 | 0.005568 | 0.074412 | 0.00 | 0.000 | 0.000 | 0.000 | 1.00 |
| has_EYE012 | 30891.0 | 0.005180 | 0.071783 | 0.00 | 0.000 | 0.000 | 0.000 | 1.00 |
| has_FAC023 | 30891.0 | 0.051148 | 0.220302 | 0.00 | 0.000 | 0.000 | 0.000 | 1.00 |
| has_NEO008 | 30891.0 | 0.001975 | 0.044394 | 0.00 | 0.000 | 0.000 | 0.000 | 1.00 |
| has_NVS005 | 30891.0 | 0.005180 | 0.071783 | 0.00 | 0.000 | 0.000 | 0.000 | 1.00 |
| has_GEN010 | 30891.0 | 0.015797 | 0.124693 | 0.00 | 0.000 | 0.000 | 0.000 | 1.00 |
| has_END011 | 30891.0 | 0.244861 | 0.430012 | 0.00 | 0.000 | 0.000 | 0.000 | 1.00 |
| has_CIR008 | 30891.0 | 0.031239 | 0.173965 | 0.00 | 0.000 | 0.000 | 0.000 | 1.00 |

Distribution of features in out-of-sample test set:

|  | count | mean | std | min | 25% | 50% | 75% | max |
|---|---|---|---|---|---|---|---|---|
| age | 19522.0 | 63.255916 | 13.163093 | 18.00 | 55.0000 | 64.000 | 72.000 | 100.00 |
| Male_sex | 19522.0 | 0.411945 | 0.492198 | 0.00 | 0.0000 | 0.000 | 1.000 | 1.00 |
| Hispanic_or_Latino_ethnicity | 19522.0 | 0.076375 | 0.265605 | 0.00 | 0.0000 | 0.000 | 0.000 | 1.00 |
| lasthbA1CInPast365Days | 19522.0 | 7.616567 | 1.217427 | 3.50 | 6.9000 | 7.500 | 8.000 | 12.70 |
| meandiastolicinPast365Days | 19522.0 | 74.077505 | 5.176443 | 44.00 | 71.0570 | 74.041 | 76.744 | 106.00 |
| meansystolicinPast365Days | 19522.0 | 131.543621 | 6.478437 | 90.00 | 130.0630 | 131.574 | 133.391 | 179.42 |
| BMI | 19522.0 | 33.938474 | 6.983243 | 13.43 | 29.6325 | 33.718 | 37.340 | 85.68 |
| meantriglyceroidsInPast365Days | 19522.0 | 168.050660 | 59.142183 | 27.00 | 136.0000 | 167.131 | 182.483 | 460.50 |
| meanldlcholesterolInPast365Days | 19522.0 | 90.381401 | 23.498278 | 10.00 | 81.2820 | 89.776 | 97.772 | 200.00 |
| meanhdlcholesterolInPast365Days | 19522.0 | 43.631889 | 9.058859 | 12.00 | 38.5000 | 43.734 | 47.646 | 98.00 |
| has_NEO024_Past12Months | 19522.0 | 0.000922 | 0.030352 | 0.00 | 0.0000 | 0.000 | 0.000 | 1.00 |
| has_RSP010_Past12Months | 19522.0 | 0.003534 | 0.059348 | 0.00 | 0.0000 | 0.000 | 0.000 | 1.00 |
| has_CIR007_Past12Months | 19522.0 | 0.120940 | 0.326066 | 0.00 | 0.0000 | 0.000 | 0.000 | 1.00 |
| has_CIR008_Past12Months | 19522.0 | 0.013062 | 0.113544 | 0.00 | 0.0000 | 0.000 | 0.000 | 1.00 |
| has_NVS018 | 19522.0 | 0.007171 | 0.084382 | 0.00 | 0.0000 | 0.000 | 0.000 | 1.00 |
| has_NEO050 | 19522.0 | 0.003330 | 0.057608 | 0.00 | 0.0000 | 0.000 | 0.000 | 1.00 |
| has_INJ026 | 19522.0 | 0.002459 | 0.049526 | 0.00 | 0.0000 | 0.000 | 0.000 | 1.00 |
| has_EXT017 | 19522.0 | 0.004047 | 0.063487 | 0.00 | 0.0000 | 0.000 | 0.000 | 1.00 |
| has_CIR034 | 19522.0 | 0.003073 | 0.055355 | 0.00 | 0.0000 | 0.000 | 0.000 | 1.00 |
| has_EYE012 | 19522.0 | 0.004200 | 0.064676 | 0.00 | 0.0000 | 0.000 | 0.000 | 1.00 |
| has_FAC023 | 19522.0 | 0.028122 | 0.165326 | 0.00 | 0.0000 | 0.000 | 0.000 | 1.00 |
| has_NEO008 | 19522.0 | 0.004354 | 0.065843 | 0.00 | 0.0000 | 0.000 | 0.000 | 1.00 |
| has_NVS005 | 19522.0 | 0.003381 | 0.058048 | 0.00 | 0.0000 | 0.000 | 0.000 | 1.00 |
| has_GEN010 | 19522.0 | 0.012652 | 0.111772 | 0.00 | 0.0000 | 0.000 | 0.000 | 1.00 |
| has_END011 | 19522.0 | 0.223850 | 0.416833 | 0.00 | 0.0000 | 0.000 | 0.000 | 1.00 |
| has_CIR008 | 19522.0 | 0.032271 | 0.176724 | 0.00 | 0.0000 | 0.000 | 0.000 | 1.00 |

Calibration coefficient $\beta$ (confidence interval): -0.012 (-0.023, -0.002)

Model metrics:

| Set | Concordance Index | Integrated Brier Score |
|---|---|---|
| Train | 0.54 | 0.142 |
| **Validation** | 0.54 | 0.141 |
| **Test (before calibration)** | 0.69 | 0.051 |
| **Test (after calibration)** | 0.69 | 0.044 |

Predicted vs Actual survival curves for test set, before and after calibration:

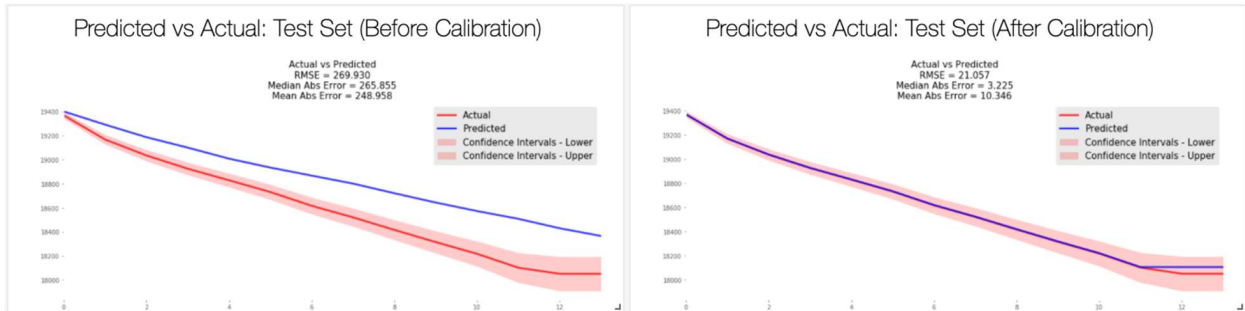

## Diabetic Neuropathy

Number of instances and proportion of class labels in training, validation and test sets:

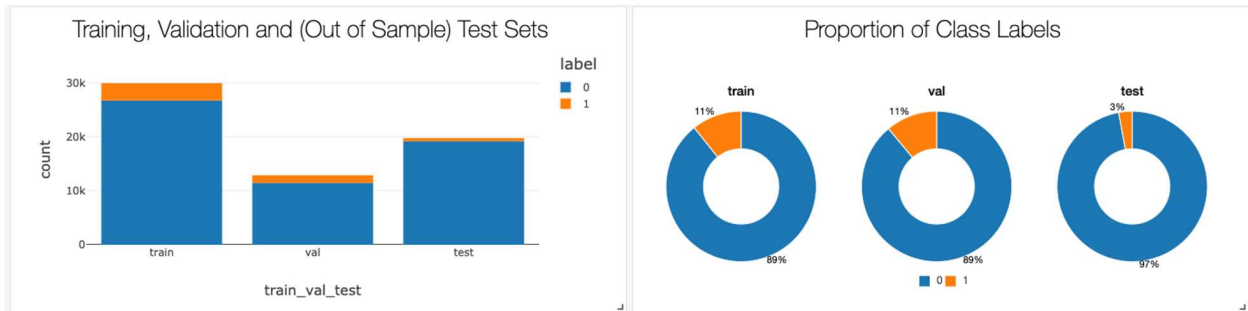

Distribution of features in training set:

|  | count | mean | std | min | 25% | 50% | 75% | max |
|---|---|---|---|---|---|---|---|---|
| age | 30020.0 | 64.242405 | 13.243568 | 18.00 | 56.000 | 66.000 | 73.000 | 102.00 |
| Male_sex | 30020.0 | 0.442971 | 0.496745 | 0.00 | 0.000 | 0.000 | 1.000 | 1.00 |
| Hispanic_or_Latino_ethnicity | 30020.0 | 0.070586 | 0.256137 | 0.00 | 0.000 | 0.000 | 0.000 | 1.00 |
| lasthbA1CInPast365Days | 30020.0 | 7.484320 | 1.180872 | 3.90 | 6.800 | 7.437 | 7.900 | 12.70 |
| meandiastolicinPast365Days | 30020.0 | 73.888056 | 8.770684 | 44.00 | 68.600 | 73.991 | 79.400 | 107.56 |
| meansystolicinPast365Days | 30020.0 | 132.179451 | 13.500032 | 82.00 | 124.120 | 131.921 | 139.000 | 180.00 |
| BMI | 30020.0 | 33.406013 | 6.671339 | 13.49 | 29.480 | 33.677 | 36.820 | 157.25 |
| meantriglyceroidsInPast365Days | 30020.0 | 166.541346 | 56.116023 | 19.00 | 141.750 | 164.470 | 180.836 | 457.00 |
| meanldicholesterolInPast365Days | 30020.0 | 88.532380 | 23.117171 | 8.00 | 81.915 | 89.011 | 98.902 | 201.00 |
| meanhdicholesterolInPast365Days | 30020.0 | 43.018117 | 9.043842 | 5.00 | 38.000 | 42.145 | 47.634 | 97.00 |
| has_INJ004_Past12Months | 30020.0 | 0.007229 | 0.084714 | 0.00 | 0.000 | 0.000 | 0.000 | 1.00 |
| has_NEO043_Past12Months | 30020.0 | 0.005863 | 0.076345 | 0.00 | 0.000 | 0.000 | 0.000 | 1.00 |
| has_NEO012_Past12Months | 30020.0 | 0.002831 | 0.053137 | 0.00 | 0.000 | 0.000 | 0.000 | 1.00 |
| has_NEO030_Past12Months | 30020.0 | 0.040640 | 0.197457 | 0.00 | 0.000 | 0.000 | 0.000 | 1.00 |
| has_CIR007_Past12Months | 30020.0 | 0.218588 | 0.413295 | 0.00 | 0.000 | 0.000 | 0.000 | 1.00 |
| has_MBD003_Past12Months | 30020.0 | 0.007328 | 0.085294 | 0.00 | 0.000 | 0.000 | 0.000 | 1.00 |
| has_CIR008_Past12Months | 30020.0 | 0.007628 | 0.087008 | 0.00 | 0.000 | 0.000 | 0.000 | 1.00 |
| has_FAC013 | 30020.0 | 0.009560 | 0.097310 | 0.00 | 0.000 | 0.000 | 0.000 | 1.00 |
| has_DIG006 | 30020.0 | 0.007795 | 0.087945 | 0.00 | 0.000 | 0.000 | 0.000 | 1.00 |
| has_NEO065 | 30020.0 | 0.004497 | 0.066910 | 0.00 | 0.000 | 0.000 | 0.000 | 1.00 |
| has_FAC004 | 30020.0 | 0.002132 | 0.046124 | 0.00 | 0.000 | 0.000 | 0.000 | 1.00 |
| has_GEN025 | 30020.0 | 0.042538 | 0.201817 | 0.00 | 0.000 | 0.000 | 0.000 | 1.00 |
| has_RSP017 | 30020.0 | 0.011492 | 0.106586 | 0.00 | 0.000 | 0.000 | 0.000 | 1.00 |
| has_NVS004 | 30020.0 | 0.010826 | 0.103486 | 0.00 | 0.000 | 0.000 | 0.000 | 1.00 |
| has_MUS037 | 30020.0 | 0.012891 | 0.112808 | 0.00 | 0.000 | 0.000 | 0.000 | 1.00 |
| has_CIR035 | 30020.0 | 0.008594 | 0.092308 | 0.00 | 0.000 | 0.000 | 0.000 | 1.00 |

Distribution of features in out-of-sample test set:

|  | count | mean | std | min | 25% | 50% | 75% | max |
|---|---|---|---|---|---|---|---|---|
| age | 19815.0 | 64.536563 | 13.452568 | 18.00 | 56.000 | 66.000 | 74.000 | 100.00 |
| Male_sex | 19815.0 | 0.440676 | 0.496481 | 0.00 | 0.000 | 0.000 | 1.000 | 1.00 |
| Hispanic_or_Latino_ethnicity | 19815.0 | 0.077921 | 0.268054 | 0.00 | 0.000 | 0.000 | 0.000 | 1.00 |
| lasthbA1CInPast365Days | 19815.0 | 7.522502 | 1.201356 | 3.60 | 6.900 | 7.437 | 7.900 | 12.70 |
| meandiastolicinPast365Days | 19815.0 | 73.814812 | 5.148202 | 43.62 | 70.866 | 73.991 | 76.775 | 104.21 |
| meansystolicinPast365Days | 19815.0 | 131.816459 | 6.264638 | 90.50 | 130.099 | 131.921 | 133.683 | 179.70 |
| BMI | 19815.0 | 33.484880 | 6.827066 | 13.43 | 29.139 | 33.600 | 37.041 | 85.68 |
| meantriglyceroidsInPast365Days | 19815.0 | 164.367480 | 56.735835 | 27.00 | 134.000 | 164.470 | 180.836 | 460.50 |
| meanldlcholesterolInPast365Days | 19815.0 | 89.276920 | 23.364085 | 10.00 | 81.915 | 89.011 | 98.902 | 201.00 |
| meanhdlcholesterolInPast365Days | 19815.0 | 43.888438 | 8.990077 | 14.00 | 39.000 | 43.333 | 47.634 | 98.00 |
| has_INJ004_Past12Months | 19815.0 | 0.004592 | 0.067614 | 0.00 | 0.000 | 0.000 | 0.000 | 1.00 |
| has_NEO043_Past12Months | 19815.0 | 0.004693 | 0.068349 | 0.00 | 0.000 | 0.000 | 0.000 | 1.00 |
| has_NEO012_Past12Months | 19815.0 | 0.004845 | 0.069438 | 0.00 | 0.000 | 0.000 | 0.000 | 1.00 |
| has_NEO030_Past12Months | 19815.0 | 0.020338 | 0.141158 | 0.00 | 0.000 | 0.000 | 0.000 | 1.00 |
| has_CIR007_Past12Months | 19815.0 | 0.129851 | 0.336148 | 0.00 | 0.000 | 0.000 | 0.000 | 1.00 |
| has_MBD003_Past12Months | 19815.0 | 0.004895 | 0.069797 | 0.00 | 0.000 | 0.000 | 0.000 | 1.00 |
| has_CIR008_Past12Months | 19815.0 | 0.013576 | 0.115724 | 0.00 | 0.000 | 0.000 | 0.000 | 1.00 |
| has_FAC013 | 19815.0 | 0.005703 | 0.075303 | 0.00 | 0.000 | 0.000 | 0.000 | 1.00 |
| has_DIG006 | 19815.0 | 0.005753 | 0.075633 | 0.00 | 0.000 | 0.000 | 0.000 | 1.00 |
| has_NEO065 | 19815.0 | 0.004996 | 0.070509 | 0.00 | 0.000 | 0.000 | 0.000 | 1.00 |
| has_FAC004 | 19815.0 | 0.000101 | 0.010046 | 0.00 | 0.000 | 0.000 | 0.000 | 1.00 |
| has_GEN025 | 19815.0 | 0.040273 | 0.196603 | 0.00 | 0.000 | 0.000 | 0.000 | 1.00 |
| has_RSP017 | 19815.0 | 0.010194 | 0.100453 | 0.00 | 0.000 | 0.000 | 0.000 | 1.00 |
| has_NVS004 | 19815.0 | 0.005753 | 0.075633 | 0.00 | 0.000 | 0.000 | 0.000 | 1.00 |
| has_MUS037 | 19815.0 | 0.009387 | 0.096432 | 0.00 | 0.000 | 0.000 | 0.000 | 1.00 |
| has_CIR035 | 19815.0 | 0.011406 | 0.106188 | 0.00 | 0.000 | 0.000 | 0.000 | 1.00 |

Calibration coefficient $\beta$ (confidence interval): -0.957 (-1.076, -0.839)

Model metrics:

| Set | Concordance Index | Integrated Brier Score |
|---|---|---|
| Train | 0.53 | 0.114 |
| **Validation** | 0.54 | 0.115 |
| **Test (before calibration)** | 0.54 | 0.040 |
| **Test (after calibration)** | 0.54 | 0.029 |

Predicted vs Actual survival curves for test set, before and after calibration:

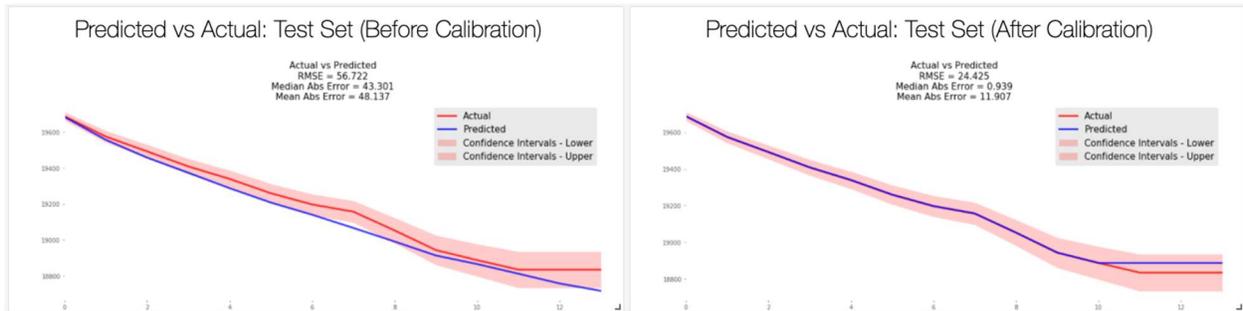

## Diabetic Retinopathy

Number of instances and proportion of class labels in training, validation and test sets:

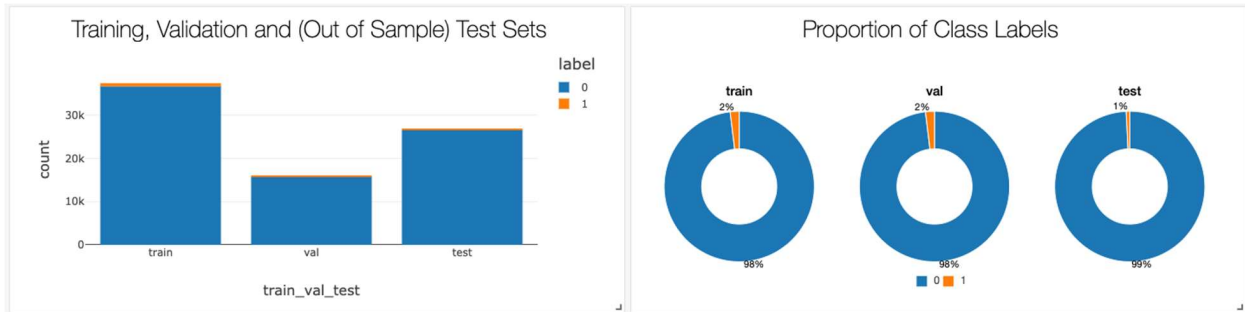

Distribution of features in training set:

|  | count | mean | std | min | 25% | 50% | 75% | max |
|---|---|---|---|---|---|---|---|---|
| age | 37393.0 | 64.558901 | 12.941430 | 18.00 | 56.000 | 66.000 | 73.000 | 102.00 |
| Male_sex | 37393.0 | 0.447945 | 0.497290 | 0.00 | 0.000 | 0.000 | 1.000 | 1.00 |
| Hispanic_or_Latino_ethnicity | 37393.0 | 0.066563 | 0.249267 | 0.00 | 0.000 | 0.000 | 0.000 | 1.00 |
| lasthbA1CInPast365Days | 37393.0 | 7.552167 | 1.220510 | 4.00 | 6.800 | 7.468 | 8.000 | 12.70 |
| meandiastolicinPast365Days | 37393.0 | 73.574956 | 8.815770 | 44.21 | 68.490 | 73.621 | 79.330 | 107.56 |
| meansystolicinPast365Days | 37393.0 | 132.126932 | 13.325088 | 81.00 | 124.080 | 131.860 | 138.940 | 180.00 |
| BMI | 37393.0 | 33.616739 | 6.690984 | 10.37 | 29.530 | 33.630 | 37.000 | 77.74 |
| meantriglyceroidsInPast365Days | 37393.0 | 169.113035 | 57.513464 | 19.00 | 140.000 | 167.109 | 182.466 | 456.50 |
| meanldlcholesterolInPast365Days | 37393.0 | 88.090462 | 23.545214 | 8.00 | 80.500 | 88.935 | 98.338 | 201.00 |
| meanhdlcholesterolInPast365Days | 37393.0 | 42.549205 | 9.268495 | 5.00 | 37.199 | 41.093 | 47.190 | 97.00 |
| has_GEN020_Past12Months | 37393.0 | 0.003316 | 0.057491 | 0.00 | 0.000 | 0.000 | 0.000 | 1.00 |
| has_CIR009_Past12Months | 37393.0 | 0.017249 | 0.130200 | 0.00 | 0.000 | 0.000 | 0.000 | 1.00 |
| has_BLD008_Past12Months | 37393.0 | 0.023347 | 0.151004 | 0.00 | 0.000 | 0.000 | 0.000 | 1.00 |
| has_CIR007_Past12Months | 37393.0 | 0.223518 | 0.416608 | 0.00 | 0.000 | 0.000 | 0.000 | 1.00 |
| has_NEO008_Past12Months | 37393.0 | 0.001524 | 0.039014 | 0.00 | 0.000 | 0.000 | 0.000 | 1.00 |
| has_CIR008_Past12Months | 37393.0 | 0.009253 | 0.095748 | 0.00 | 0.000 | 0.000 | 0.000 | 1.00 |
| has_MBD017 | 37393.0 | 0.022411 | 0.148017 | 0.00 | 0.000 | 0.000 | 0.000 | 1.00 |
| has_INJ026 | 37393.0 | 0.003985 | 0.062999 | 0.00 | 0.000 | 0.000 | 0.000 | 1.00 |
| has_GEN017 | 37393.0 | 0.036986 | 0.188729 | 0.00 | 0.000 | 0.000 | 0.000 | 1.00 |
| has_MAL004 | 37393.0 | 0.001230 | 0.035053 | 0.00 | 0.000 | 0.000 | 0.000 | 1.00 |
| has_NEO022 | 37393.0 | 0.042414 | 0.201535 | 0.00 | 0.000 | 0.000 | 0.000 | 1.00 |
| has_INJ069 | 37393.0 | 0.000963 | 0.031014 | 0.00 | 0.000 | 0.000 | 0.000 | 1.00 |
| has_RSP011 | 37393.0 | 0.067339 | 0.250611 | 0.00 | 0.000 | 0.000 | 0.000 | 1.00 |
| has_INJ023 | 37393.0 | 0.001899 | 0.043534 | 0.00 | 0.000 | 0.000 | 0.000 | 1.00 |
| has_MUS016 | 37393.0 | 0.003049 | 0.055132 | 0.00 | 0.000 | 0.000 | 0.000 | 1.00 |
| has_NVS013 | 37393.0 | 0.020137 | 0.140472 | 0.00 | 0.000 | 0.000 | 0.000 | 1.00 |

Distribution of features in out-of-sample test set:

| | count | mean | std | min | 25% | 50% | 75% | max |
|---|---|---|---|---|---|---|---|---|
| age | 26753.0 | 64.982918 | 13.204143 | 18.000 | 57.000 | 66.000 | 74.000 | 100.00 |
| Male_sex | 26753.0 | 0.442268 | 0.496665 | 0.000 | 0.000 | 0.000 | 1.000 | 1.00 |
| Hispanic_or_Latino_ethnicity | 26753.0 | 0.071319 | 0.257362 | 0.000 | 0.000 | 0.000 | 0.000 | 1.00 |
| lasthbA1CInPast365Days | 26753.0 | 7.555745 | 1.232550 | 3.500 | 6.800 | 7.463 | 8.000 | 12.70 |
| meandiastolicinPast365Days | 26753.0 | 73.432272 | 5.394946 | 43.620 | 70.624 | 73.621 | 76.480 | 107.28 |
| meansystolicinPast365Days | 26753.0 | 132.019994 | 6.519579 | 90.000 | 130.446 | 131.860 | 133.530 | 180.00 |
| BMI | 26753.0 | 33.756473 | 6.892561 | 13.430 | 29.485 | 33.600 | 37.348 | 85.68 |
| meantriglyceroidsInPast365Days | 26753.0 | 167.661845 | 58.104449 | 27.000 | 136.888 | 166.795 | 182.466 | 460.50 |
| meanldlcholesterolInPast365Days | 26753.0 | 89.047429 | 23.607218 | 10.000 | 80.000 | 88.935 | 98.338 | 201.00 |
| meanhdlcholesterolInPast365Days | 26753.0 | 43.403549 | 9.055841 | 10.333 | 39.000 | 43.000 | 47.190 | 98.00 |
| has_GEN020_Past12Months | 26753.0 | 0.005719 | 0.075409 | 0.000 | 0.000 | 0.000 | 0.000 | 1.00 |
| has_CIR009_Past12Months | 26753.0 | 0.019362 | 0.137797 | 0.000 | 0.000 | 0.000 | 0.000 | 1.00 |
| has_BLD008_Past12Months | 26753.0 | 0.008784 | 0.093313 | 0.000 | 0.000 | 0.000 | 0.000 | 1.00 |
| has_CIR007_Past12Months | 26753.0 | 0.125481 | 0.331270 | 0.000 | 0.000 | 0.000 | 0.000 | 1.00 |
| has_NEO008_Past12Months | 26753.0 | 0.002691 | 0.051809 | 0.000 | 0.000 | 0.000 | 0.000 | 1.00 |
| has_CIR008_Past12Months | 26753.0 | 0.017120 | 0.129719 | 0.000 | 0.000 | 0.000 | 0.000 | 1.00 |
| has_MBD017 | 26753.0 | 0.025418 | 0.157393 | 0.000 | 0.000 | 0.000 | 0.000 | 1.00 |
| has_INJ026 | 26753.0 | 0.002355 | 0.048471 | 0.000 | 0.000 | 0.000 | 0.000 | 1.00 |
| has_GEN017 | 26753.0 | 0.032707 | 0.177871 | 0.000 | 0.000 | 0.000 | 0.000 | 1.00 |
| has_MAL004 | 26753.0 | 0.002467 | 0.049609 | 0.000 | 0.000 | 0.000 | 0.000 | 1.00 |
| has_NEO022 | 26753.0 | 0.030688 | 0.172475 | 0.000 | 0.000 | 0.000 | 0.000 | 1.00 |
| has_INJ069 | 26753.0 | 0.001570 | 0.039592 | 0.000 | 0.000 | 0.000 | 0.000 | 1.00 |
| has_RSP011 | 26753.0 | 0.057414 | 0.232637 | 0.000 | 0.000 | 0.000 | 0.000 | 1.00 |
| has_INJ023 | 26753.0 | 0.003102 | 0.055614 | 0.000 | 0.000 | 0.000 | 0.000 | 1.00 |
| has_MUS016 | 26753.0 | 0.001383 | 0.037164 | 0.000 | 0.000 | 0.000 | 0.000 | 1.00 |
| has_NVS013 | 26753.0 | 0.016895 | 0.128882 | 0.000 | 0.000 | 0.000 | 0.000 | 1.00 |

Calibration coefficient $\beta$ (confidence interval): 0.015 (0.015, 0.015)

Model metrics:

| Set | Concordance Index | Integrated Brier Score |
|---|---|---|
| Train | 0.51 | 0.0209 |
| Validation | 0.53 | 0.0209 |
| Test (before calibration) | 0.67 | 0.0099 |
| Test (after calibration) | 0.67 | 0.0075 |

Predicted vs Actual survival curves for test set, before and after calibration:

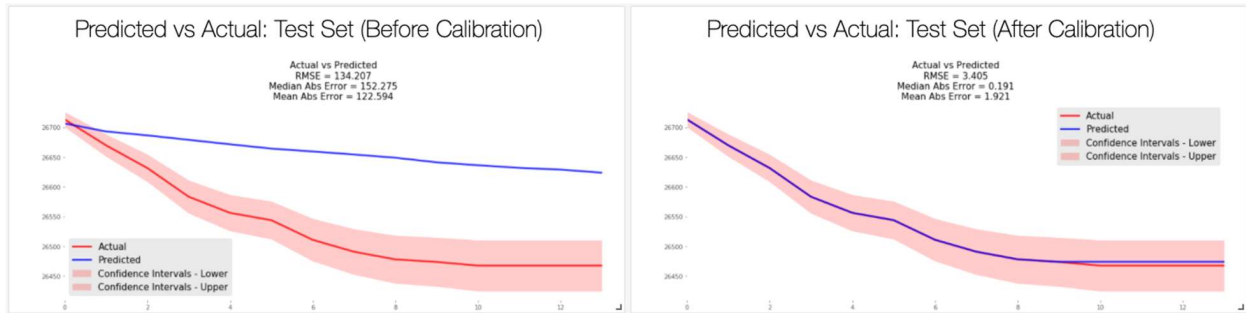

# Appendix E: Model Signatures Excluding Extremely Unbalanced Binary Features

The model signatures generated via feature selection in this evaluation included several features that were surprising or seemed out of place. It was not clear, from a clinical perspective, why these features would be found to be important for predicting the various conditions and complications.

It turned out that most of these features were binary features with extremely low proportions, < 1%, of positives. This could indicate that the models were learning from very strong associations between these features and the target variables, even though they were present only in a very small number of training instances.

A number of issues arise:

a. It may be difficult to explain to clinicians or customers why certain features are important for predicting diabetes and its complications.
b. The usefulness of these features is limited to a very small proportion (< 1%) of the proportion.
c. These feature associations may be a result of spurious correlations present in the training data set, and not universally applicable to other populations; i.e. the models may have been overfit to one population and if so, performance will decrease when they are deployed elsewhere.

In this investigation, we re-generated the model signatures for each task excluding extremely unbalanced binary features. The following procedure was used:

1. Input: The same set of encounters, features and cohort criteria as in the current iteration (Iteration 3) reported in the main body of this document.
2. Detect and remove all binary features where the instances comprise < 2% (extremely low) or > 98% (extremely high) positive values.
3. Re-run recursive feature elimination using the same procedure used in this iteration.

In the tables below, features in grey were selected in the original model signatures but not in the new ones. Conversely, features in green were selected in the new model signatures but were not in the original ones.

Observations:

1. Depending on task, between 67% to 71% of features were removed from the candidate list as they were found to be extremely imbalanced with respect to the selected cohorts.
2. Apart from the mandatory features, the other selected features were quite different for most tasks. Whether these new sets of features make sense clinically is to be validated.

3. Selected features remain largely based on diagnoses (as represented by CSSR categories), except for nephropathy, where albumin and potassium features were added to the model signature.
4. There was no detrimental effect on cross-validation scores during feature selection (which are indicative of final model performance); in fact, the cross-validation scores improved for nephropathy, neuropathy and retinopathy.

Pre-DM to DM

|  | With Full List of Candidate Features | Excluding 828 / 1,174 Extremely Imbalanced Binary Features |
|---|---|---|
| **Model Signature** | 1. has_BLD005_Past12Months (Sickle cell trait/anemia)<br>2. has_PRG029 (Uncomplicated pregnancy, delivery or puerperium)<br>3. has_NEO016_Past12Months (Gastrointestinal cancers - anus)<br>4. has_INJ057 (Effect of foreign body entering opening, subsequent encounter)<br>5. has_DIG018 (Hepatic failure)<br>6. maxhbA1CInPast365Days<br>7. has_RSP015 (Mediastinal disorders)<br>8. has_NEO017 (Gastrointestinal cancers - liver)<br>9. has_SKN003 (Pressure ulcer of skin)<br>10. has_FAC006 (Encounter for antineoplastic therapies)<br>11. Male_sex<br>12. lasthbA1CInPast365Days<br>13. has_CIR008_Past12Months (Hypertension with complications and secondary hypertension)<br>14. Hispanic_or_Latino_ethnicity<br>15. has_CIR007_Past12Months (Essential hypertension)<br>16. meanhdlcholesterolInPast365Days<br>17. meansystolicinPast365Days<br>18. BMI<br>19. meandiastolicinPast365Days<br>20. age<br>21. meanldlcholesterolInPast365Days<br>22. meantriglyceroidsInPast365Days | 1. has_DIG018 (Hepatic failure)<br>2. maxhbA1CInPast365Days<br>3. has_SKN003 (Pressure ulcer of skin)<br>4. has_FAC006 (Encounter for antineoplastic therapies)<br>5. has_SYM002 (Fever)<br>6. has_FAC023 (Organ transplant status)<br>7. Male_sex<br>8. lasthbA1CInPast365Days<br>9. has_CIR007_Past12Months (Essential hypertension)<br>10. meanhdlcholesterolInPast365Days<br>11. meansystolicinPast365Days<br>12. BMI<br>13. meandiastolicinPast365Days<br>14. age<br>15. meanldlcholesterolInPast365Days<br>16. meantriglyceroidsInPast365Days<br>17. Hispanic_or_Latino_ethnicity |
| **Cross-Validation Plot** | 22 features selected, Score = 0.77 | 17 features selected, Score = 0.77 |

## DM to Uncontrolled DM

| | **With Full List of Candidate Features** | **Excluding 794 / 1,182 Extremely Imbalanced Binary Features** |
|---|---|---|
| **Model Signature** | 1. has_CIR038 (Postprocedural or postoperative circulatory system complication)<br>2. has_END013 (Pituitary disorders)<br>3. has_MAL004_Past12Months (Nervous system congenital anomalies)<br>4. has_EXT030 (External cause codes: sequela)<br>5. has_NEO061_Past12Months (Leukemia - chronic lymphocytic leukemia (CLL))<br>6. has_MUS029 (Disorders of jaw)<br>7. has_NEO029 (Breast cancer - ductal carcinoma in situ (DCIS))<br>8. lasthbA1CInPast365Days<br>9. Hispanic_or_Latino_ethnicity<br>10. has_CIR008_Past12Months (Hypertension with complications and secondary hypertension)<br>11. has_CIR007_Past12Months (Essential hypertension)<br>12. BMI<br>13. Male_sex<br>14. age<br>15. meansystolicinPast365Days<br>16. meanhdlcholesterolInPast365Days<br>17. meandiastolicinPast365Days<br>18. meanldlcholesterolInPast365Days<br>19. meantriglyceroidsInPast365Days | 1. lasthbA1CInPast365Days<br>2. has_NEO070 (Secondary malignancies)<br>3. has_NEO070_Past12Months (Secondary malignancies)<br>4. has_DIG007 (Gastritis and duodenitis)<br>5. has_FAC022 (Acquired absence of limb or organ)<br>6. has_CIR024 (Other and ill-defined cerebrovascular disease)<br>7. has_SYM001 (Syncope)<br>8. has_CIR036 (Postthrombotic syndrome and venous insufficiency/hypertension)<br>9. has_SYM015 (General sensation/perception signs and symptoms)<br>10. Hispanic_or_Latino_ethnicity<br>11. has_CIR007_Past12Months (Essential hypertension)<br>12. BMI<br>13. age<br>14. meansystolicinPast365Days<br>15. Male_sex<br>16. meanhdlcholesterolInPast365Days<br>17. meandiastolicinPast365Days<br>18. meanldlcholesterolInPast365Days<br>19. meantriglyceroidsInPast365Days |
| **Cross-Validation Plot** | 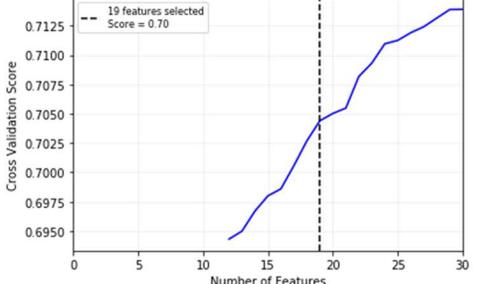 | 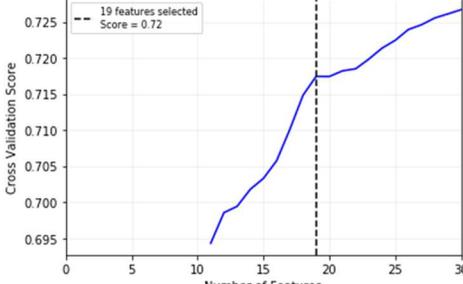 |

Pre-DM to Uncontrolled DM

|  | **With Full List of Candidate Features** | **Excluding 830 / 1,182 Extremely Imbalanced Binary Features** |
|---|---|---|
| **Model Signature** | 1. has_BLD005_Past12Months (Sickle cell trait/anemia)<br>2. has_INJ035 (Complication of internal orthopedic device or implant, initial encounter)<br>3. has_INJ014_Past12Months (Amputation of a limb, initial encounter)<br>4. has_INJ049 (Open wounds to limbs, subsequent encounter)<br>5. has_NEO022_Past12Months (Respiratory cancers)<br>6. has_INJ054 (Superficial injury; contusion, subsequent encounter)<br>7. has_NEO044 (Urinary system cancers - ureter and renal pelvis)<br>8. lasthbA1CInPast365Days<br>9. has_SKN005_Past12Months (Contact dermatitis)<br>10. has_NEO029_Past12Months (Breast cancer - ductal carcinoma in situ (DCIS))<br>11. has_BLD005 (Sickle cell trait/anemia)<br>12. has_NEO012 (Gastrointestinal cancers - esophagus)<br>13. has_INJ007 (Dislocations, initial encounter)<br>14. has_GEN018 (Inflammatory diseases of female pelvic organs)<br>15. Male_sex<br>16. has_CIR008_Past12Months (Hypertension with complications and secondary hypertension)<br>17. Hispanic_or_Latino_ethnicity<br>18. age<br>19. BMI<br>20. meanhdlcholesterolInPast365Days<br>21. has_CIR007_Past12Months (Essential hypertension)<br>22. meansystolicinPast365Days<br>23. meantriglyceroidsInPast365Days<br>24. meandiastolicinPast365Days<br>25. meanldlcholesterolInPast365Days | 1. maxhbA1CInPast365Days<br>2. has_SYM002 (Fever)<br>3. has_NEO022_Past12Months (Respiratory cancers)<br>4. lasthbA1CInPast365Days<br>5. Male_sex<br>6. Hispanic_or_Latino_ethnicity<br>7. age<br>8. meansystolicinPast365Days<br>9. BMI<br>10. meanhdlcholesterolInPast365Days<br>11. meandiastolicinPast365Days<br>12. meantriglyceroidsInPast365Days<br>13. has_CIR007_Past12Months (Essential hypertension)<br>14. meanldlcholesterolInPast365Days |
| **Cross-Validation Plot** | 25 features selected, Score = 0.81 | 14 features selected, Score = 0.80 |

## DM to Diabetic Nephropathy

| | **With Full List of Candidate Features** | **Excluding 822 / 1,182 Extremely Imbalanced Binary Features** |
|---|---|---|
| **Model Signature** | 1. has_NEO008 (Head and neck cancers - laryngeal)<br>2. has_EXT017 (External cause codes: suffocation/inhalation; initial encounter)<br>3. has_NEO024_Past12Months (Sarcoma)<br>4. has_EYE012 (Other specified eye disorders)<br>5. has_CIR034 (Chronic phlebitis; thrombophlebitis and thromboembolism)<br>6. has_NVS005 (Multiple sclerosis)<br>7. has_INJ026 (Other specified injury)<br>8. has_CIR008 (Hypertension with complications and secondary hypertension)<br>9. has_RSP010_Past12Months (Aspiration pneumonitis)<br>10. has_GEN010 (Proteinuria)<br>11. has_END011 (Fluid and electrolyte disorders)<br>12. has_NVS018 (Myopathies)<br>13. has_NEO050 (Endocrine system cancers - thyroid)<br>14. has_FAC023 (Organ transplant status)<br>15. Hispanic_or_Latino_ethnicity<br>16. Male_sex<br>17. has_CIR007_Past12Months (Essential hypertension)<br>18. lasthbA1CInPast365Days<br>19. age<br>20. BMI<br>21. meandiastolicinPast365Days<br>22. meansystolicinPast365Days<br>23. meanhdlcholesterolInPast365Days<br>24. meanldlcholesterolInPast365Days<br>25. meantriglyceroidsInPast365Days<br>26. has_CIR008_Past12Months (Hypertension with complications and secondary hypertension) | 1. has_CIR019 (Heart failure)<br>2. has_CIR008 (Hypertension with complications and secondary hypertension)<br>3. has_GEN017 (Nonmalignant breast conditions)<br>4. has_INF003 (Bacterial infections)<br>5. has_INF003_Past12Months (Bacterial infections)<br>6. has_FAC023 (Organ transplant status)<br>7. has_CIR009 (Acute myocardial infarction)<br>8. has_NEO030_Past12Months (Breast cancer - all other types)<br>9. stdalbuminInPast365Days<br>10. minalbuminInPast365Days<br>11. minpotassiumInPast365Days<br>12. has_INJ005 (Fracture of the lower limb (except hip), initial encounter)<br>13. has_FAC010 (Other aftercare encounter)<br>14. has_CIR029 (Aortic; peripheral; and visceral artery aneurysms)<br>15. has_MUS033 (Gout)<br>16. has_INF008_Past12Months (Viral infection)<br>17. has_INF007 (Hepatitis)<br>18. Hispanic_or_Latino_ethnicity<br>19. has_CIR007_Past12Months (Essential hypertension)<br>20. lasthbA1CInPast365Days<br>21. meandiastolicinPast365Days<br>22. Male_sex<br>23. meansystolicinPast365Days<br>24. age<br>25. BMI<br>26. meanldlcholesterolInPast365Days<br>27. meantriglyceroidsInPast365Days<br>28. meanhdlcholesterolInPast365Days |
| **Cross-Validation Plot** | 26 features selected, Score = 0.69 | 28 features selected, Score = 0.75 |

## DM to Diabetic Neuropathy

| | **With Full List of Candidate Features** | **Excluding 822 / 1,184 Extremely Imbalanced Binary Features** |
|---|---|---|
| **Model Signature** | 1. has_DIG006 (Gastrointestinal and biliary perforation)<br>2. has_FAC013 (Contraceptive and procreative management)<br>3. has_CIR035 (Varicose veins of lower extremity)<br>4. has_FAC004 (Encounter for prophylactic or other procedures)<br>5. has_NEO030_Past12Months (Breast cancer - all other types)<br>6. has_NEO012_Past12Months (Gastrointestinal cancers - esophagus)<br>7. has_MUS037 (Postprocedural or postoperative musculoskeletal system complication)<br>8. has_NEO043_Past12Months (Urinary system cancers - bladder)<br>9. has_NEO065 (Multiple myeloma)<br>10. has_INJ004_Past12Months (Fracture of the upper limb, initial encounter)<br>11. has_MBD003_Past12Months (Bipolar and related disorders)<br>12. has_RSP017 (Postprocedural or postoperative respiratory system complication)<br>13. has_NVS004 (Parkinson`s disease)<br>14. has_GEN025 (Other specified female genital disorders)<br>15. has_CIR008_Past12Months (Hypertension with complications and secondary hypertension)<br>16. has_CIR007_Past12Months (Essential hypertension)<br>17. Hispanic_or_Latino_ethnicity<br>18. Male_sex<br>19. lasthbA1CInPast365Days<br>20. meandiastolicinPast365Days<br>21. meansystolicinPast365Days<br>22. BMI<br>23. meanhdlcholesterolInPast365Days<br>24. meanldlcholesterolInPast365Days<br>25. age<br>26. meantriglyceroidsInPast365Days | 1. has_NEO030_Past12Months (Breast cancer - all other types)<br>2. has_SKN003 (Pressure ulcer of skin)<br>3. has_END003 (Diabetes mellitus with complication)<br>4. has_INJ012 (Open wounds to limbs, initial encounter)<br>5. has_GEN006_Past12Months (Other specified and unspecified diseases of kidney and ureters)<br>6. has_FAC022 (Acquired absence of limb or organ)<br>7. has_GEN025 (Other specified female genital disorders)<br>8. has_DIG008 (Other specified and unspecified disorders of stomach and duodenum)<br>9. has_INF002 (Septicemia)<br>10. has_CIR019 (Heart failure)<br>11. has_RSP012_Past12Months (Respiratory failure; insufficiency; arrest)<br>12. has_SYM004 (Nausea and vomiting)<br>13. lasthbA1CInPast365Days<br>14. Hispanic_or_Latino_ethnicity<br>15. has_CIR007_Past12Months (Essential hypertension)<br>16. meansystolicinPast365Days<br>17. meandiastolicinPast365Days<br>18. BMI<br>19. meanldlcholesterolInPast365Days<br>20. meanhdlcholesterolInPast365Days<br>21. meantriglyceroidsInPast365Days<br>22. age<br>23. Male_sex |
| **Cross-Validation Plot** | 26 features selected, Score = 0.58 | 23 features selected, Score = 0.74 |

## DM to Diabetic Retinopathy

|  | **With Full List of Candidate Features** | **Excluding 801 / 1,186 Extremely Imbalanced Binary Features** |
|---|---|---|
| **Model Signature** | 1. has_BLD008_Past12Months (Immunity disorders)<br>2. has_NEO022 (Respiratory cancers)<br>3. has_CIR009_Past12Months (Acute myocardial infarction)<br>4. has_INJ069 (Complication of cardiovascular device, implant or graft, subsequent encounter)<br>5. has_GEN020_Past12Months (Prolapse of female genital organs)<br>6. has_MUS016 (Stress fracture, initial encounter)<br>7. has_INJ023 (Toxic effects, initial encounter)<br>8. has_MAL004 (Nervous system congenital anomalies)<br>9. has_NVS013 (Coma; stupor; and brain damage)<br>10. has_GEN017 (Nonmalignant breast conditions)<br>11. has_INJ026 (Other specified injury)<br>12. has_NEO008_Past12Months (Head and neck cancers - laryngeal)<br>13. has_MBD017 (Alcohol-related disorders)<br>14. has_RSP011 (Pleurisy, pleural effusion and pulmonary collapse)<br>15. Hispanic_or_Latino_ethnicity<br>16. has_CIR008_Past12Months (Hypertension with complications and secondary hypertension)<br>17. lasthbA1CInPast365Days<br>18. Male_sex<br>19. has_CIR007_Past12Months (Essential hypertension)<br>20. meanhdlcholesterolInPast365Days<br>21. meandiastolicinPast365Days<br>22. age<br>23. meansystolicinPast365Days<br>24. BMI<br>25. meanldlcholesterolInPast365Days<br>26. meantriglyceroidsInPast365Days | 1. has_NEO022 (Respiratory cancers)<br>2. has_CIR001 (Chronic rheumatic heart disease)<br>3. has_GEN017 (Nonmalignant breast conditions)<br>4. has_MBD017 (Alcohol-related disorders)<br>5. has_RSP011 (Pleurisy, pleural effusion and pulmonary collapse)<br>6. has_END003 (Diabetes mellitus with complication)<br>7. has_CIR013 (Acute pulmonary embolism)<br>8. has_FAC010 (Other aftercare encounter)<br>9. has_END006 (Diabetes mellitus, due to underlying condition, drug or chemical induced, or other specified type)<br>10. has_INF008 (Viral infection)<br>11. has_INF008_Past12Months (Viral infection)<br>12. has_BLD008 (Immunity disorders)<br>13. has_SYM003 (Shock)<br>14. has_GEN010 (Proteinuria)<br>15. Hispanic_or_Latino_ethnicity<br>16. lasthbA1CInPast365Days<br>17. has_CIR007_Past12Months (Essential hypertension)<br>18. meandiastolicinPast365Days<br>19. age<br>20. meansystolicinPast365Days<br>21. meanhdlcholesterolInPast365Days<br>22. meanldlcholesterolInPast365Days<br>23. BMI<br>24. Male_sex<br>25. meantriglyceroidsInPast365Days |
| **Cross-Validation Plot** | 26 features selected, Score = 0.72 | 25 features selected, Score = 0.81 |

# Appendix F: Features for Model Training

The following table lists the relevant features for model training that are available from existing Patient Flow features.

- Many of the features also have "Unknown" or "Other" as possible values, but these are omitted from this table for brevity, as they would be treated as missing values.
- Calendar features were omitted (e.g., admission hour, day and month) as these are less relevant for predicting diabetes and its complications.

| Name | Category | Type | Description | Number of Features | Modifiable | Remarks |
|---|---|---|---|---|---|---|
| **Age** | Demographic | Integer | In years, at time of encounter | 1 | No | |
| **Ethnicity** | Demographic | Categorical | • Hispanic or Latino<br>• Not Hispanic or Latino | 1 | No | How does this overlap with Race? |
| **Race** | Demographic | Categorical | • American Indian or Alaska Native<br>• Asian<br>• Black or African American<br>• Native Hawaiian or Other Pacific Islander<br>• White or Caucasian | 1 | No | |
| **Sex** | Demographic | Categorical | • Female<br>• Male | 1 | No | |
| **Chronic Conditions** | Diagnosis | Boolean | Presence of chronic condition(s) in the last 365 days | 40 | Depends | |
| **Diagnosis Group** | Diagnosis | Boolean | Presence of diagnosis group(s) for the encounter | 358 | Depends | |
| **Albumin** | Lab | Continuous | 365-day aggregates:<br>• Min<br>• Mean<br>• Median<br>• Max<br>• Last | 5 | | |
| **Alkaline Phosphatase** | Lab | Continuous | 365-day aggregates:<br>• Min<br>• Mean<br>• Median<br>• Max<br>• Last | 5 | | |
| **Bilirubin** | Lab | Continuous | 365-day aggregates:<br>• Min<br>• Mean<br>• Median<br>• Max<br>• Last | 5 | | |
| **Bun** | Lab | Continuous | 365-day aggregates:<br>• Min<br>• Mean<br>• Median<br>• Max<br>• Last | 5 | | |

| Name | Category | Type | Description | Number of Features | Modifiable | Remarks |
|---|---|---|---|---|---|---|
| **HBA1C** | Lab | Continuous | 365-day aggregates:<br>• Min<br>• Mean<br>• Median<br>• Max<br>• Last | 5 | | |
| **Neutrophils** | Lab | Continuous | 365-day aggregates:<br>• Min<br>• Mean<br>• Median<br>• Max<br>• Last | 5 | | |
| **Potassium** | Lab | Continuous | 365-day aggregates:<br>• Min<br>• Mean<br>• Median<br>• Max<br>• Last | 5 | | |
| **Serum Bicarbonate** | Lab | Continuous | 365-day aggregates:<br>• Min<br>• Mean<br>• Median<br>• Max<br>• Last | 5 | | |
| **Sodium** | Lab | Continuous | 365-day aggregates:<br>• Min<br>• Mean<br>• Median<br>• Max<br>• Last | 5 | | |
| **WBC** | Lab | Continuous | 365-day aggregates:<br>• Min<br>• Mean<br>• Median<br>• Max<br>• Last | 5 | | |
| **Smoking** | Lifestyle | Categorical | • Current Every Day Smoker<br>• Current Some Day Smoker<br>• Former Smoker<br>• Heavy Tobacco Smoker<br>• Light Tobacco Smoker<br>• Never Assessed<br>• Never Smoker<br>• Passive Smoke Exposure - Never Smoker<br>• Smoker, Current Status Unknown | 1 | Yes | May not be useful, as definitions of each category are not clearly specified, and may not be the same for all providers |

| Name | Category | Type | Description | Number of Features | Modifiable | Remarks |
|---|---|---|---|---|---|---|
| **Specialty Received** | Specialty | Boolean | 0, 1 or more of 121 specialties for the encounter | 121 | | Is this consistently defined across providers? |
| **Specialty Requested** | Specialty | Boolean | 0, 1 or more of 121 specialties for the encounter | 121 | | Is this consistently defined across providers? |
| **Admission Type** | Utilization | Integer | • Emergency<br>• Inpatient<br>• Outpatient<br>• Not a Hospital Encounter | 1 | | May need to convert into an aggregate, e.g., number of each type in last 90, 180, 365 days |
| **Days Since Last ED Visit** | Utilization | Integer | In days | 1 | | |
| **Days Since Last Inpatient Visit** | Utilization | Integer | In days | 1 | | |
| **Discharge Disposition** | Utilization | Categorical | For the encounter<br>• home<br>• pac<br>• home-health<br>• other<br>• None | 1 | No | May need to convert into an aggregate, e.g., number of each type in last 90, 180, 365 days |
| **Is Admission** | Utilization | Boolean | For the encounter<br>• 0: outpatient<br>• 1: inpatient | 1 | No | Use Previous Inpatient / Outpatient Encounters Count instead |
| **Length of Stay** | Utilization | Integer | In days, for the encounter | 1 | No | |
| **Previous ED Visits Count** | Utilization | Integer | For days before current encounter:<br>• 90 days<br>• 180 days<br>• 365 days | 3 | | |
| **Previous Inpatient Encounters Count** | Utilization | Integer | For days before current encounter:<br>• 90 days<br>• 180 days<br>• 365 days | 3 | | Add another feature for outpatient encounters |
| **Diastolic** | Vitals | Continuous | 365-day aggregates:<br>• Min<br>• Mean<br>• Median<br>• Max<br>• SD<br>• Last | 6 | Yes | |

| Name | Category | Type | Description | Number of Features | Modifiable | Remarks |
|---|---|---|---|---|---|---|
| **Systolic** | Vitals | Continuous | 365-day aggregates:<br>• Min<br>• Mean<br>• Median<br>• Max<br>• SD<br>• Last | 6 | Yes | |
| **Glasgow Coma Scale Score** | Vitals | Integer | During encounter | 1 | | |
| **O2 Device Usage** | Vitals | Integer | • aerosolmask<br>• artificialnose<br>• bagvalvemask<br>• bipap<br>• blowby<br>• bubblecpap<br>• cpap<br>• endotrachealtube<br>• ett<br>• facetent<br>• heliox80/20<br>• highflownasalcannula<br>• hood<br>• lma<br>• misttent<br>• nasalcannula<br>• neonasalcannula<br>• neopuff<br>• noneroomair<br>• nonrebreather<br>• nonrebreathermask<br>• othercomment<br>• oxymask<br>• oxymizer<br>• pednasalcannula<br>• ramcannula<br>• simplemask<br>• tpiece<br>• trachcollar<br>• tracheostomy<br>• trachmask<br>• ventilator<br>• venturimask | 33 | | |
| **O2 Flow Flowsheet Value** | Vitals | Integer | Max during encounter | 1 | | |

# ED Utilization / Visit Non-Compliance / Medication Non-Adherence

| Name | Category | Type | Description | Number of Features | Modifiable | Mandatory |
|------|----------|------|-------------|--------------------|------------|-----------|
| **Age** | Demographic | Integer | In years, at time of appointment | 1 | No | Yes |
| **Age Above 65** | Demographic | Boolean | Whether the patient's age is above 65 | 1 | No | |
| **Ethnicity** | Demographic | Categorical | • Hispanic or Latino<br>• Not Hispanic or Latino | 1 | No | Yes |
| **Sex** | Demographic | Categorical | • Female<br>• Male | 1 | No | Yes |
| **Marital Status** | Demographic | Categorical | • Married<br>• Single<br>• Divorced<br>• Widowed | | | |
| **Insurance Status** | Demographic | Boolean | Whether the patient is insured | 1 | | |
| **Income Level** | Demographic | Categorical | | | | |
| **Educational Level** | Demographic | Categorical | | | | |
| **Employment Status** | Demographic | Categorical | | | | |
| **English Proficiency** | Demographic | Boolean | Whether the patient's primary language is English | 1 | | |
| **Distance to Clinic** | Demographic | Continuous | How far is the patient's home from the clinic | 1 | | |
| **Smoking Status** | Demographic | Categorical | • Never<br>• Former<br>• Current | | | |
| **Elixhauser Comorbidity Index** | Diagnosis | Integer | Comorbidity of the patient | 1 | | |
| **CCSR Category (12 mth)** | Diagnosis | Boolean | Presence of CCSR Category in past 12 months | 540 | Depends | |
| **CCSR Category** | Diagnosis | Boolean | Presence of CCSR Category in the patient's whole history | 540 | Depends | |
| **Albumin** | Lab | Continuous | 365-day aggregates:<br>• Min<br>• Mean<br>• Median<br>• Max<br>• Last | 5 | | |
| **Alkaline Phosphatase** | Lab | Continuous | 365-day aggregates:<br>• Min<br>• Mean<br>• Median<br>• Max<br>• Last | 5 | | |

| Name | Category | Type | Description | Number of Features | Modifiable | Mandatory |
|---|---|---|---|---|---|---|
| **Bilirubin** | Lab | Continuous | 365-day aggregates:<br>• Min<br>• Mean<br>• Median<br>• Max<br>• Last | 5 | | |
| **Bun** | Lab | Continuous | 365-day aggregates:<br>• Min<br>• Mean<br>• Median<br>• Max<br>• Last | 5 | | |
| **HBA1C** | Lab | Continuous | 365-day aggregates:<br>• Min<br>• Mean<br>• Median<br>• Max<br>• Last | 5 | | |
| **Neutrophils** | Lab | Continuous | 365-day aggregates:<br>• Min<br>• Mean<br>• Median<br>• Max<br>• Last | 5 | | |
| **Potassium** | Lab | Continuous | 365-day aggregates:<br>• Min<br>• Mean<br>• Median<br>• Max<br>• Last | 5 | | |
| **Serum Bicarbonate** | Lab | Continuous | 365-day aggregates:<br>• Min<br>• Mean<br>• Median<br>• Max<br>• Last | 5 | | |
| **Sodium** | Lab | Continuous | 365-day aggregates:<br>• Min<br>• Mean<br>• Median<br>• Max<br>• Last | 5 | | |
| **WBC** | Lab | Continuous | 365-day aggregates:<br>• Min<br>• Mean<br>• Median<br>• Max<br>• Last | 5 | | |
| **BMI** | Vitals | Continuous | BMI of the patient calculated using height and weight | 1 | | |

| Name | Category | Type | Description | Number of Features | Modifiable | Mandatory |
|---|---|---|---|---|---|---|
| AST | Lab | Continuous | 365-day aggregates:<br>• Min<br>• Mean<br>• Median<br>• Max<br>• SD<br>• Last | 6 | Yes | |
| ALT | Lab | Continuous | 365-day aggregates:<br>• Min<br>• Mean<br>• Median<br>• Max<br>• SD<br>• Last | 6 | Yes | |
| Diastolic | Vitals | Continuous | 365-day aggregates:<br>• Min<br>• Mean<br>• Median<br>• Max<br>• SD<br>• Last | 6 | Yes | |
| Systolic | Vitals | Continuous | 365-day aggregates:<br>• Min<br>• Mean<br>• Median<br>• Max<br>• SD<br>• Last | 6 | Yes | |
| LDL Cholesterol | Lab | Continuous | 365-day aggregates:<br>• Min<br>• Mean<br>• Median<br>• Max<br>• SD<br>• Last | 6 | Yes | |
| HDL Cholesterol | Lab | Continuous | 365-day aggregates:<br>• Min<br>• Mean<br>• Median<br>• Max<br>• SD<br>• Last | 6 | Yes | |
| Creatine | Lab | Continuous | 365-day aggregates:<br>• Min<br>• Mean<br>• Median<br>• Max<br>• SD<br>• Last | 6 | Yes | |

| Name | Category | Type | Description | Number of Features | Modifiable | Mandatory |
|---|---|---|---|---|---|---|
| **Triglyceroids** | Lab | Continuous | 365-day aggregates:<br>• Min<br>• Mean<br>• Median<br>• Max<br>• SD<br>• Last | 6 | Yes | |
| **Days Since Last ED Visit** | Utilization | Integer | In days | 1 | | |
| **Days Since Last Inpatient Visit** | Utilization | Integer | In days | 1 | | |
| **Days Since Last Outpatient Visit** | Utilization | Integer | In days | 1 | | |
| **Days Since Last PCP Visit** | Utilization | Integer | In days | 1 | | |
| **Previous ED Visits Count** | Utilization | Integer | For days before current appointment:<br>• 90 days<br>• 180 days<br>• 365 days | 3 | | |
| **Previous Inpatient Encounters Count** | Utilization | Integer | For days before current appointment:<br>• 90 days<br>• 180 days<br>• 365 days | 3 | | |
| **Previous Outpatient Encounters Count** | Utilization | Integer | For days before current appointment:<br>• 90 days<br>• 180 days<br>• 365 days | 3 | | |
| **Previous Outpatient Primary Care Encounters Count** | Utilization | Integer | For days before current appointment:<br>• 90 days<br>• 180 days<br>• 365 days | 3 | | |
| **Previous Outpatient Specialist Type Count** | Utilization | Integer | Number of distinct specialist type<br>For days before current appointment:<br>• 90 days<br>• 180 days<br>• 365 days | 3 | | |
| **Number of Previous No-shows** | Utilization | Integer | For days before current appointment:<br>• 90 days<br>• 180 days<br>• 365 days | 3 | | |

| Name | Category | Type | Description | Number of Features | Modifiable | Mandatory |
|---|---|---|---|---|---|---|
| **Number of Previous Late Appointments** | Utilization | Integer | For days before current appointment:<br>• 90 days<br>• 180 days<br>• 365 days | 3 | | |
| **Trend in Missed Appointments** | Utilization | Continuous | Increasing or decreasing trend with more weight given to the most recent appointments | 1 | | |
| **Past Appointment Lead Time** | Utilization | Continuous | 365-day aggregates:<br>• Min<br>• Mean<br>• Median<br>• Max<br>• SD<br>• Last | 6 | | |
| **Past Appointment Cancel Reason Count** | Utilization | Integer | e.g. Patient, Provider, Scheduled from Wait List, Error, Lack of Transportation, Weather | | | |
| **Past Kept Appointment of Day of Week** | Utilization | Continuous | Percentage of kept appointments on the same day of week as current appointment | 1 | | |
| **Provider Type** | Utilization | Categorical | • Physician<br>• Licensed practical nurse<br>• Technician<br>• Medical assistant<br>• Resident<br>• Psychologist<br>• Registered dietitian<br>• Social worker | | | |
| **Has History of Medication Side Effects** | Utilization | Boolean | Whether the patient has had any history of medication side effects | 1 | | |
| **Appointment Lead Time** | Current Appointment | Continuous | Duration between when appointment was made and actual appointment date | 1 | | |
| **Appointment Category** | Current Appointment | Categorical | • Follow Up<br>• Routine<br>• Consult<br>• Urgent<br>• Procedure<br>• Preventive | | | |
| **Appointment Hour of Day** | Current Appointment | Categorical | Hour of day of current appointment | 1 | | |

| Name | Category | Type | Description | Number of Features | Modifiable | Mandatory |
|---|---|---|---|---|---|---|
| Appointment Day of Week | Current Appointment | Categorical | Day of week of current appointment | 1 | | |
| Appointment Month | Current Appointment | Categorical | Month of current appointment | 1 | | |
| Multiple Same-day Appointments | Current Appointment | Boolean | Whether patient has multiple appointments on the same day | 1 | | |
| Estimated Appointment Length | Current Appointment | Integer | Estimated length of appointment | 1 | | |
| Appointment Rescheduled | Current Appointment | Boolean | Whether appointment was rescheduled | 1 | | |
| Appointment Change Count | Current Appointment | Integer | Number of changes | 1 | | |
| Appointment Calls Count | Current Appointment | Integer | Number of calls | 1 | | |
| Appointment Referral | Current Appointment | Boolean | Whether appointment was referred | 1 | | |
| Appointment Confirm Status | Current Appointment | Categorical | • Confirmed<br>• Not Confirmed<br>• Removed | 2 | | |
| Appointment Phone Confirmation Status | Current Appointment | Categorical | e.g. Answered not confirmed, hang up, busy, no answer | | | |
| Copayment Due | Current Appointment | Continuous | Amount of copayment due | 1 | | |
| **FEATURES SPECIFIC TO DM PATIENTS** | | | | | | |
| Has Diabetes | DM Condition | Boolean | • Past 90 days<br>• Past 365 days | 2 | | |
| Has Diabetes Nephropathy | DM Condition | Boolean | • Past 90 days<br>• Past 365 days | 2 | | |
| Has Diabetes Neuropathy | DM Condition | Boolean | • Past 90 days<br>• Past 365 days | 2 | | |
| Has Diabetes Retinopathy | DM Condition | Boolean | • Past 90 days<br>• Past 365 days | 2 | | |
| Has Diabetes Other Complications | DM Condition | Boolean | • Past 90 days<br>• Past 365 days | 2 | | |
| Time Since Diabetes Diagnosis | DM Condition | Continuous | Number of days since diabetes diagnosis | 1 | | |
| Time Since Diabetes Nephropathy Diagnosis | DM Condition | Continuous | Number of days since diabetes nephropathy diagnosis | 1 | | |
| Time Since Diabetes Neuropathy Diagnosis | DM Condition | Continuous | Number of days since diabetes neuropathy diagnosis | 1 | | |

| Name | Category | Type | Description | Number of Features | Modifiable | Mandatory |
|---|---|---|---|---|---|---|
| Time Since Diabetes Retinopathy Diagnosis | DM Condition | Continuous | Number of days since diabetes retinopathy diagnosis | 1 | | |
| Time Since Diabetes Other Complications Diagnosis | DM Condition | Continuous | Number of days since diabetes other complications diagnosis | 1 | | |
| Has Hypoglycemia Event | DM Condition | Boolean | Whether patient has experienced a hypoglycemic event | 1 | | |
| Past Hypoglycemia Event Count | DM Condition | Integer | • 90 days<br>• 180 days<br>• 365 days | 3 | | |
| Has Hyperglycemia Event | DM Condition | Boolean | Whether patient has experienced a hyperglycemic event | 1 | | |
| Past Hyperglycemia Event Count | DM Condition | Integer | • 90 days<br>• 180 days<br>• 365 days | 3 | | |
| DM-related ED Visit Count | DM Condition | Integer | • 90 days<br>• 180 days<br>• 365 days | 3 | | |
| DM-related Hospitalization Count | DM Condition | Integer | • 90 days<br>• 180 days<br>• 365 days | 3 | | |
| New Diabetes Medication | Medication | Boolean | • Whether patient has any new diabetes medication dispensed in the past 6 months | 1 | | |
| Diabetes Insulin Treatment | Medication | Boolean | Whether patient is under insulin treatment | 1 | | |
| Diabetes Insulin Type | Medication | Categorical | • Rapid acting<br>• Short acting<br>• Intermediate (NPH) acting insulin<br>• Long acting<br>Mixed action | | | |
| Insulin Treatment by Pen or Pump | Medication | Boolean | • Whether patient uses a pen-injector or pump instead of syringe or needles | 1 | | |

| Name | Category | Type | Description | Number of Features | Modifiable | Mandatory |
|---|---|---|---|---|---|---|
| Diabetes OAD Treatment | Medication | Boolean | Whether patient given treatment for each class of OAD<br>• Biguanides<br>• Sulfonylureas<br>• Meglitinides<br>• Thiazolidinediones (TZDs)<br>• Glucagonlike peptide-1 (GLP-1) agonists<br>• Dipeptidyl peptidase IV (DPP-4) inhibitors<br>Alpha-glucosidase inhibitors | 7 | | |
| Dosage Count | Medication | Integer | • Number of current dosages required in a day | 1 | | |
| Average Frequency of Current Medication | Medication | Continuous | Average frequency across all medications | 1 | | |
| Current Medication Class Count | Medication | Integer | Number of current unique medication classes | 1 | | |
| Chronic Medication Count | Medication | Integer | Number of chronically used medications | 1 | | |
| Current Medication Route Count | Medication | Integer | Number of current unique medication routes (e.g. Oral, Subcutaneous, Intramuscular) | 1 | | |
| Average Medication Length of Fill | Medication | Continuous | Average length of fill across all medications | 1 | | |
| Average Number of Available Refills | Medication | Continuous | Average number of available refills across all medications | 1 | | |
| Prescribed Medication Count | Medication | Integer | Number of unique medications prescribed in<br>• Past 90 days<br>• Past 180 days<br>Past 365 days | 3 | | |

| Name | Category | Type | Description | Number of Features | Modifiable | Mandatory |
|---|---|---|---|---|---|---|
| Prescribed Medication Class Count | Medication | Integer | Number of unique medication classes prescribed in<br>• Past 90 days<br>• Past 180 days<br>• Past 365 days | 3 | | |
| Prescriber Count | Medication | Integer | Number of prescribers in<br>• Past 90 days<br>• Past 180 days<br>• Past 365 days | 3 | | |
| Rx Fill by Mail | Medication | Boolean | • Whether the patient's prescriptions are filled by mail | 1 | | |
| Past HbA1c Tests Count | Compliance | Integer | Number of past HbA1c tests in<br>• Past 90 days<br>• Past 180 days<br>Past 365 days | 3 | | |
| Past Glucose Tests Count | Compliance | Integer | Number of past glucose tests in<br>• Past 90 days<br>• Past 180 days<br>• Past 365 days | 3 | | |
| Visit Non-Compliance | Compliance | Continuous | Patient's visit compliance rate in<br>• Past 90 days<br>• Past 180 days<br>• Past 365 days | 3 | | |
| Medication Non-Compliance | Compliance | Continuous | Patient's PDC in<br>• Past 90 days<br>• Past 180 days<br>• Past 365 days | 3 | | |

## Relevant Medications

Medications identified from [69] which may be used as features

- Glucagon-like peptide-1 agonists
- Dipeptidyl peptidase-4 inhibitor
- Acarbose or Miglitol
- Repaglinide or Nateglinide
- Metformin
- Pramlintide
- ACE-inhibitors
- Angiotensin II blockers
- Beta adrenergic blockers
- Calcium channel blockers

- Thiazides/related diuretics
- Potassium sparing diuretics
- Loop diuretics
- Alpha-adrenergic blockers
- Statin
- Bile acid resins
- Fibrates
- Niacin
- Ezetimibe
- Antidepressant
- Cholinesterase inhibitor
- Benzodiazepines
- Neuroleptics
- Memantine
- Anti-seizure meds
- Anti-arrhythmia meds
- Digoxin
- Estrogen agonist
- Anti-bisphosphonate
- Corticosteroid
- Levothyroxine
- Codeine
- Tramadol
- Quinolone
- Calcitriol
- Epoetin
- Warfarin